%% file: main.tex
\documentclass[a4paper]{report}
\usepackage[utf8]{inputenc}
\usepackage{amsmath,amssymb,amsfonts,bbm,tcolorbox}
\usepackage{graphicx,caption,subcaption}
\usepackage{cleveref}
\usepackage{pdflscape}
\usepackage{algorithm,algorithmic}
\usepackage{mathrsfs}
\usepackage{setspace} 
\usepackage{titlepic}
\usepackage{booktabs, url, color, epstopdf}
\usepackage[bottom]{footmisc}

\usepackage[margin=1.2in]{geometry}
\usepackage{pdfpages}
\usepackage{textcomp}
\usepackage[acronym,symbols,nogroupskip]{glossaries-extra}
\usepackage[sort&compress]{natbib} 
\usepackage{afterpage}
\usepackage{stringstrings}
\usepackage[tableposition=top]{caption}
\tcbuselibrary{skins}
\usepackage{float}
\usepackage{soul}
\usepackage{changepage}
\floatstyle{plaintop}
\restylefloat{table}

\counterwithin{algorithm}{chapter}
\newcommand{\tab}{\hspace*{2em}}
\newcommand{\Mod}[1]{\ (\mathrm{mod}\ #1)}

\newcommand\blankpage{%
	\null
	\thispagestyle{empty}%
	\addtocounter{page}{-1}%
	\newpage}

\graphicspath{{images/}}

\newtcolorbox[auto counter,number within=chapter]{proof}[2][]{%
  enhanced,
  fonttitle=\large,coltitle=black,titlerule=0pt,
  title={Proof~\thetcbcounter:~#2},
  #1,
  sharp corners,
  colbacktitle=white,
  colback=white,
  drop fuzzy shadow
}

\makeglossaries
\input{acronym}

\begin{document}

	\begin{titlepage}
	\begin{center}
		\vspace{3cm}
		\textbf{LEARNING ON LARGE DATASETS USING BIT-STRING TREES}
		
		\vspace{3cm}

		\textbf{PRASHANT GUPTA}\\
		\vspace{3cm}

		\includegraphics[width=4cm, height=4cm]{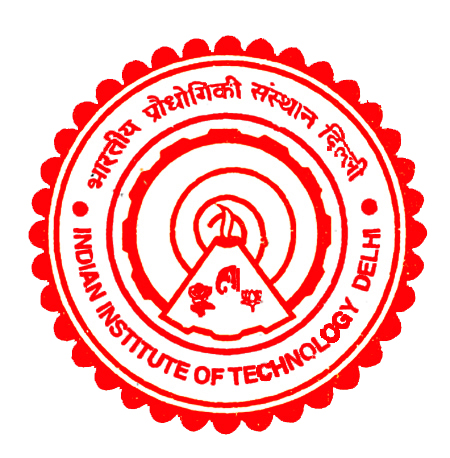}
		\\
		\vspace{5cm}
		\textbf{DEPARTMENT OF ELECTRICAL ENGINEERING}\\
		\textbf{INDIAN INSTITUTE OF TECHNOLOGY DELHI}\\
		\textbf{April 2023}\\
	\end{center}
	\afterpage{\blankpage}
\end{titlepage}	
\afterpage{\blankpage}
\begin{titlepage}
	\topskip0pt
	\vspace*{\fill}
	\textbf{\textcopyright Indian Institute of Technology Delhi (IITD), New Delhi, 2023}
	\vspace*{\fill}
\end{titlepage}
\afterpage{\blankpage}
\begin{titlepage}
	\begin{center}
		\vspace{2cm}
		\textbf{LEARNING ON LARGE DATASETS USING BIT-STRING TREES}
		
		\vspace{2cm}
		by\\
		
		\vspace{1cm}
		\textbf{PRASHANT GUPTA}\\
		
		\vspace{1cm}
		DEPARTMENT OF ELECTRICAL ENGINEERING\\
		\vspace{0.5cm}
		Submitted\\
		\vspace{0.25cm}
		\textit{in fulfillment of the requirements of the degree of Doctor of Philosophy\\
			to the}\\
		
		\vspace{3cm}
		
		\includegraphics[width=5cm, height=5cm]{iitlogo-21-3}
		\\
		\vspace{3cm}
		
		\textbf{INDIAN INSTITUTE OF TECHNOLOGY DELHI}\\
		\textbf{April 2023}\\
	\end{center}
\end{titlepage}	

\newpage
\setcounter{page}{1}

\newpage
\renewcommand{\thepage}{\roman{page}}
\input{certificate}
\addcontentsline{toc}{chapter}{Certificate}%

\includepdf{empty_page.pdf}

\renewcommand{\thepage}{\roman{page}}
\input{acknowledgements}
\addcontentsline{toc}{chapter}{Acknowledgements}%

\includepdf{empty_page.pdf}

\begin{abstract}
	\thispagestyle{plain}
	\renewcommand{\thepage}{\roman{page}}
	\addcontentsline{toc}{chapter}{Abstract}
	\setcounter{page}{5}
	\input{abstract}

\end{abstract}

\renewcommand{\thepage}{\roman{page}}
\setcounter{page}{7}
\addcontentsline{toc}{chapter}{Hindi Abstract}%
\includepdf[pagecommand={\thispagestyle{plain}}, pages={1-2}]{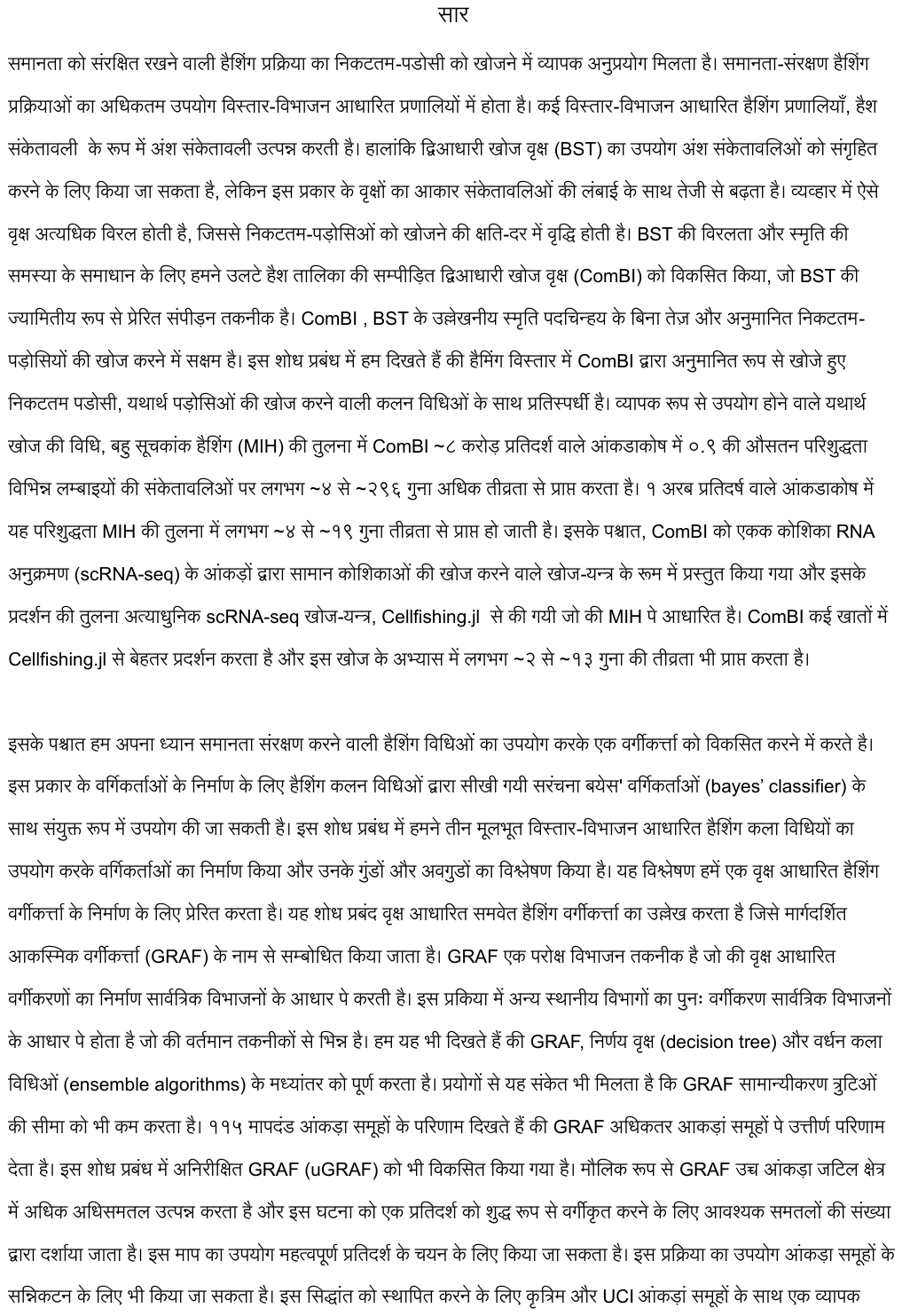}

\renewcommand{\thepage}{\roman{page}}


\tableofcontents
\includepdf[pagecommand={\thispagestyle{plain}}, pages={1}]{empty_page.pdf}

\listoffigures
\addtocontents{lof}{\protect\addcontentsline{toc}{chapter}{List of figures}}

\listoftables
\addtocontents{lot}{\protect\addcontentsline{toc}{chapter}{List of tables}}

\listofalgorithms
\addtocontents{loa}{\protect\addcontentsline{toc}{chapter}{List of algorithms}}

\includepdf[pagecommand={\thispagestyle{plain}}, pages={1}]{empty_page.pdf}

\printglossary[type=\acronymtype,nonumberlist,title=List of abbreviations]
\includepdf[pagecommand={\thispagestyle{plain}}, pages={1}]{empty_page.pdf}

\newpage
\renewcommand{\thepage}{\arabic{page}}
\setcounter{page}{1}

\chapter{Introduction}
\input{introduction}

\chapter[The ComBI: A bit-string tree for fast approximate search in hamming space]{The ComBI: A bit-string tree for fast approximate search in hamming space \footnotemark}
\footnotetext{The work presented in this chapter has been published: \textit{Gupta P, Jindal A, Sengupta D. ComBI: Compressed binary search tree for approximate k-NN searches in Hamming space. Big Data Research. 2021 Jul 15;25:100223.}}

\input{ch2combi}

\chapter{Generalized hashing classifier}

\input{ch3hashingClassifier}

\chapter[The GRAF: A bit-string tree as a hashing classifier]{The GRAF: A bit-string tree as a hashing classifier\footnotemark}
\footnotetext{Part of the work presented in this chapter has also been uploaded to a pre-print server: \textit{Gupta P, Jindal A, Jayadeva, Sengupta D. Guided Random Forest and its application to data approximation. arXiv preprint arXiv:1909.00659. 2019 Sep 2.}}

\input{ch4graf}
\includepdf[pagecommand={\thispagestyle{plain}}, pages={1}]{empty_page.pdf}

\chapter[Utilization of neighborhood learned by the bit-string trees]{Utilization of neighborhood learned by the bit-string trees\footnotemark}
\footnotetext{Part of the work presented in this chapter has also been uploaded to a pre-print server: \textit{Gupta P, Jindal A, Jayadeva, Sengupta D. Guided Random Forest and its application to data approximation. arXiv preprint arXiv:1909.00659. 2019 Sep 2.}}

\input{ch5applications}

\includepdf[pagecommand={\thispagestyle{plain}}, pages={1}]{empty_page.pdf}

\chapter[Learning nucleotide sequence context of cancer mutations and its applications in survivability]{Learning nucleotide sequence context of cancer mutations and its applications in survivability \footnotemark}
\footnotetext{The work presented in this chapter has been published: \textit{Gupta P, Jindal A, Ahuja G, Sengupta D. A new deep learning technique reveals the exclusive functional contributions of individual cancer mutations. Journal of Biological Chemistry. 2022 Jun 24:102177.}\\ A concise version of the work is also present at \textit{Gupta, P., Jindal, A. and Sengupta, D. Deep learning discerns cancer mutation exclusivity. bioRxiv:2020.04.09.022731. 2020 Apr 10.}}

\input{ch6CRCS}

\chapter{Conclusions and future work}

\input{conclusion}

\includepdf[pagecommand={\thispagestyle{plain}}, pages={1}]{empty_page.pdf}

\color{black}

\bibliographystyle{unsrt}

\begingroup
	\setlength{\bibsep}{10pt}
	\setstretch{1}
	\bibliography{main}
\endgroup


\chapter*{List of publications}
\addcontentsline{toc}{chapter}{\numberline{}List of publications}

\input{publications}

\newpage
\includepdf[pagecommand={\thispagestyle{plain}}, pages={1}]{empty_page.pdf}

\chapter*{Brief biodata of author}
\addcontentsline{toc}{chapter}{\numberline{}Brief biodata of author}

\input{biodata}

\end{document}

%% file: acronym.tex
\newacronym{combi}{ComBI}{Compressed BST of Inverted hash tables}
\newacronym{bst}{BST}{Binary Search Tree}
\newacronym{bsts}{BSTs}{Binary Search Trees}
\newacronym{mih}{MIH}{Multi-Index Hashing}
\newacronym{hbst}{HBST}{Hamming Distance Embedding BST}
\newacronym{rp}{RP}{Random Partition}
\newacronym{mrpt}{MRPT}{Multiple Random Projection Trees}
\newacronym{hwt}{HWT}{Hamming Weight Tree}
\newacronym{ibst}{IBST}{Inverted-Hash-Table BST}
\newacronym{nn}{NN}{Nearest Neighbor}
\newacronym{nns}{NNs}{Nearest Neighbors}
\newacronym{hd}{HD}{Hamming Distance}
\newacronym{lsh}{LSH}{Locality Sensitive Hashing}
\newacronym{lsb}{LSB}{Least Significant Bit}
\newacronym{msb}{MSB}{Most Significant Bit}
\newacronym{fdr}{FDR}{False Discovery Rate}
\newacronym{graf}{GRAF}{Guided Random Forest}
\newacronym{rf}{RF}{Random Forest}
\newacronym{et}{ET}{Extremely Randomized Trees}
\newacronym{ots}{OTs}{Oblique decision trees}
\newacronym{ot}{OT}{Oblique Tree}
\newacronym{oc1}{OC1}{Oblique Classifier 1}
\newacronym{cart}{CART}{Classification and Regression Trees}
\newacronym{mml}{MML}{Maximum Message Length}
\newacronym{svm}{SVM}{Support Vector Machine}
\newacronym{gto}{GTO}{Global Tree Optimization}
\newacronym{hepts}{HEPTS}{Hybrid Extreme Point Tabu Search}
\newacronym{gepsvm}{GEPSVM}{Proximal SVM with Generalized Eigenvalues}
\newacronym{mpsvm}{MPSVM}{Multi-Surface Proximal SVM}
\newacronym{pca}{PCA}{Principal Component Analysis}
\newacronym{lda}{LDA}{Linear Discriminant Analysis}
\newacronym{co2}{CO2}{Continuously Optimized Oblique}
\newacronym{wodt}{WODT}{Weighted Oblique Decision Trees}
\newacronym{lbfgs}{L-BFGS}{Limited-memory  Broyden–Fletcher–Goldfarb–Shanno Algorithm}
\newacronym{rvfl}{RVFL}{Random Vector Functional Link Network}
\newacronym{node}{NODE}{Neural Oblivious Decision Ensembles}
\newacronym{roi}{ROI}{Radius of Influence}
\newacronym{cpu}{CPU}{Central Processing Unit}
\newacronym{gpu}{GPU}{Graphics Processing Unit}
\newacronym{ram}{RAM}{Random Access Memory}
\newacronym{ada}{ADA}{Adaboost}
\newacronym{xgb}{XGB}{Extreme Gradient Boosting (XGBoost)}
\newacronym{gb}{GB}{Gradient Boosting}
\newacronym{npps}{NPPS}{Neighborhood Property-Based Pattern Selection}
\newacronym{svis}{SVIS}{Small Votes Instance Selection}
\newacronym{sv}{SV}{Support Vector}
\newacronym{crcs}{CRCS}{Continuous Representation of Codon Switches}
\newacronym{cfdna}{cfDNA}{Cell-free DNA}
\newacronym{blca}{BLCA}{Bladder Urothelial Carcinoma}
\newacronym{hcc}{HCC}{Hepatocellular Carcinoma}
\newacronym{luad}{LUAD}{Lung Adenocarcinoma}
\newacronym{eqtl}{eQTL}{Expression Quantitative Trait Loci}
\newacronym{cnn}{CNN}{Convolutional Neural Network}
\newacronym{blac}{BLAC}{Bidirectional Long Short-Term Memory with Attention $\&$ CRCS embeddings}
\newacronym{lstm}{LSTM}{Long Short-Tem Memory}
\newacronym{bilstm}{bi-LSTM}{Bidirectional Long Short-Tem Memory}
\newacronym{snv}{SNV}{Single Nucleotide Variant}
\newacronym{tsne}{tSNE}{t-Distributed Stochastic Neighbor Embedding}
\newacronym{nlp}{NLP}{Natural Language Processing}
\newacronym{umap}{UMAP}{Uniform Manifold Approximation and Projection}
\newacronym{lgd}{LGD}{Local Gene Duplication}
\newacronym{tmb}{TMB}{Tumor Mutational Burden}
\newacronym{ctdna}{ctDNA}{Circulating Tumor DNA}
\newacronym{ap}{AP}{Average Precision}
\newacronym{pr}{PR}{Precision-Recall}
\newacronym{oncokb}{OncoKB}{Oncology Knowledge Base}
\newacronym{intogen}{intOgen}{Integrative Onco Genomics}
\newacronym{cgi}{CGI}{Cancer Genome Interpreter}
\newacronym{csc}{CSC}{Cancer-Stem-Cell}
\newacronym{escc}{ESCC}{Esophageal Squamous Cell Carcinoma}
\newacronym{skcm}{SKCM}{Skin Cutaneous Melanoma}
\newacronym{uec}{UEC}{Undifferentiated Endometrial Carcinoma}
\newacronym{lusc}{LUSC}{Lung Squamous Cell Carcinoma}
\newacronym{hnsc}{HNSC}{Head and Neck Squamous Cell Carcinoma}
\newacronym{nsclc}{NSCLC}{Non-Small Cell Lung Cancer}
\newacronym{dna}{DNA}{Deoxyribonucleic Acid}
\newacronym{rna}{RNA}{Ribonucleic Acid}
\newacronym{mrna}{mRNA}{Messenger RNA}
\newacronym{indel}{indel}{Insertions \& Deletions}
\newacronym{gwas}{GWAS}{Genome-Wide Association Studies}
\newacronym{who}{WHO}{World Health Organization}
\newacronym{wgs}{WGS}{Whole Genome Sequencing}
\newacronym{wes}{WES}{Whole Exome Sequencing}
\newacronym{exac}{ExAC}{Exome Aggregation Consortium}
\newacronym{gnomad}{gnomAD}{Genome Aggregation Database}
\newacronym{dbsnp}{dbSNP}{Single Nucleotide Polymorphism Database}
\newacronym{cosmic}{COSMIC}{Catalogue Of Somatic Mutations In Cancer}
\newacronym{cbio}{cBioPortal}{cBio Cancer Genomics Portal}
\newacronym{ucsc}{UCSC}{University of California, Santa Cruz}
\newacronym{vcf}{VCF}{Variant Call Format}
\newacronym{sift}{SIFT}{Sorting Intolerant From Tolerant}
\newacronym{pph2}{PolyPhen2}{Polymorphism Phenotyping v2}
\newacronym{blacs}{BLACs}{Combined BLAC score}
\newacronym{ws}{$ws$}{window size}
\newacronym{nsr}{$nsr$}{negative sampling rate}
\newacronym{ugraf}{uGRAF}{Unsupervised GRAF}
\newacronym{scrna}{scRNA-seq}{single-cell RNA sequencing}
\newacronym{srplsh}{SRP-LSH}{Signed Random Projection LSH}
\newacronym{sph}{SPH}{Similarity Preserving Hashing}
\newacronym{dt}{DT}{Decision Trees}
\newacronym{gbm}{GBM}{Glioblastoma Multiforme}
\newacronym{ewh}{EWH}{Error Weighted Hashing}
\newacronym{bert}{BERT}{Bidirectional Encoder Representations from Transformers}
\newacronym{oob}{OOB}{Out of Bag}
\newacronym{som}{SOM}{Self-Organizing Maps}
\newacronym{dbscan}{DBSCAN}{Density-Based Spatial Clustering of Applications with Noise}

%% file: certificate.tex
\begin{center}
\LARGE{\textbf{Certificate}}\\
\end{center}
\vspace{0.3in}

\doublespacing{
\tab This is to certify that the thesis entitled \textbf{``Learning On Large Datasets Using Bit-String Trees''}, being submitted by \textbf{Prashant Gupta} for the award of the degree of \textbf{Doctor of Philosophy} to the Department of Electrical Engineering, Indian Institute of Technology Delhi, is a record of bonafide work done by him under my supervision and guidance. The matter embodied in this thesis has not been submitted to any other University or Institute for the award of any other degree or diploma.}\\

\vspace{0.6in}

\begin{adjustwidth}{-0.5cm}{}
\begin{tabular}{ll}
	\begin{minipage}[t]{.5\textwidth}
		\begin{flushleft}
			\large{\textbf{Dr. Jayadeva}}			\\
			\vspace{1mm}
			\normalsize{\textit{Professor}} \\
			\vspace{1mm}
			\text{Department of Electrical Engineering,} \\
			\vspace{1mm}
			\text{Indian Institute of Technology Delhi,} \\
			\vspace{1mm}
			\text{Hauz Khas, New Delhi - 110016,} \ \\
			\vspace{1mm}
			\text{India.}
		\end{flushleft}
	\end{minipage} &

	\begin{minipage}[t]{.55\textwidth}
		\begin{flushleft}
			\large{\textbf{Dr. Vibhor Kumar}}			\\
			\vspace{1mm}
			\normalsize{\textit{Associate Professor}} \\
			\vspace{1mm}
			\text{Department of Computational Biology,}\\
			\vspace{1mm}
			\text{Indraprashta Institute of Information Technology,} \\
			\vspace{1mm}
			\text{Okhla Phase III, Delhi - 110020,} \ \\
			\vspace{1mm}
			\text{India.}
		\end{flushleft}
	\end{minipage} \\

	\begin{minipage}[t]{.55\textwidth}
		\begin{flushleft}
		    \vspace{20mm}
			\large{\textbf{Dr. Debarka Sengupta}}			\\
			\vspace{1mm}
			\normalsize{\textit{Associate Professor}} \\
			\vspace{1mm}
			\text{Department of Computer Science \& Engineering,} \\
			\vspace{1mm}
			\text{Department of Computational Biology,}\\
			\vspace{1mm}
			\text{\textit{Head}, Center for Artificial Intelligence,}
                \vspace{1mm}
			\text{Indraprashta Institute of Information Technology,} \\
                \vspace{1mm}
			\text{Okhla Phase III, Delhi - 110020,} \\
			\vspace{1mm}
			\text{India.}\\
			\vspace{1mm}
			\normalsize{\textit{(Adj.) Associate Professor}}\\
			\vspace{1mm}
			\text{Institute of Health \& Biomedical Innovation,}\\
			\vspace{1mm}
			\text{QUT, Australia}
		\end{flushleft}
	\end{minipage} &\\

\end{tabular}
\end{adjustwidth}




%% file: acknowledgements.tex
\begin{center}
\LARGE{\textbf{Acknowledgments}}\\
\end{center}

\vspace{0.3in}

I would like to thank my supervisors Prof. Jayadeva (IITD), Assoc. Prof. Debarka Sengupta (IIITD), Assoc. Prof. Vibhor Kumar (IIITD), Research Committee Members Prof. Indra Narayan Kar (IITD), Prof. Shouri Chatterjee (IITD), and Asst. Prof. Vivekanandan Perumal (IITD); faculty members with whom I worked - Retd. Prof. Suresh Chandra (IITD), Assoc. Prof. Gaurav Ahuza (IIITD), Retd. Prof. Munishwar Nath Gupta (IITD), and Prof. Rajeev Narang (AIIMS);  friends and colleagues Aashi Jindal, Dr. Sumit Soman, Dr. Udit Kumar, Aashish Rajiv, Dr. Mayank Sharma, Dr. Shruti Sharma, Dr. Himanshu Pant, Devesh Bajpai, Ritika Bajpai, Sanjay Pandey; department staff members Rakesh Kumar, Yatindra Mani, Mukesh, and Ritwik Pahari; my parents, my parents-in-law, my wife Aashi Jindal and other family members Priyanka Gupta, Sahil Jindal, and Khushboo Mehrotra for supporting me through this journey.

I would like to extend my special thanks to my wife Aashi Jindal for being the part of both journeys, academic and life, and supporting me through the ups of downs.

\vspace{1in}

\begin{flushright}
\textit{(Prashant Gupta)}
\end{flushright}

%% file: abstract.tex
Similarity preserving hashing finds widespread application in nearest-neighbor search. The widely used form of similarity preserving hashing is space-partitioning-based hashing. Many space partitioning-based hashing techniques generate bit codes as hash codes. Although \acrfull{bsts} can be used for storing bit codes, their size grows exponentially with code length. In practice, such a tree turns out to be highly sparse, increasing the \textit{miss-rate} of nearest neighbor searches. To tackle sparsity and memory issues of \acrshort{bst}, we first developed \acrfull{combi}, a geometrically motivated compression technique for \acrshort{bsts}. \acrshort{combi} enables fast and approximate nearest neighbor searches without a significant memory footprint over \acrshort{bsts}. We show, that approximate search in \acrshort{combi} is competitive with an exact search algorithm in retrieving the nearest neighbors in a hamming space. On a database containing $\sim$80 million samples, \acrshort{combi} yields an average precision of 0.90, at $\sim$4X - $\sim$296X improvements in run-time across different code lengths when compared to \acrfull{mih}, a widely used exact search method. On a database consisting of 1 billion samples, this value of precision (0.90) is reached at $\sim$4X - $\sim$19X improvements in run-time. Next, the \acrshort{combi} has been shown as a search engine for \acrfull{scrna} data, and its performance is compared with the state-of-the-art \acrshort{scrna} search engine method, Cellfishing.jl, which is based on the \acrshort{mih}. The \acrshort{combi} outperforms Cellfishing.jl in multiple accounts. The achieved speed-up in the search is around $\sim$2 - $\sim$13.

We next shift our attention to using similarity preserving hashing to build a classifier. The learned structure of hashing algorithms is suitable to be combined with a Bayes' classifier. We explored the construction of three basic space-partitioning-based hashing algorithms and identified their pros and cons. This motivated us to build a tree-based hashing classifier. We present \acrfull{graf}, a tree-based ensemble hashing classifier that realizes global partitioning by extending the idea of building oblique decision trees with localized partitioning. We show that \acrshort{graf} bridges the gap between decision trees and boosting algorithms. Experiments indicate that it reduces the generalization error bound. Results on 115 benchmark datasets show that \acrshort{graf} yields comparable or better results on a majority of datasets. We also build an unsupervised version of \acrshort{graf}, \acrfull{ugraf}, to perform guided hashing. The \acrshort{graf} fundamentally works by generating more hyperplanes in the region of high data complexity and this phenomenon is represented by the number of planes required to classify a sample correctly. This measure can be used for importance sampling. In the next part of the thesis, this direction is explored to build a data approximator using \acrshort{graf}. An extensive empirical evaluation with simulated and UCI datasets was performed to establish the theory. The proposed methodology is compared with the two state-of-the-art importance sampling algorithms. An analogy between \acrfull{svm} and the samples marked by \acrshort{graf} as of high importance is also developed.

We then show that the learned neighborhood of a sample can be used to estimate the confusion around the sample in a scalable manner. We utilized \acrshort{ugraf} and \acrshort{combi} to estimate the per-sample classifiability. An empirical evaluation of estimated values is presented. We show how per-sample classifiability can be used to estimate cancer patient survivability. 

Cancer is a disease of the genome. Genomic changes resulting in cancer can be inherited, brought on by environmental carcinogens, or may result from random replication errors. Mutations continue to spread after the induction of carcinogenicity and significantly change cancer genomes. Most cancer-related somatic mutations are indistinguishable from germline variants or other non-cancerous somatic mutations, even though only a small subset of driver mutations have been identified and characterized thus far. Thus, such overlap makes it difficult to understand many harmful but unstudied somatic mutations. The main bottleneck results from patient-to-patient variation in mutational profiles, which makes it challenging to link particular mutations with a particular disease outcome. This thesis introduces a newly developed method called \acrfull{crcs}. This deep learning-based approach enables us to produce numerical vector representations of genetic changes, enabling a variety of machine learning-based tasks. We show how \acrshort{crcs} can be used in three different ways. First, we show how it can be used to find cancer-related somatic mutations without matched normal samples. Second, the suggested method makes it possible to find and study driver genes. Finally, we created a numerical representation of mutations by combining a sequence classifier with \acrshort{crcs}. These representations are used to score individual mutations in a tumor sample using per-sample classifiability, which was found to be predictive of patient survival in \acrfull{blca}, \acrfull{hcc}, and \acrfull{gbm}. Taken together, we propose \acrshort{crcs} as a valuable computational tool for analysis of the functional significance of individual cancer mutations.

%% file: introduction.tex
In today's data-driven age faster extraction of information from big data is imperative. At the center of the advancement of information extraction, lie similarity preserving hashing which enables the compact representation of samples~\cite{manu::ksh,manu::lsh,charikar2002similarity,semantic,manu::sh,manu::sperical,manu::ssh,manu::shl,wang2020visual}. These representations are then compared to extract similar samples. The concept of finding nearest neighbors has widespread applications in search engines, computer vision, security systems etc~\cite{oostveen2002feature,esmaeili2010robust,esmaeili2011fast,miller2005audio,cellfishing,plageras2018efficient,stergiou2018security,yu2018four,cellatlassearch}. The field of similarity search has progressed multifold with the invention of \acrfull{lsh}~\cite{lsh}. The \acrshort{lsh} is defined as:

"\textit{A family $\mathcal{H}$ for a metric space $\mathcal{M} = (M, d)$, a threshold  $R > 0$ and an approximation factor $c > 1$. It is a family of function $h:M\rightarrow S$ which maps elements from the metric space to a bucket $s \in S$. The family is called $(R, cR, P_1, P_2)$-sensitive if for any two points $p, q \in \mathbb{R}^d$.}
\begin{itemize}
	\item \textit{if $|p-q| \leq R$ then $\text{Pr}_\mathcal{H}[h(p) = h(q)] \geq P_1$}
	\item \textit{if $|p-q| \geq cR$ then $\text{Pr}_\mathcal{H}[h(p) = h(q)] \leq P_2$}"
\end{itemize}
\noindent
This definition states that the samples in the closer vicinity will get the same hash code with probability $\geq P_1$ and far away sample will get the same hash code with probability $\leq P_2$.  

With the advent of \acrshort{lsh}, there have been many algorithms in the \acrshort{lsh} family that attempt to find a better representation of samples to make the search faster and more precise. Among them, the most common are the ones that transfer the sample into hamming space and creates a bit code~\cite{manu::ksh,manu::lsh,charikar2002similarity,semantic,manu::sh,manu::sperical,manu::ssh,manu::shl,wang2020visual}. Some of them are computationally in-expensive~\cite{manu::lsh,charikar2002similarity} making them suitable for industrial applications. To perform a search in the hamming space, the developed algorithms focus on storing the indexes such that the nearest neighbors can be found by performing the minimum number of bit comparisons~\cite{manu::mih,manu::bridge,manu::ewh,lshforest}. 

Hashing algorithms that assign bit codes to every sample generally work by partitioning the spaces into smaller regions (disjoint sets, partition, and, regions would be used interchangeably in this thesis). Sample within a region gets the same bit code. Space-partitioning-based algorithms are not only used for hashing, but they also have been common in classification literature. The most common example of such algorithms is tree-based space-partitioning algorithms. There are many tree-based algorithms that have been developed that work in Euclidean space, such as \acrlong{rp} Trees (RP)~\cite{rptrees,dasgupta2013randomized}, kd-trees~\cite{kdtree}, etc. In fact, tree-based algorithms have been a topic of interest in many areas of research in artificial intelligence. Apart from classification and regression~\cite{cart,rf,et}, nearest neighbor search~\cite{balltree,kdtree,rptrees,bbdtree,lshforest}, clustering~\cite{birch,hdbscan,zhang2013agglomerative}, anomaly detection~\cite{isolationforest}, concept drift~\cite{arf} etc. have also seen advancements due to tree arrangements. These algorithms have been applied successfully in many applications such as rare event detection~\cite{fire}, finger-print analysis~\cite{do2015classifying}, remote sensing~\cite{belgiu2016random}, pedestrian detection~\cite{correia2016oblique}, time-series forecasting~\cite{qiu2017oblique}, visual tracking~\cite{zhang2017robust}, genomic data analysis~\cite{chen2012random}, internet of things (IoT)~\cite{kumar2017resource}, head pose estimation~\cite{liu2016robust} etc. The wider application of tree-based algorithms stems from the fact that the data can be analyzed at multiple levels of granularity. In general, in a tree, data is divided into groups hierarchically based on pre-defined criteria. The size of the group can be controlled by defining the height of the tree or the group size. Mostly, the samples in a group have similar properties. Now, these groups can then be interpreted in different ways based on the task at hand, for example, classification, regression, clustering, data approximation, vector quantization, nearest neighbors, etc.

\section{Scope and objectives}

This thesis focuses on the inferences that can be built using the neighborhood learned by bit string trees. \acrfull{lsh} and its variants generally produce bit codes as hashes. These hashes are then fed into another algorithm that utilizes the information stored in bit codes to find the nearest neighbors. However, the generated bit codes or hash codes can be used in varied ways to extract the inferences from the data by utilizing the learned neighborhood by the hashing technique. 

To achieve this, we first attempted to solve the major issue of the sparse neighborhood in large-bit codes. The number of possible bins increases exponentially with the increasing length of the bit code. In such cases, total possible bin counts are way more than the total number of samples in the data. Thus, many bit codes do not get assigned to any sample. In other words, there are a lot of empty bins. Thus, the probability of the query sample falling into empty bins increases significantly. Such situations increase the search time. We examine this specific case to find the structure in the bit patterns to reduce the nearest neighbor search time. 

Second, we identified that a classifier can be constructed by overlaying class label information on the bins created by hashing. We also identified that the learned structure to reduce the search time overlayed with label information could also behave as a classifier. To improve the classifier's performance, we reverse-engineered the structure's construction to ingest class information in its build strategy. We also identified that the resulting strategy to build the classifier with label information could also be used for importance sampling.

Next, we identified that the neighborhood of every sample could be used to estimate the confusion around it. The aggregation of the confusion across whole data has been termed as \textit{classifiability} in literature.. Traditionally, classifiability is computed by selecting $n$ samples in the $r$ neighborhood in euclidean space. We extend this idea in the hamming space and argue that per-sample classifiability is of more importance. We present a use-case of the classifiability in estimating cancer patient survivability. 

While attempting to estimate the survivability, we identified a drastic need for a holistic infrastructure that can numerically represent the mutation in the human genome. To achieve this, a new abstract representation of change in a codon, called codon switches, is developed by imitating the \textit{central dogma} of molecular biology. A codon switch does not have the position information, it only contains the information regarding nucleotide change. This abstract representation is then used to learn the numerical embedding of codon switches. These embeddings were then used in the downstream task of annotating the mutations based on their cancer propensity. The second output of the annotation network was a numerical representation of the mutations, which contains both positional and change information. These mutation embeddings are then used to learn the confusion in the neighborhood of every sample in the dataset to vet the mutations in cancer patients against their survival.

Thus, based on the spatial arrangement of bins we asked the following questions:

\begin{itemize}
	\item Is it possible to build a faster nearest neighbor search method?
	\item How can hashing be used as a classifier?
	\item Can we identify sample ranking strategy that can scale for large datasets?
	\item Is it possible to learn the semantic representation (embedding) of mutations?
	\item Can these learned embeddings with sample ranking be used to find the survivability of patients?
\end{itemize}

\section{Space-partitioning-based algorithms}

In this section of the introduction, we will discuss how space partitioning-based hashing algorithms are designed and nearest neighbor searches are performed in them. We will also discuss the anatomy of space-partitioning-based algorithms in the context of classification.

\subsection{Space-partitioning-based hashing and nearest neighbor search}

\begin{figure*}[!ht]
	\makebox[1\textwidth][c]{
		\resizebox{1.1\linewidth}{!}{
			\includegraphics{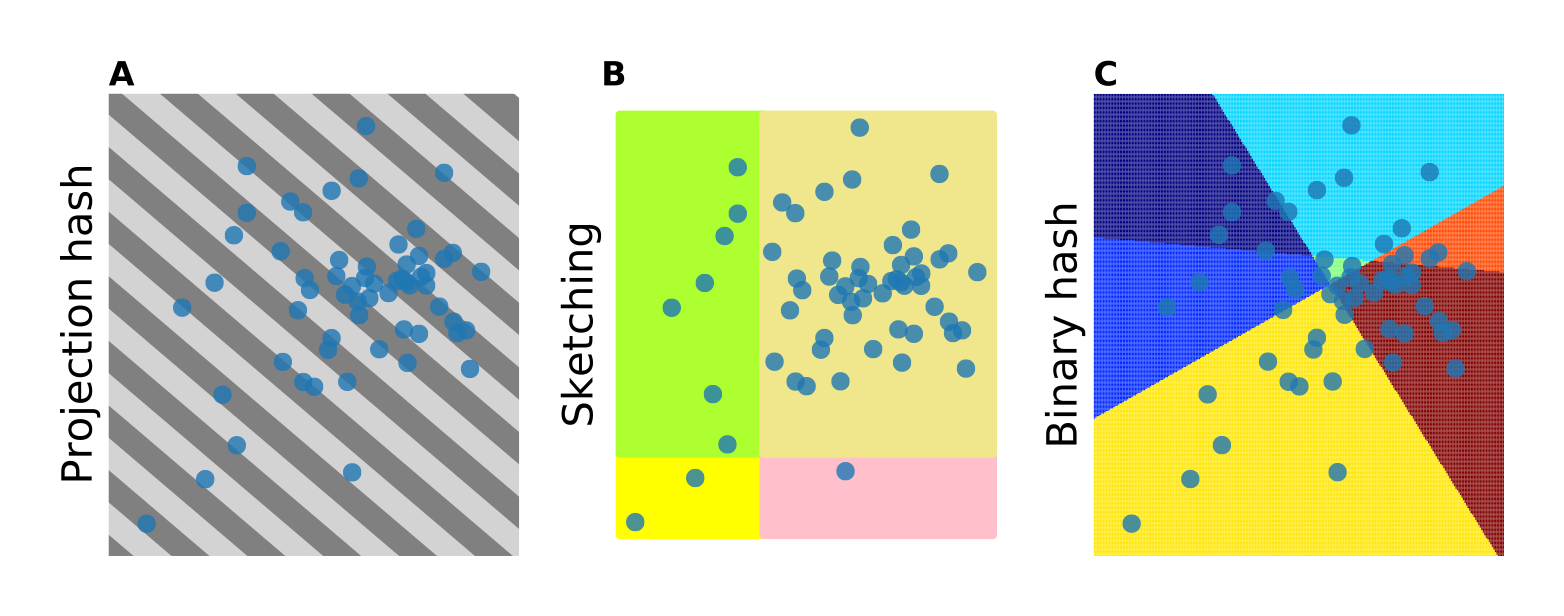}
		}
	}
	\caption{\textbf{Example of space-partitioning-based hash functions.} \textbf{A)} Projection hash which assigns hash code by quantizing the randomly projected values. \textbf{B)} Sketching, which assigns hash code by thresholding every feature. \textbf{C)} Binary hash randomly divides the space and assigns bit code to every sample based on their location in the space.}
	\label{fig:ch1:hashexample}
\end{figure*}

The definition of \acrshort{lsh} can be used to define multiple hash functions using different criteria. Here, we explore three basic ideas of designing hash functions by dividing spaces. The simplest way to hash by space-partitioning is to project the samples into a random direction and then quantize the space of the projected values to assign the hash code to every sample. All the samples falling into one quantum will be assigned the same hash code. This hashing technique is termed the projection hash~\cite{charikar2002similarity}. Figure~\ref{fig:ch1:hashexample}A presents an example of the projection hash.

Another way to divide the space is by binarizing the features. The binarization is performed by selecting a random threshold for every feature. The binarization results in a bit code. To assign hash codes of every sample, this generated bit code is converted into an integer value by weighting and adding every bit in the bit code with random integers. This idea is known as sketching~\cite{lv2006ferret,wang2007sizing}. The sketching divides the space into rectangular regions. Figure~\ref{fig:ch1:hashexample}B presents the example of sketching. The third way to divide the space is by generating the random hyperplane, which divides the space into two parts. To create the hash code, multiple such hyperplanes are generated, and the bit assignment of all these planes is concatenated. This idea is termed binary hash~\cite{manu::lsh}. Figure~\ref{fig:ch1:hashexample}C presents the example of a binary hash.

The fundamental use case of hashing is in the nearest neighbor search. Since hash codes generated by \acrshort{lsh} family algorithms are the same for nearby samples, hash code sharing enables a faster search of the nearest neighbors. However, since many of the possible hash codes may not get filled, query samples getting those codes will have no neighbors. To handle the blank hash code or missing bin problem, there are two approaches that are generally employed.

\begin{enumerate}
    \item Create multiple hash tables, search all of them and return the unique list of neighbors. This approach does not guarantee nearest neighbors and may require many hash tables to be created for convergence.
    \item Look for hash codes that have slight difference with the hash code of the query sample. A large enough tolerance in the hash code difference guarantees neighbors in this approach. This is the preferred approach in the nearest neighbor search.
\end{enumerate}

Space-partitioning-based hash algorithms that generate bit codes require the nearest neighbor search to be performed in hamming space. There are many algorithms that have been developed to perform these searches. To name a few, \acrfull{mih}~\cite{manu::mih} is the state-of-the-art to perform the exact nearest neighbor search in the hamming space with pre-computed bit codes. \acrshort{mih} divides the bit codes into smaller contiguous chunks and then creates a hash table from these chunks, and then performs a linear scan on them to retrieve the nearest neighbors. \acrfull{ewh}~\cite{manu::ewh} randomly selects the bits from the bit codes and then creates multiple small bit codes to create new hash tables. Results from any hash table is weighted to create the final nearest neighbors list. Both of these techniques or reducing the size of bit code to perform nearest neighbor search reduces the probability of a query sample falling into the empty bin. 

LSHForest~\cite{lshforest} does a prefix search on a \acrfull{bst} to perform the nearest neighbors search. Prefix search avoids looking into the empty bins. Thus, speeding up the search time. However, prefix-based searches do not consider the diversity of samples represented by the bits at the end of the bit codes. Thus, we introduce a geometrically motivated tree compression algorithm that allows us to compare only informative bits of the complete bit code to extract the nearest neighbors. The proposed algorithm searches the nearest neighbors by comparing the most informative subsequence of bits. Subsequence comparison also does not lose out on any information captured by the end bits in the bit code.

\subsection{Local space-partitioning-based algorithms - A case for classification}

The hashing algorithms discussed earlier perform global partitioning. There has also been interest in local space partitioning-based algorithms for both nearest neighbor search and building classification \& regression trees. In local partitioning algorithms, samples falling within a region are dichotomized by a new plane. These planes do not affect any other region of space. Interestingly, the \acrfull{et} and kd-tree are almost the same except for the way they interpret the regions for classification \& regression and nearest neighbor searches, respectively. \acrlong{ot}s (\acrshort{ot}s) and \acrlong{rp} Trees (\acrshort{rp}trees) also has same relation as \acrshort{et} and kd-trees.

\begin{figure*}[!ht]
	\makebox[1\textwidth][c]{
		\resizebox{1\linewidth}{!}{
			\includegraphics{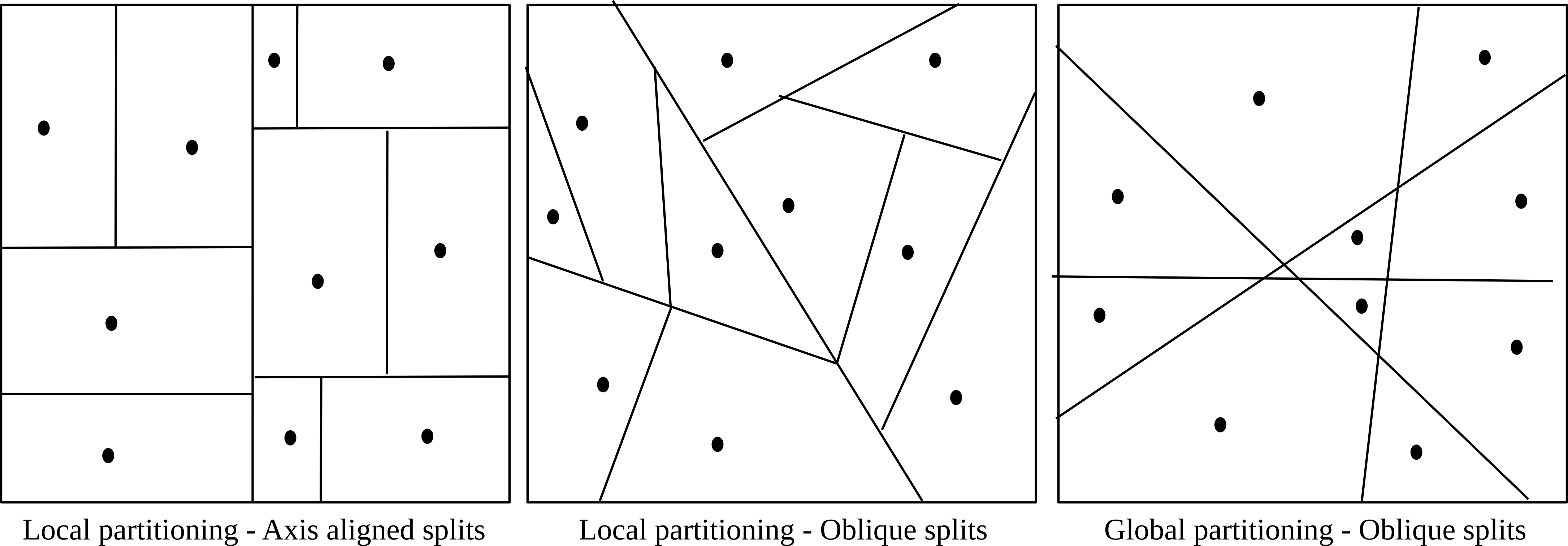}
		}
	}
	\caption{\textbf{Space-partitioning strategy example.} Local partitioning axis aligned splits - This strategy followed by \acrfull{rf}, \acrfull{et}, kd-tree etc. Local partitioning oblique splits - This strategy followed by \acrfull{ot}, \acrshort{rp}trees etc. Global partitioning oblique splits - This strategy is followed by hashing techniques and \acrfull{graf}, the classification algorithm proposed in this thesis. There is another strategy to have global partitioning axis aligned splits - This strategy is followed by sketching. }
	\label{fig:ch1:local}
\end{figure*}

All the tree-based algorithms are essentially space partitioning algorithms. Many hash functions from \acrshort{lsh} family work by dividing the spaces as well~\cite{manu::lsh,charikar2002similarity}. In a tree-based algorithm, each node represents an area in a space. These areas get divided incrementally by the continuous addition of the nodes in the tree. Most of the tree-based algorithms~\cite{balltree,kdtree,bbdtree,lshforest,cart,rf,et} divides area locally along an axis based on some criteria, known as axis-aligned splits. There are also algorithms where splits are oblique~\cite{rptrees}. Although oblique splits have a higher memory footprint than axis-aligned splits; they have found their application in various tasks where they perform better than axis-aligned~\cite{correia2016oblique,liu2016robust,do2015classifying,qiu2017oblique}. Each split, axis-aligned or oblique, can be considered a hyperplane in the space that dichotomizes it. One possible way to reduce the size of the tree is to generate global partitioning (Figure~\ref{fig:ch1:local}). This strategy effectively amounts to sharing one hyperplane in multiple nodes in different branches of the tree. We explore this idea in this thesis.

\section{Genomics of cancer}

\subsection{\textit{Central dogma} of molecular biology}

The human genome is a long sequence of 4 nucleotides namely adenine (A), cytosine (C), thymine (T), and guanine (G). It is divided into 24 chromosomes, 1 to 22, X and Y. Every human being has 23 pairs of chromosomes in every cell except germline cells. The germline cells are haploids, that is they contain a single set of all these 23 chromosomes. Among these 23 pairs, chromosomes 1 to 22 are autosomal while sex chromosomes are either XX or XY, excluding any genetic disorder, in females and males respectively. Every chromosome harbor multiple genes. Genes are the functional unit of a human genome whose activity regulates all cell behavior. 

In humans, every gene has interleaved exons and introns. The exon part of genes is involved in protein production while introns are spliced out. The production of proteins from genes can be understood in two steps:

\begin{itemize}
    \item \textbf{Transcription}: In this step, double helical structure of \acrshort{dna} is broken into two. One of the strands is then used for transcription. Here all the introns are removed and only exons are transcribed. All the transcribed fragments are concatenated to generate the \acrshort{mrna}.
    
    \item \textbf{Translation}: Resulting \acrshort{mrna}, generated in transcription step, then translated into protein. In this step, three contiguous nucleotides (codons) are converted into an amino acid, sequentially. Thus a sequence of amino acids is generated, this sequence is called protein.
\end{itemize}

This process is described as the \textit{central dogma} of biology.

\subsection{Variations in \acrshort{dna}: Cause of cancer}

\begin{figure*}[!ht]
	\makebox[1\textwidth][c]{
		\resizebox{\linewidth}{!}{
			\includegraphics{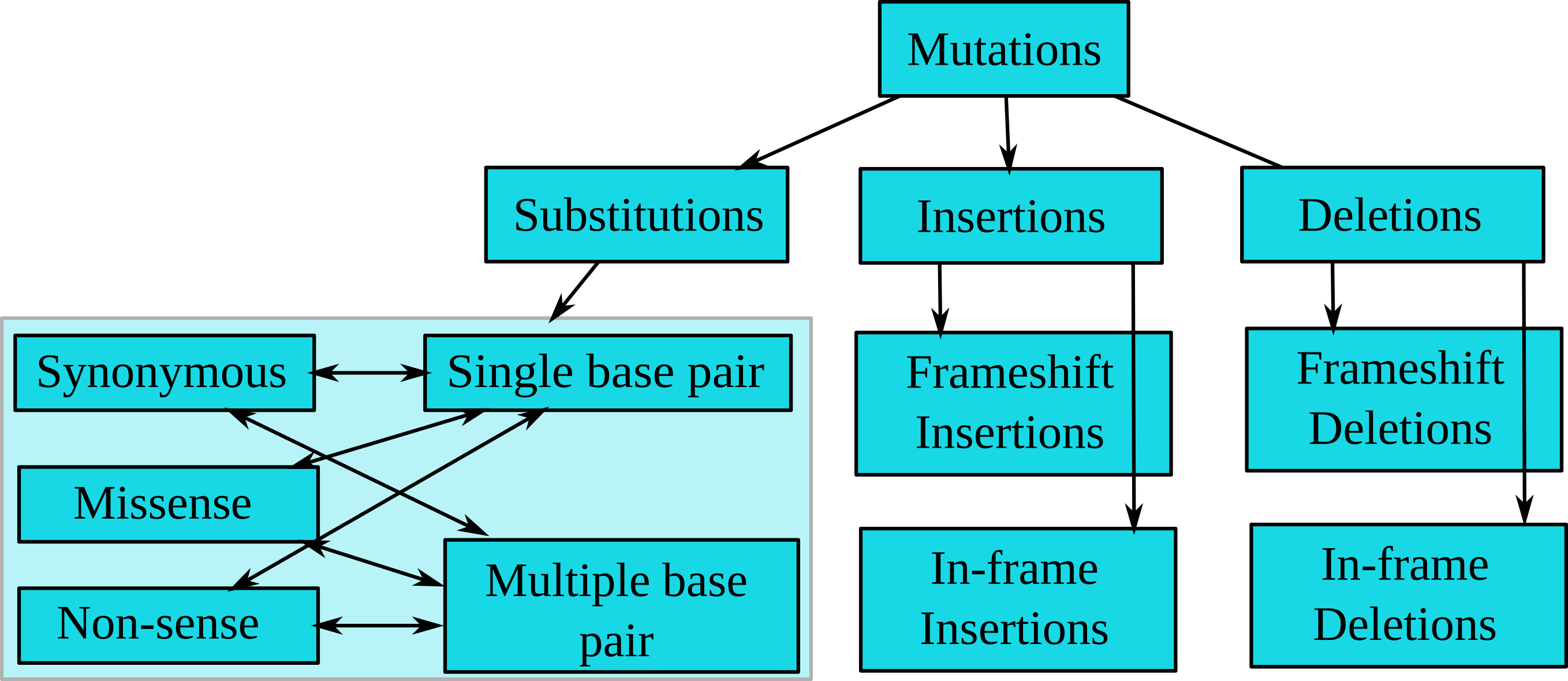}
		}
	}
	\caption{\textbf{Type of \acrshort{dna} mutations based on amino acid changes}.}
	\label{fig:ch1:mutationtype}
\end{figure*}

At the genetic level, cancer is caused by mistakes in the human genome. These mistakes get incorporated into the cell duplication process that may cause uncontrolled proliferation of cells which forms a tumor. These mistakes are called variations or mutations, or polymorphism. Based on nucleotide changes, mutations are of three types, substitutions, insertions, and deletions. Based on amino acid changes, substitutions are further divided into missense, synonymous, and non-sense mutations. Synonymous mutations do not cause any change in the resulting amino acid sequence, while missense mutations alter the amino acid. If mutations causes premature termination of translation process, it is called non-sense mutations (Figure~\ref{fig:ch1:mutationtype}). 

Substitutions can be a single base pair or multiple base pair. Single and multiple base pair substitutions are defined on the contiguous nucleotides. Insertions and deletions can be either in-frame or can cause frameshift. If the insertion and deletion are such that they do not alter the translation sequence except for the introduction or deletion of a few amino acids, they are called in-frames. Insertions and deletions that may alter the reading frame, thus creating an entirely new amino acid sequence, are called frameshift.

Based on locations, mutations can be divided into an exonic mutation that occurs in the exonic region of a gene in human \acrshort{dna}, intronic mutations that occur in the intronic regions of a gene, non-coding mutations that occur outside of genes, untranslated regions mutations that are part of gene and \acrshort{mrna} but does not code for proteins, and splice site mutations. The splice site mutations can be further divided into categories such as cryptic site activation, intron inclusion, exon skipping, etc. Apart from these, there are other kinds of mutations, such as duplication, inversion, translocation, and other complex mutations.

There are three sources from which an individual can acquire mutations in his lifetime:
\begin{enumerate}
    \item \textbf{Germline mutations} inherited from parents.
    \item \textbf{Somatic mutations} accumulate over a lifetime because of duplication errors. Somatic mutations can be further divided into two categories:
    \begin{enumerate}
        \item \textbf{Driver mutations} that affect cell fitness.
        \item \textbf{Passenger mutations} do not alter cell fitness but occurs in the cell that coincidentally or subsequently acquires a driver mutation.
    \end{enumerate}
    \item \textbf{External factors} such as exposure to environmental mutagens and radiation.
\end{enumerate}

\noindent
There are many uncharacterized somatic mutations in cancer whose roles are unclear.

\subsection{Mutational landscape of cancer}

\begin{figure*}[!ht]
	\makebox[1\textwidth][c]{
		\resizebox{\linewidth}{!}{
			\includegraphics{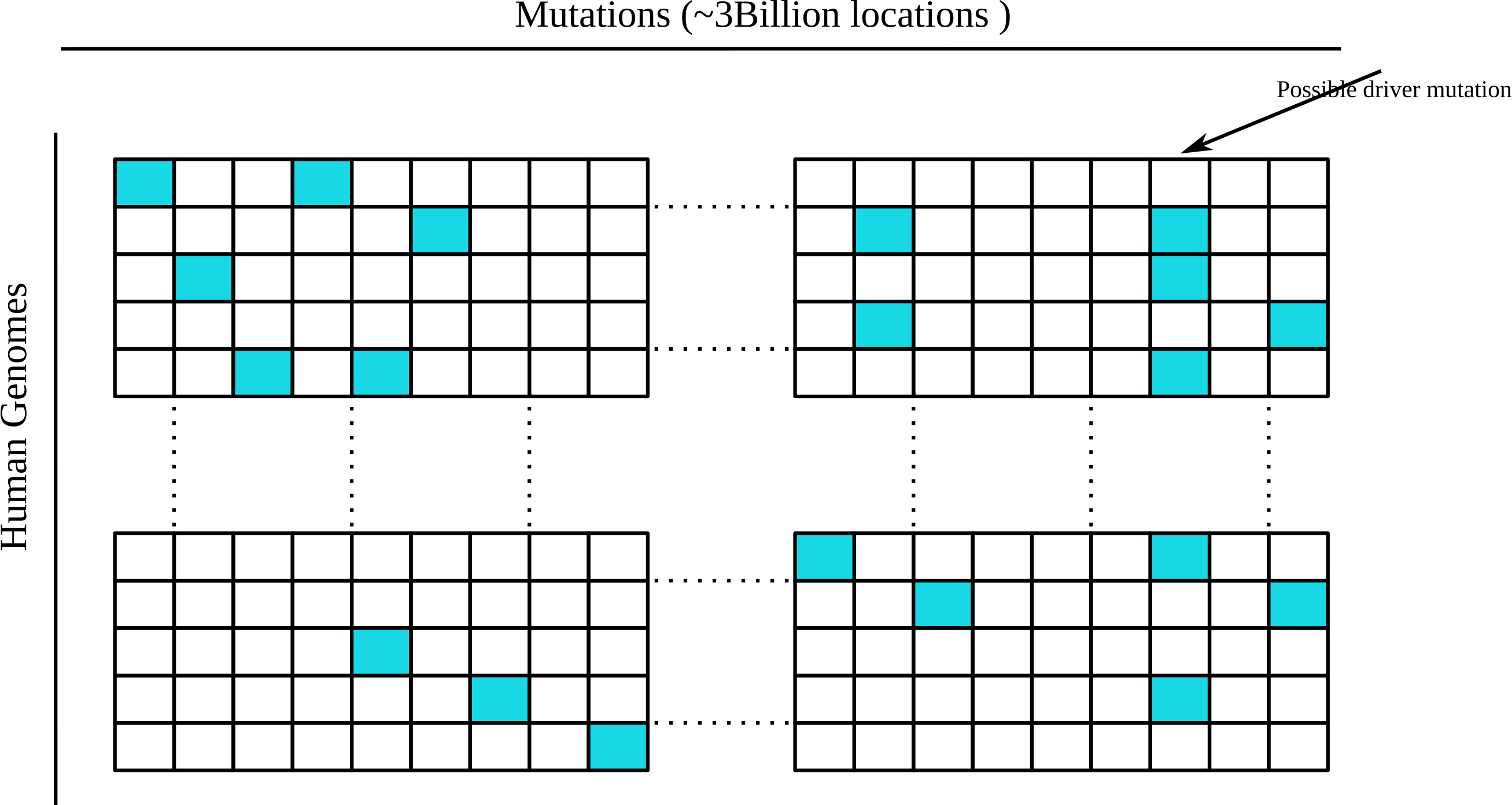}
		}
	}
	\caption{\textbf{Mutational landscape of cancer is sparse}. There are $\sim$3 billion base pairs in the human genome. Every one of them can mutate, causing a very large event space. However, the number of mutations that get mutated in an individual's lifetime is relatively small. Also between two individuals, there is much less overlap on the acquired mutations, thus making the mutation frequency in population very low.}
	\label{fig:ch1:mutationallandscape}
\end{figure*}

A human genome is $\sim$3billion base pair long. It is possible to see mutations at any location (Figure~\ref{fig:ch1:mutationallandscape}). Further combining substitution, insertion, and deletions makes the mutational space prohibitively large. Thus characterizing every single mutation is almost impossible and the mutations that have been seen frequently are characterized only. These characterized mutations become the target for therapeutic interventions in cancer. However, it is not guaranteed that these mutations will be found in every cancer patient, raising difficulty in cancer treatment. Thus, an in-silico characterization of the mutations is required to understand their role in cancer and provide possible treatment targets.

The first step toward the in-silico characterization of mutations is to find the deleteriousness or pathogenicity of a mutation~\cite{adzhubei2010method,vaser2016sift,frazer2021disease}. Depending on its position, a change in \acrshort{dna} may or may not be pathogenic. As we know, \acrshort{dna} is represented as a long sequence of four nucleotides A, C, T, and G. Thus, an analogy to put \acrshort{dna} in parallel with the language construct can be drawn. These four nucleotides become the alphabets of the language model. A small, well-defined sequence of nucleotides becomes the word. A more extended sequence of nucleotides with proper punctuation becomes the sentence. One way to put genome and language in parallel would be to assume codons as words and \acrshort{mrna} as sentences. Now, this analogy can be used to learn the mutational landscape of the genome.

Drawing analogy from \acrfull{nlp}, words representing the central theme of the documents are relatively rare. We can use word2vec~\cite{skipgram} and related techniques~\cite{devlin2018bert} to learn the numerical representation of such words. These techniques are known to put semantically similar words together. Thus we hypothesize that the mutational landscape, which is highly sparse (Figure~\ref{fig:ch1:mutationallandscape}), can be learned by making use of these techniques such that the learned embedding may preserve some similarity in the embedding space. Then these learned embeddings can be utilized in the downstream tasks. We explore this idea in the later part of the thesis. Rather than using codon as a word and \acrshort{mrna} as sentences, we define a new symbolic representation of mutations using codons. These representations are used as words, and a sequence of these symbolic representations is treated as sentences.

\section{Organization of the thesis}

\begin{figure*}[!ht]
	\makebox[1\textwidth][c]{
		\resizebox{1.1\linewidth}{!}{
			\includegraphics{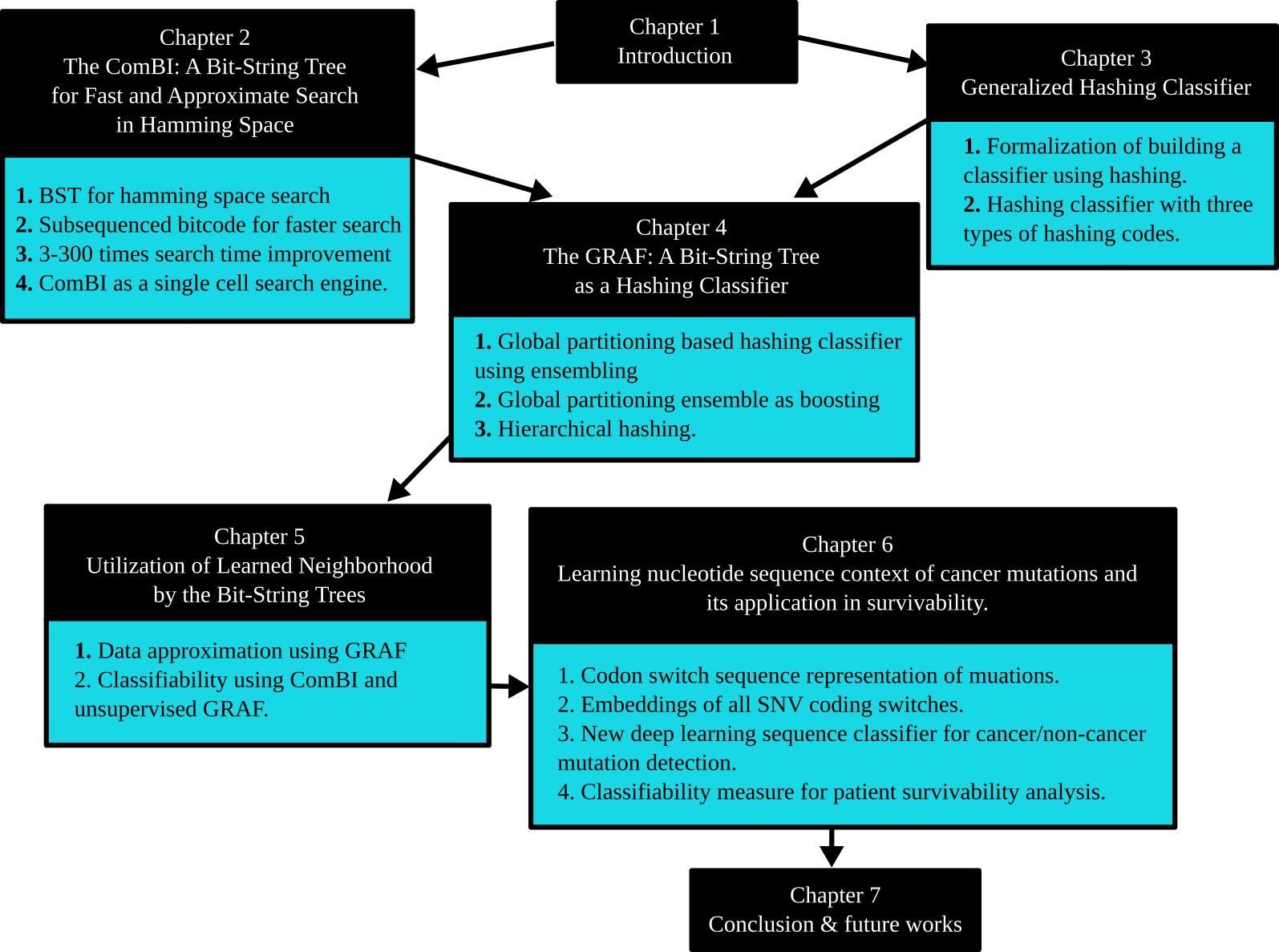}
		}
	}
	\caption{\textbf{Organization of thesis}. The chapter name, chapter flow, and headlines of every chapter in the thesis. }
	\label{fig:ch1:chapterflow}
\end{figure*}

In this thesis, we focus on learning using the bit-string trees with the broad objective of increasing the applicability of learned neighborhoods in the different tasks of machine learning. Figure~\ref{fig:ch1:chapterflow} presents the mind-map of the thesis chapter organization and their dependency. We begin with the introduction (this chapter), where we present the motivation and scope of the work. Then in Chapter 2, we discuss the applicability of \acrshort{bst} in hamming space search. Then to overcome these drawbacks of \acrshort{bst} a new geometrically motivated algorithm for fast approximate nearest neighbor search is invented. This algorithm is named as \acrfull{combi}. Extensive empirical evaluation of \acrshort{combi} is also presented. Knowledge of the neighborhood of a sample allows us to estimate the class of an unknown sample by examining its vicinity. This motivates the use of ideas behind \acrshort{combi} to solve classification tasks.  Chapter 3 presents a formal approach for building the classifiers from hashing techniques. These classifiers are termed hashing classifiers. Chapter 3 also compares the thee example of hashing classifiers and it also motivate the need to build a tree-based hashing classifier for improved performance.

It has been shown that an ensemble of trees works best as a general-purpose classifier~\cite{fernandez2014we}. Chapter 4 extends the ideas of Chapter 2 and Chapter 3 to build a tree-based ensemble classifier which can be understood as a global-partitioning-based \acrfull{rf} or a hierarchical hashing classifier. The global partitioning leads to some degree of parsimony and makes search more efficient. This algorithm is named \acrfull{graf}. An alternate representation of \acrshort{graf} can also be understood as an ensemble boosting algorithm. In the absence of label information \acrshort{graf} can also behave as a hierarchical hashing algorithm called \acrfull{ugraf}. Extensive empirical evaluation of \acrshort{graf} is also presented. We also proposed a variant of \acrshort{graf} on \acrshort{gpu}.

Chapter 5 employs the ideas of Chapter 2 and Chapter 4 to build the additional use cases of \acrshort{combi}, \acrshort{graf}, and \acrshort{ugraf}. First, we present how \acrshort{graf} can be used for importance sampling. The score generated for importance sampling is termed sensitivity. Further, we show how the duo of \acrshort{ugraf} and \acrshort{combi} can be used to quantify the neighborhood of a sample. This quantification is termed per-sample classifiability.

Next, the focus shifts to a very important and interesting application of per-sample classifiability - that to understand the relative carcinogenicity of a cancer mutation. Per-sample classifiabiity, motivated by the work in Chapter 5, is a novel thought in this domain. Chapter 6 presents the application of per-sample classifiability measures to estimate patient survivability. Although the pipeline used in Chapter 6 is somewhat different from the approaches followed thus far in the thesis, the fundamental driving thought behind the Chapter 6 is to explore the notion of per-sample classificability. To achieve this, we first developed a novel approach to represent the mutations in the human genome. This approach is termed as \acrfull{crcs}. Then for downstream analysis \acrshort{crcs} is used to train a sequence classifier. The classifier is named \acrfull{blac}. The output of \acrshort{blac} is combined with \acrshort{ugraf} and \acrshort{combi} to estimate the classifiability of the sequences. Then the per-sample classifiability measure is used to assess the survivability of patients. Chapter 7 presents the thesis's conclusion and the ideas to extend the work discussed in this thesis.

\section{Conclusion}

This thesis focuses on learning from bit-string trees and bridging the gap between hashing and classification literature. To achieve this a new nearest neighbor search technique was developed and the structural representation of the learned technique is used to build a classifier. The built classifier and structural representation of the search technique were then used for importance sampling and neighborhood quantification, respectively. The neighborhood quantification approach also finds applicability in cancer survival estimation. The performance of all the proposed methods is estimated on multiple datasets and empirical convergence of the method is also presented.

%% file: ch2combi.tex
\section{Introduction}
\label{sec:ch2:introduction}
Finding the nearest neighbors in a large database has for long challenged the big-data community. The concept of finding nearest neighbors has widespread applications in search engines, computer vision, security systems etc~\cite{oostveen2002feature,esmaeili2010robust,esmaeili2011fast,miller2005audio,cellfishing,cellatlassearch,plageras2018efficient,stergiou2018security,yu2018four}. Many methods have been proposed in the past for fast retrieval of good quality nearest neighbors. Amongst these methods, hashing and related methods have been very popular. A few of these methods focus on increasing the hashing quality~\cite{manu::ksh,manu::lsh,charikar2002similarity,semantic,manu::sh,manu::sperical,manu::ssh,manu::shl,wang2020visual}, while others focus on improving the search quality and search time~\cite{manu::mih,manu::bridge,manu::ewh}. A survey by Wang \textit{et al.}~\cite{wang2014hashing} discusses different hashing, search, modeling, and analysis techniques in this regard. Li \textit{et al.}~\cite{manu::review} presents a detailed comparison of some selected methods.

Space-partitioning-based hashing algorithms generate a binary signature (bit code) of samples in a database. Hence, searching for the nearest neighbors is performed in a hamming space. There have been several attempts to devise fast and efficient techniques to perform search in a hamming space~\cite{lsh,manu::mih,manu::ewh,manu::hbst,manu::bridge,lshforest,coveringlsh,fastcoveringlsh,onlinesearch1,onlinesearch2,gog2016fast}. Amongst them, \acrfull{mih}~\cite{manu::mih} has remained the method of choice for fast and exact search to find the nearest neighbors in a hamming space. 

\acrfull{bst}s are inherently capable of storing bit codes and performing nearest neighbor search. The search in \acrshort{bst}s are exact nearest neighbor searches. However, prohibitive memory requirements for storing \acrshort{bst}s, and the tree traversal cost for large bit codes, make them unsuitable for practical purposes. 

Some algorithms have utilized \acrshort{bst} to store bit codes and perform searches in a hamming space. LSH-Forest~\cite{lshforest} generates prefix trees. A tree is grown on a database by dividing a leaf node into two if the number of samples on the node is more than a predefined number. A combination of multiple such trees generates a forest.  \acrfull{hbst}~\cite{manu::hbst} is another method that employs \acrshort{bst} to search in a hamming space. \acrshort{hbst} finds a bit that creates a maximally balanced split of bit codes on every node. This splitting procedure continues until a specified termination criterion is met. Hyv{\"o}nen \textit{at el.}~\cite{hyvonen2016fast} proposed a variant of \acrfull{rp} trees~\cite{dasgupta2013randomized}, \acrfull{mrpt}. \acrshort{mrpt} arranges random projections hierarchically. The median value of the projection decides the split on every node. After that, a search is performed on the tree to retrieve the nearest neighbors. Eghbali \textit{et al.}~\cite{onlinesearch1,onlinesearch2} proposed \acrfull{hwt}, an online nearest neighbor search algorithm. \acrshort{hwt} is a general array tree. Each node in \acrshort{hwt} is assigned a weight vector of size $n$ based on the number of 1's in all $n$ partitions of bit codes. Thereafter, the nearest neighbor is found by iterating over the tree.

On the one hand, LSH-Forest generates a prefix tree. On the other hand, \acrshort{combi} builds a tree by selecting informative bits from the predefined bit codes. The tree growth in LSH-Forest is local. LSH-Forest and \acrshort{combi} may generate similar trees for local space partitioning-based hashing algorithms. However, \acrshort{combi} can also be used with space partitioning algorithms that utilize the global structure of the data~\cite{manu::sperical,semantic,manu::sh}. \acrshort{mrpt} can use the projections generated from different algorithms to build the tree. However, child nodes are spawned by the custom-defined criteria, the median in the original manuscript. \acrshort{combi} does not put any such restriction and can also work with algorithms that output bit code directly. Scanning a bit to create a balanced \acrshort{hbst} split may be time-consuming for big databases as it will require multiple passes. Further, an online update in \acrshort{hbst} will need a linear scan over all the leaf samples to find the best split. Whereas in \acrshort{combi}, online updates are significantly fast (requires updates of only three pointers, Section~\ref{sec:ch2:online}). \acrshort{hbst} does not enforce a requirement of points on the leaf nodes to share the bit codes. Hence, samples belonging to different bit codes may come into the same leaf. Therefore, information related to space geometry is lost in such an arrangement of bits.

\acrfull{mih}~\cite{manu::mih} also works with any algorithm that generates bit codes. \acrshort{mih} reduces the search space by dividing the bit codes into multiple chunks of the same size and then performing a search on every chunk separately. These search results are combined, and the final nearest neighbors are selected. Although \acrshort{mih}'s search strategy benefits from the hardware support (machine instructions and caching), it still scans a significant amount of the database before terminating (Section~\ref{sec:ch2:discussion}).

A binary tree constructed on bit codes can be treated as a generic search algorithm. Such trees can preserve space geometry as well. Given a query sample, its nearest neighbors can be extracted by a controlled tree traversal that can limit the scanned database size. However, a naive tree will be marred by extremely high memory requirements and search time. Section~\ref{sec:ch2:proposedmethod} discusses these drawbacks and presents a method that alleviates these issues while preserving all the desired properties of the binary tree arrangement of the bit codes.

In the context of computational biology, nearest neighbor search-based approaches have found their application in many tasks such as outlier detection~\cite{fire}, clustering~\cite{sinha2018dropclust,sinha2019dropclust2} etc. One such application is building single cell search engines~\cite{cellfishing,sato2019cellfishing}. The advent of \acrfull{scrna} has enabled the profiling of individual cells. The \acrshort{scrna} technology can perform massively parallel sequencing of cells, thus increasing the throughput. The further advancements in the technology allowed multiplexing in library preparation, thus yielding higher throughput.  

The ability to sequence individual cells demands the identity of the sequenced cell for complete characterization. A cell type may be identified by its phenotype, function, lineage, and states ~\cite{morris2019evolving}. However, the number of cells that can be sequenced by \acrshort{scrna} is prohibitively high. Further, the cell state dynamics also add to the difficulty~\cite{trapnell2015defining,mulas2021cell}. Thus, software interventions are needed to enable faster annotation. Because of cell state dynamics, a cell can have varying expression profiles. However, they all may lie in the same neighborhood. Thus nearest neighbor search in the previously annotated \acrshort{scrna} databases may be helpful in the annotation of the cell types. Cellfishing.jl~\cite{sato2019cellfishing} is one such method that facilitates the cell type annotation of cells. Of, note for nearest neighborhood search Cellfishing.jl uses \acrshort{mih}~\cite{manu::mih}. 

\section{The \acrfull{combi}}\label{sec:ch2:proposedmethod}

Let us assume a database $X$ has $N$ samples in $D$ dimensions. Further, let us assume a partition-based hashing function $f:X \mapsto \{0, 1\}^{b}$, where $b$ is the total number of bits. Assume, $f$ divides the space into small regions by introducing multiple hyperplanes using some strategy~\cite{semantic,manu::lsh,manu::ksh,manu::sh,manu::sperical}. Each hyperplane assigns a bit to the sample based on the side of the hyperplane it falls on. The concatenation of bits allocated by all hyperplanes to a sample creates its bit code. The table containing bit codes of all samples is referred to as the hash table. The table that maps a bit code to samples sharing that bit code is referred to as the inverted-hash-table. 

\subsection{Motivation}
A hash table can be arranged using a \acrshort{bst}. Each branch of a \acrshort{bst} represents a complete bit code, and the leaf on that branch contains samples sharing that bit code. Effectively, a \acrshort{bst} represents an inverted-hash-table (referred to as an \acrshort{ibst} hereafter). To construct an \acrshort{ibst}, start with a root node. If the current bit is 0 (1 respectively), go towards the left child (right child respectively) or create a new node if the left child (right child respectively) is missing. Algorithm~\ref{alg:ch2:IBSTConstruct} describes all steps to construct an \acrshort{ibst}.

\begin{algorithm}[!htb]
	\caption{Construction of \acrfull{ibst}}
	\label{alg:ch2:IBSTConstruct}
	\textbf{Input}: x: Sample x$\in $X, B: bit code of x \\
	\textbf{Parameter}: CuN: current node in \acrshort{ibst} \\
	\textbf{Output}: Constructed \acrshort{ibst}
	\begin{algorithmic}[1] 
		\STATE Assign root node of the tree to CuN
		\FOR{i $\in$ \{1,..,b\}}
		\IF{B[i] == 0}
		\IF{leftChild(CuN) is absent}
		\STATE Create new node and assign to leftChild(CuN)
		\STATE Set bitNumber(CuN) as i
		\ENDIF
		\STATE CuN = leftChild(CuN)
		\ELSE
		\IF{rightChild(CuN) is absent}
		\STATE Create new node and assign to rightChild(CuN)
		\STATE Set bitNumber(CuN) as i
		\ENDIF
		\STATE CuN = rightChild(CuN)
		\ENDIF
		\ENDFOR
		\STATE isLeaf(CuN) = \textbf{True}
		\STATE Store x at leaf
	\end{algorithmic}
\end{algorithm}

Recursive traversal can be employed to retrieve the \acrfull{nns}. Assuming $\hat{b}$ represents the bit at the current node, an additional $1$-\acrfull{hd} tolerance is considered if it proceeds in $\neg\hat{b}$ direction. A maximum \acrshort{hd} (\textit{MaH}) can be defined before the search begins to limit the search radius. The search terminates when tolerance goes beyond \textit{MaH}. If the search terminates before reaching a leaf, it will not return any neighbors. This situation will be termed as a \textit{miss}, otherwise, it is a \textit{hit}. 

For a query, it is possible not to reach any leaf or retrieve the desired number of nearest neighbors within defined \textit{MaH}. This results in another pass over the tree with an increased \textit{MaH}. A minimum \acrshort{hd} (\textit{MiH}) can be defined to avoid retrieving the same neighbors. For the first pass, $\textit{MiH}=0$ and for further passes \textit{MiH} is equal to the previous \textit{MaH} incremented by 1. The search performed by this approach is exact and returns true nearest neighbors in hamming space. For a small \acrshort{hd} range, this search can be faster compared to the linear scan of the inverted-hash-table since a tree traversal scans very few bit codes. Algorithm~\ref{alg:ch2:IBSTSearch} gives all search steps in an \acrshort{ibst}.

\begin{algorithm}[!htb]
	\caption{Search in \acrfull{ibst}}
	\label{alg:ch2:IBSTSearch}
	\textbf{Input}: \acrshort{ibst}: inverted-hash-table binary search tree, Q: query bit code\\
	\textbf{Parameter}: CuN: current node in IBST, MiH: min hamming distance, MaH: max hamming distance, CuH: current hamming distance\\
	\textbf{Output}: RS: nearest-neighbors
	\begin{algorithmic}[1] 
		\STATE HD = 0
		\IF {CuH $\geq$ MiH \textbf{and} isLeaf(CuN)}
		\STATE Append leaf content to RS
		\STATE \textbf{return}
		\ENDIF
		\IF{leftChild(CuN) is present}
		\IF{Q[bitNumber(CuN)] $\neq$ 0}
		\STATE HD = 1
		\ENDIF
		\IF{CuH+HD $\leq$ MaH}
		\STATE Call search IBST with CuN=leftChild(CuN) and CuH=CuH+HD
		\ENDIF
		\ENDIF
		\STATE HD = 0
		\IF{rightChild(CuN) is present}
		\IF{Q[bitNumber(CuN)] $\neq$ 1}
		\STATE HD = 1
		\ENDIF
		\IF{CuH+HD $\leq$ MaH}
		\STATE Call search IBST with CuN=rightChild(CuN) and CuH=CuH+HD
		\ENDIF
		\ENDIF
	\end{algorithmic}
\end{algorithm}

There are three major drawbacks in an \acrshort{ibst} approach:
\begin{itemize}
	\item The total memory required to store an \acrshort{ibst} is very high. It increases exponentially with bit code size and database size.
	\item The \textit{miss} rate increases with an increasing length of bit codes. This makes the search process via tree traversal slow for all practical purposes.
	\item The search radius denoted by \textit{MaH} may vary with query samples. Hence, the multiple passes over the tree result in the traversal of the same branch numerous times, making the search slow.
\end{itemize}

\subsection{Details of \acrshort{combi}}
 To build the intuition behind \acrshort{combi}, let us first see an example that can be generalized to any arbitrary size database. Assume a $2$-dimensional database with $N$ samples. Further, assume a hash function $h(.)$, which generates a bit code of the length of $7$. This hash function divides the space into $13$ filled regions (Figure~\ref{fig:ch2:regions}A). Each filled region is denoted by a number prefixed with $S$. Figure~\ref{fig:ch2:trees}A shows an \acrshort{ibst} constructed with these bit codes via Algorithm~\ref{alg:ch2:IBSTConstruct}. Figure~\ref{fig:ch2:regions}B shows the inclusion of empty regions into filled regions using $1$-\acrshort{nn} search (Algorithm~\ref{alg:ch2:IBSTSearch}) on the constructed \acrshort{ibst}. 

The scanning of an \acrshort{ibst} to find the nearest neighbors may involve a high \textit{miss} rate. The proposed compression method focuses on reducing the \textit{miss} rate to increase the search speed. The reduction in memory usage and reduced dependency on \textit{MaH} are by-products of the proposed method.

\subsubsection{Construction of \acrshort{combi}}
\begin{figure}[!ht]
	\makebox[1\textwidth][c]{
		\resizebox{1 \linewidth}{!}{ %
			\includegraphics[width=\linewidth]{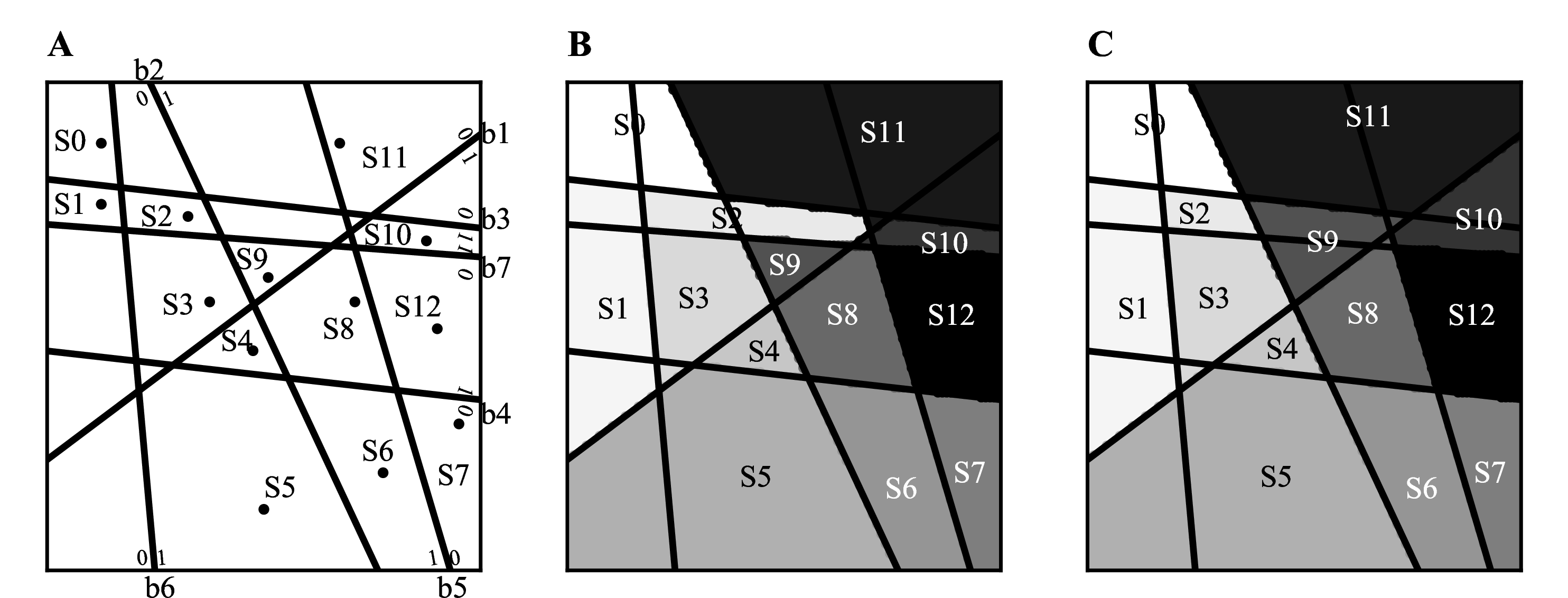}
		}
	}
	\caption{\textbf{Visualization of the resulting view after merger of regions.} \textbf{A)} Geometrical representation of space partitioning via \acrshort{lsh}. \textbf{B,C)} $1$-\acrshort{nn} approximation of space via \acrshort{ibst}, and \acrshort{combi}, respectively.}
	\label{fig:ch2:regions}
\end{figure}

For simplicity, assume a filled region with some empty regions at $1$-\acrshort{hd}. This filled region is most likely the nearest neighbor for any query sample with a bit code associated with any of these $1$-\acrshort{hd} away empty regions. \acrshort{combi} seeks to merge one of such empty regions with the filled region. This merger produces a mixed region. Further iterations will assign every empty region to some mixed region. Figure~\ref{fig:ch2:regions}C shows the resultant space after merging.  Since no empty regions are left in the space, there will be no more \textit{miss}es in the search. 

\begin{figure}[!ht]
	\makebox[1\textwidth][c]{
		\resizebox{1 \linewidth}{!}{ %
			\includegraphics[width=\linewidth]{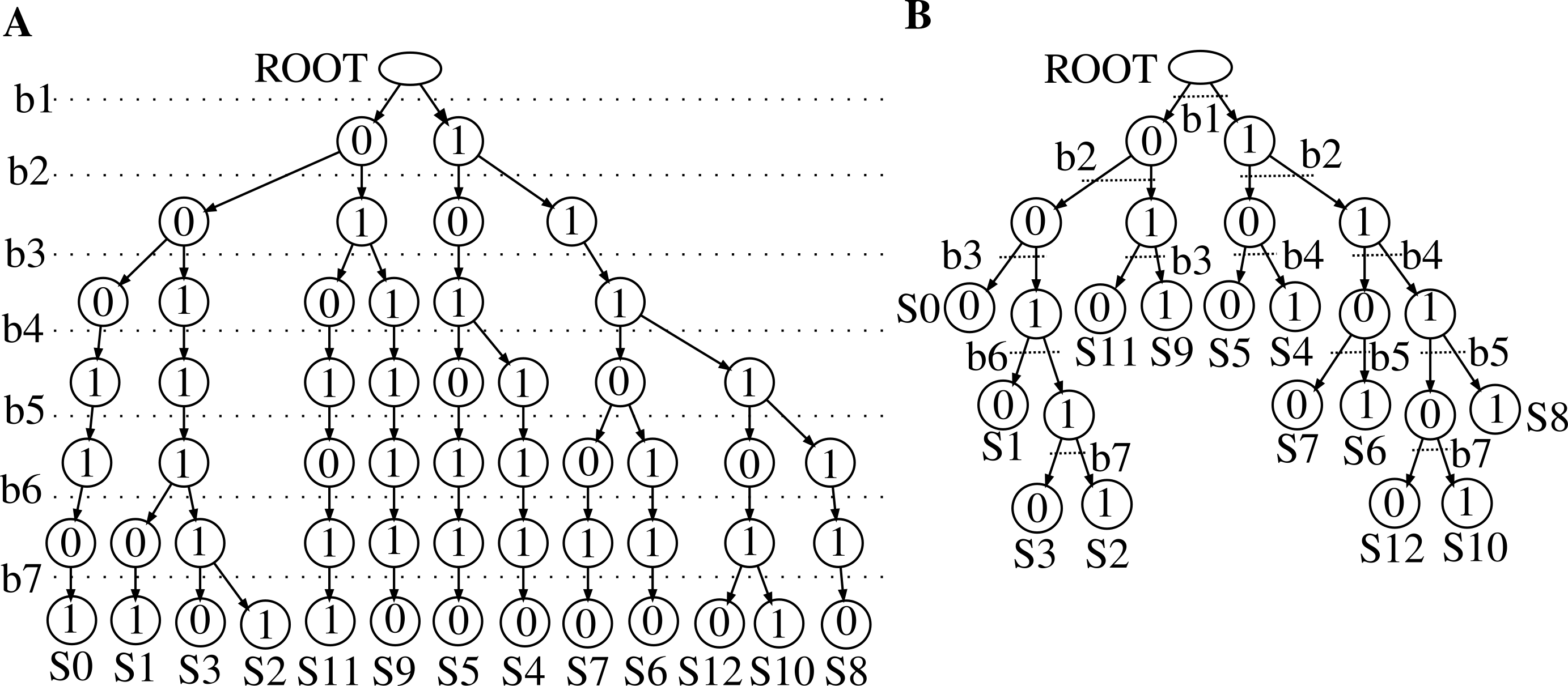}
		}
	}
	\caption{\textbf{Tree representations} \textbf{A)} \acrshort{ibst}. \textbf{B)} \acrshort{combi}.}
	\label{fig:ch2:trees}
\end{figure}

The merging of $1$-\acrshort{hd} empty regions into a filled region is equivalent to setting the corresponding bit to \textit{don't care} for these two regions. These \textit{don't care} bits can be removed from their bit code, reducing the bit code length. The number of bits marked as \textit{don't care} in a bit code may vary depending on the neighborhood structure. The removal of bits marked as \textit{don't care} is equivalent to removing a node from the \acrshort{ibst}. The recursive removal of \textit{don't care} node (bits) from an \acrshort{ibst} is called compression. Figure~\ref{fig:ch2:trees}A shows the \acrshort{ibst} and Figure~\ref{fig:ch2:trees}B shows the corresponding \acrshort{combi}. 

Algorithm~\ref{alg:ch2:IBSTCompress} gives a detailed description of the steps for compressing an \acrshort{ibst}. This algorithm utilizes the fact that if a node in the tree has two child nodes, the bit corresponding to that node can not be marked as \textit{don't care}.  For example in Figures~\ref{fig:ch2:regions}, and~\ref{fig:ch2:trees}, b7 can not be \textit{don't care} for region S2+S3 since it divides them. On the other hand, all nodes having only one child are eventually marked as \textit{don't care} and removed. For instance, b7 is \textit{don't care} for region S9. 

As bit code length increases, the number of samples per bit code decreases. Eventually, there is only one sample per bit code. In this scenario, the count of filled regions is equal to the number of samples. A further increase in bit code length does not create any new filled region, and hence, the splitting of nodes saturates. With an increase in bit code length, the depth of \acrshort{ibst}, and the count of nodes with a single child continue to grow. Since \acrshort{combi} removes nodes with a single child, it stops growing when all regions have only one sample. Therefore, \acrshort{combi} has a lower memory footprint when compared to \acrshort{ibst}.  

\begin{algorithm}[!htb]
	\caption{Construction of \acrfull{combi} by compression of \acrshort{ibst}}
	\label{alg:ch2:IBSTCompress}
	\textbf{Input}: \acrshort{ibst}: \acrlong{ibst}\\
	\textbf{Parameter}: lca: left child absent, rca: right child absent, cl: compress left side, cr: compress right side, CuN: current node in \acrshort{ibst}\\
	\textbf{Output}: \acrshort{combi}: \acrlong{combi}
	\begin{algorithmic}[1] 
		\STATE lca = rca = \textbf{True}
		\STATE cl = cr = \textbf{False}
		\IF{leftChild(CuN) is present}
		\STATE lca = \textbf{False}
		\IF{isLeaf(CuN)}
		\STATE cl = \textbf{True}
		\ELSE
		\STATE cl = Call compress \acrshort{ibst} from leftChild(CuN)
		\ENDIF
		\ENDIF
		\IF{rightChild(CuN) is present}
		\STATE rca = \textbf{False}
		\IF{isLeaf(CuN)}
		\STATE cr = \textbf{True}
		\ELSE
		\STATE cr = Call compress \acrshort{ibst} from rightChild(CuN)
		\ENDIF
		\ENDIF
		\IF{cl \textbf{and} cr} 
		\STATE \textbf{return False}
		\ENDIF
		\IF{lca \textbf{and} rca}
		\STATE \textbf{return False}
		\ELSIF{rca}
		\STATE Replace CuN with leftChild(CuN)
		\STATE \textbf{return True}
		\ELSIF{lca}
		\STATE Replace CuN with rightChild(CuN)
		\STATE \textbf{return True}
		\ELSE 
		\STATE \textbf{return False}
		\ENDIF
	\end{algorithmic}
\end{algorithm}

\subsubsection{Search in \acrshort{combi}}
Similar to \acrshort{ibst}, the search on \acrshort{combi} can be performed by a recursive traversal of the tree. However, if the search takes $\neg\hat{b}$ direction, then it is termed as a \textit{mutation} of bit $\hat{b}$. Unlike \textit{MaH}, the number of \textit{mutation}s required for a query bit code to retrieve the nearest neighbors is easy to find. Algorithm~\ref{alg:ch2:IBSTCompressedSearch} gives a detailed description of the search in \acrshort{combi}. Since every access to the tree will return some neighbors (near or far away), keeping the value of maximum \textit{mutate}, \textit{MaC}, to moderate is enough to get satisfactory results. Minimum \textit{mutate}, \textit{MiC}, serves the same purpose as \textit{MiH}. Once the nearest neighbors have been retrieved, they can be ranked based on the desired metric, and the required count can be selected from them.

\begin{algorithm}[!htb]
	\caption{Search in \acrshort{combi}}
	\label{alg:ch2:IBSTCompressedSearch}
	\textbf{Input}: \acrshort{combi}: \acrlong{combi}, Q: query bit code\\
	\textbf{Parameter}: CuN: current node in ComBI, MiC: min mutate count, MaC: max mutate count, CuC: current mutate count\\
	\textbf{Output}: RS: nearest-neighbors
	\begin{algorithmic}[1] 
		\IF {CuC $\geq$ MiC \textbf{and} CuC $\leq$ MaC \textbf{and} isLeaf(CuN)}
		\STATE Append leaf content to RS
		\STATE \textbf{return}
		\ENDIF
		\IF{Q[bitNumber(CuN)] $\neq$ 0}
		\IF{CuC $<$ MaC}
		\STATE Call search \acrshort{combi} with CuN=rightChild(CuN)
		\STATE Call search \acrshort{combi} with CuN=leftChild(CuN) and CuC=CuC+1
		\ELSIF{CuC == MaC}
		\STATE Call search \acrshort{combi} with CuN=rightChild(CuN)
		\ENDIF
		\ELSE
		\IF{CuC $<$ MaC}
		\STATE Call search \acrshort{combi} with CuN=leftChild(CuN)
		\STATE Call search \acrshort{combi} with CuN=rightChild(CuN) and CuC=CuC+1
		\ELSIF{CuC == MaC}
		\STATE Call search \acrshort{combi} with CuN=leftChild(CuN)
		\ENDIF
		\ENDIF
	\end{algorithmic}
\end{algorithm}

\subsubsection{Search in \acrshort{combi} is approximate}\label{sec:ch2:drawback}
While \acrshort{ibst} is an exact search method for any value of $MaH$, \acrshort{combi} is an approximate search method for small values of $MaC$. It is due to the greedy merging of $1$-\acrshort{hd} regions. Assume two filled regions $P$ and $Q$, and \acrshort{hd} between these two regions is $r$. Assuming $P$ is surrounded by many $1$-\acrshort{hd} empty regions. It will choose only one of them to merge at a time. In the worst case, one or more of the $1$-\acrshort{hd} empty regions of $P$ may merge with $Q$. If a query falls into any of these empty regions, its first retrieved neighbor will be at $\{r-1, r, r+1\}$-\acrshort{hd} distance.

Neighbors of $P$ may be retrieved by setting \textit{MaH} to be sufficiently high. However, making \textit{MaH} too large will slow down the search process. For $MaH=\text{longest branch length} = O(\texttt{len}(\text{bit code}))$, it will reduce \acrshort{combi} to a linear scan. A linear scan in \acrshort{combi} without low-level support (\texttt{xor}, \texttt{\_\_builtin\_popcount}) will be slower than the linear scan of bit codes. Nevertheless, constructing multiple \acrshort{combi}s can alleviate this problem at a moderate value of \textit{MaH} (Section~\ref{sec:ch2:perftuning}). The process of \acrshort{combi} construction (Algorithm~\ref{alg:ch2:IBSTCompress}) is deterministic. Hence, to construct different \acrshort{combi}s, any combination of shuffling, rotation or reverse can be applied to bit codes.

Since all trees are built on the same bit code (but the different ordering of bits), their \acrshort{hd} relative to the query sample (with harmonized order) will be the same. Hence, all retrieved samples can be ranked based on their \acrshort{hd}, and the desired count of nearest neighbors can be extracted.

\subsubsection{Convergence of search in \acrshort{combi}}
The construction of multiple trees in \acrshort{combi} is motivated by the fact, that even if a certain $1$-\acrshort{hd} away empty region from $P$ merges with a different region $Q$ for a particular ordering of bit code, it would eventually merge with $P$ for a different order of bit code. In a scenario when all considered orderings of the bit code do not merge $1$-\acrshort{hd} away empty regions with $P$, they can be retrieved by increasing the value of $MaC$. At the end of the search, the ranking step will return $P$ as the nearest neighbor for the query sample falling into $1$-\acrshort{hd} away empty region.

During empirical evaluation, it was found that four trees are enough, with adequately selected $MaC$, to get satisfactory performance in a reasonable time. The tuning methodology for the number of trees and $MaC$ is further discussed in Section~\ref{sec:ch2:perftuning}.

\subsubsection{Online construction of \acrshort{combi}}\label{sec:ch2:online}
Due to the exponential growth in the size of \acrshort{ibst}, it is not practical to construct \acrshort{combi} by compressing an \acrshort{ibst}. Conversely, Algorithm~\ref{alg:ch2:IBSTCompress} does not present a way to update the tree while inserting new bit codes.

\begin{figure}
	\makebox[1\textwidth][c]{
		\resizebox{1 \linewidth}{!}{ %
			\includegraphics[width=\linewidth]{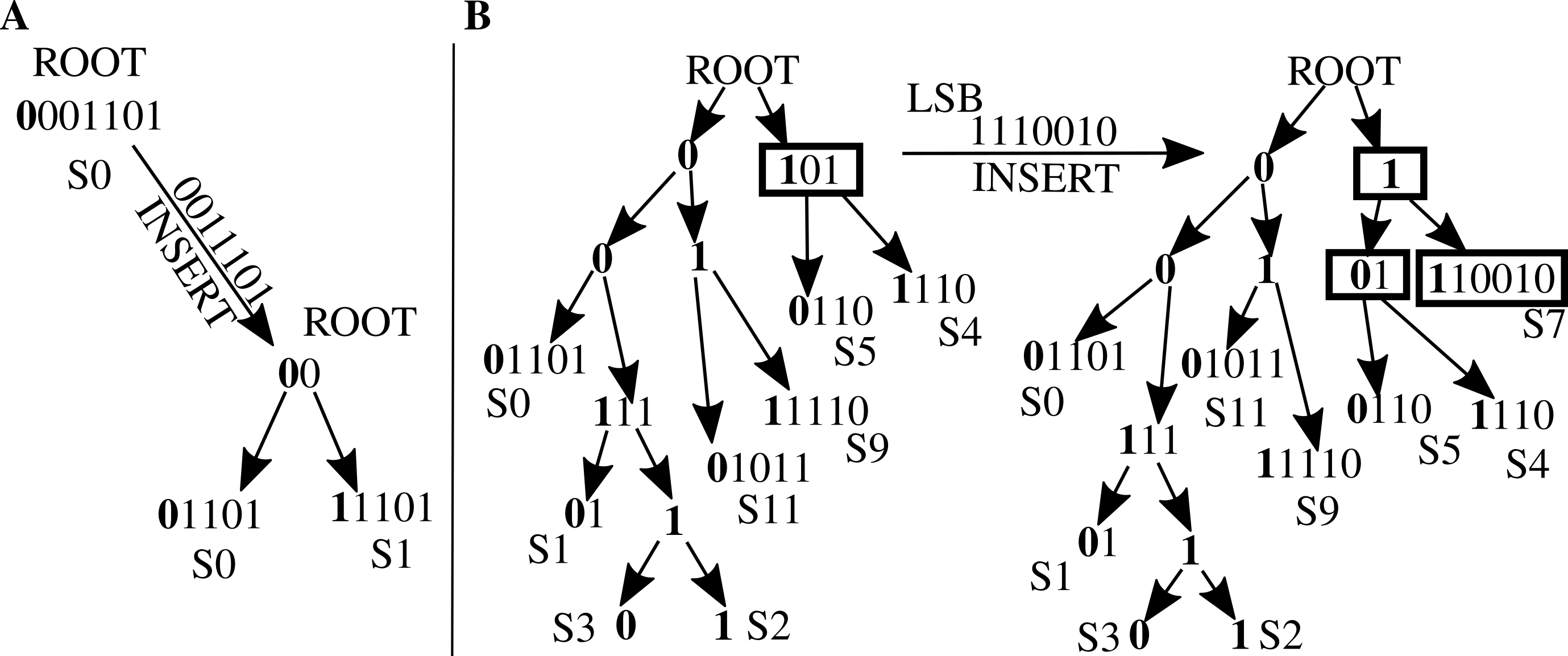}
		}
	}
	\caption{\textbf{The methodology of online \acrshort{combi} construction. The highlighted nodes are affected during insertion.} \textbf{A)} Insertion of first bit code. \textbf{B)} Update in \acrshort{combi}.}
	\label{fig:ch2:insertCode}
\end{figure}

To build \acrshort{combi} online, we start with a bit code and place it at the root of the tree (Figure~\ref{fig:ch2:insertCode}A). On arrival of a new bit code, a node will spawn its child for \textit{b} bit if it is the first bit of disagreement when scanning bits from the \acrfull{lsb} to the \acrfull{msb}. By convention, \acrshort{combi} places the child with bit 0 to the left and the child with bit 1 to the right. After spawning, restructuring of the tree falls into one of the two cases:
\begin{enumerate}
	\item \textbf{If the current node is a leaf,} the spawned children become leaf nodes.
	\item \textbf{If the current node is internal,} children of the current node become children of the spawned child that will contain the part of the bit code of the current node (Figure~\ref{fig:ch2:insertCode}B).
\end{enumerate} 

Hence, insertion in \acrshort{combi} requires the adjustment of 3 pointers. However, insertion also incurs an overhead of O(\texttt{len(}bit code\texttt{)}) \texttt{xor} computations. For every bit code, overhead also involves a linear scan on the length to find the first bit of disagreement. Thus, an insertion is of O(\texttt{len(}bit code\texttt{)}$^2$) in the worst case. Algorithm~\ref{alg:ch2:onlinealgo} gives detailed steps of the online construction of a tree in \acrshort{combi}.

Note that the tree constructed by both the algorithms (Algorithms~\ref{alg:ch2:IBSTCompress} and~ \ref{alg:ch2:onlinealgo}) will be the same. A bit of disagreement in an online tree construction algorithm will spawn two children. This is equivalent to a bit that partitions the space. Hence, a node with two children in a tree compression algorithm can be put in one-to-one correspondence with a node belonging to the bit of disagreement in an online algorithm.

\begin{algorithm}[!htb]
	\caption{Online Construction of \acrshort{combi}}
	\label{alg:ch2:onlinealgo}
	\textbf{Input}: B:bit code of x \\
	\textbf{Parameter}: CuN:current node in \acrshort{combi}, CB: local codebook\\
	\textbf{Output}: Constructed \acrshort{combi}
	\begin{algorithmic}[1] 
		\STATE assign root node of the tree to CuN
		\IF{startBit(CuN) == -1 \AND endBit(CuN) == -1}
		\STATE startBit(CuN) = 0, endBit(CuN) = len(B)
		\STATE refIndex(CuN) = size(CB), insert(CB) = B
		\ELSE
		\WHILE{True}
		\STATE xorCode = xor(CB[refIndex(CuN)], B)
		\STATE FB = firstSetBit(xorCode) // FB is first bit of disagreement
		\IF {FB $\geq$ startBit(CuN) \AND FB $<$ startBit(CuN)}
		\STATE create two nodes NN1, NN2 of type CuN
		\STATE startBit(NN1) = FB + 1
		\STATE endBit(NN1) = endBit(CuN)
		\STATE refIndex(NN1) = refIndex(CuN)
		\STATE startBit(NN2) = FB + 1
		\STATE endBit(NN2) = len(B)
		\STATE refIndex(NN2) = size(CB)
		\STATE insert(CB) = B
		\STATE endBit(CuN) = FB
		\STATE leftChild(NN1) = leftChild(CuN)
		\STATE rightChild(NN1) = rightChild(CuN)
		\IF{B[FB] == 1}
		\STATE leftChild(CuN) = NN1
		\STATE rightChild(CuN) = NN2
		\ELSE
		\STATE leftChild(CuN) = NN2
		\STATE rightChild(CuN) = NN1
		\ENDIF
		\STATE \textbf{break}
		\ELSE
		\IF{B[endBit(CuN)] == 1}
		\STATE CuN = rightChild(CuN)
		\ELSE
		\STATE CuN = leftChild(CuN)
		\ENDIF
		\ENDIF 
		
		\ENDWHILE
		\ENDIF
		
	\end{algorithmic}
\end{algorithm}

\subsubsection{Scaling \acrshort{combi} on a large data.}

\acrshort{combi} can scale to very large data sets by using multiple machines or a compute cluster. Typically, in such scenarios, all machines perform a local search on partial data sets they have, and then these results are collected by one machine to produce the final output~\cite{muja2014scalable}. A similar effect in \acrshort{combi} can be achieved by the following steps:

\begin{enumerate}
	\item \textbf{Index and Tree Construction Step}: The parameters to generate bit codes can be shared across all compute nodes in the cluster. Local \acrshort{combi} is constructed using these bit codes on every node. Algorithm~\ref{alg:ch2:combiOnComputeCluster} gives a detailed description of the construction step.
	\item \textbf{Search Step}: The query sample is shared across all compute nodes. The compute node performs a search locally and extracts local nearest neighbors. All the nodes send their information regarding retrieved samples (along with the hamming distance) to one designated node. This node ranks all the retrieved samples and selects the required count of nearest neighbors as the final output. Algorithm~\ref{alg:ch2:combiSearchOnComputeCluster} gives a detailed description of the search step.
\end{enumerate}

\begin{algorithm}[!htb]
	\caption{\acrshort{combi} construction in a Compute Cluster}
	\label{alg:ch2:combiOnComputeCluster}
	\textbf{Input}: D: Data set on a node of cluster, H: Hashing parameters\\ 
	\textbf{Parameter}: nodeID: ID of compute node in cluster.\\
	\textbf{Output}: \acrshort{combi}: Constructed \acrshort{combi} tree on local data set.
	\begin{algorithmic}[1] 
		\IF {nodeId == MASTER}
		\STATE Broadcast H to all other nodes in the cluster
		\ELSE
		\WHILE {H is not received from MASTER}
		\STATE Wait for communication from MASTER.
		\ENDWHILE
		\ENDIF
		\STATE Hash data set using H and generate bit codes.
		\STATE Create a dummy root node.
		\STATE Construct \acrshort{combi} using Algorithm~\ref{alg:ch2:IBSTConstruct} and Algorithm~\ref{alg:ch2:IBSTCompress} or Algorithm~\ref{alg:ch2:onlinealgo}.
		\IF {nodeId == MASTER}
		\STATE Wait for other cluster nodes to finish construction.
		\ELSE
		\STATE Send completion signal to MASTER node.
		\ENDIF
	\end{algorithmic}
\end{algorithm}

\begin{algorithm}[!htb]
	\caption{\acrshort{combi} search in a Compute Cluster}
	\label{alg:ch2:combiSearchOnComputeCluster}
	\textbf{Input}: \acrshort{combi}: \acrshort{combi} constructed on the node, Q: query sample, H: hashing parameters, N: nearest-neighbor count\\ 
	\textbf{Parameter}: nodeId: ID of compute node in cluster.\\
	\textbf{Output}: \acrshort{nn}: N nearest-neighbors
	\begin{algorithmic}[1] 
		
		\IF {nodeId == MASTER}
		\STATE Broadcast Q to all the other nodes in the cluster
		\ELSE
		\WHILE {Q is not received from MASTER}
		\STATE Wait for communication from the master
		\ENDWHILE
		\ENDIF
		\STATE Generate bit code for Q using H.
		\STATE Search nearest neighbors using Algorithm~\ref{alg:ch2:IBSTSearch}.
		\STATE Compute the hamming distance between the query sample and retrieved neighbors.
		
		\IF {nodeId == MASTER}
		\STATE Receive sample index and corresponding hamming distance from other nodes.
		\STATE Order all the retrieved and received samples via hamming distance.
		\STATE Select top N nearest neighbor indexes.
		\ELSE
		\STATE Send retrieved neighbors and their hamming distances to MASTER. 
		\ENDIF
		
		\IF{nodeID == MASTER}
		\STATE Broadcast selected neighbors to all other nodes
		\WHILE {Feature vectors of all selected samples are not received}
		\STATE Wait
		\ENDWHILE
		\STATE Return search result.
		\ELSE
		\STATE Wait for indexes from MASTER
		\STATE Fetch feature vector of the requested indexes and return.
		\ENDIF
		
	\end{algorithmic}
\end{algorithm}

During a $k$-\acrshort{nn} search, a compute cluster has the following communication overheads:
\begin{enumerate}
	\item From MASTER: Broadcasting the query point to all the nodes.
	\item To MASTER: Collecting the information regarding the retrieved candidate NNs.
	\item From MASTER: Sending the index list of the selected \acrshort{nns}.
	\item To MASTER: Sending the feature vector of the selected \acrshort{nns}.
\end{enumerate}
Except for retrieving feature vectors, all other communications require a small amount of data transfer. Compared to the time required for the search of N nearest neighbors, this overhead should have little impact on the overall system performance.

Since the database is divided among the nodes in a compute cluster, the \acrshort{combi} generated on a node would be smaller in size when compared to the \acrshort{combi} generated on the whole database. Hence, a search $k$-\acrshort{nn} would take less time on a compute node. This improvement in the search time would overshadow the additional time required for top $N$ nearest neighbors selection on the MASTER node. Since all compute nodes are searching for $N$ nearest neighbors in their portion of the database and the MASTER is accumulating and performing the final search, the result should vary marginally, if any, when compared with the search performed on the whole database on a single machine.

\section{Implementation, Experiments and Results}\label{sec:ch2:experiments}

\subsection{Bit code generation}\label{sec:ch2:bitcodegen}
Let us assume a database $X$ of $N$ points, in which every point is sampled identically and independently from a distribution $D$ defined on a $d$-dimensional space $\mathbb{R}^{d}$.  Further assume a family of binary hash functions $\mathcal{H}$, where each member of the family $h$ is defined as:

\begin{align}\label{eq:ch2:lsh}
h(x) = I(w^{T}x \geq 0), \text{where}
\end{align}

\noindent
$I(.)$ is an indicator function and $w$ is a normalized $d$-dimensional vector sampled from a standard normal distribution $\mathcal{N}$(\textbf{0}, \textbf{I}), where \textbf{I} is a $d\times d$ identity matrix.

To generate a hash code or bit code of $b$ length, $b$ functions were sampled from $\mathcal{H}$ and concatenated. Hence, a bit code for a sample $x\in X$, $B_{x}$, is given by, 

\begin{align}
B_{x} = h_{1}(x)h_{2}(x)...h_{b}(x), \text{where } h_{i} \in \mathcal{H} \:\: \forall i \in \{1,..,b\}
\end{align}

A hash table contains bit codes of all samples in the database.

\subsection{\acrshort{combi} implementation details}
Algorithms~\ref{alg:ch2:IBSTCompress} and~\ref{alg:ch2:IBSTCompressedSearch} were implemented in \texttt{c++}. To store a bit code, \texttt{boost::dynamic\_bitset<>} was used. All trees and variables to perform search were stored as different template initialization of \texttt{std::vector<>}. For simplicity, Algorithm~\ref{alg:ch2:IBSTCompressedSearch} was implemented using backtracking. The search performance of \acrshort{combi} was compared with \acrshort{mih}~\cite{manu::mih}, an exact search algorithm in a hamming space. For comparison, original implementation of \acrshort{mih} (\url{https://github.com/norouzi/mih}) was used. \acrshort{mih} ranks retrieved samples via hamming distances to select nearest neighbors. For fair comparison, neighbors retrieved by \acrshort{combi} were also ranked via hamming distance. Both \acrshort{combi} and \acrshort{mih} were compiled using \texttt{gcc 7.4} with \texttt{-O3} and \texttt{-march=native} flags.

\subsection{Dataset description}\label{sec:ch2:datasetdesc}

\begin{table}[!ht]
	\centering
	\begin{tabular}{lrrr}  
		\toprule
		Data set     &Database   &Query        &Data set          \\
		&Size       &Samples      &Dimension        \\
		\midrule
		SIFT-1B     & 1B        & 10K         & 128             \\
		80M-tiny    & $\sim$80M & 10K         & 384             \\
		\bottomrule
	\end{tabular}
	\caption{Dataset Description for empirical evaluation of \acrshort{combi}.}
	\label{tab:ch2:datasetdescription}
\end{table}

All experiments were performed on two large scale data sets. The description of data sets is present in Table~\ref{tab:ch2:datasetdescription}. SIFT-1B data set was downloaded from \url{http://corpus-texmex.irisa.fr/}. It has $1$ billion samples with 128 SIFT features in the database. The number of query samples in the SIFT-1B data set is $10$K. Gist features of the 80M-tiny image data set were downloaded from \url{http://horatio.cs.nyu.edu/mit/tiny/data/index.html}. This data set contains a database of size $\sim80$M samples. A set consisting of randomly selected $10$K samples from the database was kept aside as query samples. The remaining samples were used as a base database. Each image in the data set is represented via 384 Gist features. The ground truth in the hamming space up to 100 \acrshort{nns} was also created by an exhaustive search for both data sets.

\subsection{Experiment design}\label{sec:ch2:experimentDesign}
All experiments were performed with $64$, $128$, and $256$ lengths of bit codes. To generate bit codes of base and query images, \acrshort{lsh} was used (Section~\ref{sec:ch2:bitcodegen}). Only one hash table was constructed for every data set. In the current study, the performance of \acrshort{combi} is compared with \acrshort{mih}~\cite{manu::mih}. The comparison was made for the tasks of retrieving 1, 10, and 100 nearest neighbors.

\acrshort{mih} has one hyperparameter, the number of bit splits, $s$, to be tuned. This value was tuned for 64 bits, 128 bits, and 256 bits long bit codes in the range of 2-4, 4-8, and 8-10, respectively. For all data sets, the value of $s$ resulting in the lowest search time is selected.

\acrshort{combi} has two hyperparameters, the number of trees ($T$) and the number of mutates ($m$). The values of these hyperparameters are intuitive and can be easily decided for the required task. A complete analysis of performance tuning of \acrshort{combi} is present in Section~\ref{sec:ch2:perftuning}. For all experiments, the maximum number of trees in \acrshort{combi} was fixed at 4. One tree was built with the original bit code. The other tree used the reversed order of bit code. For the remaining two trees, the bit code was circularly rotated to the right from the middle. If the need arises, more trees can also be generated by different combinations of rotation and reversal. For the 80M-tiny image data set, all trees were generated by a pipeline of Algorithm~\ref{alg:ch2:IBSTConstruct} and Algorithm~\ref{alg:ch2:IBSTCompress}. The trees for the SIFT-1B data set were created by employing Algorithm~\ref{alg:ch2:onlinealgo} due to the enormous size of the database. The experiments with randomization of bit codes were also performed. However, a combination of reversal and rotation was found to converge on a small number of trees. The number of mutations ($m$) was varied from 0 to 6 for all data sets. Note, the value of $m$ has a direct relationship with the length of bit code and the size of the database or the amount of compression achieved by \acrshort{combi} w.r.t. \acrshort{ibst}. In effect, higher compression implies a larger value of $m$. In context of Algorithm~\ref{alg:ch2:IBSTCompressedSearch}, the implementation of \acrshort{combi} assumes $MiC=0$ and $MaC=m$.

Table~\ref{tab:ch2:prec09} and Table~\ref{tab:ch2:prec095} reports the selected hyperparameters for different tasks for both \acrshort{mih} and \acrshort{combi}.

\subsection{Performance metric}

Let us assume for a query sample $q$; an algorithm is tasked to find $k$ \acrshort{nns}. Assume that the returned set of $k$ neighbors for query $q$ is represented by $Q^{k}_{q}$ and $|Q^{k}_{q}|=k$. Further, in the ground truth, $k^{th}$ nearest neighbor is present at $d$-\acrshort{hd}. Then, a subset of ground truth containing samples up to $d$-\acrshort{hd} is given by ${G}^{k}_{q}$.

\subsubsection{Nearest samples in a hamming space}\label{sec:ch2:hdperformance}
The retrieval precision for $q$ in a hamming space is denoted by $H_{(p@k)_{q}}$, and can be computed as:

\begin{align}\label{eq:ch2:hamperf}
H_{(p@k)_{q}}=\frac{|Q^{k}_{q} \cap G^{k}_{q}|}{k}
\end{align}

\noindent
and average precision $\overline{H_{p@k}}$ across all the query samples is given as $\frac{1}{N}\sum_{q=1}^{q=N}H_{(p@k)_{q}}$, where $N$ is the total number of query samples.

\subsubsection{False discovery rate}

The \acrfull{fdr} of query $q$ for $k$ nearest neighbors is given by 

\begin{align}\label{eq:ch2:fdrq}
FDR_{q} = \frac{|Q^{k}_{q} \setminus G^{k}_{q}|}{k}
\end{align}

\noindent
If there are $N$ queries, then \acrshort{fdr} is given by

\begin{align}\label{eq:ch2:fdr1}
\acrshort{fdr} = \frac{1}{N}\sum_{q=1}^{q=N}FDR_{q}
\end{align}

\noindent
Replacing the value of (\ref{eq:ch2:fdrq}) in (\ref{eq:ch2:fdr1}) and rearranging

\begin{align}\label{eq:ch2:fpmv}
\acrshort{fdr} = \frac{1}{N*k} \sum_{q=1}^{q=N}|Q^{k}_{q} \setminus G^{k}_{q}|
\end{align}

\noindent
Using the identity, $|A \setminus B| = |A \setminus (A \cap B)| = |A| - |A \cap B|$ in (\ref{eq:ch2:fpmv}). 

\begin{align}\label{eq:ch2:fvmv1}
\acrshort{fdr} = \frac{1}{N*k} \sum_{q=1}^{q=N} |Q^{k}_{q}|-|Q^{k}_{q} \cap G^{k}_{q}|
\end{align}

\noindent
Since $k$ nearest neighbors are retrieved, rearranging (\ref{eq:ch2:fvmv1}) and replacing values from (\ref{eq:ch2:hamperf}) gives

\begin{align}\label{eq:ch2:fdr}
\acrshort{fdr} = 1 - \frac{1}{N}\sum_{q=1}^{q=N} H_{(p@k)_{q}} = 1 -  \overline{H_{p@k}}
\end{align}

\subsubsection{Speedup}
The speedup is computed as \acrshort{mih}-time/\acrshort{combi}-time. In a particular case, when \acrshort{combi}-time $\leq$ \acrshort{mih}-time, then the value of speedup $\geq$ 1.

\subsection{Performance comparison}

\acrfull{mih}~\cite{manu::mih} is an exact search algorithm in a hamming space. Thus, its precision (\ref{eq:ch2:hamperf}) is found to be 1 for every setting. To compare the performance of \acrshort{combi} and \acrshort{mih}, speed-up of \acrshort{combi} over \acrshort{mih} was analyzed first. Then, the quality of nearest neighbors were analysed.

\subsubsection{Speed-up analysis}

The performance of the search algorithms is evaluated in terms of the time taken by the algorithm to reach a specified precision. Since both \acrshort{mih} and \acrshort{combi}, rank the \acrshort{nns} in a hamming space, the retrieval speed of the algorithm in a hamming space were compared. As shown in Table~\ref{tab:ch2:prec09}, ComBI can achieve at least 0.90 precision (\ref{eq:ch2:hamperf}) in relatively lesser time. Notably, the time taken by \acrshort{combi} is almost $\sim296$ times lesser in retrieving 1 \acrshort{nn} for 64-bit long bit codes for the 80M-tiny image data set. In general, \acrshort{combi} is at least $\sim$4 times faster for the 80M-tiny image data set. For the SIFT-1B data set, \acrshort{combi} is at least $\sim$3 times faster in reaching a precision of 0.90. Table~\ref{tab:ch2:prec095} shows the time taken by \acrshort{combi} to achieve at least 0.95 precision (\ref{eq:ch2:hamperf}) on both the data sets. In this case, \acrshort{combi} is at least $\sim$3 times faster for the 80M-tiny data set and $\sim$2 times faster for the SIFT-1B data set. All reported values of time are inclusive of the ranking time.

All experiments on the SIFT-1B data set were run on a single core of a workstation with Intel Xeon E7-4820 CPUs with 1.9 GHz, 1024 GB DDR4 Synchronous 2133 MHz RAM, and Ubuntu 14.04 LTS Operating System with the 4.4.0-142-generic kernel. All experiments on the 80M-tiny data set were run on a single core of a workstation with Intel Xeon Gold 6148 CPUs with a clock speed of 2.5 GHz, 756 GB DDR4 Synchronous 2666 MHz RAM, and Ubuntu 18.04 LTS Operating System with the 4.15.0-20-generic kernel.

\begin{table*}[!ht]
	\makebox[1 \textwidth][c]{
		\resizebox{1 \linewidth}{!}{ %
			\centering
			\begin{tabular}{lrr|rrrrr|rrr|rr}
				\toprule
				&       &       &   &   & \acrshort{combi}     &                       &           &   & \acrshort{mih}       &               &           & Access\\ 
				\cmidrule{4-11}
				Data set & bits  & \acrshort{nns}   & T & m & time(s)   & $\overline{H_{p@k}}$  & Access \% & s & time(s)   & Access \%     & speedup   & Ratio\\
				\midrule
				SIFT-1B&64&1&2&1&1.9e-03&0.953&0.00014&2&3.7e-02&0.00281&19.57&19\\
				&&10&3&1&4.3e-03&0.916&0.00015&2&5.7e-02&0.00767&13.26&52\\
				&&100&2&2&2.0e-02&0.913&0.00079&2&1.1e-01&0.02057&5.16&26\\
				&128&1&2&3&1.2e-01&0.922&0.00185&4&3.1e-01&0.06729&2.60&36\\
				&&10&3&3&1.7e-01&0.936&0.00259&4&8.6e-01&0.14507&5.07&56\\
				&&100&2&4&5.5e-01&0.936&0.01291&4&1.4e+00&0.30566&2.55&23\\
				&256&1&3&4&8.9e-01&0.930&0.01822&8&3.7e+00&0.55323&4.20&30\\
				&&10&3&4&8.8e-01&0.917&0.01822&8&6.6e+00&1.01405&7.51&55\\
				&&100&3&5&3.5e+00&0.963&0.09186&8&1.1e+01&1.77733&3.20&19\\
				\midrule
				80M-tiny&64&1&3&0&1.3e-05&0.920&0.00012&2&3.8e-03&0.04642&296.12&372\\
				&&10&1&1&6.9e-05&0.913&0.00181&2&9.6e-03&0.13051&139.00&72\\
				&&100&1&2&8.0e-04&0.940&0.01385&2&2.3e-02&0.32416&29.08&23\\
				&128&1&2&2&2.5e-03&0.907&0.00304&4&6.5e-02&0.87325&26.31&286\\
				&&10&2&3&1.8e-02&0.973&0.03122&4&1.5e-01&2.02797&8.52&64\\
				&&100&3&3&2.6e-02&0.921&0.04239&4&2.7e-01&3.59069&10.41&84\\
				&256&1&3&4&1.2e-01&0.915&0.23904&8&5.0e-01&5.41580&4.36&22\\
				&&10&4&4&1.6e-01&0.942&0.33002&8&1.0e+00&11.04096&6.19&33\\
				&&100&3&5&4.5e-01&0.952&1.05529&8&1.5e+00&15.92497&3.40&15\\
				\bottomrule
			\end{tabular}
	}}
	\caption{The column of \textit{time(s)} contains an average of the nearest neighbor search time for all query samples. For \acrshort{combi}, the results are reported for a configuration that takes minimum time to achieve a precision of $\geq$0.90 in a hamming space. For \acrshort{mih}, a configuration with the least time is selected. }
	\label{tab:ch2:prec09}
\end{table*}

\begin{table*}[!ht]
	\makebox[1 \textwidth][c]{
		\resizebox{1 \linewidth}{!}{ %
			\begin{tabular}{lrr|rrrrr|rrr|rr}
				\toprule
				&       &       &   &   & \acrshort{combi}     &                       &           &   & \acrshort{mih}       &               &           & Access\\ 
				\cmidrule{4-11}
				Data set & bits  & \acrshort{nns}   & T & m & time(s)   & $\overline{H_{p@k}}$  & Access \% & s & time(s)   & Access \%     & speedup   & Ratio\\
				\midrule
				SIFT-1B&64&1&2&1&1.9e-03&0.953&0.00014&2&3.7e-02&0.00281&19.57&19\\
				&&10&1&2&9.1e-03&0.955&0.00069&2&5.7e-02&0.00767&6.24&11\\
				&&100&4&2&3.8e-02&0.977&0.00093&2&1.1e-01&0.02057&2.74&21\\
				&128&1&3&3&1.6e-01&0.954&0.00259&4&3.1e-01&0.06729&1.93&26\\
				&&10&4&3&2.2e-01&0.965&0.00339&4&8.6e-01&0.14507&3.87&42\\
				&&100&3&4&7.8e-01&0.971&0.01792&4&1.4e+00&0.30566&1.79&17\\
				&256&1&4&4&1.2e+00&0.972&0.02457&8&3.7e+00&0.55323&3.23&22\\
				&&10&4&4&1.1e+00&0.964&0.02457&8&6.6e+00&1.01405&5.76&41\\
				&&100&3&5&3.5e+00&0.963&0.09186&8&1.1e+01&1.77733&3.20&19\\
				\midrule
				80M-tiny&64&1&1&1&6.8e-05&0.981&0.00181&2&3.8e-03&0.04642&55.46&25\\
				&&10&2&1&1.6e-04&0.963&0.00187&2&9.6e-03&0.13051&60.21&69\\
				&&100&2&2&1.9e-03&0.978&0.01515&2&2.3e-02&0.32416&12.26&21\\
				&128&1&4&2&4.7e-03&0.954&0.00533&4&6.5e-02&0.87325&13.95&163\\
				&&10&2&3&1.8e-02&0.973&0.03122&4&1.5e-01&2.02797&8.52&64\\
				&&100&2&4&9.0e-02&0.986&0.19594&4&2.7e-01&3.59069&3.02&18\\
				&256&1&4&4&1.6e-01&0.957&0.33002&8&5.0e-01&5.41580&3.16&16\\
				&&10&3&5&4.4e-01&0.982&1.05529&8&1.0e+00&11.04096&2.32&10\\
				&&100&3&5&4.5e-01&0.952&1.05529&8&1.5e+00&15.92497&3.40&15\\
				\bottomrule
			\end{tabular}
	}}
	\caption{The column of \textit{time(s)} contains an average of the nearest neighbor search time for all query samples. For \acrshort{combi}, the results are reported for a configuration that takes minimum time to achieve a precision of $\geq$0.95 in a hamming space. For \acrshort{mih}, a configuration with the least time is selected.}
	\label{tab:ch2:prec095}
\end{table*}

\subsubsection{Quality of approximate search}
With increasing values of the number of trees ($T$) and the mutation count ($m$), an approximate search of \acrshort{combi} tends towards an exact search but, its search time also increases. Hence, within the given time limits, an approximate search is performed. It is important to note that in many applications, an exact search in a hamming space is not always required. For example, in the image retrieval task, it can assumed that similar objects form clusters. To find some similar images for a query object, it is enough to select images from within the cluster of the query object. In such situations, if the search stops in the proximity of the exact \acrshort{nns} for the query object, results will still be satisfactory. That is, in a hamming space, if the difference between \acrshort{hd} of approximate \acrshort{nns} and exact \acrshort{nns} of a query object is not too significant, results will still be satisfactory. 

Assuming that $k$ \acrshort{nns} are required for an query object $q$ and $k^{th}$ \acrshort{nn} is $d_{q}$-\acrshort{hd} away from $q$. Let us assume that $Q^{k}_{q}$ represents the set of selected $k$ neighbors by \acrshort{combi} for query $q$. Further, assume that $O^{k}_{q} \subseteq Q^{k}_{q}$ represents the set of neighbors farther than $d_{q}$-\acrshort{hd} away from $q$. Notice that $O^{k}_{q}$ is equal to ($Q^{k}_{q} \setminus G^{k}_{q}$) in (\ref{eq:ch2:fdrq}).  Assume that a sample $o \in O^{k}_{q}$ is $k_{i}$-\acrshort{hd} away from $k^{th}$ \acrshort{nn}, 
then a multi-set $\Lambda_{q}$ can be defined as 

\begin{align}\label{eq:ch2:lp}
\Lambda_{q} = \{k_{i}:F_{i}| \forall k_{i} > 0 \}, \text{ where}
\end{align}

\noindent
$F_{i}$ is frequency of $k_{i}$ for elements in $O^{k}_{q}$. For $N$ query samples these values can be accumulated in a multi-set $\Lambda$ as

\begin{align}\label{eq:ch2:cardmultiset}
\Lambda = \uplus_{q=1}^{q=N}\Lambda_{q}.
\end{align}

\noindent
Notice that the cardinality of multi-set $\Lambda$ is 

\begin{align*}
    \#\Lambda = \sum_{q=1}^{q=N}\#\Lambda_{q} = \sum_{q=1}^{q=N}|O^{k}_{q}| = \sum_{q=1}^{q=N}|Q^{k}_{q} \setminus G^{k}_{q}|
\end{align*}

\noindent
Extracting the value of $Q^{k}_{q} \setminus G^{k}_{q}$ from (\ref{eq:ch2:fpmv}).

\begin{align}
\#\Lambda = N*k*FDR, \text{ where}
\end{align}

\noindent
$FDR$ is defined in (\ref{eq:ch2:fdr}).

\begin{figure}[!ht]
	\makebox[1 \textwidth][c]{
		\resizebox{1 \linewidth}{!}{ %
			\includegraphics[width=\linewidth]{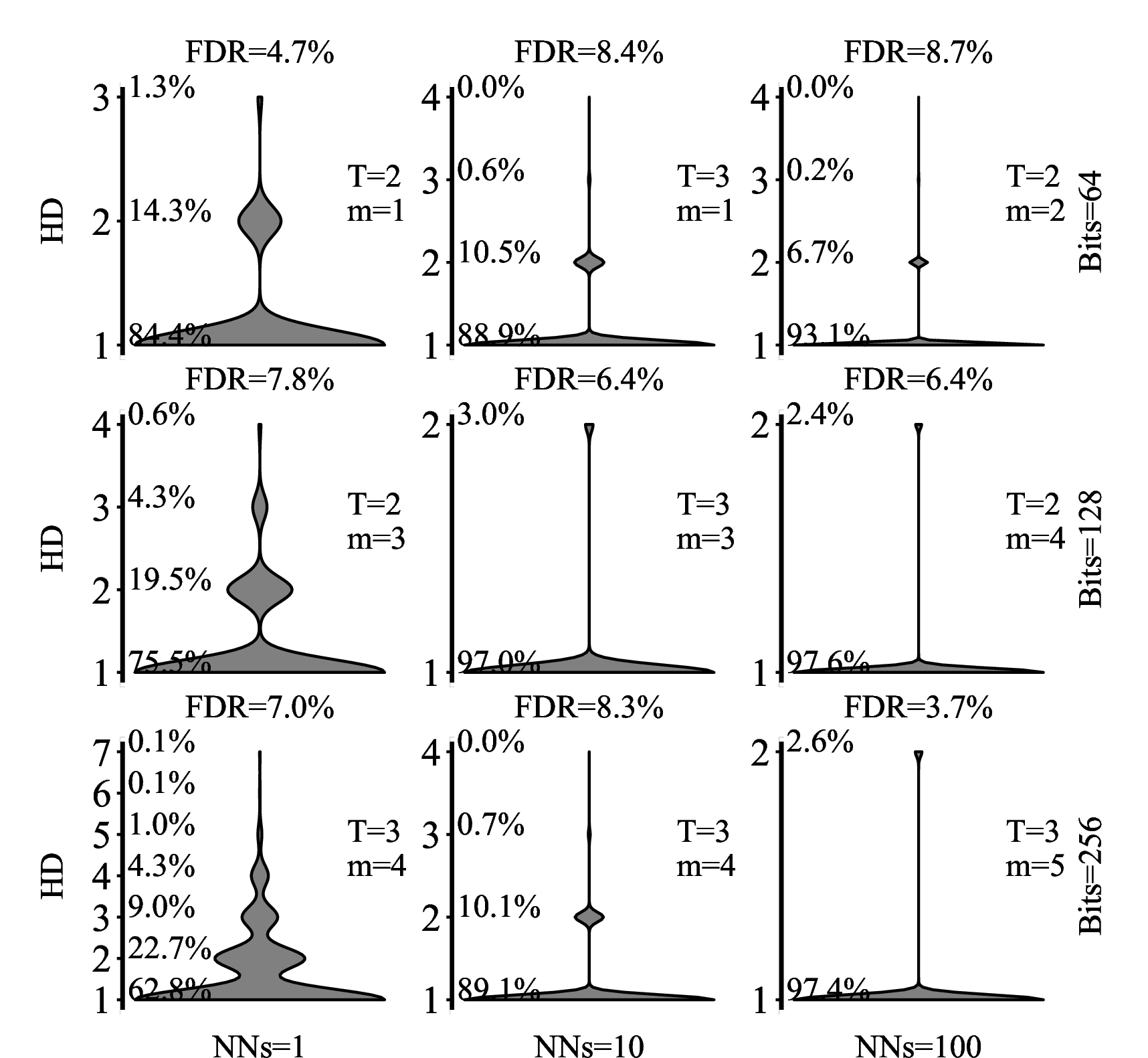}
	}}
	\caption{\textbf{Distribution of hamming distances (\acrshort{hd}) of returned neighbors by \acrshort{combi} from the farthest exact \acrshort{nn}}. \acrshort{fdr} represents the fraction of neighbors returned by \acrshort{combi} that are not part of the exact \acrshort{nn} set. Each violin plot corresponds to the configuration shown in Table~\ref{tab:ch2:prec09} for the combination of bits and \acrshort{nns} for the SIFT-1B data set.}
	\label{fig:ch2:approx}
\end{figure}

Each violin plot in Figure~\ref{fig:ch2:approx} plots $\Lambda$ for different combinations of bit-lengths and \acrshort{nns}. Each bubble in the violin plot corresponds to a $k_{i}$. The size of the bubble and the corresponding number represents the fraction of approximate nearest neighbors at $k_{i}$ distance from the farthest exact \acrshort{nn}. The plots correspond to the configurations mentioned in Table~\ref{tab:ch2:prec09} for the SIFT-1B data set. 

It is evident from Figure~\ref{fig:ch2:approx} that most of the samples amongst the sample that are not exact \acrshort{nns} ($FDR$) are only $1$-\acrshort{hd} away from the farthest exact \acrshort{nn}. For instance, \acrshort{combi} achieved a precision of 0.953 ($FDR$=4.7\%) in retrieving 1-\acrshort{nn} for 64 bit long code (Table~\ref{tab:ch2:prec09}). However, out of these 4.7\% neighbors, 84.4\%, 14.3\%, and 1.3\% were at $1$-\acrshort{hd}, $2$-\acrshort{hd}, and $3$-\acrshort{hd} away, respectively, from the farthest exact \acrshort{nn}. Similar trends are visible for other configurations as well.

\begin{figure}[!ht]
	\makebox[1 \textwidth][c]{
		\resizebox{1 \linewidth}{!}{ %
			\includegraphics[width=\linewidth]{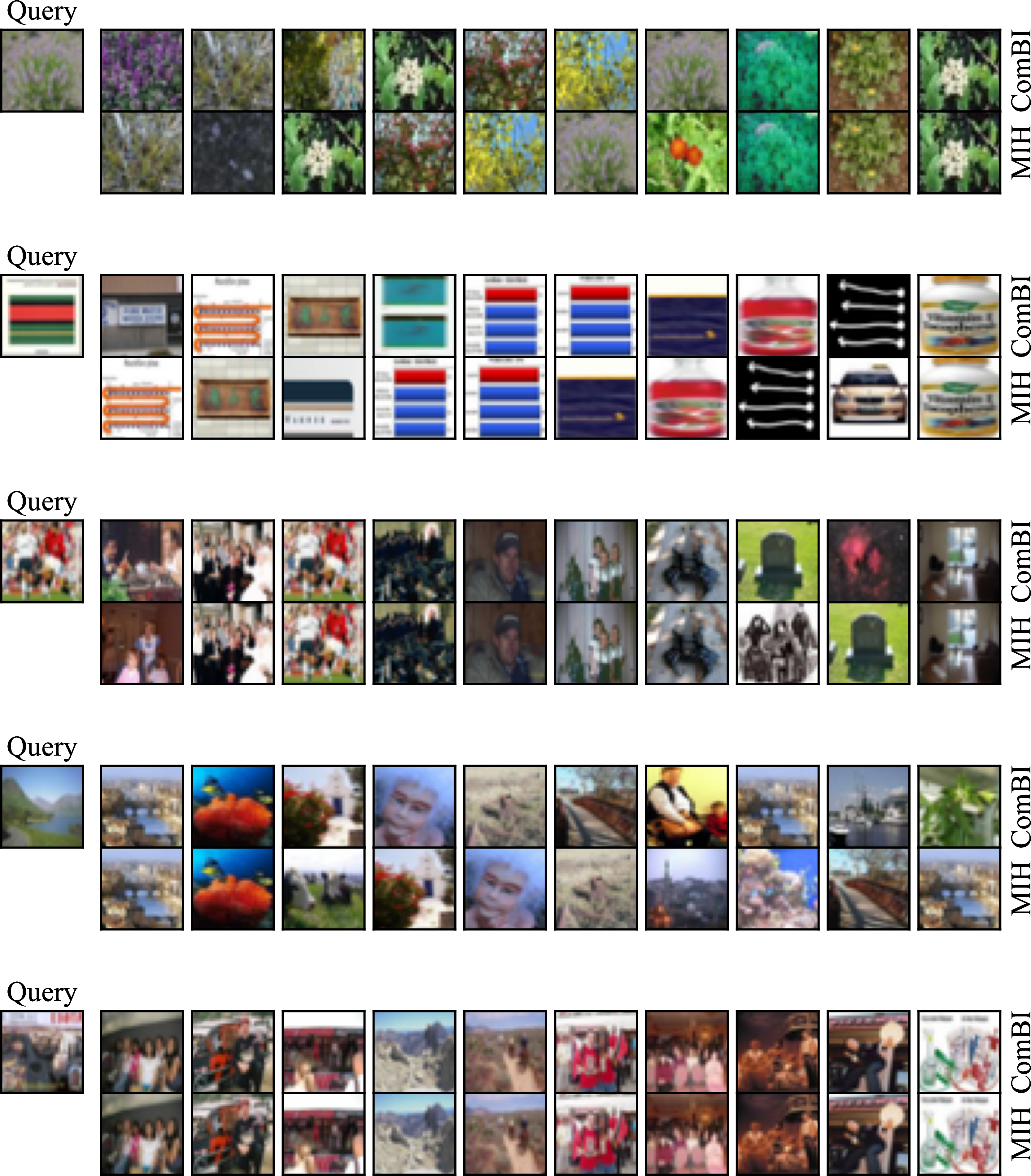}
	}}
	\caption{\textbf{Visualization of approximate search}. Visualization of top 10 nearest neighbors returned by \acrshort{combi} and \acrshort{mih} on 5 random samples for 256 bits long bit code.}
	\label{fig:ch2:visual}
\end{figure}

Figure~\ref{fig:ch2:approx} suggests that the neighbors returned by \acrshort{combi} , if are not exact \acrshort{nns}, are not very far from exact \acrshort{nns}. Hence, this suggests that an approximate search in \acrshort{combi} can be of practical use. For a subset of samples, a visual comparison of the search results is shown in Figure~\ref{fig:ch2:visual}.

\subsection{Comments on performance tuning}\label{sec:ch2:perftuning}

\begin{figure}
	\makebox[1 \textwidth][c]{
		\resizebox{1 \linewidth}{!}{ %
			\includegraphics[width=\linewidth]{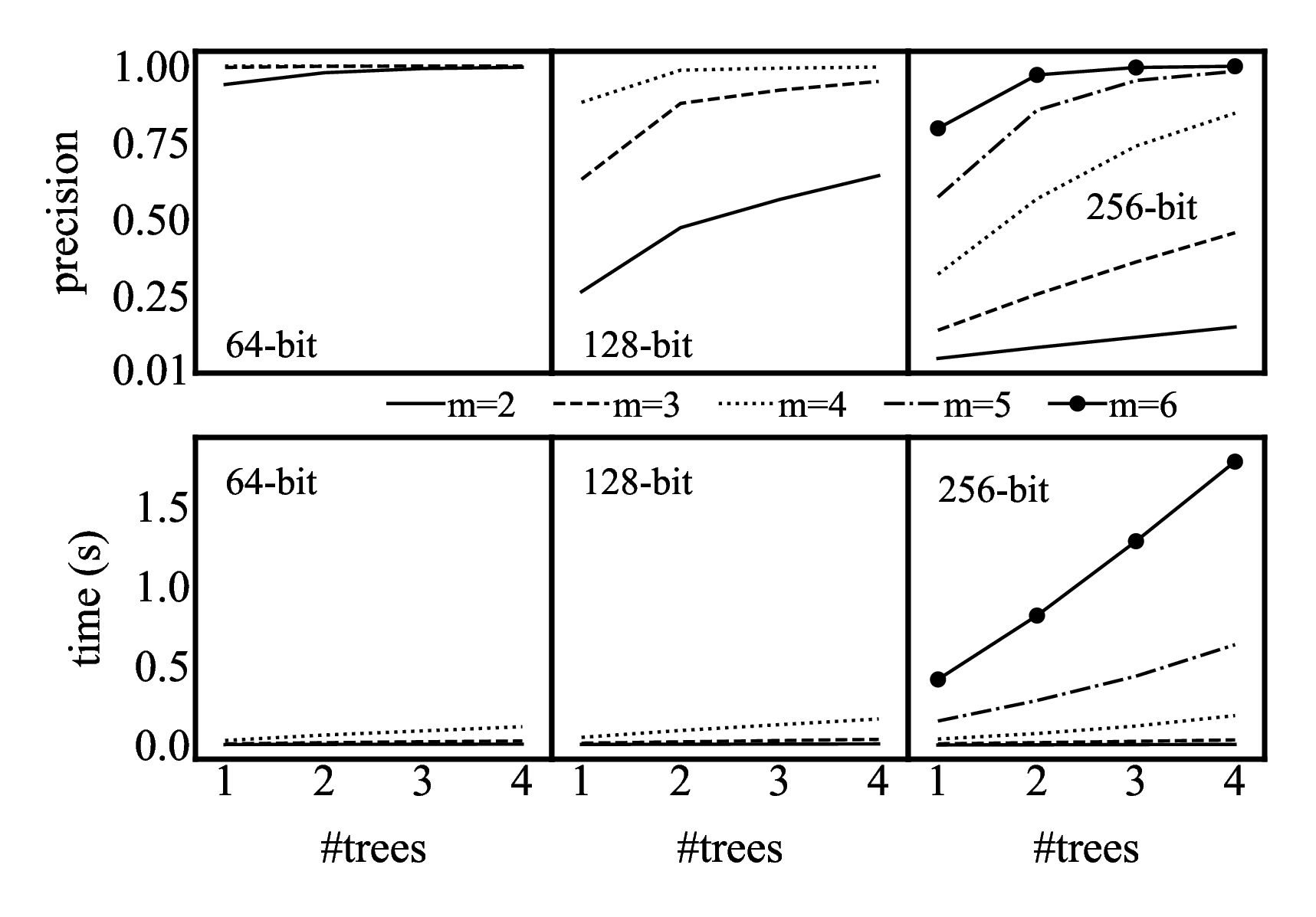}
		}
	}
	\caption{\textbf{Convergence on hyperparameters on 80-M tiny dataset}. Impact of tunable parameters of \acrshort{combi} ($T$ and $m$), on precision and time, for 100 \acrshort{nns} on the 80M-tiny image data set.}
	\label{fig:ch2:res}
\end{figure}

\begin{figure}
	\makebox[1 \textwidth][c]{
		\resizebox{1 \linewidth}{!}{ %
			\includegraphics[width=\linewidth]{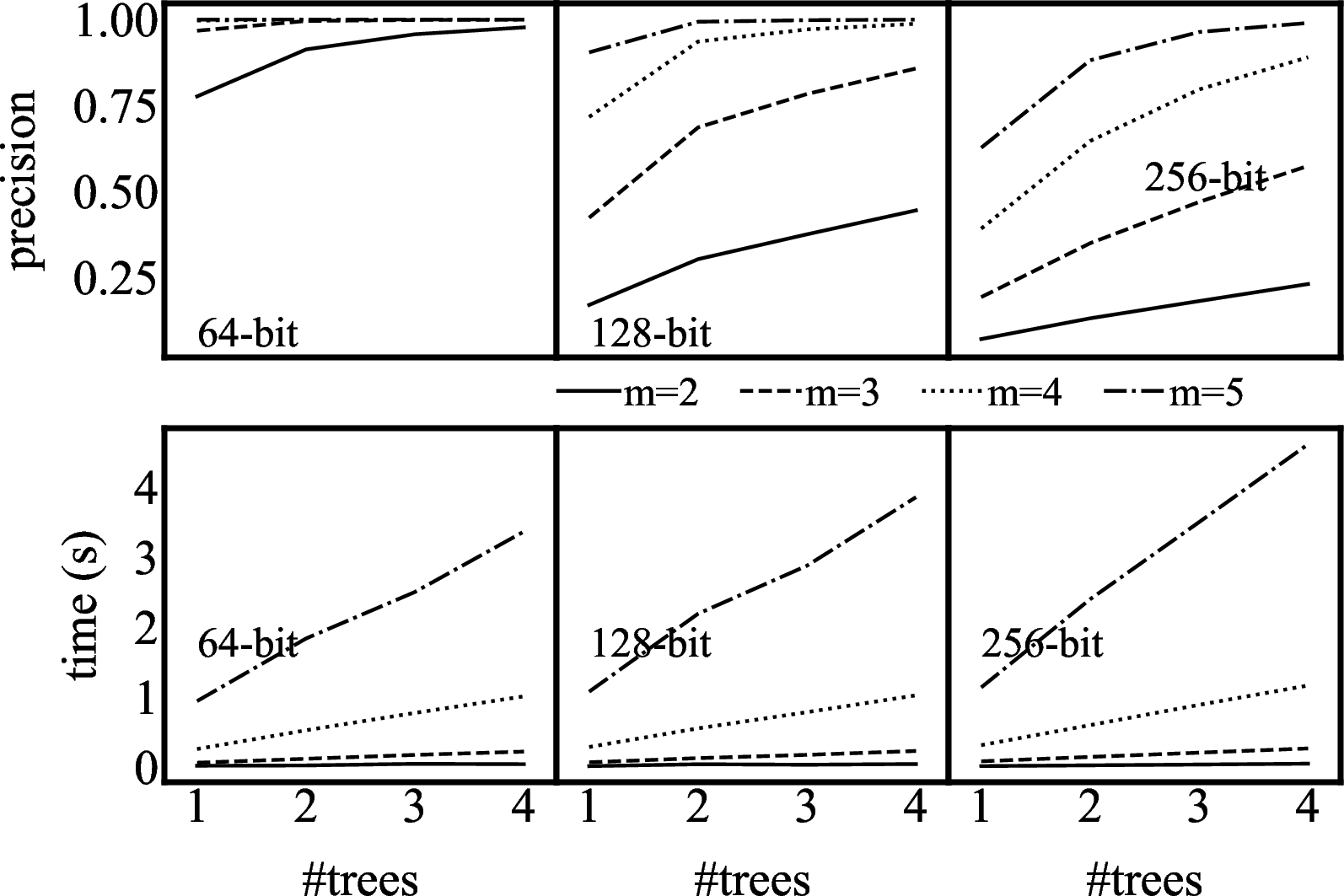}
		}
	}
	\caption{\textbf{Convergence on hyperparameters on SIFT-1B dataset}. Impact of tunable parameters of \acrshort{combi} ($T$ and $m$), on precision and time, for 100 \acrshort{nns} on the SIFT-1B dataset.}
	\label{fig:ch2:res1}
\end{figure}

The performance of \acrshort{combi} depends on two parameters, namely the number of trees ($T$) and mutate count ($m$). Increasing the value of either one of them increases the quality of nearest neighbors at the expense of time. Figure~\ref{fig:ch2:res} shows an increase in precision and time with increasing $T$ on different $m$ and bit code lengths for 100 nearest neighbors. Increasing $m$ improves precision more than an increase in $T$. However, the required search time increases. One desirable approach will be to decide a value of $m$ for a given combination of bit code length and desired nearest neighbor count and then increase $T$ until the precision threshold is reached or allocated search time is exhausted. As shown in Figures~\ref{fig:ch2:res} and~\ref{fig:ch2:res1}, a carefully selected value of $m$ does not require numerous trees to achieve the desired precision threshold in a reasonable time.

\subsection{Reduction in memory usage in comparison to \acrshort{ibst}}

\acrshort{combi} has a reduced memory footprint in comparison to \acrshort{ibst}. For 80M-tiny image data set, the total number of nodes in \acrshort{ibst} of 64, 128, and 256 bits reduced by a factor of $\sim$9, $\sim$32, and $\sim$99, respectively, during compression via Algorithm~\ref{alg:ch2:IBSTCompress} (Table~\ref{tab:ch2:nodecount}). A Similar trend is also visible in the SIFT-1B data set, where the compression ratio varies from $\sim$9 to $\sim$103.

\begin{table}[!ht]
	\centering
	\begin{tabular}{lrrrr}  
		\toprule
		Data set     & bits 	            & node count 		& node count 	& compression\\
		&                   & \acrshort{ibst} 		        & \acrshort{combi} 		& ratio\\
		& 					& 				    &               &(\acrshort{ibst}:\acrshort{combi})\\
		\midrule
		SIFT-1B     &64                 & 14,186,204,429    & 1,543,318,609 & 9.19:1\\
		&128                & 77,469,678,644    & 1,979,484,773 & 39.14:1\\
		&256                & 205,861,251,662   & 1,997,233,859 & 103.07:1\\
		\midrule
		80M-tiny    &64 		        & 649,305,347 		& 75,723,215 	& 8.58:1\\
		&128 	            & 4,508,768,077		& 143,237,939 	& 31.48:1 \\
		&256 	            & 14,228,334,090 	& 145,000,417 	& 98.13:1\\
		\bottomrule
	\end{tabular}
	\caption{A comparison of node counts in \acrshort{ibst} and \acrshort{combi}.}
	\label{tab:ch2:nodecount}
\end{table}

\subsection{ComBI as single cell search engine}\label{sec:ch2:singlecell}


\subsubsection{Pre-processing and hashing of gene expressions}

Gene expression data is generally presented as a matrix where every row represents a gene, and every column represents a cell type. Multiple pre-processing steps need to be performed before feeding them into the hashing algorithm for index creation to control for artifacts. For proper comparison, the steps proposed by Cellfishing.jl~\cite{sato2019cellfishing} were utilized for pre-processing and hashing. The pre-processing steps include:

\begin{itemize}
    \item Dropping the low-abundance and low-variance genes.
    \item Normalize every cell by total count to remove the library size artifact.
    \item Log transform ($\log(1+x)$) of every value $x$ in the filtered and normalized data.
    \item Extraction of top 50 principal components of the transformed data. This step is crucial as it reduces the hash computation time and helps mitigate the batch effect.

\end{itemize}

\begin{figure}
	\centering
	\makebox[1 \textwidth][c]{
		\resizebox{1.1 \linewidth}{!}{
			\includegraphics[width=\linewidth,keepaspectratio]{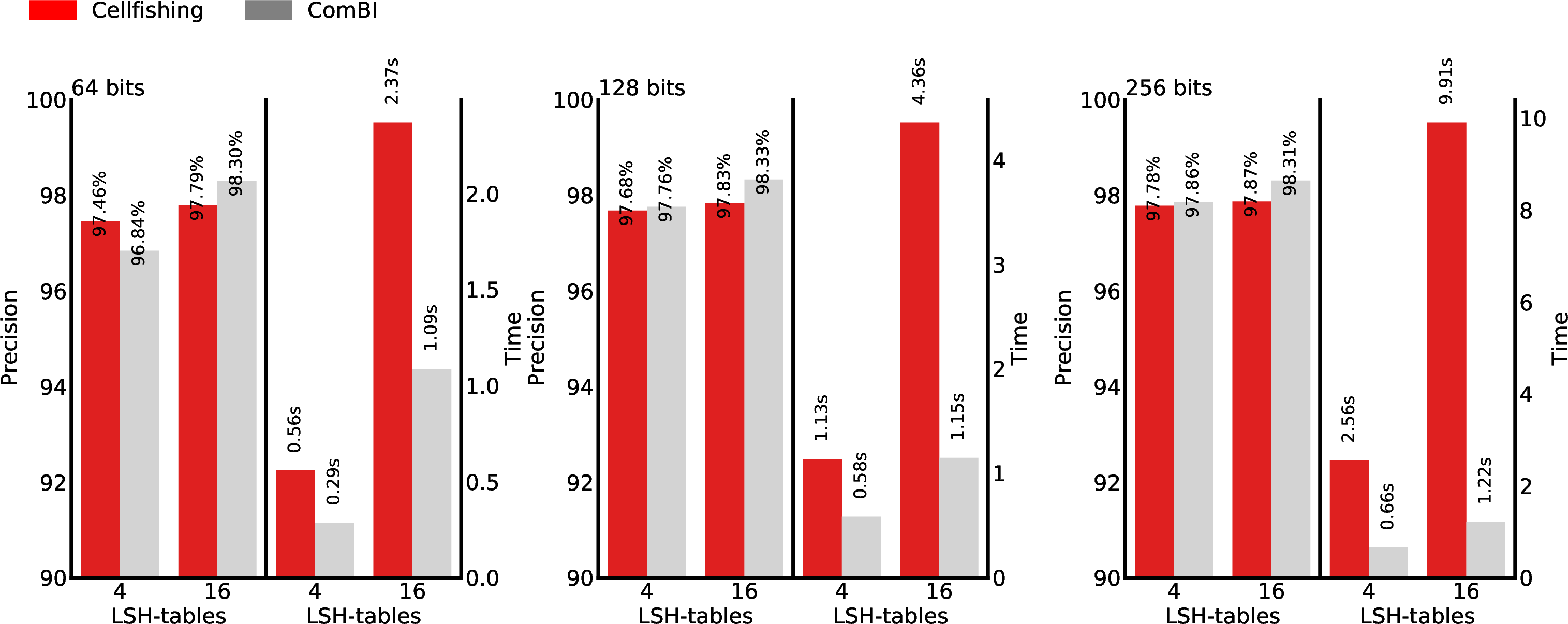}
		}
	}
	\caption{\textbf{Performance of \acrshort{combi} and Cellfishing.jl on baron2016 dataset}. The dataset has 8569 samples. With increasing length of bit code and higher number of tables \acrshort{combi} has better performance. For 64, 128, and 256 bits \acrshort{combi} has $\sim$2, $\sim$2, and $\sim$4 times speed-up in search time with 4 tables, respectively. Similarly, \acrshort{combi} has speed-up of $\sim$2, $\sim$3.8, and $\sim$8 for 64, 128, and 256 bits with 16 tables, respectively.}
	\label{fig:ch2:baron}
\end{figure}

After these pre-processing steps, the resulting matrix had the size of \texttt{No of cells} $\times 50$. Now, the resulting matrix was hashed using \acrshort{lsh}. The random vectors were orthogonalized to generate the hash codes. Assume that the required bit code length is $T$ and the number of features in the data that need to be hashed are $d$. Then if $T>d$, then $\lceil\frac{T}{d}\rceil$ batches of at most $d$ random vectors of $d$ dimensions were generated and then orthogonalized by QR decomposition. Then all the batches are concatenated, and final hashing is performed by following the steps discussed in Section~\ref{sec:ch2:bitcodegen}.

\begin{figure}
	\centering
	\makebox[1 \textwidth][c]{
		\resizebox{1.1 \linewidth}{!}{
			\includegraphics[width=\linewidth,keepaspectratio]{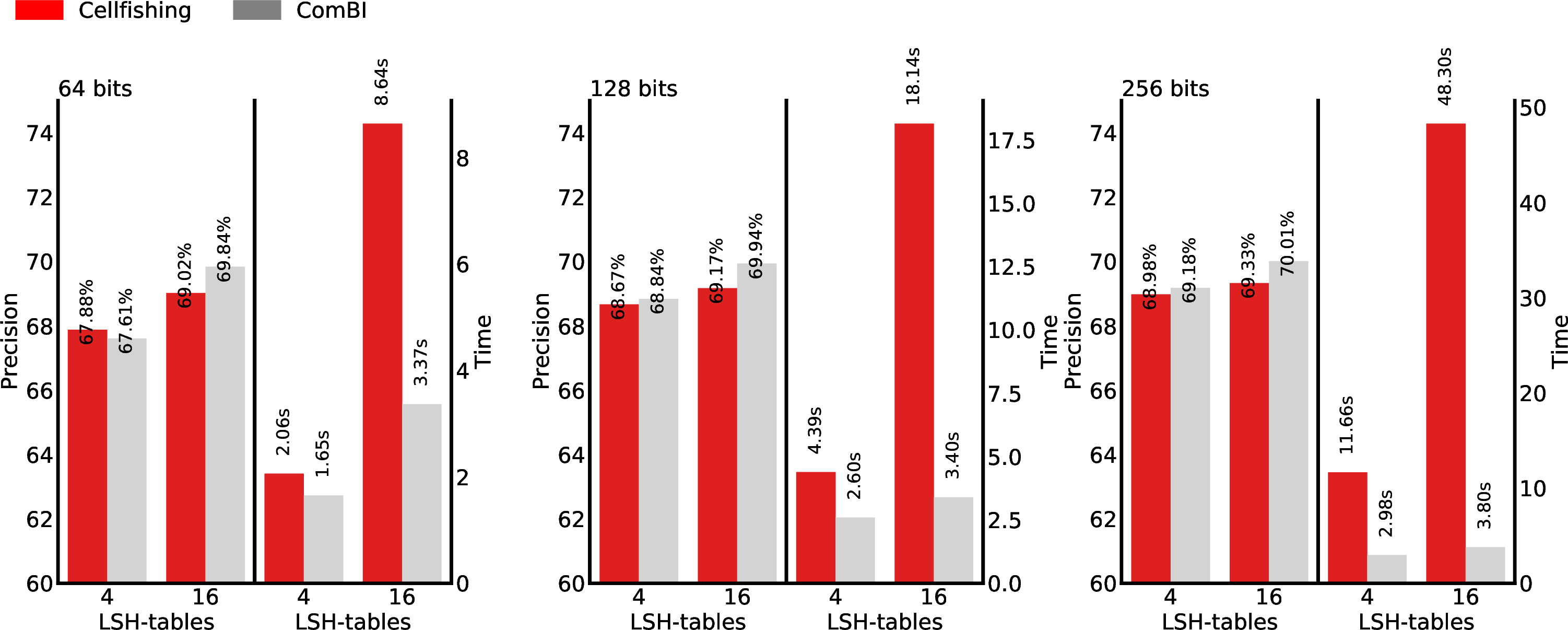}
		}
	}
	\caption{\textbf{Performance of \acrshort{combi} and Cellfishing.jl on plass2018 dataset}. The dataset hash 21612 samples. With increasing length of bit code and higher number of tables \acrshort{combi} has better performance. For 64, 128, and 256 bits \acrshort{combi} has $\sim$1.24, $\sim$1.68, and $\sim$4 times speed-up in search time with 4 tables, respectively. Similarly, \acrshort{combi} has speed-up of $\sim$2.5, $\sim$5.3, and $\sim$12.7 for 64, 128, and 256 bits with 16 tables, respectively.}
	\label{fig:ch2:plass}
\end{figure}

\subsubsection{Experimental setup for comparison}

This section compares \acrshort{combi} with Celfishing.jl~\cite{sato2019cellfishing} on three datasets. All three datasets have cell-type annotations. Baron \textit{et al.}~\cite{baron2016single} annotated the sequenced cells via hierarchical clustering followed by cell marker genes. The human cell and their annotations from the study were utilized for comparison. A total of 8569 cells were sequenced from humans in Baron's dataset. Plass \textit{et al.}~\cite{plass2018cell} sequenced more than 20,000 planarian cells. They utilized a combination of computational and experimental procedures to annotate and validate the annotated cell types. Shekhar \textit{et al.}~\cite{shekhar2016comprehensive} sequenced almost 25,000 mouse retinal bipolar cells and presented a systematic molecular methodology for cell type annotation. 

Both the algorithms were compared on bit codes of lengths 64, 128, and 256. 4 and 16 hash tables were generated for every length of bit code. Every dataset is divided into 5 folds. The test split of a fold was used for performance evaluation, and the train split was used to create the database. In every test split of a fold, 10 nearest neighbors were retrieved for a query sample. The quality of Nearest neighbors was evaluated by computing the mapping percentage of the query sample cell type with the nearest neighbors' cell type. All the test splits across folds were concatenated, and final performance was reported.

\begin{figure}
	\centering
	\makebox[1 \textwidth][c]{
		\resizebox{1.1 \linewidth}{!}{
			\includegraphics[width=\linewidth,keepaspectratio]{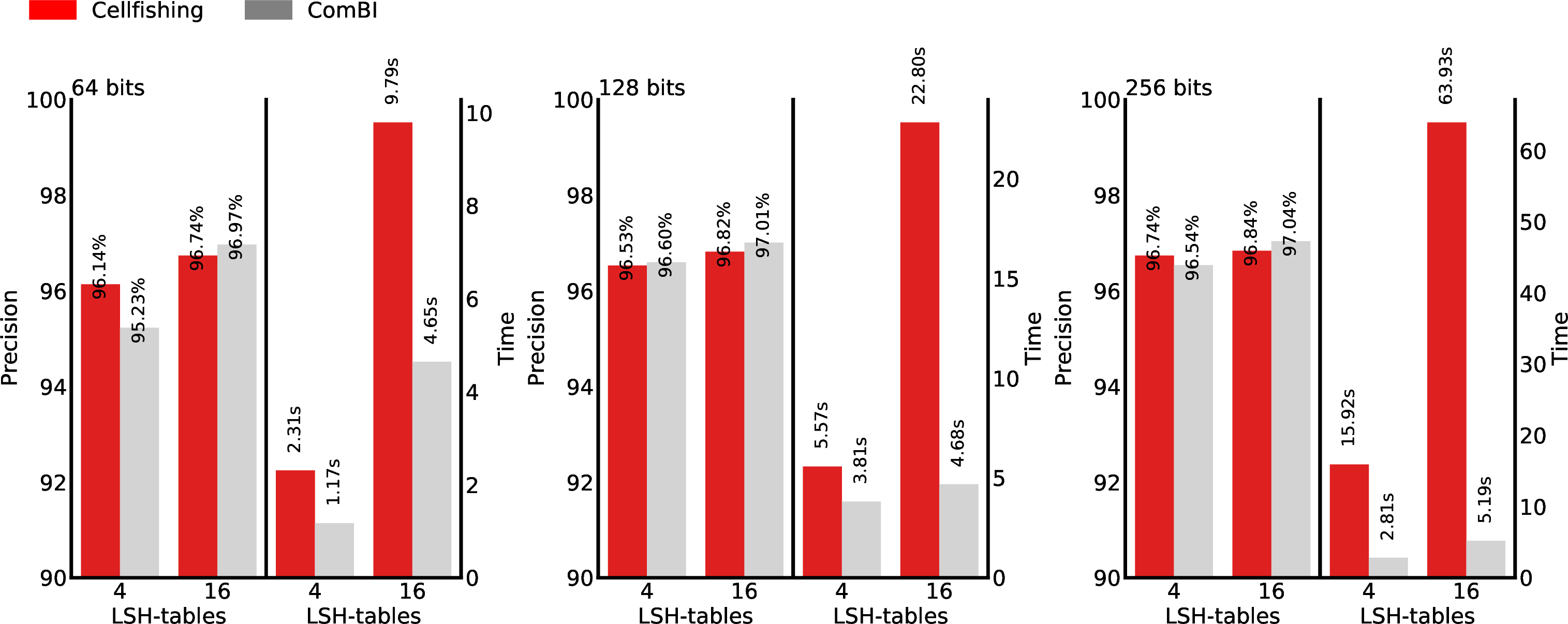}
		}
	}
	\caption{\textbf{Performance of \acrshort{combi} and Cellfishing.jl on shekhar2016 dataset}. The dataset hash 27499 samples. With increasing length of bit code and higher number of tables \acrshort{combi} has better performance. For 64, 128, and 256 bits \acrshort{combi} has $\sim$2, $\sim$1.4, and $\sim$5.6 times speed-up in search time with 4 tables, respectively. Similarly, \acrshort{combi} has speed-up of $\sim$2, $\sim$5, and $\sim$12.3 for 64, 128, and 256 bits with 16 tables, respectively.}
	\label{fig:ch2:shekhar}
\end{figure}

\subsubsection{Results}
Figure~\ref{fig:ch2:baron}, Figure~\ref{fig:ch2:plass}, and Figure~\ref{fig:ch2:shekhar} shows the performance of both the methods on baron2016~\cite{baron2016single}, plass2018~~\cite{baron2016single}, and shekhar2016~\cite{shekhar2016comprehensive} datasets, respectively. In terms of speed \acrshort{combi} outperforms Cellfishing.jl~\cite{sato2019cellfishing} significantly for more tables while their matching performance is almost the same. The difference between the speed becomes significant with larger bit codes. This experiment reinforces that the tree structure of \acrshort{combi} helps find the bit codes of nearest neighbors efficiently.

\section{Discussion}\label{sec:ch2:discussion}

Implementation of \acrshort{ibst} (Algorithm~\ref{alg:ch2:IBSTConstruct}), though straightforward, is not practical in many scenarios due to high memory usage. The Nearest neighbors' search in \acrshort{ibst} is effectively a linear scan of inverted-hash-table while matching prefixes. When a bit code length is large, the filled regions represent a tiny fraction of all possible bit codes. It increases the hamming distance between the nearest neighbors. As a result, \acrshort{ibst} will require to match longer prefixes before determining \textit{hit} or \textit{miss}. In the absence of hardware support for prefix matching, a linear scan of inverted-hash-table yields faster searches.

\acrshort{combi} typically features fewer nodes relative to \acrshort{ibst}. Hence, the memory requirement is not very high. From a tree point of view, compressing an \acrshort{ibst} is equivalent to removing all nodes with only one child. Now, \acrshort{combi} can be searched for nearest neighbors using recursive traversal of \acrshort{bst} (Algorithm~\ref{alg:ch2:IBSTCompressedSearch}). Search in every branch of \acrshort{combi} will terminate at a leaf, thereby guaranteeing to return some neighbors in all search attempts. These returned neighbors can be ranked by the desired metric to find the nearest neighbors. This compression of \acrshort{ibst} reduces \textit{miss} rate to 0 at the expense of an exact search (Section~\ref{sec:ch2:drawback}).

A linear scan of inverted-hash-table for longer bit codes is still slower for practical use. To overcome this, \acrshort{mih}~\cite{manu::mih} divides the bit code into small fragments. It results in reduced search space and increased collision. However, breaking bit codes into tiny fragments may turn reduced search space into full search space due to increased collision. \acrshort{mih}, therefore, requires an appropriate choice of the number of fragments. \acrshort{mih} also accesses a significant number of database points to find the nearest neighbors. To find 1, 10, and 100 for 256 long bit codes, \acrshort{mih} accesses 5.5\%, 11\%, and 16\% of the database, respectively, for 80M-tiny image data set. Similarly, for the SIFT-1B, it accesses 0.5\%, 1\%, and 1.2\% of the database for 256 bits long bit code. Table~\ref{tab:ch2:prec09} and Table~\ref{tab:ch2:prec095} give a comparison between the percentages of database accessed by both \acrshort{combi} and \acrshort{mih}.

Results reported in Table~\ref{tab:ch2:prec09} and~\ref{tab:ch2:prec095} were generated using a \texttt{c++} implementation of Algorithm~\ref{alg:ch2:IBSTCompressedSearch}. This implementation utilizes backtracking. \acrshort{combi} reaches a precision of 0.90 in significantly less time. It is at least $\sim$4X faster than \acrshort{mih} for bit codes of length 256 for the 80M-tiny image data set. This speedup may be attributed to the fact that \acrshort{combi} searches a small fraction of the database compared to \acrshort{mih}. For 256 bits, only 0.23\%, 0.33\%, and 1.05\% of the database were scanned to find 1, 10, and 100 \acrshort{nns}, respectively. This is $\sim$15 to $\sim$33 times smaller than the size of the database searched by \acrshort{mih}. For 64 bits, the search time of \acrshort{combi} is significantly lesser. \acrshort{combi} returns 1$^{st}$ nearest neighbor in $\sim$296X lesser time. For 10 and 100 \acrshort{nns}, the speedup is in the range of $\sim$29X-$\sim$139X. \acrshort{combi} accesses 0.00012\%, 0.0018\%, and 0.014\% of the database, which is $\sim$22 to $\sim$372 times lower than that of \acrshort{mih}. The superior performance of \acrshort{combi} is also attributable to an incremental increase in search space. Similarly, for the SIFT-1B data set, \acrshort{combi} was able to achieve $\sim$19X speedup in retrieving 1$^{st}$ nearest neighbor for 64 bits. In doing so, ComBI scanned 0.00014\% of the database, which is $\sim$19 times lower than the size of the database scanned by \acrshort{mih}. For 10 and 100 \acrshort{nns}, scanned database sizes are 0.00015\% and 0.00079\%, which is $\sim$52 and $\sim$26 times smaller, respectively, than that of \acrshort{mih}.  For longer bit codes, \acrshort{combi} is at least $\sim$3X faster in achieving the precision 0.90. Table~\ref{tab:ch2:prec09} and Table~\ref{tab:ch2:prec095} gives the percentage size of search database and ratio of access size of \acrshort{mih} and \acrshort{combi}. 

In comparison to \acrshort{mih}, \acrshort{combi} has a higher memory footprint. \acrshort{combi} is required to store multiple trees along with bit codes. However, it is a matter of trade-off between space and speed. If a specific application needs to run on a memory constraint platform, but the delayed response is acceptable, \acrshort{mih} can be used. In speed-sensitive applications, \acrshort{combi} can be a method of choice.

\section{Conclusion}

The chapter started with the motivation for arranging the bit codes generated from space-partitioning-based hashing algorithms as a binary search tree. The resulting tree was called \acrfull{ibst}. The \acrshort{ibst}s had many issues such as increased \textit{miss}-rates, higher memory footprint, and heavy dependency on the radius of the search. The chapter presented a geometrically motivated heuristic, \acrfull{combi}, to mitigate the issues. The chapter also presented an extensive empirical evaluation of \acrshort{combi} to establish its superiority. The chapter ended with a use-case of \acrshort{combi} as a single cell search engine.

Knowledge of the neighbourhood of a sample allows us to estimate the class of an unknown sample by examining its vicinity. This motivates the use of ideas behind \acrshort{combi} to solve classification tasks. We expand this idea in the next chapter and discuss novel ways for the same.

%% file: ch3hashingClassifier.tex
\section{Introduction}
Traditionally, similarity preserving hashing (\acrshort{sph}) has been utilized to perform nearest neighbor searches. \acrshort{sph} works by grouping samples in close vicinity. To perform the grouping, \acrshort{sph} assigns a hash code to every sample, which is generated by approximating some distance measure between two samples. These groups are then used to retrieve the nearest neighbors. In the presence of target labels, these groups can be used to assign class labels to the query samples. $k$-nearest neighbor classifier is the closest relative to \acrshort{sph} to perform classification by utilizing similar concepts. 

The alternative view of "grouping the samples" is "partitioning of the space". Tree-based classification algorithms such as \acrfull{dt}, \acrfull{rf}, \acrfull{et}, \acrfull{ot}, etc. performs the guided or supervised partitioning of space using the label information. \acrshort{sph}s are predominantly unsupervised~\cite{manu::lsh,charikar2002similarity,semantic,manu::sh,manu::sperical,manu::shl,coveringlsh,fastcoveringlsh,gog2016fast}, however, there have been some attempts to build semi-supervised and supervised hashing algorithms~\cite{manu::ksh,manu::ssh,jiang2018deep,zhang2014supervised,jiang2018asymmetric}. Some of these algorithms perform hashing in the original space, such as projection hash~\cite{charikar2002similarity}, semantic hash~\cite{semantic}, spectral hash~\cite{manu::sh} etc. On the other hand, some of these algorithms first project the data into higher dimensional space and then perform hashing such as kernel hash~\cite{manu::ksh}, deep learning-based hashing algorithms~\cite{jiang2018deep,jiang2018asymmetric} etc. However, in both cases notion of space partitioning remains intact. Each region in the partitioned space is treated as a bin, and samples in that region are grouped together. Then the strategy followed by tree-based classifiers to assign class labels to each sample can be adopted to build a hashing classifier.

In this chapter, we follow the above notion and formalize the idea of building a hashing based classifier. To discuss the approach, the focus is on the hashing algorithms that assign hash codes in the original space. Projection hash~\cite{charikar2002similarity}, sketching~\cite{lv2006ferret,wang2007sizing}, and binary hash~\cite{manu::lsh} are used for case studies because these are extremely fast hashing algorithms and have been widely used. Pros, cons, and remedies for these classifiers are also discussed.

\section{Hashing classifier}\label{sec:ch3:hcmethod}
Let $\mathbb{R}^n$ denote the n-dimensional Euclidean space. Let $X \subseteq \mathbb{R}^n$ denote the input space, and let $Y$ denote the labels corresponding to a set of $C$ classes $\{1,..,C\}$. Let a set $S$ contain $N$ samples drawn from a population characterized by a probability distribution function $D$ over $X \times Y$. Further, assume that number of samples belonging to a class $c$ is given by $N_{c}$. Thus the given dataset is

\begin{align}\label{eq:ch3:sampleSpace}
S = \{(x^{(i)},y_i): x^{(i)}\in X, y_i \in Y, (i=1,2,..,N)\}.
\end{align}

Let a set $H$ represent a family of \textbf{similarity preserving} hash functions. Say, $h \in H$ is a function which generates a hash code $h(x) = code_{x}$ for sample $x$. Assume a set $\Omega_{p} \subseteq S$ such that for any two samples in the set their hash codes are same, i.e., $\forall a_{1}, a_{2} \in \Omega_{p}, code_{a1} = code_{a2}$. Further, any two proper subsets of $S$ are disjoint, i.e., $(\forall \Omega_{1}, \Omega_{2} \subset S)\Omega_{1}\neq \Omega_{2} \implies \Omega_{1} \cap \Omega_{2} = \phi$, and size of set $\Omega_{p}$ is $|\Omega_{p}| = n_{p}$.

The frequentist probability for a sample to fall into a set $\Omega_{p}$ is given by
\begin{align}
    p(\Omega_{p}) = \frac{n_{p}}{N}.
\end{align}

Further, assuming $n_{p} = \sum_{c\in C}n_{p_{c}}$ where $n_{p_{c}}$ is count of samples belonging to class $c$ that are present in set $\Omega_{p}$. Then, the probability of a sample from class $c$ to be included in set $\Omega_{p}$ is given by
\begin{align}
    p(\Omega_{p}|c) = \frac{n_{p_{c}}}{N_{c}},
\end{align}

if $p(c) = \frac{N_{c}}{N}$, then
\begin{align}\label{eq:ch3:bayesClassifier}
    p(c|\Omega_{p}) = \frac{p(\Omega_{p}|c)p(c)}{p(\Omega_{p})}
    =\frac{n_{p_{c}}}{n_{p}}.
\end{align}

Equation~(\ref{eq:ch3:bayesClassifier}) suggests that a sample falling into set $\Omega_{p}$ can be classified by the local estimate of the probabilities inside the set. However, empirical evaluation and geometric motivation suggest that this formulation works poorly in the case of imbalance. To handle imbalance, these values can be normalized with class proportions.

\begin{align}\label{eq:ch3:bayesClassifierImbalance}
    p(c|\Omega_{p}) = \frac{n_{p_{c}}\Big(\prod_{j\in C\setminus\{c\}}N_{j}\Big)}{\sum_{j\in C}n_{p_{j}}\Big(\prod_{k\in C\setminus\{j\}}N_{k}\Big)}.
\end{align}

To build a hashing classifier, an ensemble of such classifiers is created.

\subsection{General idea}

\begin{figure}[!ht]
	\centering
	\makebox[1 \textwidth][c]{
		\resizebox{1 \linewidth}{!}{
			\includegraphics[width=\linewidth,height=6cm,keepaspectratio]{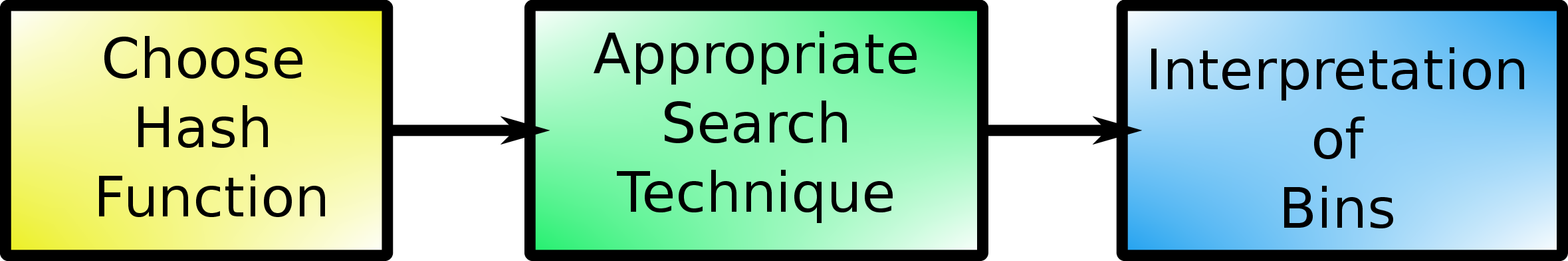}
		}
	}
	\caption{\textbf{Components of a hashing classifier}. In summary, first, select a hashing function, then select the search technique, and finally identify the appropriate method to interpret the bins to perform the task.}
	\label{fig:ch3:hcm}
\end{figure}

In general, a hashing classifier has three components (Figure.~\ref{fig:ch3:hcm}); 

\begin{enumerate}
    \item A similarity preserving hash
    \item An algorithm which retrieves samples in neighborhood of a query samples. Say $\mathcal{A}$.
    \item An interpreter for retrieved samples to classify query samples. Say $\mathcal{B}$.
\end{enumerate}

Let an ensemble of L hashing classifier are created, $h_{l} \in H \forall l \in \{1, ..., L\}$. Further assume for a given test sample $x$, its hash code $h_{l}(x)$ is given by $code_{x}^{(l)}$. For a given hashing classifier $l$ assume a set $\Phi_{l}$ containing hash codes of all samples in set $S$, i.e., $\Phi_{l} = \{code^{(l)}_{t}|\forall t \in S\}$.

Further assume, a retrieval algorithm $\mathcal{A}$ which returns a set $R_{code^{(l)}_{x}}$ of all the samples in $r$-neighborhood for sample $x$ , i.e.

\[R_{code^{(l)}_{x}} = \mathcal{A}(code^{(l)}_{x}, \Phi_{l}, r) = \{code^{(l)}_{t}|t \in S \land t \in r\text{-neigh}(code^{(l)}_{x}, \Phi_{l})\},\]

\noindent
and count of returned sample is $k = |R_{code^{(l)}_{x}}|$.

To compute the probability of class assignment, say, an algorithm $\mathcal{B}$ is used, which estimates the class-wise probability on $r$-neigh. Then, the class label of a test sample, $\hat{c}$ is given by

\begin{align}\label{eq:ch3:predict}
\hat{c} = \arg\max_{c\in C}\Big(\frac{1}{L} \sum_{i=1}^{L}\log(1 + \mathcal{B}(R_{code^{(l)}_{x}}, c))\Big), \text{where } c\in C.
\end{align}

\subsubsection{Choice of $\mathcal{A}$}\label{sec:ch3:choiceA}

    The algorithm $\mathcal{A}$ is utilized to retrieve the nearest neighbors of a query sample. Its choice depends on the hashing algorithms used (Figure~\ref{fig:ch3:hcm}). The major factors which drive the choice of $\mathcal{A}$ are:

	\begin{enumerate}
		\item The index structure is a major factor in deciding the choice of $\mathcal{A}$. It decides the data structure used to store the indexes for faster retrieval. In turn, the data structure decides the appropriate search algorithm for faster nearest-neighbor retrieval. If the output of the hashing algorithm lies on a number/straight line (Section~\ref{sec:ch3:projhash}), AVL Trees, Red-Black Trees, etc. can be employed to store the indexes. If output is in the form of bit code, specific tree-based structures like \acrshort{combi} or dictionary like structure \acrshort{mih} etc. can be used.
		\item Speed of the \acrfull{nn} search is also an important deciding factor. Depending on the underlying hardware performance, the retrieval algorithm may change. If the retrieval algorithm supposes to run on a hard disk, the tree-based index arrangement will have faster search performance. For smaller datasets where index structure can fit in memory, dictionary-based searches will result in faster performance (Section~\ref{sec:ch2:discussion}).
		\item Another factor for consideration while choosing the $\mathcal{A}$ is the precision of \acrlong{nn} search. Exact \acrshort{nn}-search algorithms have 100\% precision for \acrshort{nn}-search. Still, an approximate \acrlong{nn} search algorithm may be more desirable for classification if it has a lower search time and the majority of the returned samples are from the query/test sample neighborhood. 
		\item It is possible that the hash value assigned to the query sample was not generated for any sample in the training data. In such cases, the $\mathcal{A}$ should return with some neighbors, i.e., its miss-rate should be 0\footnote{The miss-rate is defined as an average number of attempts when a search algorithm fails to return any neighbor. These neighbors need not be the \acrfull{nns}}.    
	\end{enumerate}

\subsubsection{Choice of $\mathcal{B}$}\label{sec:ch3:choiceB}

    The algorithm $\mathcal{B}$ describes how to set the parameters and interpret the result of algorithm $\mathcal{A}$ (Section~\ref{sec:ch3:choiceA}). The output of these algorithms are the probabilities by which a sample belongs to different classes. Assuming that the miss-rate of $\mathcal{A}$ is zero, we discuss two ways of defining $\mathcal{B}$ below.

    \begin{itemize}
    \item \textbf{$1$-neigh - Weighted decay}: Assuming that $\mathcal{A}$ is configured to return a set with exact match $\Omega_{p}$, i.e., the hash code of query sample $x$ is the same as that of at least one of the training samples ($code^{(l)}_x \in \Phi_{l}$). Then, we can assume that $x$ is very similar to the elements in $\Omega_{p}$, i.e. $x \sim \Omega_{p}$. Thus the returned probabilities are given by
        \begin{align}
            \mathcal{B}(R_{code^{(l)}_{x}}, c) = p(c|\Omega_{p}),
        \end{align}
    
    where $\Omega_{p}$ is partition of space associated with $code^{(l)}_{x}$. To handle the \textit{miss} condition, i.e., query sample hash code is absent from $\Phi_{l}$, i.e. $code^{(l)}_x \notin \Phi_{l}$, $\mathcal{A}$ returns neighboring sets that meets a pre-defined criteria. To elaborate, assume that the hash codes are defined so that they have \textit{total ordering}. Say, the hash code of the query sample bifurcates the ordering. Say, $mi$ represents the maximum element in the left set, i.e.,

	\[ mi = \arg\max_{ code^{(l)}_{t} \in \Phi_{l}}\{dist(code^{(l)}_{t}, code^{(l)}_{x})|dist(code^{(l)}_{t}, code^{(l)}_{x}) < 0\} \] 
	
	and $ma$ represents the minimum element in the right set, i.e.,  
	
	\[ma = \arg\min_{code^{(l)}_{t} \in \Phi_{l}}\{dist(code^{(l)}_{t}, code^{(l)}_{x})|dist(code^{(l)}_{t}, code^{(l)}_{x}) > 0\}\], 
	
	Then, $\mathcal{A}$ returns a set of neighbors as  $R_{code^{(l)}_{x}} = \{mi, ma\}$.
	Assuming two sets $\Omega^{(l)}_{j}$ and $\Omega^{(l)}_{k}$, if $mi\in\Omega^{(l)}_{j}$ and $ma\in\Omega^{(l)}_{k}$, Then $\mathcal{B}(R_{code^{(l)}_{x}}, c)$ is given by the linear \textbf{weighted sum} of the probabilities associated with the sets $\Omega^{(l)}_{j}$ and $\Omega^{(l)}_{k}$ (\ref{eq:ch3:1neighweightedsum}).
	
	\begin{align}\label{eq:ch3:1neighweightedsum}
	    \mathcal{B}(R_{code^{(l)}_{x}}, c) = \frac{|dist(mi, code^{(l)}_{x})|p(c|\Omega^{(l)}_{k})+|dist(ma, code^{(l)}_{x})|p(c|\Omega^{(l)}_{j})}{|dist(mi, code^{(l)}_{x})| + |dist(ma, code^{(l)}_{x})|}
	\end{align}

    \item \textbf{$k$-neigh $\forall k \geq 1$ - Exponential decay}: Assuming that, $\mathcal{A}$ is configured to return $k$-\acrshort{nns}, here \acrshort{nns} are the samples from train dataset\footnote{Other possibility in this approach is to use retrieve $k$ nearest sets in place of $k$ nearest samples.}. To compute the class probabilities of query samples, the returned probabilities are exponentially weighted by the distance of the query hash code and nearest neighbor hash code. Here $D$ denotes the diameter of $\Phi_{l}$
    
        \begin{align}
            \mathcal{B}(R_{code^{(l)}_{x}}, c) = \sum_{r\in R_{code^{(l)}_{x}}} p(c|\Omega^{(l)}_{r}) \exp \Bigg(-\lambda \Big(\frac{dist(r, code^{(l)}_{x})}{D}\Big)^{2} \Bigg)
        \end{align}

\noindent
where, $\Omega^{(l)}_{r}$ is partition associated with $r\in R_{code^{(l)}_{x}}$

\end{itemize}

\subsection{Some sample hashing classifier}

This section describes three basic hashing algorithms and the choices of $\mathcal{A}$ and $\mathcal{B}$ to use them as classifiers. 

\begin{figure}[!ht]
	\centering
	\makebox[1 \textwidth][c]{
		\resizebox{1 \linewidth}{!}{
			\includegraphics[width=\linewidth,height=6cm,keepaspectratio]{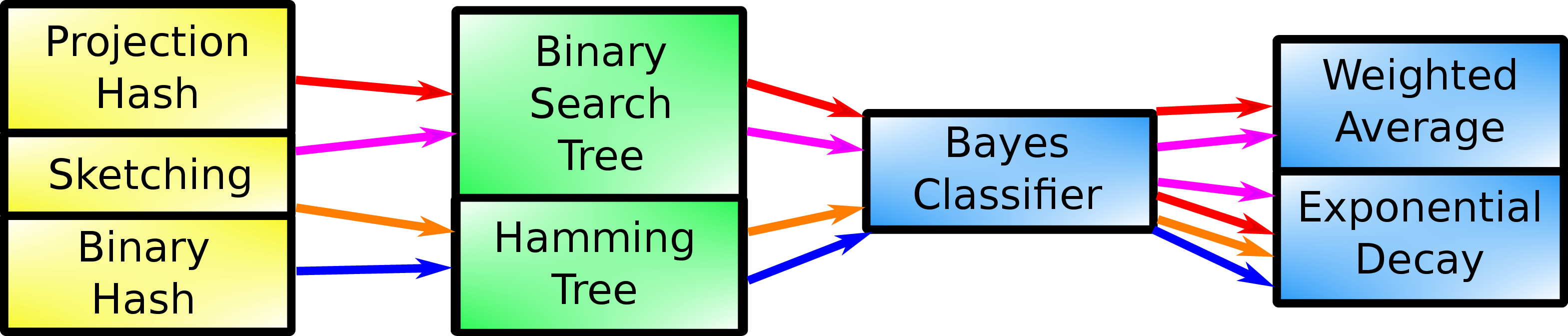}
		}
	}
	\caption{\textbf{Example hashing classifiers.} An illustration of components to build hashing classifiers for three possible hashing methods, namely projection-based hashing, sketching, and binary hashing. The projection-based hashing can be converted into a classifier by choosing \acrshort{bst} as the search technique. The bins can be interpreted with weighted average or exponential decay. Similarly, a binary search tree or hamming tree can be employed to convert the sketching technique into a classification. The bins from the sketching technique can be interpreted with the weighted average of exponential decay. The binary hash can be associated with a hamming tree for nearest neighbor search, and exponential decay can be employed to interpret bins. }
	\label{fig:ch3:hcm1}
\end{figure}

\subsubsection{Sketching-based classifiers}\label{sec:ch3:sketching}

\noindent 
\textbf{Description of hash function}

Assume, for an estimator $l \in \{1,..,L\}$, a weight vector $w^{(l)} \in \mathbb{I^{+}}^{(n)}$ where every feature of the weight vector was sampled from an discrete uniform distribution $U(0, \infty)$ of non-negative numbers.

\begin{align}\label{eq:ch3:sketchweight}
    w^{(l)}_{j} \in \mathbb{I^{+}} \cup \{0\} \land w^{(l)}_{j} \sim U(0, \infty)  \:\: \forall j \in \{1,..,n\}
\end{align}

\noindent
Assuming $m_{j}$ and $M_{j}$ represent the minimum and maximum value of a feature $j$ in the train dataset, then a threshold value is sampled uniformly from a distribution $U(m_{j}, M_{j})$

\begin{align}\label{eq:ch3:sketchth}
    th^{(l)}_{j} \sim U(m_{j}, M_{j}) \:\: \forall j \in \{1,..,n\}
\end{align}

\noindent
Alternatively, the threshold for a feature can also be sampled from the normal distribution parameterized by mean $\mu_{j}$ and standard deviation $\sigma_{j}$ of the feature.  
\begin{align}
    th^{(l)}_{j} \sim N(\mu_{j}, \sigma_{j}) \:\: \forall j \in \{1,..,n\}
\end{align}

Now to generate the hash code for a sample (train or test/query) $x^{(i)}\in\mathbb{R}^n$, first, its bit code is generated by thresholding every feature with the corresponding chosen threshold in (\ref{eq:ch3:sketchth}).

\begin{align}\label{eq:ch3:featureBinarization}
	(B^{(l)}_{x^{(i)}})_{j} = \mathbbm{1}(x^{(i)}_{j} >= th^{(l)}_{j}) \:\: \forall j \in \{1,..,n\},
\end{align}

\noindent
then the weight vectors (\ref{eq:ch3:sketchweight}) is linearly combined by using the bit code from (\ref{eq:ch3:featureBinarization}) as coefficients. The linear combination is then modulated by a prime number $p$. Thus hash code for sample $x^{(i)}\in\mathbb{R}^n$ generated for the estimator $l$, $code^{(l)}_{x^{(i)}}$, is given by 

\begin{align}\label{eq:ch3:sketching}
	code^{(l)}_{x^{(i)}} = \Bigg(\sum_{j=1}^{j=n}\Big((B^{(l)}_{x^{(i)}})_{j} * w^{(l)}_{j}  \Big)\Bigg) \Mod{p}
\end{align}
where $p \in P$ and $P$ is set of all prime numbers. 

Note that the weights in (\ref{eq:ch3:sketchweight}) are generated from an unbounded distribution; thus, hash codes will be very large numbers without modulation. However, modulation, as a side effect, distorts the neighborhood of samples because numbers with far apart bit code may hash to the same bucket due to the cyclic nature of modulation. To avoid distortion of the neighborhood in hashed space due to wrap-around, we can sample the weight values from a bounded distribution whose upper bound is given by $\lfloor\frac{p}{n}\rfloor$.

\begin{align}\label{eq:ch3:sketchingnowraparound}
    w^{(l)}_{j} \sim U(0, \lfloor\frac{p}{n}\rfloor) \:\: \forall j \in \{1,..,n\}
\end{align}

\noindent
Proof to show that sampling weight vector using (\ref{eq:ch3:sketchingnowraparound}) does not result in the wraparound present in box~\ref{prfbox:ch3:sketchingwraparound}.

\begin{proof}[float=ht,label=prfbox:ch3:sketchingwraparound]{Proof of (\ref{eq:ch3:sketchingnowraparound})}

Assume that all the bits in the bit-vector $(B^{(l)}_{x^{(i)}})_{j}$ is set to 1.
\[(B^{(l)}_{x^{(i)}})_{j} = 1\:\: \forall j \in {1,..,n}\]
and all elements of the weight vector $w^{(l)}$ attains the maximum value from the distribution
\[w^{(l)}_{j} = \sup\{x:\forall x \in X \sim  U(0, \lfloor\frac{p}{n}\rfloor)\} = \frac{p-n}{n}\]
Thus the hash vector for sample $x^{(i)}$ is given by
\begin{align*}
    code^{(l)}_{x^{(i)}} = \sum_{j=1}^{j=n}\Big((B^{(l)}_{x^{(i)}})_{j} * w^{(l)}_{j}\Big) = \frac{p-n}{n} \sum_{j=1}^{j=n}B^{(l)}_{x^{(i)}})_{j} = \frac{p-n}{n} \sum_{j=1}^{j=n}1 = p-n
\end{align*}

This is the maximum possible value of $code^{(l)}_{x^{(i)}}$.

since $p-n < p$, thus
\[ p - n \equiv (p - n) \Mod{p}\]

Thus, there will be no wraparound.

\end{proof}

\bigbreak
\noindent
\textbf{Choice of $\mathcal{A}$:}
Based on the construction of the sketch-based hashing, there are two ways to select the algorithm $\mathcal{A}$.
\begin{itemize}

    \item \textbf{Feature binarization:} As an intermediate step of hash code generation \ref{eq:ch3:featureBinarization} binarizes all features. Thus, all the features are first mapped to a \textbf{hamming space}. Search techniques developed for binary features can be employed in the search for similar codes. To make sure that the search is successful for every query sample, we can choose exact search algorithms like \acrfull{mih}~\cite{manu::mih} or approximate search algorithms like \acrfull{combi}~\cite{mycombi}. 
    
    \item \textbf{Linear combination of binarized features:}
    The sketching algorithm combines all the binarized features (\ref{eq:ch3:sketching}), thus mapping hamming space onto a \textbf{straight line}. Binary search tree-based data structures (AVL Tree, Red-black Tree) can be used to store and search the nearest neighbors. If a query sample generated hash code does not exist in the tree, the predecessor and successor of the hash code can be used to approximate the search.
    
\end{itemize}
\bigbreak
\noindent
\textbf{Choice of $\mathcal{B}$:} It is possible to use both the choices of $\mathcal{B}$ discussed in Section~\ref{sec:ch3:choiceB}. With \textbf{feature binarization}, a better choice would be to use \textbf{exponential decay} since, in the case of unseen query hash, all the possible samples up to a certain \acrfull{hd} will be returned. With \textbf{linear combination} both \textbf{weighted decay} and \textbf{exponential decay} can be used. However, for search efficiency, returning the immediate predecessor and successor would be more desirable, thus making \textbf{weighted decay} a default choice for \textbf{linear combination}.

\bigbreak
\noindent
\textbf{Pros, cons and remedies:} In the practical implementation of sketching features, spaces are sub-spaced or super-spaced, depending upon the properties of the data. An estimate of the number of features is important since it decides the total number of possible bins. The modulo operation controls the number of bins only if we create the linear combination of bit codes. As a general rule of thumb, the number of generated bins should be large enough to learn the neighborhood density properly but not too sensitive to over-fit on the noise, and each sample gets its separate bin. The following discusses some of the pros and cons of sketching techniques and their possible remedies.

\begin{itemize}
    \item Pros:
    \begin{itemize}
    	\item For the moderate number of features, feature binarization makes hashing insensitive for noise.
    	\item Modulo of the number generated from the linear combination of bit-vector by prime number limits the number of possible bins.
    \end{itemize}
    \item Cons:
    \begin{itemize}
    	\item Binarization of features limits total number of possible bins to $2^n$. This is unwanted in dataset with high variablility and small number of features.
    	\item For very high dimensional data, feature bit-vector becomes too large. It accentuate the problem of unseen hash code of query samples.
    \end{itemize}
    \item Remedies:
    \begin{itemize}
    	\item For low dimensional data super-spacing can be utilized to increase the number of bins.
    	\item For high dimensional data sub-spacing can be utilized to decrease the number of bins.
        \item Decoupling the number of bins from the dimension size.
    	\item Hash functions where number of bins can be controlled explicitly.
        \item Hash function which can incrementally create the newer bins as needed. One example would be hierarchical hashing.
    \end{itemize}
\end{itemize}

\subsubsection{Projection hash-based classifiers}\label{sec:ch3:projhash}

\noindent 
\textbf{Description of hash function}

Assume, for an estimator $l \in \{1,..,L\}$ and a given \texttt{bin-width}, a weight vector $w^{(l)} \in \mathbb{R}^{(n)}$ where every feature of the weight vector was sampled from uniform distribution $U(m_{j}, M_{j})$. Here, $m_{j}$ and $M_{j}$ are the minimum and maximum values of the feature $j$, respectively.

\begin{align}
    w^{(l)}_{j} \sim U(m_{j}, M_{j})  \:\: \forall j \in \{1,..,n\}
\end{align}

\noindent
The bias value for estimator $l$ is sampled from the uniform distribution $U[-\texttt{bin-width}, \texttt{bin-width}]$.

\begin{align}
    b_{l} \sim U[-\texttt{bin-width}, \texttt{bin-width}]
\end{align}

\noindent
Then hash code, $code^{(l)}_{x^{(i)}}$, of sample $x^{(i)}\in\mathbb{R}^n$ for estimator $l$ is given by

\begin{align}\label{eq:ch3:projectionHash}
	code^{(l)}_{x^{(i)}} = \left\lfloor \frac{b_{l} + \sum_{j=1}^{j=n}\Big(x^{(i)}_{j} * w^{(l)}_{j}  \Big)}{\texttt{bin-width}} \right\rfloor.
\end{align}

Alternatively, weight vectors can also be sampled from the normal distribution $N(\mu_{j}, \sigma_{j})$ where $\mu_{j}$ is mean value of the feature $j$ and $\sigma_{j}$ is standard deviation of the feature $j$.
\begin{align}
    w^{(l)}_{j} \sim N(\mu_{j}, \sigma_{j}) \:\: \forall j \in \{1,..,n\}
\end{align}

\bigbreak
\noindent
\textbf{Choice of $\mathcal{A}$ and $\mathcal{B}$:} 

Projection hash projects all the samples on a number line. Hence, there is a total ordering in the hash codes. Thus, binary search tree-based arrangement for hash codes will be a choice for $\mathcal{A}$. We can employ an AVL or a Red-black tree for the nearest neighbor search. In case the query sample is assigned an unseen hash code, the predecessor and successor of the hash code can approximate the class-wise probability distribution for the query sample. Thus, the choice of $\mathcal{B}$ would be \textbf{weighted decay}.

\bigbreak
\noindent
\textbf{Pros, cons and remedies:} The parameter \texttt{bin-width} is a double-edged sword. Compared with sketching-based classifiers~\ref{sec:ch3:sketching}, where the number of bins is dependent on the number of features used to hashed a sample, \texttt{bin-width} provides more control over the number of bins by defining the quantization size of the projected numbers. However, choosing a proper quantization size is not easy.

\begin{itemize}
    \item Pros:
    \begin{itemize}
    	\item bin-width provides explicit control over number of bins, hence can scale to arbitrary variability in data.
    \end{itemize}
    \item Cons:
    \begin{itemize}
    	\item Choosing \texttt{bin-width} value too low increases the sensitivity of hashing by a significant amount; thus, it becomes very easy to overfit the training data. This also renders the algorithm prone to noise.
    	\item Size of \texttt{bin-width} does not follow a linear trend with the accuracy of the system. Hence, it is not easy to calibrate the appropriate value for \texttt{bin-width}.
    	\item For a small value of \texttt{bin-width}, there is also a high probability for a query sample to get an unseen hash code.
    	\item Using a large value of \texttt{bin-width} will result in very few bins thus, it becomes difficult for the algorithm to learn the data distribution.
    	\item Quantization of \texttt{bin-width} creates fix length bins that may not always be able to learn the full extent of data distribution. For example, a too big bin size may place samples from different classes into the same bin, and a too small bin size may overfit to noise.
    \end{itemize}
    \item Remedies:
    \begin{itemize}
    	\item Use fix number of bins to control noise.
    	\item Remove dependency on bin-width and create a mechanism for dynamic bin size. Hierarchical hashing/binning may solve this problem.
    \end{itemize}
\end{itemize}

\subsubsection{Binary hashing-based classifiers}

\noindent 
\textbf{Description of hash function}

Assume, for an estimator $l \in \{1,..,L\}$, a weight vector $w^{(l)}_{j} \in \mathbb{R}^{(n)}$ where every feature of weight vector was sampled from uniform distribution $U(m_{j}, M_{j})$. Here, $m_{j}$ and $M_{j}$ are minimum and maximum values of the feature $j$, respectively.

\begin{align}
	w^{(l)}_{j} \sim U(m_{j}, M_{j})  \:\: \forall j \in \{1,..,n\}
\end{align}

All the samples in the train datasets are first projected on the weight vector. Assuming a train sample is given by $x^{(i)}$ then its projection $proj^{(l)}_{x^{(i)}}$ on the weight vector $w^{(l)}_{j}$ of estimator $l$ is given by

\begin{align}
	proj^{(l)}_{x^{(i)}} = \sum_{j=1}^{j=N} x^{(i)}_{j} * w^{(l)}_{j} \:\: \forall i \in \{1..N\}
\end{align}

\noindent
Now the origin is moved to the mean of the projected numbers. 

\begin{align}
    \bar{proj}^{(l)}_{x^{(i)}} = proj^{(l)}_{x^{(i)}} - \frac{1}{N}\sum_{k=1}^{k=N}proj^{(l)}_{x^{(k)}}
\end{align}

\noindent
Now every sample is assigned a bit, if the resulting number is positive it is given the bit $1$ otherwise $0$.

\begin{align}
	code^{(l)}_{x^{(i)}} = \mathbbm{1}\Big( \bar{proj}^{(l)}_{x^{(i)}} ) >= 0  \Big)
\end{align}

This exercise is repeated for every estimator $l$, and their resulting code is concatenated to get the final hash code. Here we assume that all the estimators have been placed in fixed but arbitrary order i.e. $code_{x^{(i)}} = (code_{x^{(i)}})$. Then, for any sample $x^{(i)}$ its code $code_{x^{(i)}} \in \{0, 1\}^{l}, l \in \mathbb{N}$ is assigned as  

\begin{align}\label{eq:ch3:binaryHash(a)}
    code_{x^{(i)}} = \bigcup_{r=1}^{r=l}code^{(r)}_{x^{(i)}} 
\end{align}

Alternatively, the weight vector can also be sampled from normal distribution $N(\mu_{j}, \sigma_{j} \forall j \in \{1..n\}$ where $\mu_{j}$ and $\sigma_{j}$ are mean and standard deviation of feature $j$, respectively. Other choices are standard normal distribution $N(0, 1)$ and uniform distribution $U(-1, 1)$. 
\begin{align}
	w^{(l)}_{j} \sim N(\mu_{j}, \sigma_{j}) \:\: \text{or } w^{(l)}_{j} \sim N(0, 1) \:\: \text{or } w^{(l)}_{j} \sim U(-1, 1) \forall j \in \{1,..,n\}
\end{align}

\bigbreak
\noindent
\textbf{Choice of $\mathcal{A}$ and $\mathcal{B}$:} 
In the binary hash, every sample is represented with a bit code. Thus search techniques in the hamming space can be a choice of $\mathcal{A}$. However, in the binary hash, all the estimators need to be used to assign a hash code to a sample. Since, in case a query sample is assigned an unseen bit code, which is very likely for large values of $l$, getting $k$-nearest neighbor would be more sensible. It will return all the samples at a minimum of $1$-\acrshort{hd} away. Thus \textbf{exponential decay} will be a valid choice for $\mathcal{B}$. 

Since all the estimators $l$ are used to construct the one hash code for a single sample. Thus, multiple trials are needed to increase the stability of the result. Consequently, a small ensemble of binary hash classifiers is created. Across all the elements of ensemble choice of $\mathcal{A}$ and $\mathcal{B}$ remains the same. In the end output of $\mathcal{B}$ is combined by creating a voting classifier.

\bigbreak
\noindent
\textbf{Pros, cons and remedies:} Random generation of planes causes the space to be divided into multiple dynamically shaped regions. Although vanilla binary hash does not give much control over the bin size, it is possible to modify the hyperplane generation technique to gain explicit control of bin size. The following points discuss the method's pros, cons, and possible remedies.

\begin{itemize}
    \item Pros:
    \begin{itemize}
    	\item Less sensitive to noise because of binarization.
    	\item Dynamic bin size.
    \end{itemize}
    \item Cons:
    \begin{itemize}

    	\item For larger bit code, high probability of sample falling into empty bins.
    	
    	\item For the larger bit code, there are many bins that are empty, thus exact Nearest Neighbor search becomes difficult and is very time consuming.

    \end{itemize}
    \item Remedies:
    \begin{itemize}
    	\item Use approximate nearest neighbor search algorithm.
    	\item Use hashing methodology to guide test sample such that it always fall into one of the filled bins.
    \end{itemize}
\end{itemize}

\subsection{Need for a tree arrangement}

\begin{figure}[!ht]
	\centering
	\makebox[1 \textwidth][c]{
		\resizebox{1.1 \linewidth}{!}{
			\includegraphics[width=\linewidth,height=6cm,keepaspectratio]{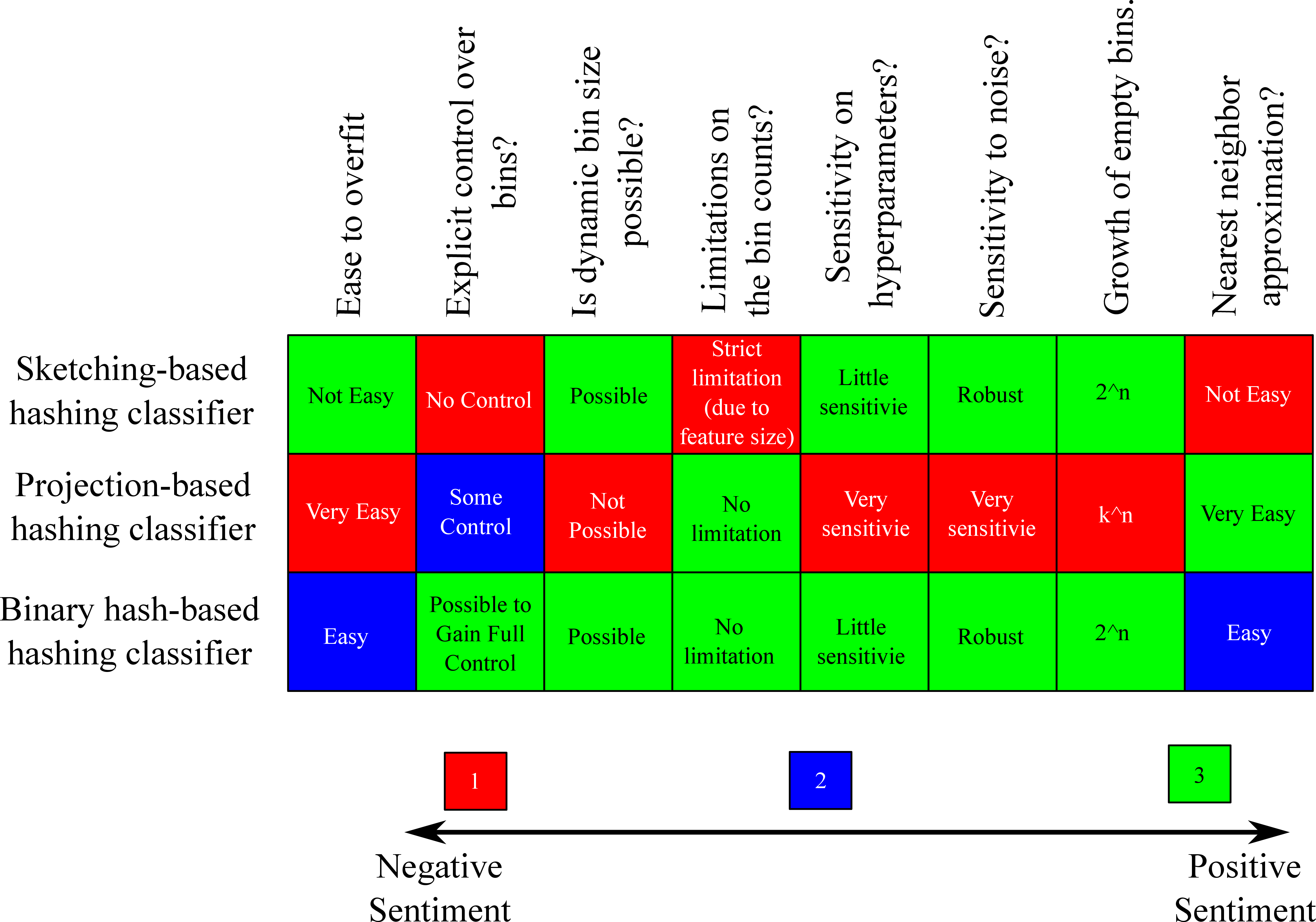}
		}
	}
	\caption{\textbf{Qualitative comparison of hashing classifiers.} Among the three example classifier, the binary hash-based hashing classifier has the most desired properties. Sketching-based hashing classifier is at the second. Projection-based hashing classifier is the most undesirable.}
	\label{fig:ch3:needtree}
\end{figure}

Figure~\ref{fig:ch3:needtree} summarizes the pros and cons of the three probable hashing classifiers discussed in the chapter. In summary, a classifier is desirable if it is not prone to over-fit and data noises. Further, it provides explicit control over the bins and can create dynamic size bins; the growth rate of the empty bins should be lower and there should not be any limit on how many bins can be created. It should not be very sensitive to the hyperparameters, and the nearest neighbor search should be easy and non-time consuming.

Based on these parameters, the projection hash is not suitable for classification. It is extremely prone to over-fitting and does not create dynamic-size bins. The classifier's performance is extremely dependent on the value of \texttt{bin-width} hyperparameter. With a small value of \texttt{bin-width} the projection-based hashing classifier is extremely likely to model noises. On the other hand, sketching-based hashing classifiers are not prone to over-fitting and have very little sensitivity to hyperparameters. The problem with the sketching-based hashing classifiers is related to bin control and bin counts. The total number of bins in this hashing technique is governed by the number of dimensions of the data. One way to increase the possible number of bins is by super-spacing the data. However, including every dimension in super-space will double the number of possible bins in space. Thus increasing the probability of the test/query sample falling into the empty bin, increasing search time.

The binary hash classifier is the most robust in the presence of noise and provides complete control over the bin size and count. Multiple techniques have been developed to perform a search in the hamming space, which can be utilized here to construct the neighborhood of the test sample to assign class labels. The issue with the binary hash-based classifier is its tendency to over-fit since it requires a large number of hyperplanes to be generated to model the neighborhood. Thus increasing the possibility of empty bins and time in the nearest neighbor search. To alleviate this issue, a guided approach is needed to generate the planes so neighborhoods can be learned more effectively by keeping the bit code length in check.

Interestingly, a super-spaced sketching-based hashing classifier can be understood as the flattened version of \acrshort{rf}~\cite{rf} or \acrshort{et}~\cite{geurts2006extremely}. If every selected feature for thresholding in \acrshort{rf} and \acrshort{et} is stored in an array (as hyperplanes), it becomes a sketching hash function. Although, in this arrangement, the local information preserved in the tree structure is lost, and every threshold in the array can be considered to have a global effect. Similarly, the oblique tree hyperplane on every node can be considered as a hash function that only has the local effect. Storing these hyperplanes in the array results in information loss about the planes' locality.

In reverse, the array of hashing planes can be arranged in tree form. Such an arrangement can be constructed by considering the hyperplanes' global effect. Based on the criteria discussed above, the tree arrangement of hashing planes has most of the desired properties of a classification tree:
\begin{enumerate}
    \item Tree-arrangement of planes provides explicit control over bins. The bins can be divided as and when needed, and their size can also be controlled by ingesting class information in the division criteria.
    \item Although tree arrangements are easy to over-fit, their growth can be controlled to increase the generalizability of the classifier. Ensemble of trees can also be used to increase generalizability. These criteria can also be used to handle noises in the data.
    \item As discussed in Chapter 2, the binary tree arrangement of bit codes can be used for nearest neighbor search. Thus, the tree arrangement of hashing planes has the inherent property to retrieve nearest neighbors.
    \item Tree arrangement provides an added benefit. It stops the query samples from falling into the empty bin. Thus every attempt to search for nearest neighbors will return at least one neighbor.
\end{enumerate}

In the next chapter, the idea of building tree-based hashing classifier from the binary hashing will be formalized.

\subsection{Conclusion}

This chapter started with a revision of the definition of the Bayes classifier. Then the method to build classifiers using the hashing techniques is formalized, and a detailed discussion on $\mathcal{A}$-$\mathcal{B}$ formalization of hashing classifier is presented. Then three examples were presented to build the hashing classifier using sketching-based hashing, projection-based hashing, and binary hashing. In the last section of the chapter, a discussion on the desirable properties of hashing classifiers is presented, and a motivation to arrange hashing plane in a tree is discussed. 

It has been shown that an ensemble of trees works best as a general-purpose classifier~\cite{fernandez2014we}. In the next chapter, we extend ideas explored in this chapter with this motivation in mind. In particular, we focus on efficient space partitioning, in which a hyperplane used to split one region can be used in another region as well. This leads to some degree of parsimony, and makes search more efficient.

%% file: ch4graf.tex
\section{Introduction}
In supervised learning, one aims to learn a classifier that generalizes well on unknown samples ~\cite{dietterich2000ensemble}. As commonly understood, a classifier should have an error rate better than a random guess. If a classifier performs slightly better than a coin toss, it is termed a weak classifier. In ensemble learning, several weak classifiers are trained, and during prediction, their decisions are combined to generate a weighted or unweighted (voting) prediction for test samples. The motivation is that the classifiers' errors are uncorrelated; hence, the combined error rate is much lower than individual ones~\cite{rf}.

It has been shown that an ensemble of trees works best as a general-purpose classifier~\cite{fernandez2014we}. Amongst several known methods for constructing ensembles, \textit{Bagging} and \textit{Boosting} are widely used. For every tree, bagging generates a new subset of training examples~\cite{rf}. Boosting assigns higher weights to misclassified samples while building an instance of a tree~\cite{friedman2001greedy,chen2016xgboost}. With either strategy, a tree in an ensemble is constructed by a recursive split of the data into two parts at every node. The split can be axis-aligned, in which the split is based on a feature ~\cite{rf,et}, or oblique, where a combination of features is used~\cite{murthy1993oc1,murthy1994system} for every split.

Axis-aligned trees perform well with redundant features~\cite{menze2011oblique,wickramarachchi2016hhcart}, while oblique splits yield shallower trees~\cite{zhang2017robust}. However, memory and computational requirements are higher for oblique trees. Hence, the literature focuses on finding better splits to create shallower oblique trees. Shallower trees tend to generalize better.

Despite these limitations, oblique trees have been widely used in diverse tasks across various domains. Do \textit{et al.}~\cite{do2015classifying} apply oblique trees to fingerprint dataset classification. Qiu \textit{et al.}~\cite{qiu2017oblique} used them for time-series forecasting, Zhang \textit{et al.}~\cite{zhang2017robust} for visual tracking, and Correia and Schwartz~\cite{correia2016oblique} for pedestrian detection.

In this work, we propose \acrfull{graf}~\cite{gupta2019guided}, which extends the outlook of a plane generated for a certain region to other regions as well. \acrshort{graf} iteratively draws random hyperplanes and corrects each impure region to increase the purity values of resultant regions. Unlike other methods, a hyperplane in \acrshort{graf} is not constrained to the region it is generated for but is shared across all possible regions. Sharing planes across regions reduces the number of separating hyperplanes in trees, reducing the memory requirement. The idea of \acrshort{graf} is inspired by the hierarchical arrangement of hashing planes.

The resultant regions (or leaf nodes) in \acrshort{graf} are represented with variable length codes. This tree construction process bridges the gap between boosting and decision trees, where every tree represents a high variance instance. We show that \acrshort{graf} outperforms state-of-the-art bagging and boosting-based algorithms, like Random Forest~\cite{rf} and Gradient Boosting~\cite{friedman2001greedy}, on several datasets.

\begin{figure*}[!ht]
	\centering
	\makebox[1 \textwidth][c]{
		\resizebox{1 \linewidth}{!}{
			\includegraphics[width=1.1\linewidth,height=2.5in,keepaspectratio]{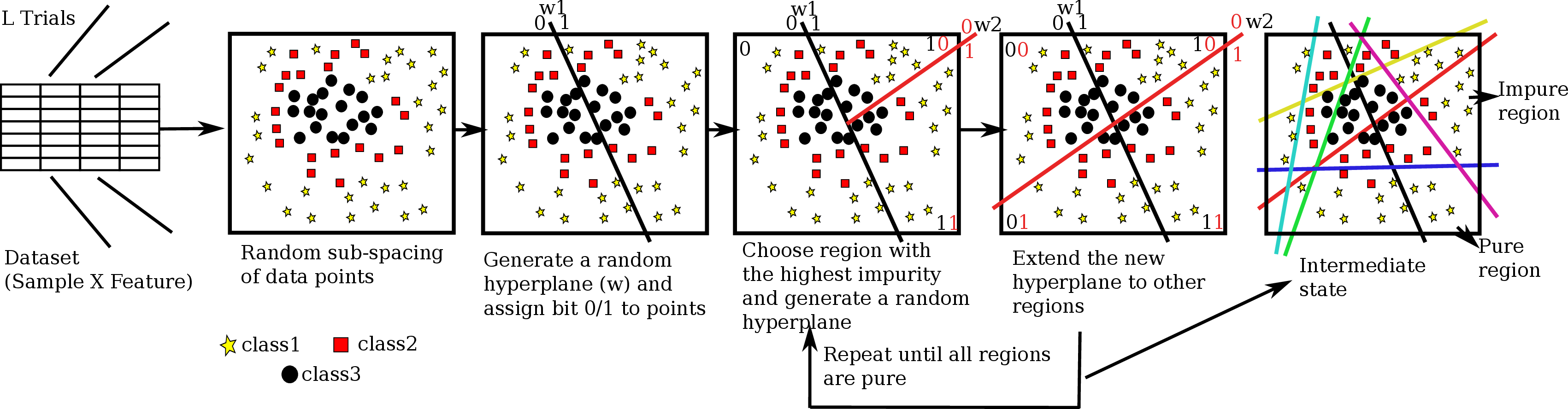}
		}
	}
	\caption{\textbf{An overview of the creation of high variance instances in \acrshort{graf}}. Every instance consists of sub-spacing the dataset in a uniformly sampled feature space. A random hyperplane is generated for the sub-spaced samples. It assigns a bit 0/1 to every sample. A pure (impure) region is a region containing all (some) samples of the same class. Amongst these regions, the most impure region affects the generation of the next hyperplane. This hyperplane is extended to the other region as well, if it improves the purity of subsequent regions in that space. This generation of hyperplanes is continued until all regions are maximally purified. At an intermediate stage, regions are either pure or impure. To increase the confidence of classification, the above process is repeated to create $L$ high variance instances.}
	\label{fig:ch4:ga}
\end{figure*}

\section{Related Work}\label{sec:ch4:relatedWork}

The construction of tree-based classifiers has been an active area of research. Classifiers may differ by the number of trees being generated; single decision tree~\cite{breiman1984classification} \textit{vs.} forest algorithms~\cite{rf}; by the type of splits on nodes of the tree - axis-aligned splits~\cite{rf,geurts2006extremely} \textit{vs.}  oblique splits~\cite{murthy1993oc1,murthy1994system}. Tree-based algorithms may vary in terms of size - fixed size~\cite{bennett1998support} \textit{vs.} top-to-down built, or error correction methodology - misclassification~\cite{martinelli1990pyramidal,deb2002binary} \textit{vs.} residual error correction~\cite{friedman2001greedy,chen2016xgboost}. Amongst all criteria, the type of split at a node has attracted the most attention. Two notable methods for axis-aligned splits are \acrfull{rf}~\cite{rf}, and \acrfull{et}~\cite{geurts2006extremely}. \acrshort{rf} searches for the best split using uniformly spread thresholds in the range of every feature at a node. \acrshort{et} generates random threshold for every feature and the select the threshold which gives the best split.

\acrfull{ots} generate splits that are not aligned with the feature axes. Since \acrshort{ots} consider multiple features simultaneously, the search space increases exponentially. Consequently, an exhaustive search to find the optimal oblique split is impractical. Researchers have used greedy or optimization-based approximations to select the best possible split. Thus, many oblique tree variants have been proposed that vary in terms of how separating hyperplanes are generated to create splits. Murthy \textit{et al.}~\cite{murthy1993oc1,murthy1994system} proposed a \acrfull{oc1}, which refines CART's~\cite{breiman1984classification} strategy of optimal split selection by employing a combination of axis-aligned and oblique splits~\cite{murthy1994system}. Tan and Dowe~\cite{tan2004mml,tan2006decision} suggested selecting an oblique split based on \acrfull{mml}~\cite{wallace1968information} criterion. To induce decision trees, Bennet and Blue~\cite{bennett1998support} introduced a \acrfull{svm} based formulation, called \acrlong{gto} - \acrshort{svm} (\acrshort{gto}/\acrshort{svm}). It uses \textit{\acrlong{hepts}} (\acrshort{hepts})~\cite{blue1998hybrid} to approximately solve a non-convex problem. In later studies, Takahashi and Abe~\cite{takahashi2002decision} proposed a top-to-down approach to learning decision trees with \acrshort{svm}s. In another study, Wang \textit{et al.}~\cite{wang2006improved} proposed alternative criteria to group samples based on the separability of classes. The class with the highest separability is considered one group, and other classes are grouped to generate the split on the node. Manwani and Sastry~\cite{manwani2011geometric} suggest an alternative based on a variant of proximal \acrshort{svm}, \acrfull{gepsvm}~\cite{mangasarian2005multisurface}. Zhang \textit{et al.}~\cite{zhang2014oblique,zhang2017robust} also used \acrfull{mpsvm} to grow decision trees. \textit{Rotation Forests}~\cite{rodriguez2006rotation,kuncheva2007experimental} used principal components of high variance to obtain the direction of split. Menze \textit{et al.}~\cite{menze2011oblique} proposed two models, one with \acrfull{lda} like projections and another with ridge regression to obtain the split. \acrfull{co2} Forest~\cite{norouzi2015co2,norouzi2015efficient} optimizes a objective function based on latent variable \acrlong{svm}~\cite{yu2009learning} to select an oblique split. Katuwal \textit{et al.}~\cite{katuwal2020heterogeneous} suggest selecting the splitting criteria using different kinds of linear classifiers, viz. \acrshort{svm}, \acrshort{mpsvm}, \acrshort{lda}, etc., on every node. This gives heterogeneous nature to the \acrshort{ots}.

In all the methods mentioned above, correction is limited to the region for which the split has been generated for every new split.  To the best of our knowledge, \acrshort{graf} is the first attempt to extend the plane to share it with other nodes explicitly.

The tree-based algorithms have also been used in other areas such as Nearest Neighbor Search~\cite{liu2006new,rptrees}, outlier/anomaly detection~\cite{isolationforest}, matrix imputation~\cite{stekhoven2012missforest} etc. There has also been some attempt to integrate neural networks with trees~\cite{kontschieder2015deep,popov2019neural}. In another effort, Katuwal \textit{et al.}~\cite{katuwal2018ensemble,katuwal2018enhancing} has proposed to combine \acrfull{rvfl} with trees to create an ensemble.

As discussed above, many variants of \acrfull{ots} are different in how they generate the splits. However, most of the split criteria defined above are local in nature. In the vast history of \acrshort{ots}, there has been no attempt to connect hashing with the tree-based classifiers. In the next part of the chapter, \acrshort{graf} is built that generates the global splits. In the later parts, \acrshort{graf} has been put into the context of hashing classifiers and boosting classifiers.

\section{\acrfull{graf}}\label{sec:ch4:GRAF}

Similar to the Section~\ref{sec:ch3:hcmethod}, let $\mathbb{R}^n$ denote the n-dimensional Euclidean space. Let $X \subseteq \mathbb{R}^n$ denote the input space, and let $Y$ denote the labels corresponding to a set of $C$ classes $\{1,..,C\}$. Let a set $S$ contain $N$ samples drawn from a population characterized by a probability distribution function $D$ over $X \times Y$. Thus the given dataset is

\begin{align}\label{eq:ch4:sampleSpace}
S = \{(x^{(i)},y_i): x^{(i)}\in X, y_i \in Y, (i=1,2,..,N)\}.
\end{align}

Let us assume that $T$ high variance classifier instances are constructed on the dataset $S$. The training of an instance involves the introduction of random hyperplanes in a forward stage-wise fashion. At a given step, a combination of these hyperplanes divides $S$ into a finite number (say $P$) of disjoint regions whose union is $S$. To be specific, a single hyperplane classifier will divide $S$ into two disjoint regions (say $\Omega_1$ and $\Omega_2$), and a combination of $d$ hyperplane classifiers will divide $S$ into at most $2^d$ regions. Let the $p$th region ($1\leq p\leq P$) be denoted by $\Omega_p$. Thus, $S=\cup_{p=1}^{P}\Omega_p$ and $\Omega_i\cap\Omega_j=\emptyset \text{ for } i\neq j$. Let $n_p$ denote the number of samples in the region $\Omega_p$. Obviously, $n_p>0$, otherwise $\Omega_p$ will be an empty region and hence, have no contribution.

For each sample in the region $\Omega_p$, we generate a bit '0' or '1' such that the weights $w^{(p)} = (w_1^{(p)},...,w_n^{(p)}) \in \mathbb{R}^n$ and the bias $b^{(p)}\in \mathbb{R}$ dichotomizes the region $\Omega_p$. This is achieved by using a mapping $\lambda_p:X \to \{0, 1\}$ such that for the sample point $(x^{(i)},y_i)$ in $\Omega_p$,

\begin{align}\label{eq:ch4:bitAssignment}
\lambda_{p}(x^{(i)}) = \mathbbm{1}{\left(\sum_{j=1}^{n}(w_{j}^{(p)}x^{(i)}_{j}) + bias^{(p)} > 0\right)}.
\end{align}

Here $\mathbbm{1}(.)$ denotes the indicator function and $x^{(i)}_j$ is the $j^{th}$ component of the vector $x^{(i)}$.

We now introduce the following notations for $j=1,2,..,n$.

\begin{align}\label{eq:ch4:minEquation}
m_{j}^{(p)} = \min_{1\leq i \leq n_{p}}(x^{(i)}_j\: :\: (x^{(i)}, y_i) \in \Omega_p),
\end{align}
\begin{align}\label{eq:ch4:maxEquation}
M_{j}^{(p)} = \max_{1\leq i \leq n_{p}}(x^{(i)}_j\: :\: (x^{(i)}, y_i) \in \Omega_p),
\end{align}
\begin{align}\label{eq:ch4:meanEquation}
\mu_{j}^{(p)} = \frac{1}{n_{p}}(\sum_{i=1}^{n_{p}}x^{(i)}_j,\: (x^{(i)}, y_i) \in \Omega_p),
\end{align}
\begin{align}\label{eq:ch4:weightGeneration}
w_{j}^{(p)} \sim U(m_{j}^{(p)}+\varepsilon, M_{j}^{(p)}-\varepsilon),
\end{align}

where (\ref{eq:ch4:minEquation}), (\ref{eq:ch4:maxEquation}), and (\ref{eq:ch4:meanEquation}) represents the minimum value, maximum value, and mean value of a feature $j$ in the region $p$, respectively. Then we define bias as

\begin{align}\label{eq:ch4:biasP}
bias^{(p)} = -\sum_{j}w_{j}^{(p)}\mu_{j}^{(p)},
\end{align}
where $U(a, b)$ denotes the uniform distribution of a random variable over the interval [a,b].

The mapping $\lambda_p:X \to \{0, 1\}$ as defined at (\ref{eq:ch4:bitAssignment}) above assigns a code comprising of 0s and 1s for every sample in $\Omega_p$. A region $\Omega_p$ is said to be \textbf{pure} if it contains samples of the same class, or if samples from different classes can not be separated further. On the other hand, the region $\Omega_p$ is said to be \textbf{impure} if it contains samples of different classes, that can be further dichotomized by the addition of new hyperplanes (Figure~\ref{fig:ch4:ga}).

Let $\mathcal{F} = \{\Omega_1,\Omega_2,..,\Omega_P\}$. We now introduce a mapping $Z:\mathcal{F} \to \mathbb{R}$ such that for $1\leq p \leq P$,

\begin{align}\label{eq:ch4:labeldef}
Z(\Omega_p) = \left(1 - \sum_{c=1}^{C}(\frac{n_{p_c}}{N_c})^2 \times (\sum_{c=1}^{C}\frac{n_{p_c}}{N_c})^{-2}\right) \times n_{p},
\end{align}

where $N_c$ denotes the total samples of class $c$, and $n_{p_c}$ denotes the samples of class $c$ in region $\Omega_p$.

The function $Z$ as defined at (\ref{eq:ch4:labeldef}) is the weighted Gini impurity function whose value $Z(\Omega_p)$ quantifies the impurity associated with the region $\Omega_p$. Also $Z(S) = \sum_{p=1}^{P}Z(\Omega_p)$ defines the total overall impurity of the space $S$.


We next proceed to discuss the process of hyperplane generation, which is a greedy approach. In this process, we choose the most impure region $\Omega^{*}$ which is obtained as 

\begin{align}\label{eq:ch4:mostImpurePartitions}
\Omega^{*} = \arg\max_{\Omega_p \in \mathcal{F}_1} Z(\Omega_p),\:\text{where}
\end{align}
\begin{align}\label{eq:ch4:impurePartitions}
\mathcal{F}_1 = \{\Omega_p: \Omega_p\in \mathcal{F}, Z(\Omega_p)> 0 \text{ and } \exists j \text{ such that } ((m_{j}^{(p)}\neq M_{j}^{(p)})\}
\end{align}

consists of only impure regions that can be divided.

Let region $\Omega^{*}$ be divided into regions $\Omega^{*}_0$ and $\Omega^{*}_1$, where

\begin{align}\label{eq:ch4:partition1}
\Omega^{*}_0 = \{(x^{(i)}, y_i): \lambda^{*}(x^{(i)}) = 0\: \forall (x^{(i)}, y_i) \in \Omega^{*}\},
\end{align}

and

\begin{align}\label{eq:ch4:partition2}
\Omega^{*}_1 = \{(x^{(i)}, y_i): \lambda^{*}(x^{(i)}) = 1\: \forall (x^{(i)}, y_i) \in \Omega^{*}\}.
\end{align}

In (\ref{eq:ch4:partition1}) and (\ref{eq:ch4:partition2}), the mapping $\lambda^{*}$ is generated as for $\lambda_p$ defined at (\ref{eq:ch4:bitAssignment}). The mapping $\lambda_p$ is defined for all $\Omega_p$, and $\Omega^{*}$ is one of the $\Omega_p$'s from the family of $\mathcal{F}_1$.

The effect of the hyperplane corresponding to $\lambda^{*}$ is extended to other impure regions as well. For the region $\Omega_p \in \mathcal{F}_1 \setminus {\Omega^{*}}$, we define

\begin{align}\label{eq:ch4:remPartition1}
\Omega^{*}_{p_0} = \{(x^{(i)}, y_i): \lambda^{*}(x^{(i)}) = 0,\: (x^{(i)}, y_i) \in \Omega_p\},
\end{align}

and

\begin{align}\label{eq:ch4:remPartition2}
\Omega^{*}_{p_1} = \{(x^{(i)}, y_i): \lambda^{*}(x^{(i)}) = 1,\: (x^{(i)}, y_i) \in \Omega_p\},
\end{align}
so that $\Omega_p = \Omega^{*}_{p_0} \cup \Omega^{*}_{p_1}$ for $\Omega_p\in \mathcal{F}_1$ but $\Omega_p \neq \Omega^{*}$.

Next, $K$ different hyperplanes are generated via the procedure described in  (\ref{eq:ch4:minEquation})-(\ref{eq:ch4:biasP}) for the given region $\Omega^{*}$ as chosen from (\ref{eq:ch4:mostImpurePartitions}). These are denoted by $\langle w^{(k)},x\rangle + b^{(k)} = 0$, $k=1,2,..,K$. For each of these hyperplanes, the steps proposed in  (\ref{eq:ch4:partition1})-(\ref{eq:ch4:partition2}), and (\ref{eq:ch4:remPartition1})-(\ref{eq:ch4:remPartition2}), are performed, and $Z^{(k)}(S)$ is computed for $k=1,2,..,K$. Here $Z^{(k)}(S)$ is the notation used for $Z(S)$ with respect to the $k$th hyperplane $\langle w^{(k)},x\rangle + b^{(k)} = 0$, $k=1,2,..,K$. Let

\begin{align}\label{eq:ch4:minHyperplaneImpurity}
Z^{(l)}(S) = \min_{k=1,2,..,K}(Z^{(k)}(S)).
\end{align}

We choose the hyperplane $\langle w^{(l)},x\rangle + b^{(l)} = 0$ and any tie in (\ref{eq:ch4:minHyperplaneImpurity}) is broken arbitrarily.

We subsequently update the family of impure regions $\mathcal{F}_1$ to take into account new nonempty impure regions. This gives a newly updated family of impure regions. 

The process is repeated until no impure region is left to be further dichotomized.

Once the above process is completed, all pure regions are collected in the family $\Tilde{\mathcal{F}}$. Thus

\begin{align}\label{eq:ch4:allPureFilledPartitions}
\Tilde{\mathcal{F}} = \{\Omega_p :\Omega_p \in \mathcal{F}, Z(\Omega_p) = 0 \text{ or } m_{j}^{(p)}=M_{j}^{(p)} \forall j \in \{1,..,n\} \}.
\end{align}

Every pure region $\Omega_p$ in the family $\Tilde{\mathcal{F}}$ is assigned a code that is shared by every sample in the region. Here we assume that all regions have been placed in an arbitrary but fixed order $\Bar{\mathcal{F}}=(\Tilde{\mathcal{F}})$, then for any sample $(x^{(i)}, y_{i})\in S$, its $code_{x^{(i)}}\in\{0, 1\}^{r}, r \in \mathbb{N}$ is assigned as

\begin{align}\label{eq:ch4:codeAssiginment}
code_{x^{(i)}} = (\lambda^{p}(x^{(i)}):\forall \Omega_p \in \Bar{\mathcal{F}}),
\end{align}

where $r$ is the total number of hyperplanes. These steps are equivalent to hashing and (\ref{eq:ch4:codeAssiginment}) assigns hash codes to every sample.

The proportion of samples from different classes in resultant regions yields their probability. For a given test sample, these probabilities are combined across all instances, and it is associated with the class having the highest probability. Let us assume $f$ that maps every pure region (represented by its unique code) to the posterior probabilities of finding a class $c \in Y$ in the given region. In other words, let $f:\{0,1\}^r \times Y \to \mathbb{R}$, then

\begin{align}\label{eq:ch4:probAssignment1}
f{(code_{x^{(i)}}, y_{i})} = \frac{\hat{f}{(code_{x^{(i)}}, y_{i})} \times IF_{y_i}}{\sum_{c=1}^{C}IF_c \times \hat{f}{(code_{x^{(i)}}, c)}},
\end{align}

where

\begin{align}\label{eq::probAssignment}
\hat{f}{(code_{x^{(i)}}, y_{i})} =  \frac{|\{y_{j}:(y_{j}=y_{i})\land(code_{x^{(j)}}=code_{x^{(i)}}) \}\:\forall j \in \{1,..,N\}|}{|\{y_{j}:code_{x^{(j)}}=code_{x^{(i)}}\}\:\forall j \in \{1,..,N\}|},
\end{align}

and $IF_c$ denote the weight associated with a class $c$ such that abundant classes have smaller weights, and vice-versa.

\begin{align}
IF_c = \frac{N}{|\{y_{j}:y_{j}=c\}\:\forall j \in \{1,..,N\}|}\: \forall c \in \{1,...,C\}.
\end{align}

Let us define $h_t$ such that $h_t:X \times Y \to \mathbb{R}$, $\forall t \in \{1,...,{T}\}$. Further, we define $h_t$ as follows, that maps every pure region to its posterior probabilities.

\begin{align}\label{eq:ch4:totalHypothesis}
h_t(x^{(i)}, y_i) = f(code_{x^{(i)}}, y_i)\: \forall (x^{(i)}, y_i) \in X \times Y
\end{align}

The above steps outline the construction of one high variance classifier instance. It is well established in the literature, that an ensemble of such high variance instances, in general, tends to yield better generalization on test samples~\cite{schapire1998boosting}. Our proposed method \acrshort{graf} creates several such high variance instances.

Next, we define $h$ such that it maps a sample to a class. This is done by using a consensus for prediction, that can be reached by computing the joint probability of predictions returned by each high variance classifier. We, therefore, define $h:X\to Y$ given by

\begin{align}\label{eq:ch4:prediction}
h(x^{(i)}) = \arg\max_{y_i \in Y}\sum_{t=1}^{T}\log_2\left(1+h_t(x^{(i)}, y_i)\right).
\end{align}

It should be noted, that when all regions for sample $x^{(i)}$ contain only one class $c$, then $h_t(x^{(i)},c)$ is 1 for $c$ and 0 for remaining classes. Hence, $h(x^{(i)})$ is equivalent to a voting classifier.

Given an ensemble of instances $h_1$, $h_2$,...,$h_T$, \acrshort{graf} optimizes the margin function as follows

\begin{align}\label{eq:ch4:treeMargin}
mg(x^{(i)}, y_i) = \mathbbm{1}{\left(h(x^{(i)}) = y_i\right)} - \max_{y_j \in Y \setminus{y_i}}\mathbbm{1}{\left(h(x^{(i)})=y_j\right)}.
\end{align}

Hence, the margin over the complete set of samples $X \times Y$ is defined as

\begin{align}\label{eq:ch4:treeExpectedMargin}
mg = \textbf{E}_{X, Y}mg(x^{(i)}, y_i).
\end{align}

\section{\acrshort{graf} as $\mathcal{A}$-$\mathcal{B}$ formulation}

\subsection{Choice of $\mathcal{A}$}

Similar to binary hash, \acrshort{graf} assigns a bit code to every sample (\ref{eq:ch4:codeAssiginment}). These similarity-preserving hash codes are generated by ingesting the class information, thus these hash codes are able to separate classes efficiently. For datasets having complex decision boundaries, the resulting bit code can be quite large which triggers all the issues associated with large bit codes. As discussed in Chapter 2, these bit codes can be arranged in a tree for faster retrieval. Section~\ref{sec:ch4:implementation} explores this idea in detail. The tree arrangement of bit codes then can be used to retrieve the nearest neighbors. Alternatively, any retrieval algorithm in hamming space such as \acrfull{combi}~\cite{mycombi} or \acrfull{mih}~\cite{manu::mih} can be employed.

\subsection{Choice of $\mathcal{B}$}

Assuming that, $\mathcal{A}$ is tree version of \acrshort{graf}, then for a sample $x$, \acrshort{graf} returns a partition $\Omega_{p}^{(l)}$ for an estimator $l \in \{1,..,L\}$, where $L$ is the total number of estimators. Assume that the associated bit code is $code_{x}^{(l)}$, i.e., $R_{code_{x}^{(l)}} = \Omega_{code_{x}^{(l)}}$.

Thus, probability estimator is given by,

\begin{align}
    \mathcal{B}(R_{code_{x}^{(l)}}, c) = p(c|\Omega_{p}^{(l)}).
\end{align}

Then equation~(\ref{eq:ch3:predict}) can be used to predict the class.

\section{Implementation details}\label{sec:ch4:implementation}

\begin{figure}[!ht]
	\centering
	\makebox[1 \textwidth][c]{
		\resizebox{1 \linewidth}{!}{
			\includegraphics[width=\linewidth,height=6cm,keepaspectratio]{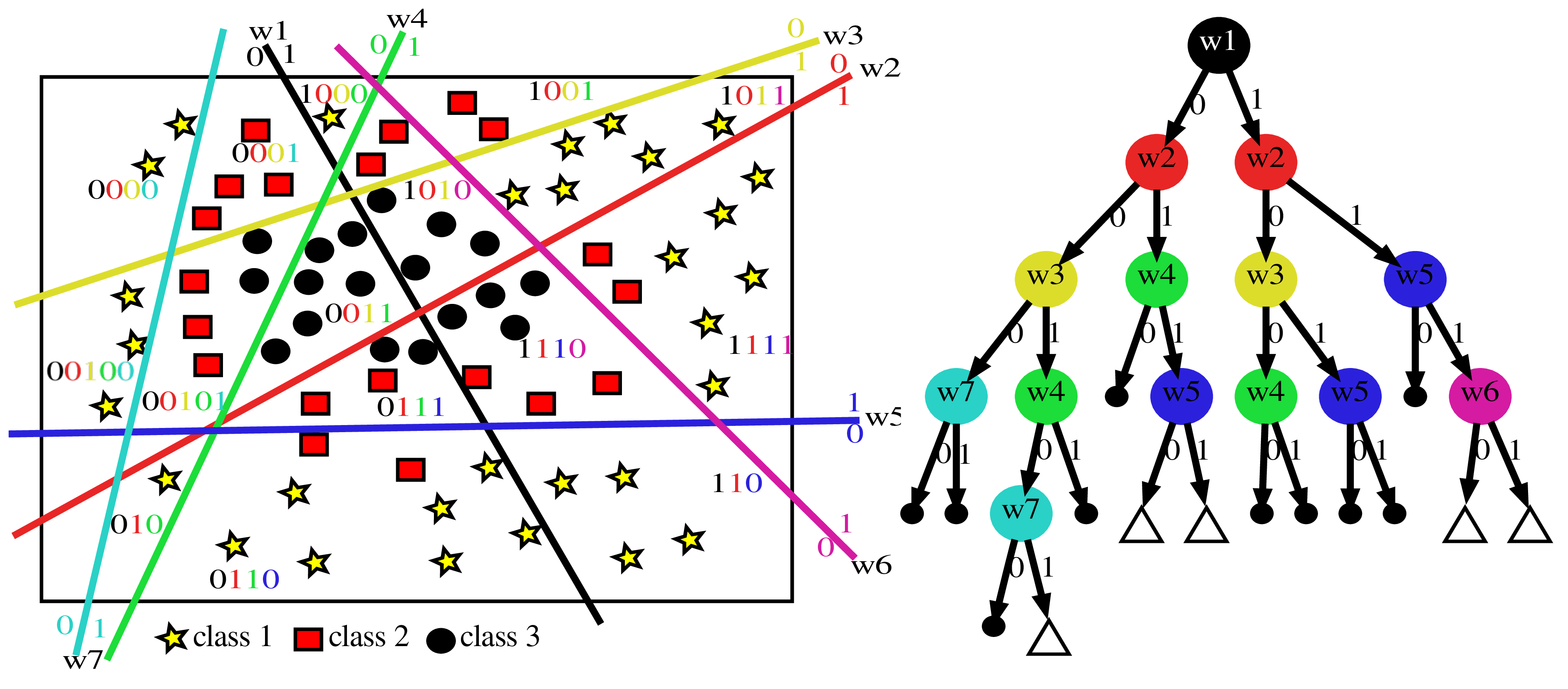}
		}
	}
	\caption{\textbf{The division of space in \acrshort{graf} is represented by a tree.} A region containing a subset of samples is defined by its unique combination of hyperplanes. However, these hyperplanes may affect the formation of other regions. The process terminates once space is maximally divided such that the impurity in any region cannot be reduced any further. Every resultant region corresponds to a leaf node in the tree, represented by a dot in the figure. (A triangle denotes an impure region that may be dichotomized further.)   }
	\label{fig:ch4:spacePartitioning}
\end{figure}

\acrfull{graf} creates an ensemble classifier by repeatedly dichotomizing the input space. To build one classifier instance from a given set $S$ of samples, a subset of $M$ features is uniformly sampled from the given set of features $n$. Samples are then projected into this M-dimensional sub-space, denoted by $X_M$. To facilitate efficient implementation, the additive construction of an instance is represented as a tree from the beginning. The tree is represented by its collection of regions (Figure~\ref{fig:ch4:spacePartitioning}). At the $0$-th height, $\Omega_{root}$ consist of all samples, $(x^{'})^{(i)} \in X_M$ and hence, the hyperplane $w^{\textit{(height)}}$ and $bias^{\textit{(height)}}$ is generated by considering all samples. At every height, the most impure region $\Omega^{*}$ (whole space at root), affects the generation of $w^{\textit{(height)}}$ and $bias^{\textit{(height)}}$. For $\Omega^{*}$, $K$ such hyperplanes are generated, and the effect of these hyperplanes is extended to other impure regions as well. The hyperplane whose inclusion yields the lowest overall space impurity $Z(S)$ is selected. Empty, pure, and impure regions may exist at each given height. The number of these regions is given by $\sum_{i=0}^{M}\binom{height}{i}$ (for $height < i$, $\binom{height}{i} = 0$), i.e., it is a polynomial in $height$ of the order of $M$ ($\mathcal{O}(height^{M})$). Thus, the number of filled (pure and impure) regions is \\$\mathcal{O}(\min(N, \sum_{i=0}^{M}\binom{height}{i}))$. For further processing, only impure regions need to be considered. Hence, $\mathcal{F}_1$ consists of only impure regions. The most impure region $\Omega^{*} \in \mathcal{F}_1$ defines the distribution of the next random weight vector $w^{(\textit{height})}$ to be included at next height. Even though $w^{(\textit{height})}$ almost surely dichotomizes the region $\Omega^{*}$, it may or may not dichotomize other remaining regions in  $\mathcal{F}_1$. To avoid creating empty regions, bit assignment is skipped for the non-dichotomized region at a given height. Hence, the resultant $code(j)$ for sample $x^{(j)}$ in region $\Omega_j$, formed by the concatenation of bits is of variable length. Once all impure regions have been fixed, leaf nodes represent the posterior probabilities of a class. The above procedure is repeated for the construction of other trees, with a different random sub-space of features of length $M$. Algorithm~\ref{algo:ch4:grafalgo} represents this process systematically.

\subsection{Heuristic for region search}
A naive implementation of the scanning-regions part of the algorithm will require scanning all the impure regions, which would incur an excessive overhead. \acrshort{graf} employs a heuristic to limit the number of impure regions to be scanned.

The \acrfull{roi} of a region $\Omega_{p}$ is defined as

\begin{align}\label{eq:ch4:roi}
\acrshort{roi}_{p} = \max\Bigg(\sqrt{\sum_{j=1}^{j=n}(m^{(p)}_{j} - \mu^{(p)}_{j})^{2}}, \sqrt{\sum_{j=1}^{j=n}(M^{(p)}_{j} - \mu^{(p)}_{j})^{2}}\Bigg)
\end{align}

\begin{figure}[!ht]
	\centering
	\makebox[1 \textwidth][c]{
		\resizebox{0.5 \linewidth}{!}{
			\includegraphics[width=\linewidth,height=6cm,keepaspectratio]{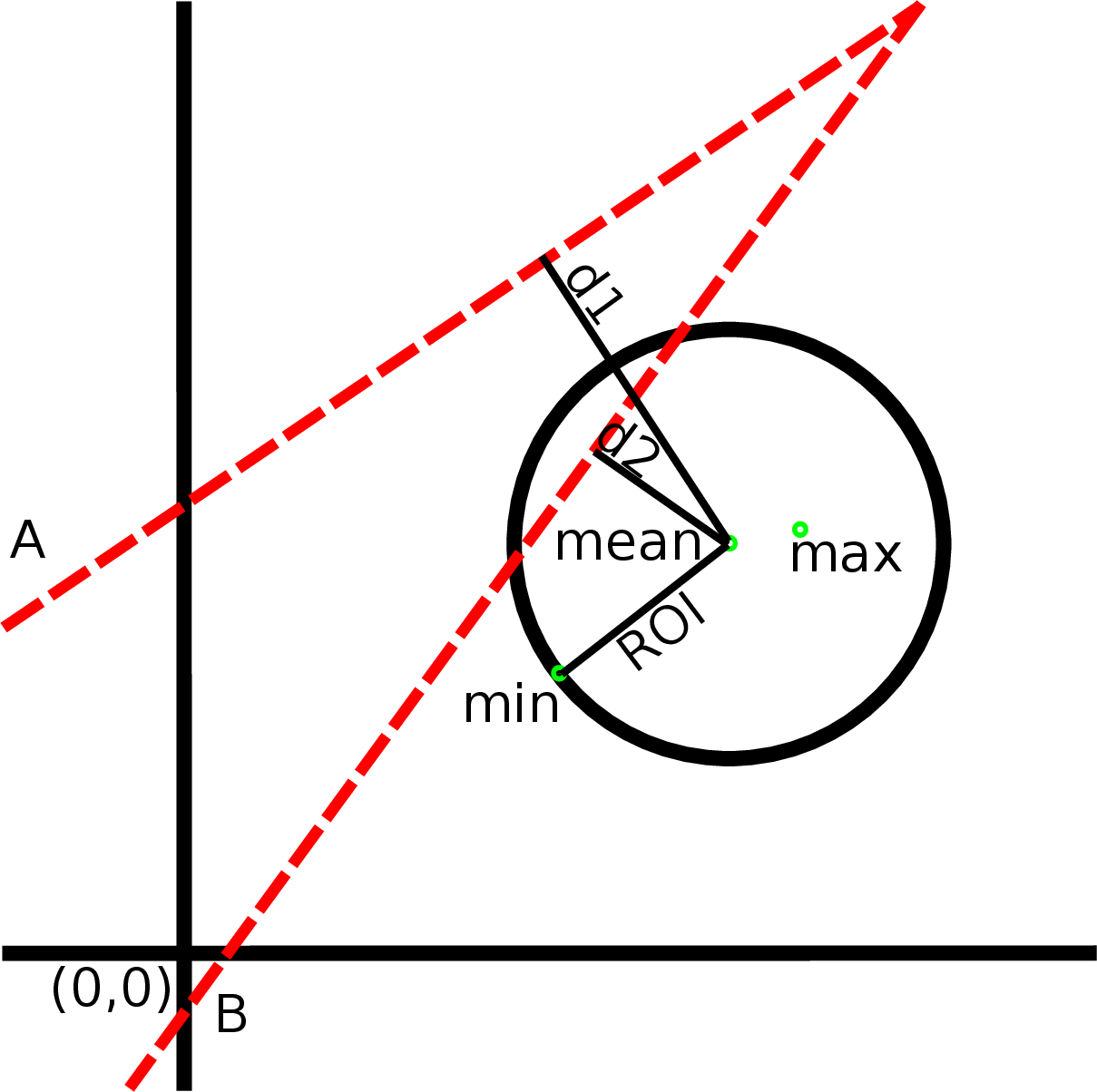}
		}
	}
	\caption{\textbf{A heuristic for faster run-time of GRAF.} The perpendicular distance of the mean point from plane A (d1) is greater than \acrfull{roi}. Hence, Plane A does not dichotomize the region. The perpendicular distance of the mean point from plane B (d2) is less than \acrshort{roi}. Hence, plane B may dichotomize the region. If the perpendicular distance is equal to \acrshort{roi}, it is considered as not dichotomized.}
	\label{fig:ch4:roi}
\end{figure}

A region is scanned for a split if the perpendicular distance (referred as $pdist$ in Algorithm~\ref{algo:ch4:grafalgo}) of the hyperplane to the mean (\ref{eq:ch4:meanEquation}) is less than \acrshort{roi} (\ref{eq:ch4:roi}). In Figure~\ref{fig:ch4:roi}, the min corner of the region is farther away from the mean, and hence \acrshort{roi} is defined as the distance between these two points. Two hyperplanes A and B are shown, where the perpendicular distance of A from mean (d1) is greater than the \acrshort{roi}, and hence this plane is guaranteed not to split the regions. Therefore,  while scanning for hyperplane A, this region will be skipped. When the perpendicular distance of B from the mean (d2) is less than the \acrshort{roi}, hyperplane B may or may not split the region. Hence, the region will be scanned for hyperplane B.

\begin{algorithm}[!htb]
	\caption{Pseudocode of \acrshort{graf}}\label{algo:ch4:grafalgo}
	\begin{algorithmic}[1]  
		\STATE \textbf{Input:} Dataset $X\times Y$ containing $N$ samples of $n$ features
		\STATE \hspace{10mm} $T$ - total number of trees
		\STATE \hspace{10mm} $M$ - feature subspace size ($\leq n$)
		\STATE \hspace{10mm} $K$ - trials to search the most suitable hyperplane
		\STATE \textbf{for} $t=1$ to $T$ \textbf{do}
		\STATE \hspace{3mm} choose $M$-dimensional feature subspace $X_{M}$
		\STATE \hspace{3mm} $height \gets 0$
		\STATE \hspace{3mm} Create $\Omega_{root}$, a region of whole data $X_{M}$
		\STATE \hspace{3mm} $\Omega_{root}.lc \gets \varnothing$, $\Omega_{root}.rc \gets \varnothing$, $\Omega_{root}.bit \gets \varnothing$ 
		\STATE \hspace{3mm} $\Omega_{root}.p \gets \varnothing$, $\Omega_{root}.h \gets height$
		\STATE \hspace{3mm} $\Omega_{root}.roi \gets ROI_{root}$ (\ref{eq:ch4:roi})
		\STATE \hspace{3mm} $\mathcal{F}_{1} \gets \{ \Omega_{root} \}$, $\Omega^{*} \gets \Omega_{root}$
		\STATE \hspace{3mm} \textbf{while} $|\mathcal{F}_{1}| > 0$ \textbf{do}
		\STATE \hspace{6mm} $W, b \gets$  generate $K$ hyperplanes for $\Omega^{*}$ (\ref{eq:ch4:weightGeneration}, \ref{eq:ch4:biasP})
		\STATE \hspace{6mm} \textbf{for} $k \in \{1,..,K\}$ \textbf{do}
		\STATE \hspace{9mm} $\mathcal{F}^{k} \gets \varnothing$
		\STATE \hspace{9mm} split $\Omega^{*}$ into $(\Omega^{*}_{0})^{k}$ and $(\Omega^{*}_{1})^{k}$ (\ref{eq:ch4:partition1}, \ref{eq:ch4:partition2})
		\STATE \hspace{9mm} $(\Omega^{*}_{0})^{k}.bit \gets 0$, $(\Omega^{*}_{1})^{k}.bit \gets 1$
		\STATE \hspace{9mm} $(\Omega^{*}_{0})^{k}.p \gets \Omega^{*}$, $(\Omega^{*}_{1})^{k}.p \gets \Omega^{*}$
		\STATE \hspace{9mm} $(\Omega^{*}_{0})^{k}.h, (\Omega^{*}_{1})^{k}.h \gets height + 1$
		\STATE \hspace{9mm} $\mathcal{F}^{k} \gets \mathcal{F}^{k}\cup\{(\Omega^{*}_{0})^{k}, (\Omega^{*}_{1})^{k}\}$
		\STATE \hspace{9mm} \textbf{for} $\Omega_{p} \in \mathcal{F}_{1} \setminus \{\Omega^{*}\}$
		\STATE \hspace{12mm} \textbf{if} $\Omega_{p}.roi > pdist(W^{k}, \mu^{p})$ \textbf{then}
		\STATE \hspace{15mm} split $\Omega_{p}$ into $(\Omega^{*}_{p0})^{k}$ and $(\Omega^{*}_{p1})^{k}$ (\ref{eq:ch4:remPartition1}, \ref{eq:ch4:remPartition2})
		\STATE \hspace{15mm} \textbf{if} $|\Omega^{*}_{p0}| > 0$ \& $|\Omega^{*}_{p1}| > 0$  \textbf{then}
		\STATE \hspace{18mm} $(\Omega^{*}_{p0})^{k}.p \gets \Omega_{p}$, $(\Omega^{*}_{p1})^{k}.p \gets \Omega_{p}$
		\STATE \hspace{18mm} $(\Omega^{*}_{p0})^{k}.bit \gets 0$, $(\Omega^{*}_{p1})^{k}.bit \gets 1$
		\STATE \hspace{18mm} $(\Omega^{*}_{p0})^{k}.h, (\Omega^{*}_{p1})^{k}.h \gets height + 1$
		\STATE \hspace{18mm} $\mathcal{F}^{k} \gets \mathcal{F}^{k}\cup\{(\Omega^{*}_{p0})^{k}, (\Omega^{*}_{p1})^{k}\}$
		\STATE \hspace{15mm} \textbf{else}
		\STATE \hspace{18mm} $\mathcal{F}^{k} \gets \mathcal{F}^{k} \cup \{\Omega_{p}\}$
		\STATE \hspace{12mm} \textbf{else}
		\STATE \hspace{15mm} $\mathcal{F}^{k} \gets \mathcal{F}^{k} \cup \{\Omega_{p}\}$
		\STATE \hspace{9mm} compute impurity of resultant partition of $S$ as $Z^{k}(S) = \sum_{\Omega_{p}\in\mathcal{F}^{k}}Z(\Omega_{p})$
		\STATE \hspace{6mm} $bestK \gets \arg \min_{k \in \{1,..,K\}}Z^{k}(S)$
		\STATE \hspace{6mm} $w^{(height)} \gets W^{(bestK)}$
		\STATE \hspace{6mm} $bias^{(height)} \gets b^{(bestK)}$
		\STATE \hspace{6mm} $\mathcal{F}_{1} \gets \mathcal{F}^{(bestK)}$
		\STATE \hspace{6mm} \textbf{for} $\Omega_{p} \in \mathcal{F}_{1}$ \textbf{do}
		\STATE \hspace{9mm} \textbf{if} $\Omega_{p}.bit = 0$ \textbf{then} $\Omega_{p}.p.lc \gets \Omega_{p}$
		\STATE \hspace{9mm} \textbf{if} $\Omega_{p}.bit = 1$ \textbf{then} $\Omega_{p}.p.rc \gets \Omega_{p}$
		\STATE \hspace{6mm} $height \gets height + 1$
	\end{algorithmic}
\end{algorithm}

\subsection{\acrshort{cpu} vs \acrshort{gpu} implementation}
For each impure region in $\mathcal{F}_1$, the division of region (\ref{eq:ch4:partition1}-\ref{eq:ch4:partition2}) requires a multiplication of two matrices of size $n_p \times M$ and $M \times K$. Matrix multiplication is computationally intensive, requiring $\mathcal{O}(n_p\times M\times K)$ \acrshort{cpu} operations. \acrlong{gpu}s (\acrshort{gpu}s) can significantly reduce matrix multiplication time via parallel computation.

\acrshort{graf}'s \acrshort{gpu} implementation differs slightly from the \acrshort{cpu} one. To avoid massive data transfer between the host's \acrshort{ram} and the \acrshort{gpu} co-processor, all training samples ($N \times M$) are stored in the \acrshort{gpu}'s \acrshort{ram} before initiating the training process. Upon selection of $\Omega^{*}$, the generated weight matrix of size $M \times K$ and the bias vector of length $K$ is sent to the \acrshort{gpu}, and a region assignment matrix of size $N \times K$ is retrieved. All impure regions from $\mathcal{F}_1$ are then scanned, to find the overall reduction in impurity ($Z(S)$) to select the best hyperplane.

\subsection{Time Complexity}\label{sec:ch4:timeComplexity}

To analyze the worst-case time complexity, assume a dataset where the neighborhood of each sample consists of examples from different classes. Further, assume that full trees are grown and that there are $N$ samples with $M$ dimensions. In this case, all leaf nodes will contain only one sample. Hence, there will be $N$ leaf nodes in the tree.

\subsubsection{Training time complexity of a tree}

Let us first assume that balanced trees are grown. In this case, the maximum number of impure regions at any time would be $N/2$. In the worst case, each hyperplane will only divide the region for which it was generated. The scanning of the region will take $\mathcal{O}(\sum_{i=1}^{i=(N/2)-1}(K\times N))$ time, until the maximum number of impure regions is created. Subsequently generated hyperplanes will "purify" at least one region. This will take $\mathcal{O}(\sum_{i=1}^{i=N/2}(K\times N))$ time. Hence, the total time spent in scanning will be $\mathcal{O}(K\times(N^2-N))\equiv\mathcal{O}(K\times N^2)$. Therefore, the total number of generated weights will be $N-1$ (total number of non-leaf nodes). The total time spent in matrix multiplication will be $\mathcal{O}((N\times M\times K+K)\times(N-1))$. Hence, total train time complexity $\mathcal{O}((N\times M\times K+K)\times(N-1) + K\times N^2)$.

In another scenario, assume that extremely skewed trees are generated. The maximum number of impure regions at any time will be $1$. In this case, the total number of generated weights will be $N-1$, and the total train time complexity is given as $\mathcal{O}((N\times M\times K+K)\times(N-1)+K)$.

The above-mentioned cases represent extreme scenarios. In practice, the training time complexity of \acrshort{graf} will lie somewhere in-between. Let the total number of generated weights be denoted by $TW$. Since weights are shared between regions, the value of $TW$ will be much smaller than $N-1$, and matrix multiplication time will reduce to $\mathcal{O}((N\times M\times K+K)\times TW)$. Similarly, the maximum number of impure regions at any instance is much smaller than $N/2$, since samples from similar classes tend to cluster. This reduces the total number of leaf nodes, which in turn reduces the maximum number of non-leaf nodes needed to be searched at any instance. The time required to scan impure regions can be reduced further by \acrshort{roi} heuristic. With the \acrshort{roi} heuristic, only a fraction of impure regions needs to be scanned to compute the quality of a hyperplane. However, this value is still upper bounded by $\mathcal{O}(K\times N^2)$. 

Hence, the worst case train time complexity of \acrshort{graf} for a \acrshort{cpu} implementation is $\mathcal{O}((N\times M\times K+K)\times TW + K\times N^2)$. Since matrix multiplication can be parallelized with \acrshort{gpu}s, the time complexity for a \acrshort{gpu} implementation is given by $\mathcal{O}(C_1 TW + C_2 + K\times N^2)$, where $C_1$ and $C_2$ are overheads for weight transfer, and data transfer, respectively.

\subsubsection{Testing time complexity of a tree}
The worst-case test time complexity of \acrshort{graf} is defined as the total time taken to reach a leaf node. For a given test sample, it is equal to $\mathcal{O}(max\_tree\_height\times M)$ for a \acrshort{cpu} implementation. For a \acrshort{gpu} implementation, it is $\mathcal{O}($ $max\_tree\_height + C_1)$, where $C_1$ is data transfer overhead.

\subsection{Model Size}\label{sec:ch4:modelsize}
The model size of \acrshort{graf} corresponds to the amount of information needed to make predictions. Since \acrshort{graf} uses a binary tree data structure, every internal/non-leaf ($TNL$) node will have exactly two child nodes. In addition, it also contains information about the index of weight to decide which path to traverse. Each leaf node ($TL$) also contains label information. Hence, the total model size (for a tree) of \acrshort{graf} is given by $TW\times (M+1) + TNL \times 3 + TL$.

\subsection{Space Complexity}
The scenario as described in Section~\ref{sec:ch4:timeComplexity} is followed to discuss the space complexity of \acrshort{graf}. In addition to the space required to store a dataset, \acrshort{graf} requires $\mathcal{O}(N)$ space to store temporary regions spawned in every trial. To perform $K$ trials, the total space requirement is $\mathcal{O}(K\times N)$. \acrshort{graf} also needs to store the tree in memory. As discussed in Section~\ref{sec:ch4:modelsize},  the total space required to store a tree is $TW\times (M+1) + TNL \times 3 + TL$, and hence, the total space complexity of GRAF is $\mathcal{O}(K\times N + TW\times (M+1) + TNL \times 3 + TL)$.

\section{Relationship of \acrshort{graf} with boosting}\label{sec:ch4:graf_adaboost}

As shown in Algorithm~\ref{algo:ch4:bsummary}, the construction of a high variance instance of a classifier can be abstracted as a boosting algorithm~\cite{freund1997decision}. Assuming that the weight of each sample is initially $1$, a random hyperplane is generated (\ref{eq:ch4:bitAssignment}). This generated hyperplane divides the region into two parts. Based on their impurity (\ref{eq:ch4:labeldef}) sample weights are updated to focus on the region under consideration. All the samples in that region are assigned a weight of $1$, while the remaining samples are assigned a weight of $0$. A new random hyperplane is generated (\ref{eq:ch4:weightGeneration}) based on the weight distribution of samples. However, this new plane is extended to other regions as well. The combination of all these planes (hypotheses) increases confidence, and hence, eventually creates a strong learner.

\begin{algorithm*}[!ht]
	\caption{High variance instance of \acrshort{graf} as boosting}\label{algo:ch4:bsummary}
	\begin{algorithmic}[1]
		\STATE \textbf{Input:}$(x^{(1)}, y_1)$,..,$(x^{(N)}, y_N)$; $x^{(i)} \in X$, $y_i \in \{1,..,C\}$, $C$ denotes the total unique classes and $N$ denotes the total training samples.
		\STATE \hspace{10mm}$Z:\mathcal{F}\to\mathbb{R}$ where $\Omega \in \mathcal{F}$ constitutes a set of points with same code.
		\STATE \hspace{10mm}$Y=\{1,..,C\}$
		\STATE \textbf{Initialize:} $P(i) \gets 1\: \forall i \in \{1,..,N\}$
		\STATE \hspace{15mm}$\textit{code}(i) \gets \varnothing\: \forall i \in \{1,..,N\}$
		\STATE \textbf{until} $\sum_{i=1}^{i=N} P(i) = 0$ \textbf{do}
		\STATE \hspace{8mm}Choose a random hypothesis using $P(i)$, such that $\lambda: X \to \{0,1\}$
		\STATE \hspace{8mm}$\textit{code}(i) \gets \textit{code}(i) \cup \{\lambda(x^{(i)})\}\: \forall i \in \{1,..,N\}$
		\STATE \hspace{8mm}Let $\Omega{_i} \gets \{(x^{(j)}, y_j): code(j) = code(i)\:\forall j \in \{1,..,N\}\}\:\forall i \in \{1,..,N\}$
		\STATE \hspace{8mm}$\omega \gets \arg\max_{i \in \{1,..,N\}} Z(\Omega_{i})$
		\STATE \hspace{8mm}Update $P(i) \gets \mathbbm{1}{\left(\Omega_{i}=\Omega_{\omega}\right)}\: \forall i \in \{1,..,N\}$
	\end{algorithmic}
\end{algorithm*}

\section{Feature selection using \acrshort{graf}}

\acrshort{graf} is an oblique split classifier. Thus, the direct estimation of feature contribution in the split improvement is not feasible. Alternatively, an \acrfull{oob} estimation can be performed to calculate the feature importance. The procedure works as follow:

Let us assume that $T$ high variance classifier instances are supposed to be constructed on the dataset $S$ (\ref{eq:ch4:sampleSpace}). For every high variance classifier $t \in T$, the dataset $S$ is divided into two non overlapping subsets $S1_{t}$ and $S2_{t}$ i.e. $S=S1_{t}\cup S2_{t}$ and $S1_{t} \cap S2_{t} = \phi$. Assuming that the larger set $S1_{t}$ ($|S1_{t}| > |S2_{t}|$) is used for training the high variance classifier $t$. The remaining set $S2_{t}$ can be used to compute the performance of the trained tree. This estimation is call \acrfull{oob} estimation.

Assuming that there are $n$ features in the data and out of which $M$ were selected to build a tree $t$ of \acrshort{graf}. The performance of a tree $t$ on $S2_{t}$ is given by $P_{S2_{t}}$. Now, to compute the importance of every feature, we will permute one feature of the \acrshort{oob} set at a time and compute the reduction in performance. Assume that when feature $f$ is permuted, the performance of tree $t$ on \acrshort{oob} set is give by $P_{S2_{t}}^{-f}$. Thus the reduction in the performance $\Delta P_{S2_{t}}^{-f}$ is given by

\begin{align}
    \Delta P_{S2_{t}}^{-f} = P_{S2_{t}} - P_{S2_{t}}^{-f},
\end{align}

This process is repeated for every feature and reduction in performance is computed and normalized.

\begin{align}
    \overline{\Delta P_{S2_{t}}^{-f}} = \begin{cases} \frac{\exp(\Delta P_{S2_{t}}^{-f})}{\sum_{f \in M}\exp(\Delta P_{S2_{t}}^{-f})}, & \text{if } f\in M \\ 0 & \text{otherwise}, \end{cases} \:\:\:\:\:\forall f \in n
\end{align}

These performance reduction scores are then accumulated for all the trees $t \in T$. Assuming that overall performance improvement, when feature $f$ is permutated, is given by $\Delta P^{f}$, then

\begin{align}
    \Delta P^{f} = \sum_{t \in T} \overline{\Delta P_{S2_{t}}^{-f}}
\end{align}

Thus the feature importance of a feature $f$ is given by

\begin{align}
    F^{f} = \frac{\exp(\Delta P^{f})}{\sum_{f \in n} \exp(\Delta P^{f}) }
\end{align}


\section{The \acrfull{ugraf}}\label{sec:ch4:ugraf}

Similar to Section~\ref{sec:ch4:GRAF}, let $\mathbb{R}^n$ denote the n-dimensional Euclidean space and $X \subseteq \mathbb{R}^n$ denote the input space. Let a set $S$ contain $N$ samples drawn from a population characterized by a probability distribution function $D$ over $X$. Thus the given dataset is

\begin{align}
    S = \{x^{(i)}\in X : i=1,2,..,N\}.
\end{align}

Note that this dataset does not have the label information. Further, let us assume that $T$ high variance classifier instances are constructed on the dataset $S$ and assume that the maximum number of samples allowed per region is $M$. The definition of a region is the same as discussed in Section~\ref{sec:ch4:GRAF}. For every sample the bit assignment is performed by (\ref{eq:ch4:bitAssignment}).

In contrast to Section~\ref{sec:ch4:GRAF}, A region $\Omega_p$ is considered for further division only if $|\Omega_p| >= M$. 

The next candidate partition is the biggest region $\Omega^{*}$ which is obtained as

\begin{align}
    \Omega^{*} = \arg\max_{\Omega_p \in \mathcal{F}_1} |\Omega_p|,\:\text{where}
\end{align}

where $\mathcal{F}_1$ is defined in (\ref{eq:ch4:impurePartitions}).

Let region $\Omega^{*}$ be divided into regions $\Omega^{*}_0$ and $\Omega^{*}_1$, where

\begin{align}\label{eq::ch4_1::partition1}
    \Omega^{*}_0 = \{x^{(i)}: \lambda^{*}(x^{(i)}) = 0\: \forall x^{(i)} \in \Omega^{*}\},
\end{align}

and

\begin{align}\label{eq::ch4_1::partition2}
    \Omega^{*}_1 = \{x^{(i)}: \lambda^{*}(x^{(i)}) = 1\: \forall x^{(i)} \in \Omega^{*}\}.
\end{align}
 
In (\ref{eq::ch4_1::partition1}) and (\ref{eq::ch4_1::partition2}), the mapping $\lambda^{*}$ is generated as for $\lambda_p$ defined at (\ref{eq:ch4:bitAssignment}). 

The rest of the steps to generate the bit code is same as discussed in Section~\ref{sec:ch4:GRAF}. \textbf{These $code_{x^{(i)}}$ will be the hash codes of the samples.} The fundamental use of \acrshort{ugraf} is to generate guided hashing. However, the idea presented here can also be used to identify clusters in the data in the hamming space.

KMeans~\cite{lloyd1982least,macqueen1965some}, \acrfull{som}~\cite{kohonen1990self}, \acrfull{dbscan}~\cite{ester1996density} and its hierarchical version~\cite{mcinnes2017accelerated}, etc. clustering algorithms works in the euclidean space. The performance of such algorithms deteriorates with increasing dimensionality of the data (the curse of dimensionality). The remedy for such a situation is to decrease the dimensionality of the data with ad-hoc algorithms and then perform clustering. Following this queue, locality-sensitive hashing methods can transform high dimensional into hamming space and can approximate the neighborhood information from the euclidean spaces. However, in the hamming space, these algorithms can not be applied. Thus, we propose to use \acrshort{combi} or \acrshort{ugraf} to perform clustering in the hamming spaces.

The locality-sensitive hashing algorithms that generate full-length bit codes can be arranged into tree structures using \acrshort{combi}. Traversing the tree can then identify the closest samples in hamming space to generate the neighborhood snapshot. Utilizing \acrshort{ugraf} gives a tree, which can be traversed to generate the neighborhood snapshot. This neighborhood snapshot can then be used to perform clustering. A detailed algorithm for the same is proposed in Section~\ref{sec:ch7:clustering}.

\section{Simulation Study}\label{sec:ch4:simulation}

We designed a simulation study to examine design aspects of \acrshort{graf}, such as oblique hyperplanes for dichotomization, and extension of the hyperplane. It is known that axis-aligned decision trees do not generalize well for tasks with high concept variation~\cite{perez1996learning,bengio2010decision}. To emulate a high concept variation task, samples were generated near the vertices of a $n$ dimensional hypercube as per Algorithm~\ref{algo:ch4:simulationData}. For a binary classification task, the parity function was considered. A label 1 is assigned to a sample if it is generated near a vertex having an odd number of 1s, and a label 0 otherwise. For a multi-class classification task, the label is assigned as the total number of 1's in the neighboring vertex.

\begin{algorithm}[!ht]
	\caption{Simulation Data for GRAF benchmarking}\label{algo:ch4:simulationData}
	\begin{algorithmic}
		\STATE \textbf{Input:} $n$ dimension of hypercube.
		\STATE \textbf{Initialize:} 
		\STATE \hspace{4mm} $sample\_per\_vertex \gets [3, 4, 5]$
		\STATE \hspace{4mm} $all\_coords \gets$ all vertices of n dimensional hypercube
		\STATE \hspace{4mm} $mean\_0$, $mean\_1$, $stdev\_0$ and $stdev\_1$ of size $n$
		\STATE \textbf{Output:} $generated\_data \gets []$ 
		\STATE \textbf{Run:}
		\STATE \textbf{for} $i \in \{1,..,n\}$ \textbf{do}
		\STATE \hspace{4mm}$mean\_0_i$, $stdev\_0_i \sim \mathcal{U}[-0.5, 0.5)$
		\STATE \hspace{4mm}$mean\_1_i$, $stdev\_1_i \sim \mathcal{U}[0.5, 1.5)$
		\STATE \textbf{for} $coord \in all\_coords$ \textbf{do}
		\STATE \hspace{4mm} $c \gets $ \text{select one number randomly from } $sample\_per\_vertex$
		\STATE \hspace{4mm} \textbf{for} $j \in \{1,..,c\}$ \textbf{do}
		\STATE \hspace{8mm} $gen\_sample \gets $ array of size $n$
		\STATE \hspace{8mm} $ct \gets 0$
		\STATE \hspace{8mm} \textbf{for} $bit \in coord$ \textbf{do}
		\STATE \hspace{12mm} \textbf{if} $bit = 0$ \textbf{then}
		\STATE \hspace{16mm} $gen\_sample_{ct}$ $\sim$ $\mathcal{N}$($mean\_0_{ct}$, $stdev\_0_{ct}$) until $-0.5 <$ $gen\_sample_{ct}$ $< 0.5$
		\STATE \hspace{12mm} \textbf{if} $bit = 1$ \textbf{then}
		\STATE \hspace{16mm} $gen\_sample_{ct}$ $\sim$ $\mathcal{N}$($mean\_1_{ct}$, $stdev\_1_{ct}$) until $0.5 <$ $gen\_sample_{ct}$ $< 1.5$
		\STATE \hspace{12mm} $ct \gets ct + 1$
		\STATE \hspace{8mm} $generated\_data.append(gen\_samples)$
	\end{algorithmic}
\end{algorithm}

The number of features ($n$) is varied from 3 to 15 (since very few samples can be generated when only 2 features are used). In effect, the total number of samples vary from $\sim25$ - $\sim115,000$ (Table~\ref{tab:ch4:simulation_data}). For a multiclass example with $n$ features, $n+1$ classes are possible. For a given configuration (binary or multiclass) with $n$ features, 10 different datasets were generated. For every dataset, the train-test split consisted of 70-30\% of the total samples. 

\begin{table}[!ht]
	\centering
	\scalebox{1}{
		\begin{tabular}{||c|c|c|c|c||}
			\hline
			Features & Classes & Train samples & Test Samples & PC(v=0.9)\\
			\hline\hline
			3 & 2,4 & 18.1$\pm$0.700 & 8.6$\pm$0.489&3,1.7\\
			4 & 2,5 & 38.9$\pm$1.044 & 17.4$\pm$0.663&4,1.5\\
			5 & 2,6 & 77.8$\pm$1.887  & 34.3$\pm$0.900&5,2.1 \\
			6 & 2,7 & 155.6$\pm$1.685 & 67.3$\pm$0.900&6,2.2 \\
			7 & 2,8 & 312.9$\pm$3.477 & 134.6$\pm$1.497&7,2.6 \\
			8 & 2,9 & 626.5$\pm$5.463 & 269.3$\pm$2.452&8,2.7 \\
			9 & 2,10 & 1256.5$\pm$9.729 & 539.1$\pm$4.346&8,2.9 \\
			10 & 2,11 & 2515.4$\pm$10.312 & 1078.7$\pm$4.647&9,3.3 \\
			11 & 2,12 & 5024.7$\pm$15.408 & 2154.1$\pm$6.730&10,3.3 \\
			12 & 2,13 & 10032.9$\pm$10.540 & 4300.6$\pm$4.652&11,3.6 \\
			13 & 2,14 & 20072.4$\pm$36.546 & 8603.1$\pm$15.776&12,3.8 \\
			14 & 2,15 & 40129.0$\pm$41.613 & 17198.8$\pm$17.713&13,4 \\
			15 & 2,16 & 80302.7$\pm$68.444 & 34416.3$\pm$29.312&14,4.5 \\
			\hline\hline
		\end{tabular}
	}
	\caption{A simulation study to discuss the design aspects of \acrshort{graf}. The number of features varied from 3 to 15. For a given value of the feature, both binary and multiclass examples were generated. For every configuration, 10 different trials were performed to generate samples. The total number of samples vary from $\sim25$ - $\sim115,000$ across all trials. The train-test split consists of 70-30\% of the total samples. The total number of principal components which explains 90\% of the total variance in the dataset differs when it is projected on a random matrix.}
	\label{tab:ch4:simulation_data}
\end{table}

\begin{figure}[!ht]
	\centering
	\makebox[1 \textwidth][c]{
		\resizebox{1.1 \linewidth}{!}{
			\includegraphics[width=\linewidth,keepaspectratio]{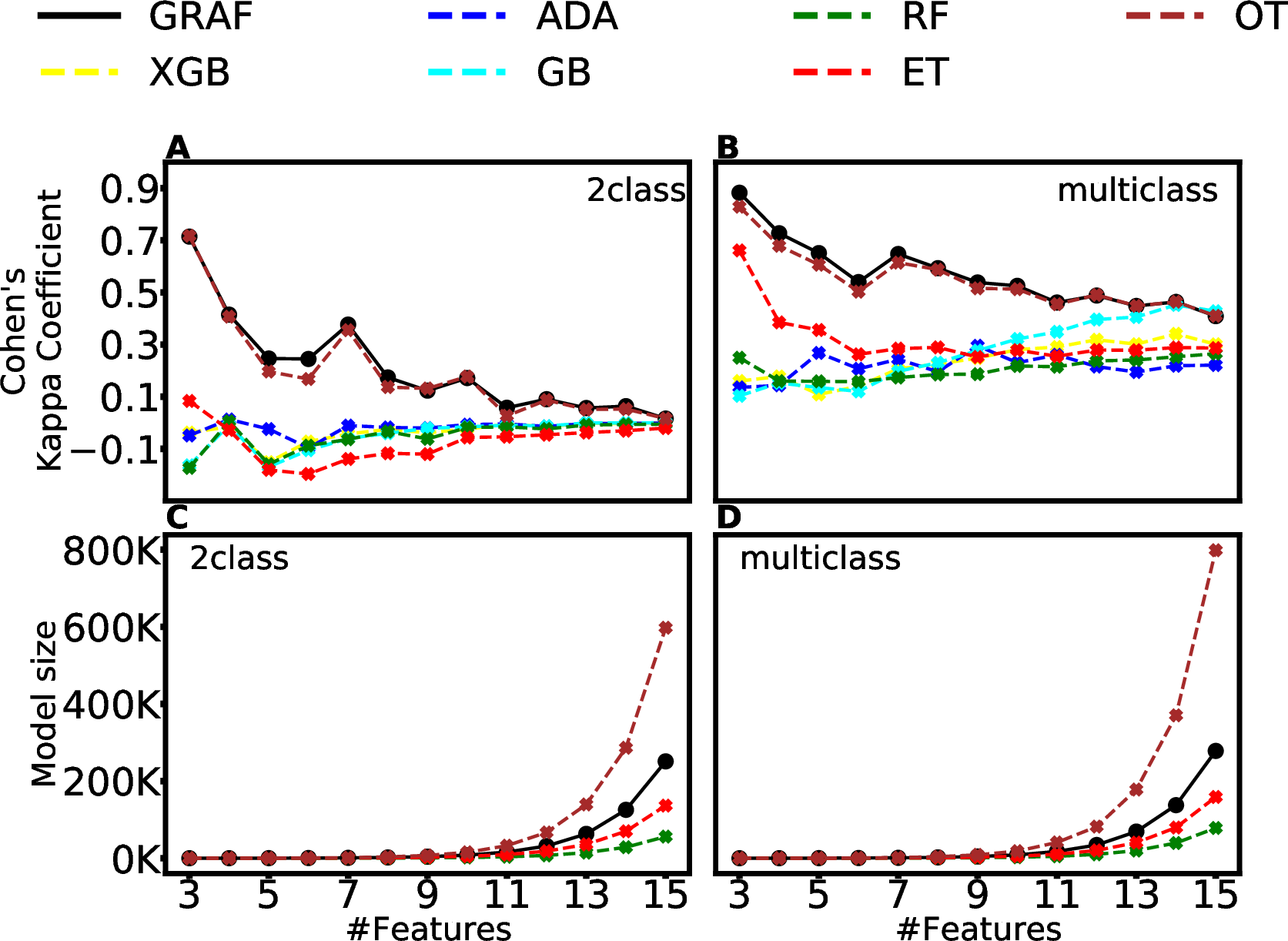}
		}
	}
	\caption{\textbf{The performances and model size comparison of methods on simulated binary and multiclass examples with high concept complexity.} The high concept complexity means that all the features are independent of each other. The number of features varies from 3 to 15. \textbf{A}, \textbf{B)} For both binary and multiclass examples, \acrshort{graf} has the highest values of Cohen's kappa coefficients, closely followed by \acrfull{ot}. \textbf{C}, \textbf{D)} However, for similar performance measures, the overall model size of \acrshort{ot} is much higher when compared with \acrshort{graf}.}
	\label{fig:ch4:simulationIndividual}
\end{figure}

For comparison, $100$ trees were generated for every method, and the entire feature space was considered for every tree. For all experiments, $K$ (for \acrshort{graf}) was equal to $M$ and $M=n$. For a given feature ($n$) and label information (binary or multi-class), the performance of a method was evaluated using Cohen's kappa coefficient for every trial, and averaged across all trials. For both binary and multiclass cases, the performance of \acrshort{graf} supersedes others, closely followed by \acrfull{ot}~\cite{menze2011oblique} (Figure~\ref{fig:ch4:simulationIndividual}A, B). This is primarily because when concept variation is high, all features are independent and relevant. Thus, axis-aligned decision trees suffer because they consider only a single feature at a time to define a region. The performances of all others such as \acrfull{ada}~\cite{freund1997decision}, \acrfull{rf}~\cite{rf}, \acrfull{xgb}~\cite{chen2016xgboost}, \acrfull{gb}~\cite{friedman2001greedy}, and \acrfull{et}~\cite{et} are comparable to each other. The model size of \acrshort{et}, \acrshort{rf}, \acrshort{graf}, and \acrshort{ot} has also been compared. For decision trees, the model size is mainly affected by factors such as the total number of internal/non-leaf nodes (TNL), the total number of leaf nodes (TL), and the total weights generated (TW). Non-leaf nodes contain threshold information, links to both child nodes, and the feature used for the split. The leaf nodes contain label information. For \acrshort{et}, \acrshort{rf}, and \acrshort{ot}, the total number of weights is equal to the total number of non-leaf nodes in the tree. The overall model size for \acrshort{et} and \acrshort{rf} is $TNL\times 4+TL$. For \acrshort{ot}, the weight vector $w$ lies in $\mathbb{R}^n$. Hence, the model size of \acrshort{ot} is $TNL\times (n+1)+TNL\times2+TL$. However, for \acrshort{graf}, since weights are shared between different regions, the total number of weights is much smaller than the total number of non-leaf nodes. \acrshort{graf}'s model size is therefore $TW\times (n+1)+TNL\times 3+TL$. \acrshort{graf}'s model size is significantly smaller than \acrshort{ot}'s, for comparable performance (Figure~\ref{fig:ch4:simulationIndividual}C, D).

\begin{figure}[!ht]
	\centering
	\makebox[1 \textwidth][c]{
		\resizebox{1.1 \linewidth}{!}{
			\includegraphics[width=\linewidth,keepaspectratio]{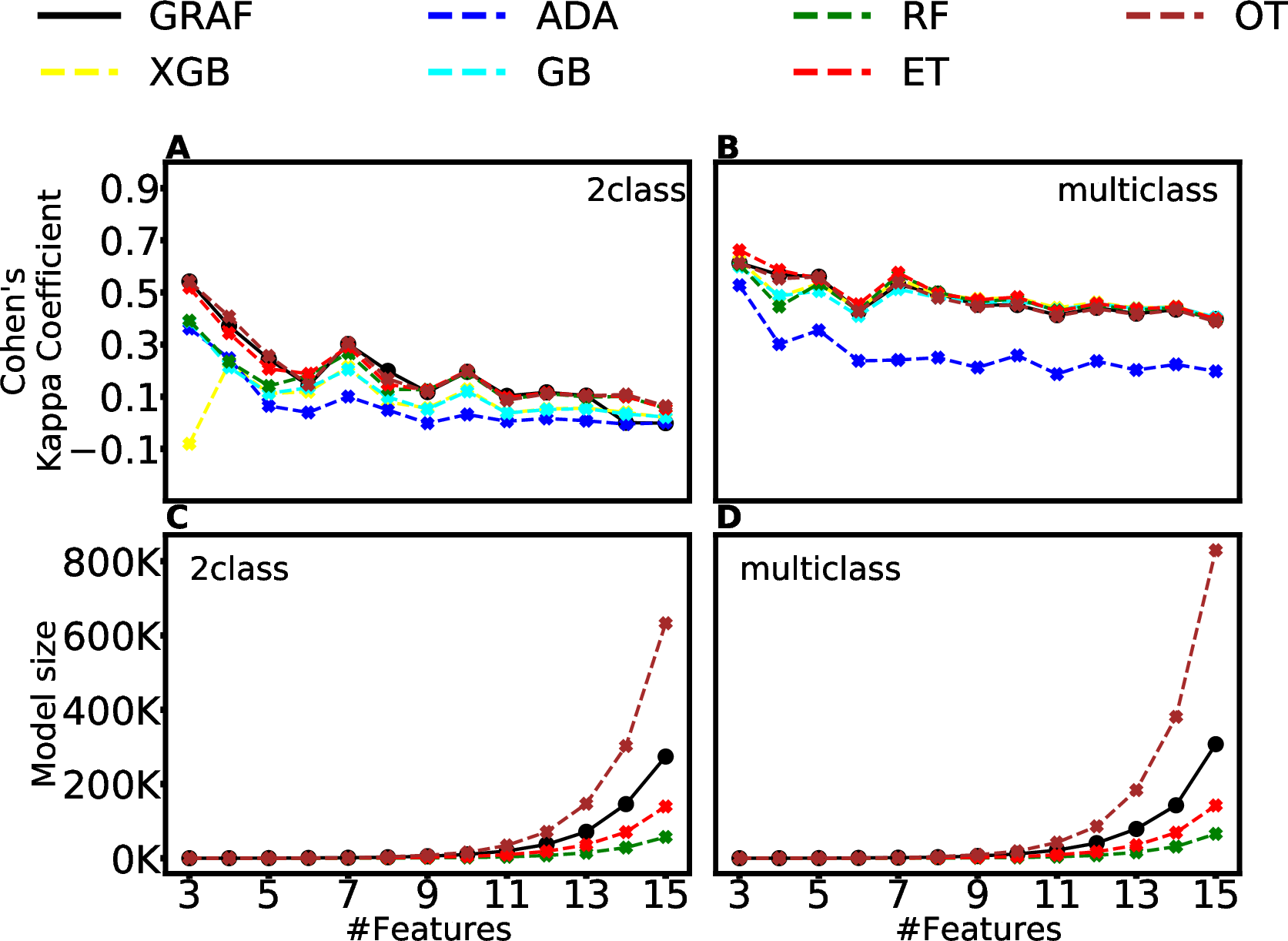}
		}
	}
	\caption{\textbf{The performances and model size comparison of methods on simulated binary and multiclass examples with low concept complexity.}. The low concept complexity means that only a few features are relevant and independent. The number of features varies from 3 to 15. \textbf{A}, \textbf{B)} For both binary and multiclass examples. In these settings performances of all methods are comparable. \textbf{C}, \textbf{D)} The trend in the model size is the same as the high concept complexity datasets. }
	\label{fig:ch4:simulationProjection}
\end{figure}

The previous simulation study's essence was to establish that a scenario where all features are independent and relevant \acrshort{graf} shows satisfactory performance along with competitive model size. In addition, it is imperative to evaluate the performances of methods when all features are not necessarily independent. For this, the samples in the previous study are projected using a random matrix. The resultant dataset's overall variance is explained with a few principal components (Table~\ref{tab:ch4:simulation_data}). For instance, when $15$ features are used to generate a simulated dataset, $14$ principal components are needed to explain 90\% of the total variance in the dataset.
On the other hand, when the same dataset is projected using a random matrix, less than $5$ principal components are adequate. For this scenario, similar experiments were performed. For this case, almost all methods have comparable performances (Figure~\ref{fig:ch4:simulationProjection}). In other words, \acrshort{graf} performs satisfactorily in both scenarios. 

\begin{figure}[!ht]
	\centering
	\makebox[1 \textwidth][c]{
		\resizebox{1.1 \linewidth}{!}{
			\includegraphics[width=\linewidth,keepaspectratio]{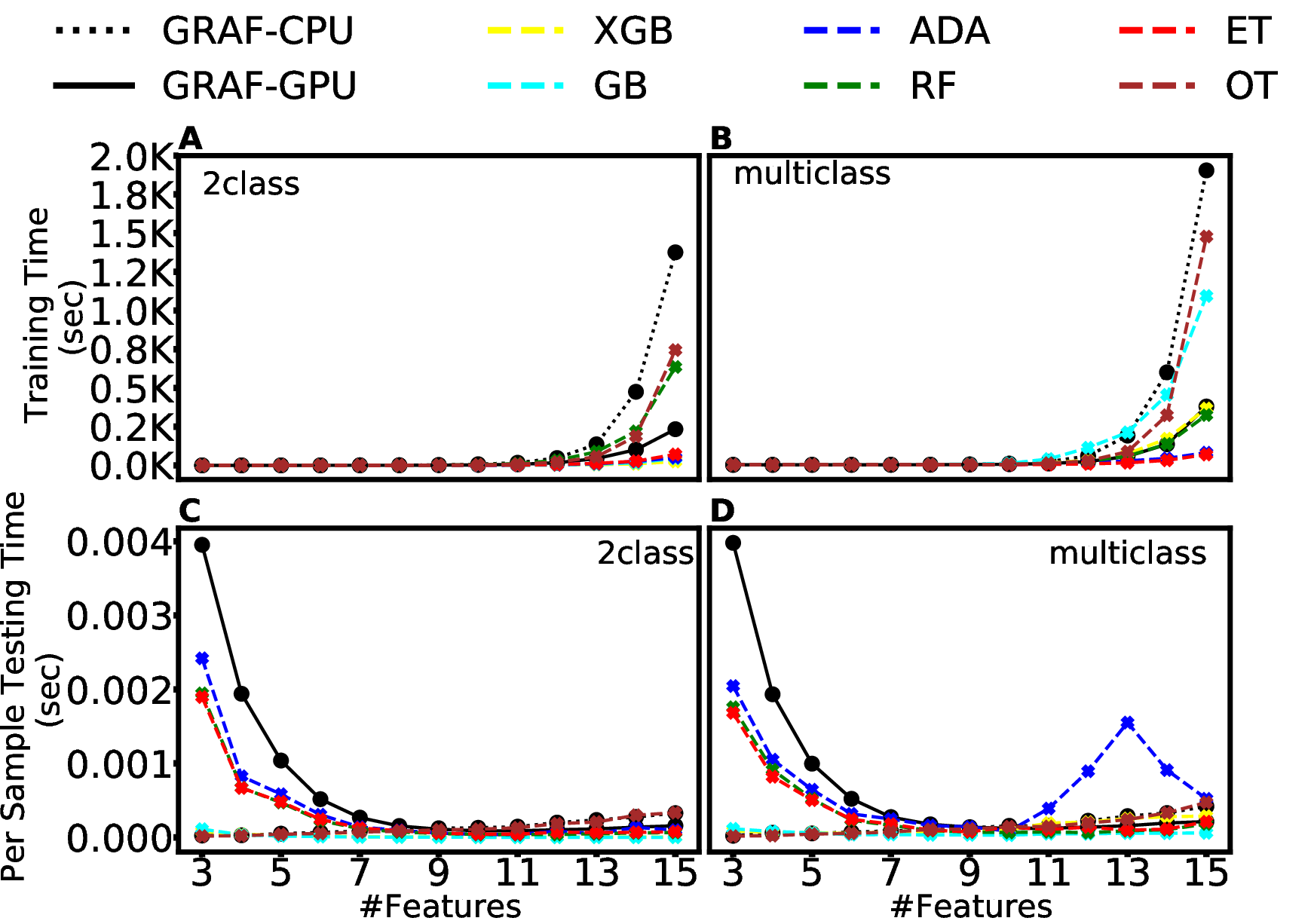}
		}
	}
	\caption{\textbf{The run-time complexity analysis of high concept complexity datasets.} The training and testing time of different methods is compared on a simulated dataset. \textbf{A}, \textbf{B)} \acrshort{graf}'s \acrshort{gpu} implementation significantly reduces the training time for both binary and multiclass examples. \textbf{C}, \textbf{D)} \acrshort{graf}'s testing time is comparable with other methods. }
	\label{fig:ch4:simulationTimeIndividual}
\end{figure}

\begin{figure}[!ht]
	\centering
	\makebox[1 \textwidth][c]{
		\resizebox{1.1 \linewidth}{!}{
			\includegraphics[width=\linewidth,keepaspectratio]{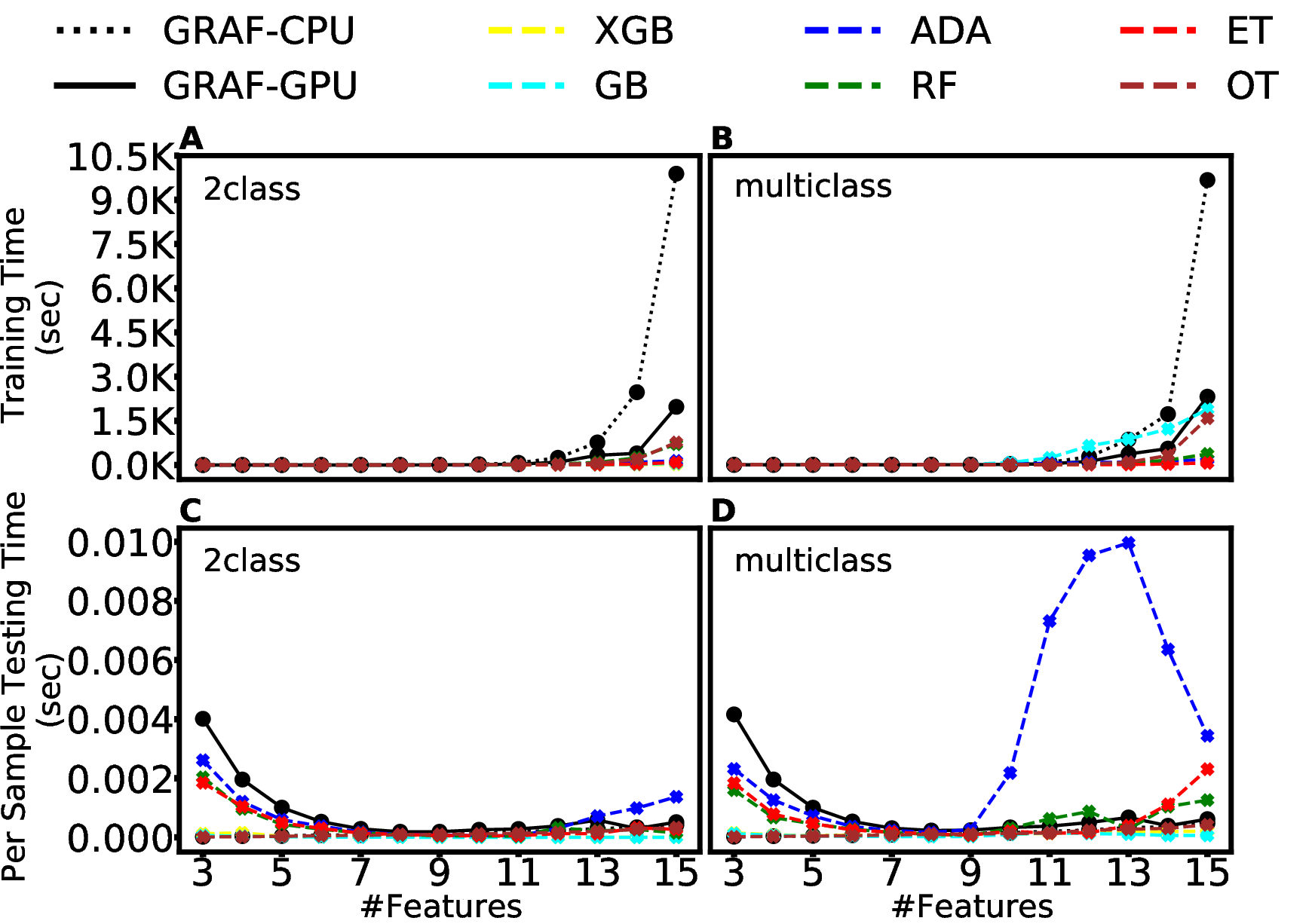}
		}
	}
	\caption{\textbf{The run-time complexity analysis of low concept complexity datasets.} The training and testing times of different methods are compared on a simulated dataset projected by using a random matrix. \textbf{A}, \textbf{B)} The \acrshort{gpu} implementation of \acrshort{graf} significantly reduces its training time for both binary and multiclass examples. \textbf{C}, \textbf{D)} The testing time of \acrshort{graf} is comparable with other methods. }
	\label{fig:ch4:simulationTimeProjection}
\end{figure}

The other important criterion for comparing different methods is their run time complexity. As discussed in Section~\ref{sec:ch4:timeComplexity}, \acrshort{graf}'s \acrshort{gpu} train time (\acrshort{graf}-\acrshort{gpu}) is considerably lower than its \acrshort{cpu} counterpart, because \acrshort{graf} involves matrix multiplication. Hence, we compare the training and test time complexity of both implementations of \acrshort{graf} with \acrshort{ot}, \acrshort{et}, \acrshort{rf}, \acrshort{gb}, \acrshort{ada} and \acrshort{xgb} on simulated dataset (Figure~\ref{fig:ch4:simulationTimeIndividual}) and simulated dataset after projection (Figure~\ref{fig:ch4:simulationTimeProjection}). As shown in Figure~\ref{fig:ch4:simulationTimeIndividual}A and B, the training time of \acrshort{graf}-\acrshort{gpu} is considerably smaller than \acrshort{ot} and \acrshort{gb}, and competitive with \acrshort{rf} and \acrshort{xgb}. \acrshort{graf}-\acrshort{gpu}'s test time (Figure~\ref{fig:ch4:simulationTimeIndividual}C, D) is higher for smaller datasets because the data transfer overhead overshadows the speed gain from parallelization while being considerably smaller for larger datasets.  

In all the above experiments, the number of trials for \acrshort{graf} is equal to the number of features in the dataset. It was also observed, that the performance of \acrshort{graf} without trials is slightly lower when compared with its trial counterpart. However, the training time is significantly lower. For the cases where features are independent and informative, the training time of \acrshort{graf} is as fast as \acrshort{et}.

Performance measures reported in this article are recorded on a workstation with 40 cores using Intel\textregistered Xeon\textregistered E7-4800 (Haswell-EX/Brickland Platform) \acrshort{cpu}s with a clock speed of 1.9 GHz, 1024 GB DDR4-1866/2133 ECC \acrshort{ram} and Ubuntu 14.04.5 LTS operating system with the 4.4.0-38-generic kernel. The time taken by each algorithm has been measured by running it on a single core. For computation on GPUs, a 12GB NVIDIA Tesla K80 GPU was used.

This simulation study explains that cases where features are independent and relevant, oblique partitions (\acrshort{graf}, \acrshort{ot}) fair well in comparison to axis-aligned (\acrshort{rf}, \acrshort{et}) partitions (Figure~\ref{fig:ch4:simulationIndividual}A, B). However, in the cases where the intrinsic dimensionality of data is smaller than the number of features, all methods have comparable performance (Figure~\ref{fig:ch4:simulationProjection}A, B). These results are concordant with the previously observed results~\cite{menze2011oblique,zhang2017robust}. Between \acrshort{graf} and \acrshort{ot}, \acrshort{graf} has a smaller model size. This is because in \acrshort{graf}, hyperplanes are shared between multiple regions, while in \acrshort{ot}, each hyperplane does local partitioning. Therefore, \acrshort{graf} has fewer hyperplanes and hence, a smaller model size.  However, \acrshort{et} and \acrshort{rf} have lower model size in comparison to \acrshort{graf} (Figure~\ref{fig:ch4:simulationIndividual}C and D, \ref{fig:ch4:simulationProjection}C and D). In the first case, the training time of \acrshort{graf}-\acrshort{gpu} is lower in comparison to \acrshort{ot} and \acrshort{rf} (Figure~\ref{fig:ch4:simulationTimeIndividual}A, B) but in a later case, the training time of \acrshort{graf}-\acrshort{gpu} is the highest (Figure~\ref{fig:ch4:simulationTimeProjection}A, B). All methods have equivalent testing time (Figure~\ref{fig:ch4:simulationTimeIndividual}C and D, \ref{fig:ch4:simulationTimeProjection}C and D). Considering all these aspects, it may be concluded that for the first case, \acrshort{graf} can be a choice of method for both binary and multiclass cases. 

\section{Results}\label{sec:ch4:results}

\subsection{Data generation with Weka for Bias-variance tradeoff}\label{sec:ch4:dataGeneration}

In order to examine the bias-variance tradeoff, 6 different binary and multi-class datasets with different numbers of centroids were generated using Weka~\cite{witten1999weka}. The RandomRBF data generator was selected to simulate the data. A detailed description of this class is available at \url{http://weka.sourceforge.net/doc.dev/weka/datagenerators/classifiers/classification/RandomRBF.html}. In order to generate the data set, the number of features '-a' was set to 10, the number of centroids '-C' was selected from \{10, 20, 50\}, and the number of classes '-c' was selected from \{2, 5\}. For each dataset, a total of 10000 samples '-n' were generated. The commands to generate the data from weka with seed '-S' 1 are given below:

\begin{verbatim}
java -Xmx128m -classpath $PWD:weka.jar weka.datagenerators.classifiers.
classification.RandomRBF -r weka.datagenerators.classifiers.classification.
RandomRBF-datafile -S 1 -n 10000 -a 10 -c 2 -C 10
\end{verbatim}

\begin{verbatim}
java -Xmx128m -classpath $PWD:weka.jar weka.datagenerators.classifiers.
classification.RandomRBF -r weka.datagenerators.classifiers.classification.
RandomRBF-datafile -S 1 -n 10000 -a 10 -c 5 -C 10
\end{verbatim}

\begin{verbatim}
java -Xmx128m -classpath $PWD:weka.jar weka.datagenerators.classifiers.
classification.RandomRBF -r weka.datagenerators.classifiers.classification.
RandomRBF-datafile -S 1 -n 10000 -a 10 -c 2 -C 20
\end{verbatim}

\begin{verbatim}
java -Xmx128m -classpath $PWD:weka.jar weka.datagenerators.classifiers.
classification.RandomRBF -r weka.datagenerators.classifiers.classification.
RandomRBF-datafile -S 1 -n 10000 -a 10 -c 5 -C 20
\end{verbatim}

\begin{verbatim}
java -Xmx128m -classpath $PWD:weka.jar weka.datagenerators.classifiers.
classification.RandomRBF -r weka.datagenerators.classifiers.classification.
RandomRBF-datafile -S 1 -n 10000 -a 10 -c 2 -C 50
\end{verbatim}

\begin{verbatim}
java -Xmx128m -classpath $PWD:weka.jar weka.datagenerators.classifiers.
classification.RandomRBF -r weka.datagenerators.classifiers.classification.
RandomRBF-datafile -S 1 -n 10000 -a 10 -c 5 -C 50
\end{verbatim}

\subsection{Bias-variance tradeoff}\label{sec:ch4:bias-variance}

In order to understand the behavior of a classifier, it is imperative to study its bias-variance tradeoff. A classifier with a low bias has a higher probability of predicting the correct class than any other class, i.e., the predicted output is much closer to the true output. On the other hand, the classifier with low variance indicates that its performance does not deviate for a given test set across several different models. There are several methods to evaluate bias-variance tradeoff for 0-1 loss on classification learning~\cite{breiman1996bias,kohavi1996bias,domingos2000unified,james2003variance}. Of these, we use the definitions of Kohavi \& Wolpert~\cite{kohavi1996bias} for bias-variance decomposition (\ref{eq:ch4:vote}-\ref{eq:ch4:err}).

\begin{align}\label{eq:ch4:vote}
p^{(i)}_j=\frac{1}{R}\sum_{r=1}^{r=R}\mathbbm{1}{(\hat{y_{i}}=j}) 
\end{align}
\begin{align}\label{eq:ch4:bias}
bias^2 = \frac{1}{N_t}\Bigg(\sum_{i=1}^{i=N_t}\sum_{j=1}^{j=C}\big((\mathbbm{1}{(y_{i}=j)} - p^{(i)}_j)^{2} - \frac{p^{(i)}_j*(1-p^{(i)}_j)}{R-1}\big)\Bigg)
\end{align}
\begin{align}\label{eq:ch4:variance}
variance = 1 - \frac{1}{N_t}\sum_{i=1}^{i=N_t}\sum_{j=1}^{j=C}(p^{(i)}_j)^{2}
\end{align}
\begin{align}\label{eq:ch4:err}
err = \frac{1}{R}\sum_{r=1}^{r=R}\Big(1-\frac{1}{N_t}\sum_{i=1}^{i=N_t}\mathbbm{1}{(y_{i}=\hat{y_{i}})}\Big)
\end{align}

For the analysis of the bias-variance tradeoff, $N/2$ samples were set aside as the test set. From the remaining dataset, $R$ overlapping training sets of the same size $N_m$ were created, and $R$ models were trained. For every model, the estimate $\hat{y_i}$ is obtained for every instance $i$ in the test set, whose size is denoted by $N_t$.

\begin{figure}[!ht]
	\centering
	\makebox[1 \textwidth][c]{
		\resizebox{1 \linewidth}{!}{
			\includegraphics[width=\linewidth,keepaspectratio]{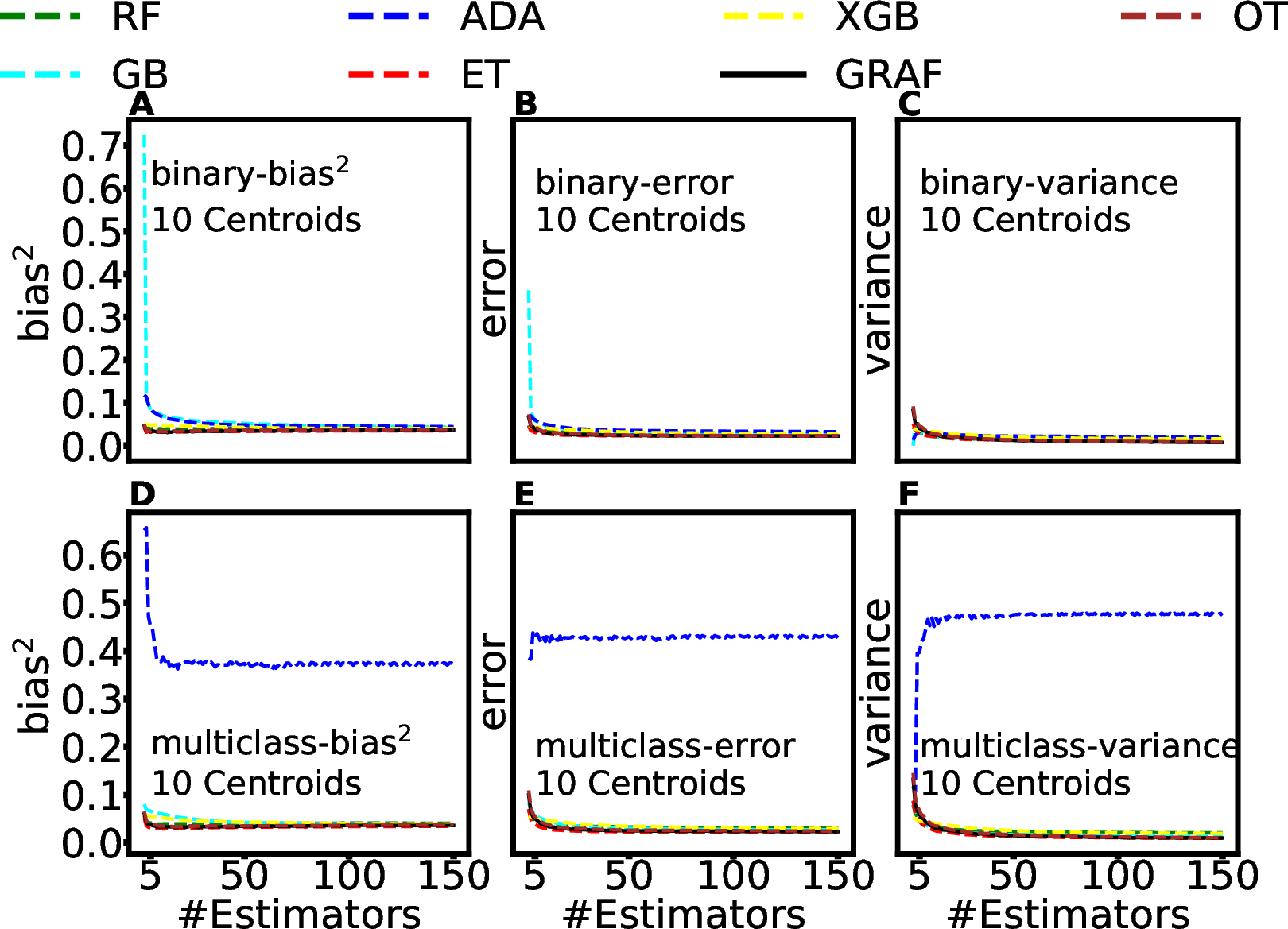}
		}
	}
	\caption{\textbf{Bias-variance analysis with an increasing number of estimators (trees) in a classifier.} For both binary \textbf{A} - \textbf{C)} and multi-class \textbf{D} - \textbf{F)} datasets with 10 centroids, the number of estimators is increased from $2$ to $150$, while fixing the number of dimensions to be sampled $(M = n/2)$. As the number of estimators is increased, bias, error, and variance rapidly saturate.}
	\label{fig:ch4:increasingL}
\end{figure}

Two different studies were performed to evaluate the performance of \acrshort{graf} in terms of bias and variance decomposition. First, the effect of different values of hyper-parameters (namely, number of trees and feature sub-space size) on the bias, variance, and misclassification error rate was analyzed. Second, the trends of bias and variance were observed for increasing train set sizes and compared with different classifiers. To perform these analyses, 6 different binary and multi-class datasets with a different number of centroids from \{10, 20, 50\} were simulated with Weka~\cite{witten1999weka} (Section~\ref{sec:ch4:dataGeneration}). Each dataset consisted of 10000 samples and 10 features (generated using RandomRBF class), while other parameters were set as default. To create the test set, $5000$ samples were randomly selected. For a given train dataset size ($200 \leq N_m \leq 2500$), 50 models were generated by repeatedly sampling without replacement, from the remaining dataset.

\begin{figure}[!ht]
	\centering
	\makebox[1 \textwidth][c]{
		\resizebox{1 \linewidth}{!}{
			\includegraphics[width=\linewidth,keepaspectratio]{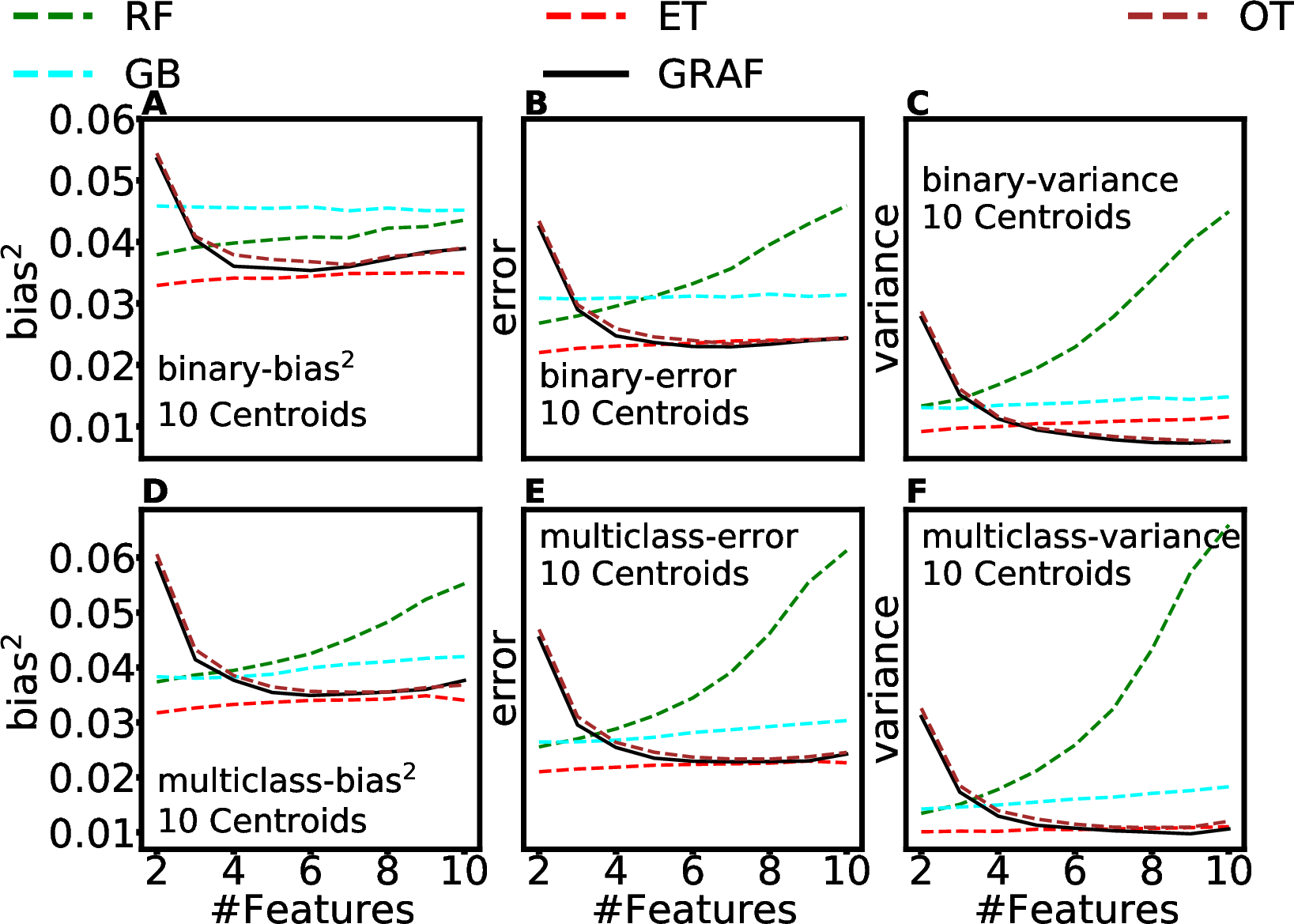}
		}
	}
	\caption{\textbf{Bias-variance analysis with an increasing number of dimensions (features) selected from a given feature space in a classifier.} For both binary \textbf{A} - \textbf{C)} and multi-class \textbf{D} - \textbf{F)} datasets with 10 centroids, $M$  is increased from $2$ to $10$, while fixing the number of estimators to be assembled $(L = 100)$. For \acrshort{graf}, when the dimension of the sub-space is large enough to distinguish samples of different classes, bias and variance saturate and converge to their minimum. With increasing dimensionality of the sub-space, misclassification error continues to decrease and rapidly saturates to its minimum.}
	\label{fig:ch4:increasingM}
\end{figure}

The effect of increasing the number of trees from 2 to 150 for 10 centroids is illustrated in Figure~\ref{fig:ch4:increasingL} (similar trends were also observed for 20 and 50 centroids). For intermediate values of tree numbers, bias-variance curves saturate to their minima, and hence, the average misclassification converges to its minimum. It implies that higher accuracies can be achieved well before all trees are used~\cite{ho1998random}.  Figure~\ref{fig:ch4:increasingM} highlights the effect of increasing the number of randomly selected dimensions/features for 10 centroids (similar trends were also observed for 20 and 50 centroids). This figure shows that a subset of features, in general, may be enough to generate the desired results. However, the selected sub-space must be large enough to distinguish the samples in this sub-space. For these experiments $N_m$ was set to $2500$.

\begin{figure}[!ht]
	\centering
	\makebox[1 \textwidth][c]{
		\resizebox{1 \linewidth}{!}{
			\includegraphics[width=\linewidth,keepaspectratio]{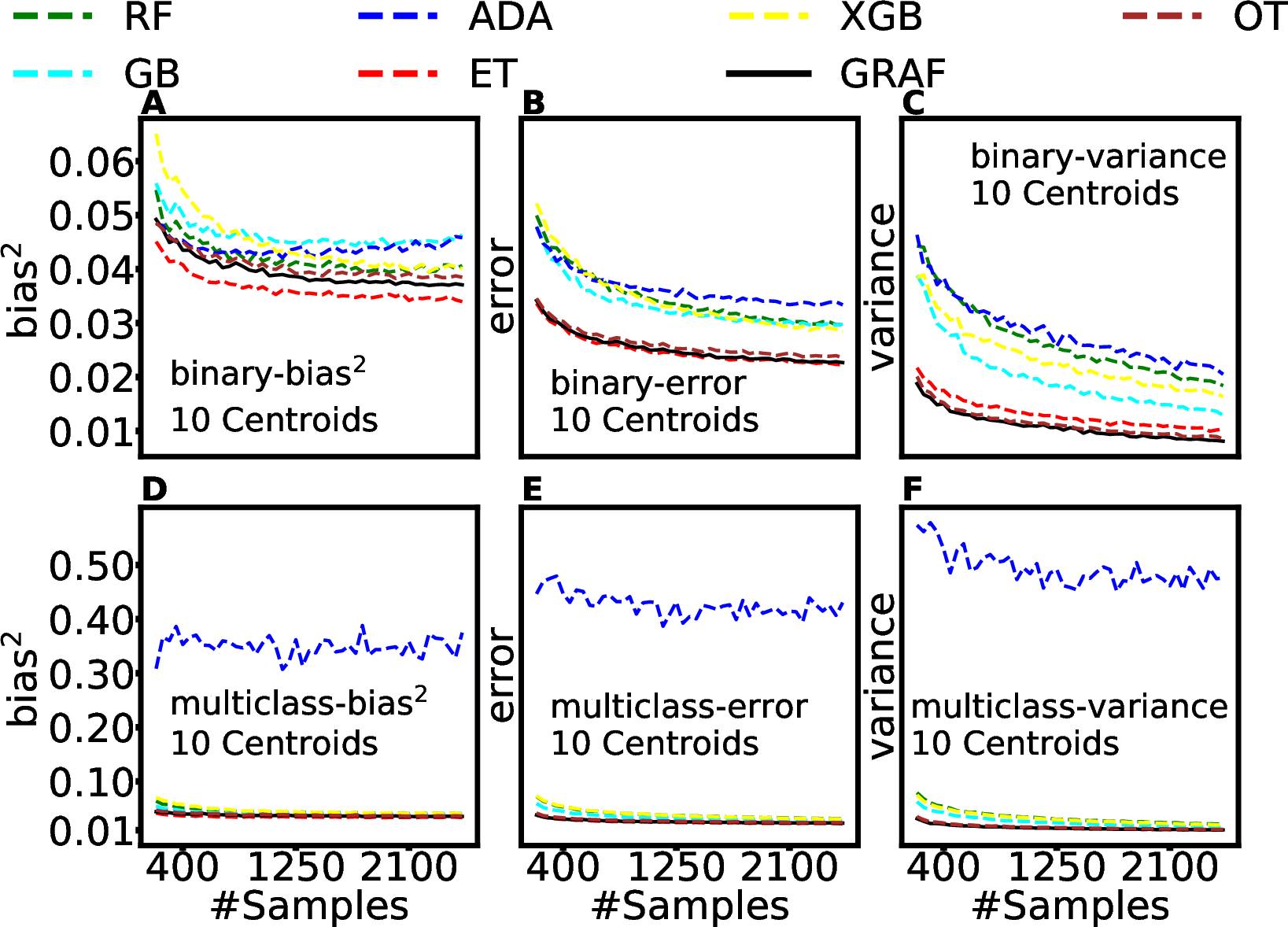}
		}
	}
	\caption{\textbf{Bias-variance analysis with increasing samples in a training set.} For both binary \textbf{A} - \textbf{C)} and multi-class \textbf{D} - \textbf{F)} datasets with 10 centroids, the number of samples is increased from $200$ to $2500$, while fixing the number of dimensions to be sampled $(M = n/2)$ and the number of estimators as $L=100$. As the cardinality of the training set is increased, bias-variance continues to decrease, and the misclassification error continues to decrease and may saturate to its minimum.}
	\label{fig:ch4:comparison50}
\end{figure}

In a different study, the influence of an increasing number of training samples ($200 \leq N_m \leq 2500$) is illustrated in Figure~\ref{fig:ch4:comparison50} for a dataset with 10 centroids (similar trends were also observed for 20 and 50 centroids). Bias and variance decrease with an increase in the size of the training set. In general, \acrshort{graf} was found to have the least variance, and the lowest or comparable misclassification errors on test samples, when compared with other methods (default values of hyper-parameters are used, $L=100$ and $M=5$).

\subsection{Performance comparison on UCI datasets}\label{sec:ch4:perfuci}

The performance of \acrshort{graf} has been evaluated on 115 UCI datasets~\cite{Dua:2017} and compared against \acrfull{rf}~\cite{rf}, \acrfull{gb}~\cite{friedman2001greedy}, \acrfull{ada}~\cite{freund1997decision}, \acrfull{et}~\cite{et}, \acrfull{xgb}~\cite{chen2016xgboost}, and \acrfull{ot}~\cite{menze2011oblique}. Statistics of all 115 datasets are available in Table~\ref{tab:ch4:UCIdataStat}. The total number of samples across all datasets varies from 24 to $\sim 130$k. The count of features across datasets varies from 3 to 262. For comparison, we used the strategy as defined in Fernandez-Delgado \textit{et al.}~\cite{fernandez2014we}\footnote{Fernandez-Delgado et al. concluded that random forest is the best performing algorithm after comparing 179 classifiers. These results may be found at \url{http://persoal.citius.usc.es/manuel.fernandez.delgado/papers/jmlr/ data.tar.gz}}. They use four-fold cross-validation on the whole dataset to compute the performance. The training dataset contains 50\% of the total samples. 

The hyper-parameters are tuned using 5-fold cross-validation on the training dataset. For all methods, the number of estimators is tuned from $\{100, 200, 500, 1000, 2000\}$. For \acrshort{graf}, \acrshort{rf}, \acrshort{gb}, \acrshort{et}, and \acrshort{ot}, the number of dimensions to be selected ($M$) has been tuned from $\{\log_2(n), \sqrt{n}, n/2, n\}$, and the node is further split only if it has minimum samples, tuned between 2 and 5. For \acrshort{graf} and \acrshort{ot}, the number of trials (hyperplane search) $K$ is set to the value of $M$.

On the tuned hyperparameters, as discussed above, the average of the test sets' Cohen's kappa score across 4-folds cross-validation were computed. The same has been tabulated in Table~\ref{tab:ch4:kappaUCI}. For every dataset, the method with the highest score has been highlighted. On 33 datasets, \acrshort{graf} outperforms all other methods. On 87, 66, 77, 71, 101, and 77 datasets, \acrshort{graf}'s performance is either better than or comparable with \acrshort{ot}, \acrshort{et}, \acrshort{rf}, \acrshort{gb}, \acrshort{ada}, and \acrshort{xgb}, respectively. 

As discussed in Section~\ref{sec:ch4:simulation}, oblique partitioning-based trees have a better performance where features are independent and relevant in comparison to axis-aligned partitioning-based trees. To reinforce this, we extend this analysis to UCI datasets as well. Table A11 contains the information about number of principal components (PC) required to explain the 50\%, 70\% and 90\% variance in columns PC(v=0.5), PC(v=0.7) and PC(v=0.9), respectively. \acrshort{graf} has improved performance on datasets (PC(v=0.9)/total features) with a large number of components to explain the high variance, such as adult (12/14), balance-scale (4/4), bank (13/16), congressional-voting (11/16), mammographic (4/5), statlog-australian-credit (11/14), titanic (3/3), waveform (15/21), wine-quality-red (7/11), yeast (7/10), led-display (6/7), etc. when compared with \acrshort{et} and \acrshort{rf}. On the other hand, \acrshort{graf} has either poor or comparable performance on miniboone (2/50), musk-1 (23/66), musk-2 (26/166), statlog-landsat (4/36), plant-margin (25/64), plant-shape (2/64), plant-texture (20/64), etc.

Finally, we analyze the statistical significance of the results. For this, we first subject the results to the Friedman ranking test. In the analysis, the average ranks of 3.07, 3.49, 3.66, 3.82, 3.94, 4.16, and 5.87 were obtained by \acrshort{graf}, \acrshort{et}, \acrshort{ot}, \acrshort{gb}, \acrshort{xgb}, \acrshort{gb}, and \acrshort{ada}, respectively. With 115 datasets and 7 methods, the test statistic of the Friedman test was 117.6689. Assuming a significance level of $0.05$ with 6 degrees of freedom, the value of $\chi^{2}_{6}(0.05)=12.592$ is lesser than the test statistic. Hence, we reject the null hypothesis that all method's performances are similar.

\begin{figure}
	\centering
	\makebox[1 \textwidth][c]{
		\resizebox{1 \linewidth}{!}{
			\includegraphics[width=\linewidth,keepaspectratio]{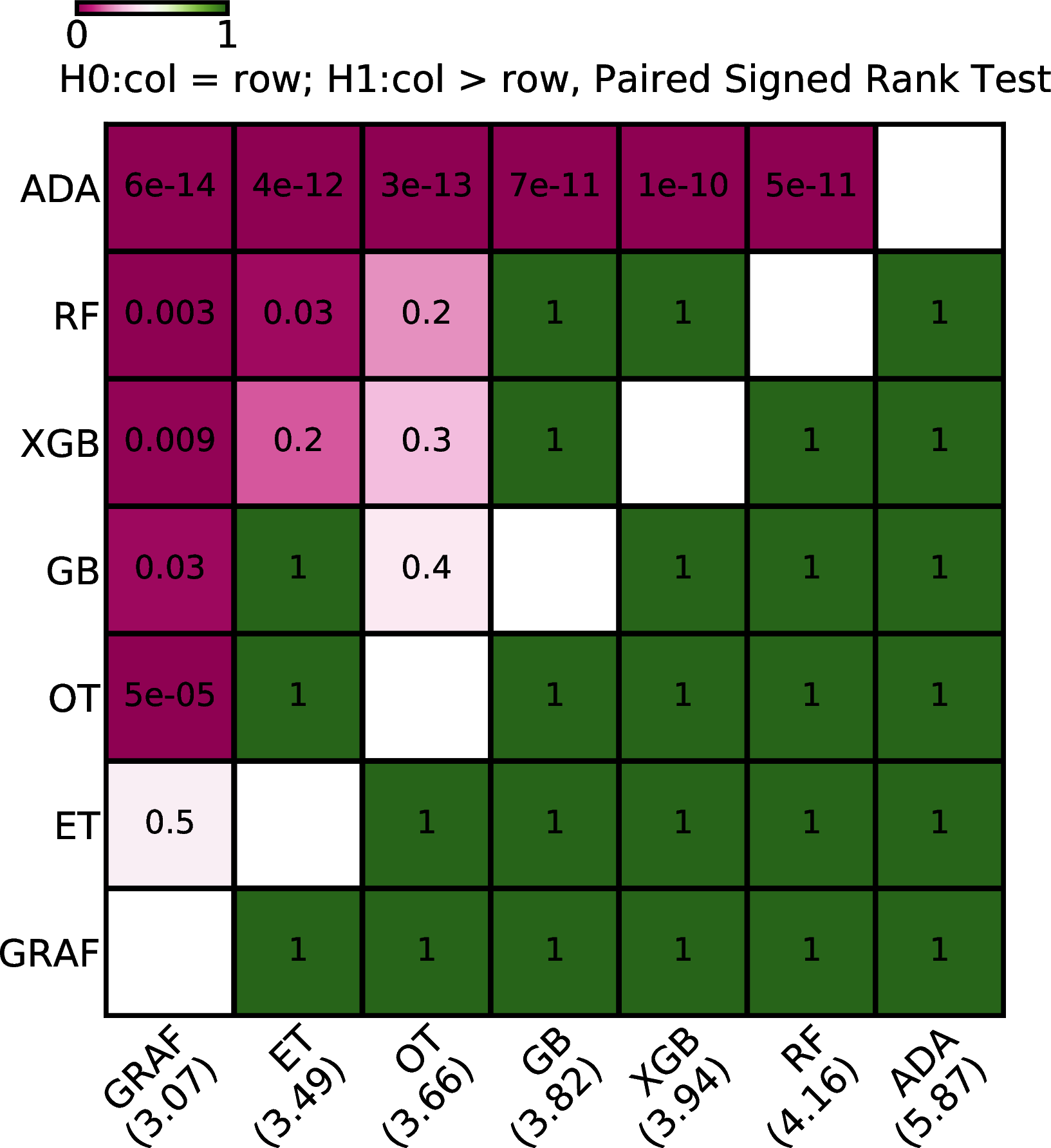}
		}
	}
	\caption{\textbf{One-sided paired Wilcoxon signed-rank test on Cohen's kappa score.} Each method is paired with every other method, and p-value was computed for the null hypothesis '$\text{left method} = \text{right method}$'. Null hypothesis is rejected in favour of hypothesis '$\text{left method} > \text{right method}$', if the corrected p-value is below a certain significance level. The method on the left side (of comparison) is placed on the x-axis, and the method on the right side is placed on the y-axis. Each cell represents the corrected p-value. Hence, every column represents the significance of the kappa score for a method when compared with other methods. Suppose the corrected p-value is less than a certain significance level in a cell. In that case, the null hypothesis is rejected, and the method on the x-axis will be assumed to perform better than the corresponding method on the y-axis. The numerals in the x-axis represent the average Friedman ranking of the method.}
	\label{fig:ch4:pavluechoenkappa}
\end{figure}

\begin{landscape}
	\begin{center}
		\singlespacing
		\begin{longtable}{||c|c|c|c|c|c|c|c||}
		    \caption{Data statistics of 115 UCI datasets. The total number of samples across all datasets varies from 24 to $\sim$130k. The count of features across all datasets varies from 3 to 262.}\\
			\hline
			Dataset & nFeatures & nClasses & nSamples & imbalance & PC(v=0.5) & PC(v=0.7) & PC(v=0.9) \\ [0.5ex] 
			\hline\hline
			abalone&8&3&4177&N&1.0&1.0&2.0\\
			\hline
			acute-inflammation&6&2&120&N&2.0&3.0&4.0\\
			\hline
			acute-nephritis&6&2&120&Y&2.0&3.0&4.0\\
			\hline
			adult&14&2&32561&Y&6.0&9.0&12.0\\
			\hline
			arrhythmia&262&13&452&Y&12.0&25.0&55.0\\
			\hline
			audiology-std&59&18&171&Y&10.0&16.0&26.0\\
			\hline
			balance-scale&4&3&625&Y&2.0&3.0&4.0\\
			\hline
			bank&16&2&4521&Y&6.0&9.0&13.0\\
			\hline
			blood&4&2&748&Y&1.0&2.0&2.0\\
			\hline
			breast-cancer&9&2&286&Y&3.0&5.0&7.0\\
			\hline
			breast-cancer-wisc&9&2&699&Y&1.0&2.0&6.0\\
			\hline
			breast-cancer-wisc-diag&30&2&569&Y&2.0&3.0&7.0\\
			\hline
			breast-cancer-wisc-prog&33&2&198&Y&2.0&4.0&9.0\\
			\hline
			breast-tissue&9&6&106&Y&1.0&2.0&3.0\\
			\hline
			car&6&4&1728&Y&3.0&5.0&6.0\\
			\hline
			cardiotocography-10clases&21&10&2126&Y&3.0&6.0&11.0\\
			\hline
			cardiotocography-3clases&21&3&2126&Y&3.0&6.0&11.0\\
			\hline
			chess-krvk&6&18&28056&Y&3.0&4.0&5.0\\
			\hline
			chess-krvkp&36&2&3196&N&9.0&16.0&26.0\\
			\hline
			congressional-voting&16&2&435&Y&4.0&7.0&11.0\\
			\hline
			conn-bench-sonar-mines-rocks&60&2&208&N&4.0&8.0&20.0\\
			\hline
			conn-bench-vowel-deterding&11&11&528&N&3.0&4.0&7.0\\
			\hline
			connect-4&42&2&67557&Y&9.0&17.0&31.0\\
			\hline
			contrac&9&3&1473&Y&3.0&5.0&7.0\\
			\hline
			credit-approval&15&2&690&Y&4.0&7.0&11.0\\
			\hline
			cylinder-bands&35&2&512&Y&6.0&12.0&21.0\\
			\hline
			dermatology&34&6&366&Y&3.0&8.0&16.0\\
			\hline
			echocardiogram&10&2&131&Y&3.0&5.0&7.0\\
			\hline
			energy-y1&8&3&768&Y&2.0&3.0&5.0\\
			\hline
			energy-y2&8&3&768&Y&2.0&3.0&5.0\\
			\hline
			fertility&9&2&100&Y&3.0&5.0&7.0\\
			\hline
			glass&9&6&214&Y&2.0&3.0&5.0\\
			\hline
			haberman-survival&3&2&306&Y&2.0&2.0&3.0\\
			\hline
			hayes-roth&3&3&132&Y&2.0&2.0&3.0\\
			\hline
			heart-cleveland&13&5&303&Y&4.0&6.0&10.0\\
			\hline
			heart-hungarian&12&2&294&Y&3.0&6.0&9.0\\
			\hline
			heart-switzerland&12&5&123&Y&4.0&6.0&9.0\\
			\hline
			heart-va&12&5&200&Y&3.0&5.0&8.0\\
			\hline
			hepatitis&19&2&155&Y&4.0&7.0&13.0\\
			\hline
			hill-valley&100&2&606&N&1.0&1.0&1.0\\
			\hline
			horse-colic&25&2&300&Y&5.0&10.0&18.0\\
			\hline
			ilpd-indian-liver&9&2&583&Y&2.0&3.0&5.0\\
			\hline
			image-segmentation&18&7&210&N&1.0&3.0&6.0\\
			\hline
			ionosphere&33&2&351&Y&4.0&8.0&16.0\\
			\hline
			iris&4&3&150&N&1.0&1.0&2.0\\
			\hline
			led-display&7&10&1000&Y&3.0&4.0&6.0\\
			\hline
			lenses&4&3&24&Y&2.0&3.0&4.0\\
			\hline
			letter&16&26&20000&N&3.0&6.0&10.0\\
			\hline
			libras&90&15&360&N&3.0&4.0&7.0\\
			\hline
			low-res-spect&100&9&531&Y&1.0&2.0&4.0\\
			\hline
			lung-cancer&56&3&32&Y&4.0&7.0&11.0\\
			\hline
			lymphography&18&4&148&Y&4.0&7.0&12.0\\
			\hline
			magic&10&2&19020&Y&2.0&4.0&6.0\\
			\hline
			mammographic&5&2&961&Y&2.0&3.0&4.0\\
			\hline
			miniboone&50&2&130064&Y&1.0&1.0&3.0\\
			\hline
			molec-biol-promoter&57&2&106&N&10.0&16.0&27.0\\
			\hline
			molec-biol-splice&60&3&3190&Y&24.0&37.0&51.0\\
			\hline
			monks-1&6&2&124&N&3.0&4.0&6.0\\
			\hline
			monks-2&6&2&169&Y&3.0&4.0&6.0\\
			\hline
			monks-3&6&2&122&N&3.0&4.0&5.0\\
			\hline
			mushroom&21&2&8124&N&4.0&7.0&13.0\\
			\hline
			musk-1&166&2&476&Y&3.0&7.0&23.0\\
			\hline
			musk-2&166&2&6598&Y&3.0&9.0&26.0\\
			\hline
			nursery&8&5&12960&Y&4.0&6.0&8.0\\
			\hline
			oocytes\_merluccius\_nucleus\_4d&41&2&1022&Y&1.0&1.0&3.0\\
			\hline
			oocytes\_merluccius\_states\_2f&25&3&1022&Y&2.0&3.0&5.0\\
			\hline
			oocytes\_trisopterus\_nucleus\_2f&25&2&912&Y&2.0&3.0&5.0\\
			\hline
			oocytes\_trisopterus\_states\_5b&32&3&912&Y&1.0&2.0&5.0\\
			\hline
			optical&62&10&3823&N&8.0&15.0&30.0\\
			\hline
			ozone&72&2&2536&Y&2.0&4.0&12.0\\
			\hline
			page-blocks&10&5&5473&Y&2.0&3.0&5.0\\
			\hline
			parkinsons&22&2&195&Y&1.0&2.0&6.0\\
			\hline
			pendigits&16&10&7494&N&3.0&4.0&8.0\\
			\hline
			pima&8&2&768&Y&3.0&4.0&6.0\\
			\hline
			pittsburg-bridges-MATERIAL&7&3&106&Y&3.0&4.0&6.0\\
			\hline
			pittsburg-bridges-REL-L&7&3&103&Y&2.0&4.0&6.0\\
			\hline
			pittsburg-bridges-SPAN&7&3&92&Y&2.0&4.0&6.0\\
			\hline
			pittsburg-bridges-T-OR-D&7&2&102&Y&3.0&4.0&6.0\\
			\hline
			pittsburg-bridges-TYPE&7&6&105&Y&2.0&4.0&6.0\\
			\hline
			planning&12&2&182&Y&3.0&4.0&5.0\\
			\hline
			plant-margin&64&100&1600&N&4.0&8.0&25.0\\
			\hline
			plant-shape&64&100&1600&N&1.0&1.0&2.0\\
			\hline
			plant-texture&64&100&1599&N&6.0&13.0&30.0\\
			\hline
			post-operative&8&3&90&Y&3.0&4.0&6.0\\
			\hline
			ringnorm&20&2&7400&N&10.0&14.0&18.0\\
			\hline
			seeds&7&3&210&N&1.0&1.0&3.0\\
			\hline
			semeion&256&10&1593&N&16.0&36.0&103.0\\
			\hline
			soybean&35&18&307&Y&5.0&10.0&19.0\\
			\hline
			spambase&57&2&4601&Y&15.0&26.0&41.0\\
			\hline
			spect&22&2&79&Y&3.0&6.0&11.0\\
			\hline
			spectf&44&2&80&N&2.0&3.0&10.0\\
			\hline
			statlog-australian-credit&14&2&690&Y&4.0&7.0&10.0\\
			\hline
			statlog-german-credit&24&2&1000&Y&7.0&11.0&18.0\\
			\hline
			statlog-heart&13&2&270&Y&4.0&6.0&10.0\\
			\hline
			statlog-image&18&7&2310&N&2.0&4.0&8.0\\
			\hline
			statlog-landsat&36&6&4435&Y&2.0&2.0&4.0\\
			\hline
			statlog-shuttle&9&7&43500&Y&3.0&4.0&6.0\\
			\hline
			statlog-vehicle&18&4&846&N&1.0&2.0&5.0\\
			\hline
			steel-plates&27&7&1941&Y&3.0&5.0&10.0\\
			\hline
			synthetic-control&60&6&600&N&1.0&4.0&18.0\\
			\hline
			teaching&5&3&151&N&2.0&3.0&5.0\\
			\hline
			thyroid&21&3&3772&Y&7.0&11.0&16.0\\
			\hline
			tic-tac-toe&9&2&958&Y&4.0&5.0&7.0\\
			\hline
			titanic&3&2&2201&Y&2.0&2.0&3.0\\
			\hline
			twonorm&20&2&7400&N&8.0&13.0&18.0\\
			\hline
			vertebral-column-2clases&6&2&310&Y&1.0&2.0&4.0\\
			\hline
			vertebral-column-3clases&6&3&310&Y&1.0&2.0&3.0\\
			\hline
			wall-following&24&4&5456&Y&5.0&10.0&18.0\\
			\hline
			waveform&21&3&5000&N&2.0&6.0&15.0\\
			\hline
			waveform-noise&40&3&5000&N&10.0&19.0&29.0\\
			\hline
			wine&13&3&178&Y&2.0&4.0&7.0\\
			\hline
			wine-quality-red&11&6&1599&Y&3.0&4.0&7.0\\
			\hline
			wine-quality-white&11&7&4898&Y&3.0&5.0&8.0\\
			\hline
			yeast&8&10&1484&Y&3.0&5.0&7.0\\
			\hline
			zoo&16&7&101&Y&2.0&4.0&8.0\\
			\hline
			
			\label{tab:ch4:UCIdataStat}
		\end{longtable}
	\end{center}
\end{landscape}

\begin{landscape}
	\begin{center}
		\singlespacing
		\begin{longtable}{||c|c|c|c|c|c|c|c||}
			\caption{The performance of methods is compared on 115 UCI datasets using Cohen's kappa coefficient.}\\
			\hline
			Dataset & GRAF & OT & ET & GB & ADA & RF & XGB \\
			\hline\hline
			abalone&0.488$\pm$0.005&\textbf{0.492$\pm$0.016}&0.484$\pm$0.009&0.469$\pm$0.007&0.458$\pm$0.008&0.483$\pm$0.016&0.466$\pm$0.013\\
			\hline
			acute-inflammation&\textbf{1.000$\pm$0.000}&\textbf{1.000$\pm$0.000}&\textbf{1.000$\pm$0.000}&\textbf{1.000$\pm$0.000}&\textbf{1.000$\pm$0.000}&\textbf{1.000$\pm$0.000}&\textbf{1.000$\pm$0.000}\\
			\hline
			acute-nephritis&\textbf{1.000$\pm$0.000}&\textbf{1.000$\pm$0.000}&\textbf{1.000$\pm$0.000}&\textbf{1.000$\pm$0.000}&\textbf{1.000$\pm$0.000}&\textbf{1.000$\pm$0.000}&\textbf{1.000$\pm$0.000}\\
			\hline
			adult&0.602$\pm$0.005&0.600$\pm$0.004&0.571$\pm$0.005&\textbf{0.631$\pm$0.003}&0.625$\pm$0.004&0.594$\pm$0.002&0.630$\pm$0.004\\
			\hline
			arrhythmia&0.403$\pm$0.058&0.342$\pm$0.026&\textbf{0.628$\pm$0.046}&0.567$\pm$0.023&0.258$\pm$0.049&0.614$\pm$0.032&0.582$\pm$0.020\\
			\hline
			audiology-std&\textbf{0.843$\pm$0.039}&0.741$\pm$0.061&0.785$\pm$0.075&0.800$\pm$0.064&0.605$\pm$0.071&0.785$\pm$0.083&0.792$\pm$0.071\\
			\hline
			balance-scale&0.842$\pm$0.027&0.850$\pm$0.029&0.755$\pm$0.021&0.860$\pm$0.044&\textbf{0.887$\pm$0.028}&0.763$\pm$0.028&0.807$\pm$0.022\\
			\hline
			bank&\textbf{0.465$\pm$0.052}&0.382$\pm$0.040&0.317$\pm$0.048&0.391$\pm$0.018&0.338$\pm$0.034&0.391$\pm$0.057&0.405$\pm$0.023\\
			\hline
			blood&0.265$\pm$0.029&\textbf{0.291$\pm$0.044}&0.233$\pm$0.085&0.238$\pm$0.078&0.089$\pm$0.008&0.233$\pm$0.089&0.212$\pm$0.068\\
			\hline
			breast-cancer&\textbf{0.438$\pm$0.037}&0.433$\pm$0.043&0.353$\pm$0.059&0.278$\pm$0.134&0.283$\pm$0.101&0.336$\pm$0.079&0.319$\pm$0.135\\
			\hline
			breast-cancer-wisc&0.953$\pm$0.016&\textbf{0.956$\pm$0.018}&0.950$\pm$0.015&0.931$\pm$0.020&0.944$\pm$0.018&0.947$\pm$0.013&0.934$\pm$0.028\\
			\hline
			breast-cancer-wisc-diag&\textbf{0.947$\pm$0.023}&0.935$\pm$0.027&0.928$\pm$0.027&0.928$\pm$0.023&0.920$\pm$0.020&0.909$\pm$0.018&0.936$\pm$0.023\\
			\hline
			breast-cancer-wisc-prog&\textbf{0.431$\pm$0.042}&0.418$\pm$0.048&0.325$\pm$0.071&0.322$\pm$0.121&0.212$\pm$0.211&0.263$\pm$0.153&0.241$\pm$0.172\\
			\hline
			breast-tissue&0.706$\pm$0.085&\textbf{0.718$\pm$0.067}&0.647$\pm$0.097&0.590$\pm$0.134&0.451$\pm$0.061&0.684$\pm$0.111&0.613$\pm$0.134\\
			\hline
			car&0.941$\pm$0.020&0.922$\pm$0.012&0.968$\pm$0.009&0.986$\pm$0.013&0.719$\pm$0.008&0.975$\pm$0.006&\textbf{0.987$\pm$0.009}\\
			\hline
			cardiotocography-10clases&0.800$\pm$0.015&0.800$\pm$0.011&0.841$\pm$0.017&0.868$\pm$0.011&0.637$\pm$0.038&0.841$\pm$0.012&\textbf{0.874$\pm$0.011}\\
			\hline
			cardiotocography-3clases&0.791$\pm$0.031&0.773$\pm$0.024&0.867$\pm$0.018&\textbf{0.885$\pm$0.023}&0.709$\pm$0.014&0.848$\pm$0.022&0.878$\pm$0.025\\
			\hline
			chess-krvk&0.694$\pm$0.002&0.626$\pm$0.003&0.858$\pm$0.004&\textbf{0.911$\pm$0.001}&0.120$\pm$0.006&0.855$\pm$0.003&0.907$\pm$0.002\\
			\hline
			chess-krvkp&0.955$\pm$0.017&0.953$\pm$0.016&\textbf{0.994$\pm$0.004}&0.993$\pm$0.003&0.940$\pm$0.011&0.990$\pm$0.007&0.991$\pm$0.004\\
			\hline
			congressional-voting&0.212$\pm$0.032&\textbf{0.215$\pm$0.036}&0.003$\pm$0.033&0.048$\pm$0.064&0.031$\pm$0.039&0.030$\pm$0.042&0.035$\pm$0.046\\
			\hline
			conn-bench-sonar-mines-rocks&0.697$\pm$0.070&0.667$\pm$0.020&\textbf{0.765$\pm$0.049}&0.605$\pm$0.049&0.582$\pm$0.055&0.608$\pm$0.088&0.747$\pm$0.088\\
			\hline
			conn-bench-vowel-deterding&0.975$\pm$0.013&0.975$\pm$0.012&\textbf{0.979$\pm$0.018}&0.944$\pm$0.012&0.562$\pm$0.019&0.958$\pm$0.018&0.871$\pm$0.022\\
			\hline
			connect-4&0.677$\pm$0.003&0.678$\pm$0.004&0.663$\pm$0.002&0.750$\pm$0.003&0.499$\pm$0.003&0.620$\pm$0.002&\textbf{0.763$\pm$0.005}\\
			\hline
			contrac&0.275$\pm$0.013&0.273$\pm$0.022&0.242$\pm$0.035&0.294$\pm$0.047&0.281$\pm$0.026&0.261$\pm$0.040&\textbf{0.302$\pm$0.050}\\
			\hline
			credit-approval&\textbf{0.784$\pm$0.048}&0.769$\pm$0.039&0.720$\pm$0.071&0.768$\pm$0.052&0.717$\pm$0.048&0.754$\pm$0.017&0.738$\pm$0.042\\
			\hline
			cylinder-bands&0.560$\pm$0.022&0.557$\pm$0.034&0.594$\pm$0.026&0.601$\pm$0.062&0.489$\pm$0.067&0.583$\pm$0.051&\textbf{0.642$\pm$0.046}\\
			\hline
			dermatology&\textbf{0.976$\pm$0.011}&\textbf{0.976$\pm$0.011}&\textbf{0.976$\pm$0.006}&0.969$\pm$0.006&0.917$\pm$0.022&0.972$\pm$0.010&0.962$\pm$0.006\\
			\hline
			echocardiogram&\textbf{0.630$\pm$0.110}&0.603$\pm$0.048&0.579$\pm$0.023&0.524$\pm$0.098&0.606$\pm$0.041&0.578$\pm$0.038&0.518$\pm$0.154\\
			\hline
			energy-y1&0.912$\pm$0.015&0.908$\pm$0.019&0.945$\pm$0.030&0.935$\pm$0.011&0.682$\pm$0.010&\textbf{0.950$\pm$0.006}&0.945$\pm$0.004\\
			\hline
			energy-y2&0.848$\pm$0.015&0.852$\pm$0.013&0.835$\pm$0.031&\textbf{0.871$\pm$0.021}&0.785$\pm$0.007&0.842$\pm$0.020&0.846$\pm$0.018\\
			\hline
			fertility&0.335$\pm$0.230&\textbf{0.351$\pm$0.203}&0.218$\pm$0.251&0.285$\pm$0.358&0.000$\pm$0.000&0.201$\pm$0.206&0.229$\pm$0.138\\
			\hline
			glass&\textbf{0.737$\pm$0.113}&0.718$\pm$0.109&0.679$\pm$0.049&0.667$\pm$0.054&0.462$\pm$0.074&0.699$\pm$0.040&0.689$\pm$0.073\\
			\hline
			haberman-survival&\textbf{0.249$\pm$0.106}&0.235$\pm$0.101&0.091$\pm$0.024&0.098$\pm$0.054&0.172$\pm$0.107&0.049$\pm$0.060&0.241$\pm$0.054\\
			\hline
			hayes-roth&0.754$\pm$0.050&0.754$\pm$0.050&0.742$\pm$0.069&0.741$\pm$0.072&\textbf{0.778$\pm$0.062}&0.754$\pm$0.052&0.765$\pm$0.090\\
			\hline
			heart-cleveland&\textbf{0.323$\pm$0.051}&0.295$\pm$0.050&0.304$\pm$0.027&0.279$\pm$0.089&0.245$\pm$0.060&0.285$\pm$0.067&0.256$\pm$0.065\\
			\hline
			heart-hungarian&\textbf{0.696$\pm$0.066}&0.686$\pm$0.068&0.653$\pm$0.063&0.613$\pm$0.074&0.624$\pm$0.037&0.653$\pm$0.070&0.597$\pm$0.042\\
			\hline
			heart-switzerland&0.250$\pm$0.061&\textbf{0.253$\pm$0.107}&0.121$\pm$0.033&0.088$\pm$0.054&0.134$\pm$0.105&0.132$\pm$0.094&0.096$\pm$0.048\\
			\hline
			heart-va&\textbf{0.175$\pm$0.040}&0.148$\pm$0.061&0.073$\pm$0.072&0.110$\pm$0.054&-0.023$\pm$0.055&0.153$\pm$0.042&0.098$\pm$0.081\\
			\hline
			hepatitis&\textbf{0.648$\pm$0.086}&0.618$\pm$0.094&0.366$\pm$0.133&0.370$\pm$0.122&0.495$\pm$0.130&0.407$\pm$0.120&0.249$\pm$0.153\\
			\hline
			hill-valley&-0.029$\pm$0.064&-0.029$\pm$0.064&0.064$\pm$0.034&0.050$\pm$0.041&0.089$\pm$0.051&0.071$\pm$0.023&\textbf{0.104$\pm$0.042}\\
			\hline
			horse-colic&0.674$\pm$0.086&0.669$\pm$0.074&0.691$\pm$0.046&\textbf{0.704$\pm$0.032}&0.664$\pm$0.087&0.684$\pm$0.096&0.630$\pm$0.088\\
			\hline
			ilpd-indian-liver&0.252$\pm$0.018&\textbf{0.269$\pm$0.043}&0.237$\pm$0.050&0.205$\pm$0.062&0.221$\pm$0.041&0.146$\pm$0.021&0.185$\pm$0.079\\
			\hline
			image-segmentation&0.921$\pm$0.025&0.921$\pm$0.019&\textbf{0.932$\pm$0.028}&0.893$\pm$0.051&0.615$\pm$0.073&0.916$\pm$0.029&0.899$\pm$0.061\\
			\hline
			ionosphere&\textbf{0.881$\pm$0.011}&0.868$\pm$0.011&0.866$\pm$0.029&0.866$\pm$0.043&0.804$\pm$0.047&0.818$\pm$0.048&0.836$\pm$0.029\\
			\hline
			iris&0.949$\pm$0.018&\textbf{0.959$\pm$0.041}&0.949$\pm$0.018&0.949$\pm$0.018&0.929$\pm$0.018&0.919$\pm$0.029&0.939$\pm$0.020\\
			\hline
			led-display&\textbf{0.725$\pm$0.013}&\textbf{0.725$\pm$0.013}&0.681$\pm$0.031&0.718$\pm$0.023&0.694$\pm$0.018&0.704$\pm$0.026&0.720$\pm$0.021\\
			\hline
			lenses&0.662$\pm$0.239&0.583$\pm$0.433&0.583$\pm$0.433&0.583$\pm$0.433&\textbf{0.762$\pm$0.274}&0.583$\pm$0.433&0.583$\pm$0.433\\
			\hline
			letter&0.953$\pm$0.001&0.939$\pm$0.002&\textbf{0.973$\pm$0.001}&0.966$\pm$0.002&0.348$\pm$0.018&0.964$\pm$0.002&0.964$\pm$0.001\\
			\hline
			libras&\textbf{0.848$\pm$0.020}&0.836$\pm$0.037&0.833$\pm$0.029&0.729$\pm$0.021&0.327$\pm$0.080&0.792$\pm$0.021&0.714$\pm$0.040\\
			\hline
			low-res-spect&0.829$\pm$0.037&0.812$\pm$0.022&0.857$\pm$0.022&\textbf{0.872$\pm$0.033}&0.684$\pm$0.037&0.860$\pm$0.028&0.866$\pm$0.034\\
			\hline
			lung-cancer&0.309$\pm$0.093&0.327$\pm$0.291&\textbf{0.360$\pm$0.151}&0.319$\pm$0.211&0.339$\pm$0.160&0.219$\pm$0.136&0.160$\pm$0.216\\
			\hline
			lymphography&\textbf{0.814$\pm$0.109}&0.804$\pm$0.124&0.631$\pm$0.067&0.748$\pm$0.092&0.509$\pm$0.065&0.721$\pm$0.113&0.778$\pm$0.123\\
			\hline
			magic&0.686$\pm$0.005&0.665$\pm$0.008&0.711$\pm$0.002&\textbf{0.727$\pm$0.005}&0.655$\pm$0.007&0.714$\pm$0.007&0.721$\pm$0.007\\
			\hline
			mammographic&0.663$\pm$0.006&\textbf{0.670$\pm$0.012}&0.572$\pm$0.025&0.632$\pm$0.028&0.593$\pm$0.052&0.584$\pm$0.026&0.632$\pm$0.023\\
			\hline
			miniboone&0.752$\pm$0.003&0.745$\pm$0.002&0.852$\pm$0.001&0.873$\pm$0.002&0.817$\pm$0.004&0.843$\pm$0.001&\textbf{0.875$\pm$0.001}\\
			\hline
			molec-biol-promoter&0.827$\pm$0.084&0.827$\pm$0.100&\textbf{0.865$\pm$0.084}&0.827$\pm$0.064&0.731$\pm$0.139&0.808$\pm$0.038&0.827$\pm$0.114\\
			\hline
			molec-biol-splice&0.714$\pm$0.025&0.698$\pm$0.030&0.926$\pm$0.020&0.937$\pm$0.004&0.888$\pm$0.006&0.922$\pm$0.014&\textbf{0.938$\pm$0.012}\\
			\hline
			monks-1&0.807$\pm$0.121&0.791$\pm$0.147&0.806$\pm$0.177&\textbf{0.935$\pm$0.079}&0.314$\pm$0.072&0.790$\pm$0.084&0.807$\pm$0.222\\
			\hline
			monks-2&0.558$\pm$0.086&\textbf{0.561$\pm$0.092}&0.431$\pm$0.151&0.496$\pm$0.128&0.000$\pm$0.000&0.188$\pm$0.120&0.173$\pm$0.156\\
			\hline
			monks-3&\textbf{0.917$\pm$0.055}&\textbf{0.917$\pm$0.055}&0.817$\pm$0.128&0.833$\pm$0.100&0.900$\pm$0.075&0.900$\pm$0.033&0.900$\pm$0.075\\
			\hline
			mushroom&\textbf{1.000$\pm$0.000}&\textbf{1.000$\pm$0.000}&\textbf{1.000$\pm$0.000}&\textbf{1.000$\pm$0.000}&\textbf{1.000$\pm$0.000}&\textbf{1.000$\pm$0.000}&\textbf{1.000$\pm$0.000}\\
			\hline
			musk-1&0.805$\pm$0.057&\textbf{0.810$\pm$0.061}&0.779$\pm$0.080&0.794$\pm$0.072&0.778$\pm$0.066&0.766$\pm$0.083&0.650$\pm$0.093\\
			\hline
			musk-2&0.919$\pm$0.008&0.914$\pm$0.006&0.946$\pm$0.005&\textbf{0.982$\pm$0.002}&0.968$\pm$0.005&0.915$\pm$0.004&0.970$\pm$0.006\\
			\hline
			nursery&0.956$\pm$0.005&0.945$\pm$0.005&0.996$\pm$0.001&\textbf{1.000$\pm$0.000}&0.742$\pm$0.006&0.995$\pm$0.001&\textbf{1.000$\pm$0.000}\\
			\hline
			oocytes\_merluccius\_nucleus\_4d&0.431$\pm$0.055&0.390$\pm$0.079&\textbf{0.562$\pm$0.074}&0.544$\pm$0.069&0.475$\pm$0.036&0.474$\pm$0.087&0.529$\pm$0.052\\
			\hline
			oocytes\_merluccius\_states\_2f&0.821$\pm$0.016&0.805$\pm$0.026&\textbf{0.836$\pm$0.025}&0.822$\pm$0.020&0.758$\pm$0.033&0.823$\pm$0.025&0.827$\pm$0.021\\
			\hline
			oocytes\_trisopterus\_nucleus\_2f&0.595$\pm$0.032&0.559$\pm$0.027&\textbf{0.639$\pm$0.033}&0.605$\pm$0.025&0.550$\pm$0.025&0.619$\pm$0.028&0.619$\pm$0.038\\
			\hline
			oocytes\_trisopterus\_states\_5b&0.840$\pm$0.032&0.824$\pm$0.022&0.839$\pm$0.021&0.857$\pm$0.031&0.613$\pm$0.031&0.839$\pm$0.017&\textbf{0.864$\pm$0.013}\\
			\hline
			optical&0.973$\pm$0.006&0.959$\pm$0.006&\textbf{0.983$\pm$0.002}&0.981$\pm$0.004&0.872$\pm$0.012&0.981$\pm$0.004&0.974$\pm$0.002\\
			\hline
			ozone&\textbf{0.256$\pm$0.050}&0.045$\pm$0.080&-0.001$\pm$0.001&0.025$\pm$0.045&0.000$\pm$0.000&-0.001$\pm$0.001&0.213$\pm$0.056\\
			\hline
			page-blocks&0.824$\pm$0.020&0.795$\pm$0.021&0.848$\pm$0.021&0.852$\pm$0.023&0.524$\pm$0.083&0.845$\pm$0.023&\textbf{0.856$\pm$0.024}\\
			\hline
			parkinsons&0.767$\pm$0.117&0.741$\pm$0.121&\textbf{0.786$\pm$0.069}&0.721$\pm$0.085&0.659$\pm$0.171&0.692$\pm$0.097&0.752$\pm$0.056\\
			\hline
			pendigits&0.990$\pm$0.002&0.990$\pm$0.002&\textbf{0.994$\pm$0.002}&0.992$\pm$0.001&0.770$\pm$0.008&0.989$\pm$0.001&0.990$\pm$0.002\\
			\hline
			pima&0.448$\pm$0.030&\textbf{0.458$\pm$0.026}&0.427$\pm$0.031&0.423$\pm$0.044&0.383$\pm$0.012&0.451$\pm$0.060&0.437$\pm$0.042\\
			\hline
			pittsburg-bridges-MATERIAL&0.846$\pm$0.051&0.828$\pm$0.070&\textbf{0.849$\pm$0.077}&0.736$\pm$0.098&0.721$\pm$0.051&0.735$\pm$0.085&0.695$\pm$0.056\\
			\hline
			pittsburg-bridges-REL-L&0.611$\pm$0.047&\textbf{0.626$\pm$0.084}&0.573$\pm$0.083&0.387$\pm$0.147&0.505$\pm$0.049&0.456$\pm$0.110&0.414$\pm$0.119\\
			\hline
			pittsburg-bridges-SPAN&\textbf{0.534$\pm$0.164}&0.512$\pm$0.178&0.435$\pm$0.131&0.445$\pm$0.088&0.282$\pm$0.051&0.348$\pm$0.081&0.366$\pm$0.115\\
			\hline
			pittsburg-bridges-T-OR-D&0.266$\pm$0.249&0.282$\pm$0.340&0.296$\pm$0.346&\textbf{0.503$\pm$0.212}&0.234$\pm$0.234&0.318$\pm$0.191&0.356$\pm$0.229\\
			\hline
			pittsburg-bridges-TYPE&0.541$\pm$0.100&0.541$\pm$0.100&\textbf{0.565$\pm$0.122}&0.483$\pm$0.117&0.249$\pm$0.098&0.539$\pm$0.131&0.437$\pm$0.075\\
			\hline
			planning&\textbf{0.104$\pm$0.073}&0.082$\pm$0.089&0.024$\pm$0.043&-0.020$\pm$0.046&0.000$\pm$0.000&0.001$\pm$0.089&-0.098$\pm$0.036\\
			\hline
			plant-margin&0.841$\pm$0.017&0.816$\pm$0.008&\textbf{0.885$\pm$0.007}&0.708$\pm$0.010&0.360$\pm$0.030&0.859$\pm$0.007&0.711$\pm$0.006\\
			\hline
			plant-shape&0.655$\pm$0.013&0.587$\pm$0.013&\textbf{0.665$\pm$0.011}&0.456$\pm$0.017&0.192$\pm$0.014&0.642$\pm$0.018&0.533$\pm$0.029\\
			\hline
			plant-texture&0.811$\pm$0.008&0.788$\pm$0.007&\textbf{0.846$\pm$0.007}&0.516$\pm$0.300&0.407$\pm$0.021&0.838$\pm$0.011&0.718$\pm$0.012\\
			\hline
			post-operative&-0.069$\pm$0.193&\textbf{-0.032$\pm$0.238}&-0.115$\pm$0.126&-0.215$\pm$0.161&-0.132$\pm$0.099&-0.091$\pm$0.114&-0.081$\pm$0.171\\
			\hline
			ringnorm&\textbf{0.968$\pm$0.002}&\textbf{0.968$\pm$0.002}&0.965$\pm$0.003&0.958$\pm$0.008&0.962$\pm$0.005&0.918$\pm$0.006&0.958$\pm$0.005\\
			\hline
			seeds&0.913$\pm$0.054&0.899$\pm$0.052&\textbf{0.942$\pm$0.046}&0.906$\pm$0.043&0.492$\pm$0.012&0.913$\pm$0.035&0.906$\pm$0.043\\
			\hline
			semeion&0.939$\pm$0.017&0.937$\pm$0.015&0.948$\pm$0.016&\textbf{0.951$\pm$0.013}&0.768$\pm$0.022&0.947$\pm$0.017&0.916$\pm$0.015\\
			\hline
			soybean&0.924$\pm$0.012&0.920$\pm$0.016&\textbf{0.942$\pm$0.023}&0.905$\pm$0.030&0.722$\pm$0.062&0.927$\pm$0.023&0.916$\pm$0.026\\
			\hline
			spambase&0.891$\pm$0.008&0.886$\pm$0.004&0.908$\pm$0.007&\textbf{0.910$\pm$0.002}&0.891$\pm$0.009&0.906$\pm$0.003&0.906$\pm$0.010\\
			\hline
			spect&0.267$\pm$0.067&0.295$\pm$0.312&0.081$\pm$0.150&0.079$\pm$0.176&\textbf{0.376$\pm$0.095}&0.246$\pm$0.159&0.083$\pm$0.136\\
			\hline
			spectf&0.600$\pm$0.100&\textbf{0.675$\pm$0.109}&0.550$\pm$0.087&0.300$\pm$0.187&0.400$\pm$0.122&0.450$\pm$0.112&0.500$\pm$0.158\\
			\hline
			statlog-australian-credit&0.171$\pm$0.094&\textbf{0.180$\pm$0.050}&0.153$\pm$0.043&0.162$\pm$0.049&-0.005$\pm$0.018&0.122$\pm$0.055&0.164$\pm$0.084\\
			\hline
			statlog-german-credit&0.422$\pm$0.041&0.411$\pm$0.036&0.437$\pm$0.014&0.435$\pm$0.051&0.377$\pm$0.018&\textbf{0.446$\pm$0.027}&0.436$\pm$0.025\\
			\hline
			statlog-heart&0.748$\pm$0.034&\textbf{0.764$\pm$0.026}&0.672$\pm$0.053&0.696$\pm$0.057&0.718$\pm$0.063&0.726$\pm$0.089&0.598$\pm$0.053\\
			\hline
			statlog-image&0.972$\pm$0.006&0.964$\pm$0.008&0.984$\pm$0.006&0.980$\pm$0.004&0.826$\pm$0.033&0.977$\pm$0.006&\textbf{0.985$\pm$0.003}\\
			\hline
			statlog-landsat&0.876$\pm$0.005&0.872$\pm$0.007&0.879$\pm$0.005&0.878$\pm$0.005&0.629$\pm$0.041&0.877$\pm$0.009&\textbf{0.886$\pm$0.003}\\
			\hline
			statlog-shuttle&\textbf{0.999$\pm$0.000}&0.998$\pm$0.000&\textbf{0.999$\pm$0.000}&\textbf{0.999$\pm$0.000}&0.997$\pm$0.001&\textbf{0.999$\pm$0.000}&\textbf{0.999$\pm$0.000}\\
			\hline
			statlog-vehicle&0.639$\pm$0.016&0.637$\pm$0.019&0.659$\pm$0.024&0.687$\pm$0.030&0.475$\pm$0.027&0.671$\pm$0.018&\textbf{0.706$\pm$0.015}\\
			\hline
			steel-plates&0.710$\pm$0.012&0.703$\pm$0.006&\textbf{0.751$\pm$0.011}&0.740$\pm$0.016&0.441$\pm$0.053&0.724$\pm$0.008&0.738$\pm$0.009\\
			\hline
			synthetic-control&0.984$\pm$0.008&0.978$\pm$0.016&0.986$\pm$0.007&\textbf{0.988$\pm$0.007}&0.600$\pm$0.038&0.984$\pm$0.006&0.960$\pm$0.013\\
			\hline
			teaching&0.517$\pm$0.050&\textbf{0.527$\pm$0.077}&0.478$\pm$0.088&0.489$\pm$0.106&0.361$\pm$0.071&0.517$\pm$0.069&0.478$\pm$0.075\\
			\hline
			thyroid&0.691$\pm$0.036&0.694$\pm$0.026&0.954$\pm$0.022&0.977$\pm$0.010&0.951$\pm$0.006&\textbf{0.989$\pm$0.004}&0.987$\pm$0.006\\
			\hline
			tic-tac-toe&0.958$\pm$0.008&0.951$\pm$0.008&0.977$\pm$0.008&0.974$\pm$0.012&0.944$\pm$0.015&\textbf{0.979$\pm$0.012}&0.972$\pm$0.012\\
			\hline
			titanic&0.445$\pm$0.029&0.445$\pm$0.029&0.427$\pm$0.008&0.427$\pm$0.008&\textbf{0.453$\pm$0.003}&0.427$\pm$0.008&0.427$\pm$0.008\\
			\hline
			twonorm&0.959$\pm$0.008&\textbf{0.960$\pm$0.007}&0.957$\pm$0.004&0.948$\pm$0.004&0.949$\pm$0.007&0.949$\pm$0.006&0.950$\pm$0.005\\
			\hline
			vertebral-column-2clases&\textbf{0.653$\pm$0.037}&0.650$\pm$0.039&0.651$\pm$0.082&0.539$\pm$0.122&0.573$\pm$0.086&0.572$\pm$0.083&0.562$\pm$0.093\\
			\hline
			vertebral-column-3clases&0.762$\pm$0.025&\textbf{0.767$\pm$0.018}&0.738$\pm$0.056&0.691$\pm$0.059&0.540$\pm$0.155&0.740$\pm$0.062&0.734$\pm$0.060\\
			\hline
			wall-following&0.924$\pm$0.007&0.919$\pm$0.005&0.977$\pm$0.006&\textbf{0.997$\pm$0.002}&0.919$\pm$0.021&0.994$\pm$0.001&0.995$\pm$0.002\\
			\hline
			waveform&\textbf{0.808$\pm$0.012}&0.798$\pm$0.021&0.786$\pm$0.020&0.779$\pm$0.011&0.765$\pm$0.023&0.771$\pm$0.013&0.769$\pm$0.014\\
			\hline
			waveform-noise&0.776$\pm$0.010&0.775$\pm$0.011&\textbf{0.803$\pm$0.009}&0.795$\pm$0.008&0.752$\pm$0.009&0.794$\pm$0.014&0.785$\pm$0.013\\
			\hline
			wine&\textbf{0.991$\pm$0.015}&\textbf{0.991$\pm$0.015}&\textbf{0.991$\pm$0.015}&\textbf{0.991$\pm$0.015}&0.904$\pm$0.046&0.974$\pm$0.028&0.974$\pm$0.029\\
			\hline
			wine-quality-red&\textbf{0.518$\pm$0.016}&0.512$\pm$0.021&0.492$\pm$0.031&0.419$\pm$0.022&0.255$\pm$0.007&0.494$\pm$0.026&0.434$\pm$0.016\\
			\hline
			wine-quality-white&\textbf{0.532$\pm$0.012}&0.529$\pm$0.015&0.523$\pm$0.011&0.512$\pm$0.006&0.090$\pm$0.020&0.511$\pm$0.008&0.502$\pm$0.006\\
			\hline
			yeast&0.508$\pm$0.042&0.502$\pm$0.029&0.485$\pm$0.034&0.496$\pm$0.020&0.189$\pm$0.055&\textbf{0.519$\pm$0.027}&0.505$\pm$0.027\\
			\hline
			zoo&\textbf{0.986$\pm$0.024}&\textbf{0.986$\pm$0.024}&\textbf{0.986$\pm$0.024}&\textbf{0.986$\pm$0.024}&0.918$\pm$0.061&\textbf{0.986$\pm$0.024}&\textbf{0.986$\pm$0.024}\\
			\hline
			\hline
			AVERAGE&\textbf{0.685$\pm$0.043}&0.675$\pm$0.051&0.673$\pm$0.047&0.663$\pm$0.055&0.550$\pm$0.046&0.663$\pm$0.047&0.660$\pm$0.052\\
			\hline
			\label{tab:ch4:kappaUCI}
		\end{longtable}
	\end{center}
\end{landscape}

  Now, we perform one-sided paired Wilcoxon signed-rank tests for every method to further demonstrate the statistical significance of the results. The six p-values for each method from the Wilcoxon test were corrected using the Bonferroni method~\cite{chen2017general}. Figure~\ref{fig:ch4:pavluechoenkappa} shows that at a significance level of $0.05$, \acrshort{graf} is significantly better than all other methods except for \acrshort{et}. Further, the methods have been arranged in increasing order of their Friedman ranking on the x-axis of Figure~\ref{fig:ch4:pavluechoenkappa}.

\section{Conclusion}
In this chapter, we presented a supervised approach to constructing random forests, termed as \acrfull{graf}. This approach is inspired by the hierarchical arrangement of hyperplanes. \acrshort{graf} repeatedly draws random hyperplanes to partition the data. It uses successive hyperplanes to correct impure partitions to the extent feasible so that the overall purity of resultant partitions increases. The resultant partitions (or leaf nodes) are represented with variable length codes. This guided tree construction bridges the gap between boosting and decision trees, where every tree represents a high variance instance. The idea to put the \acrshort{graf} in $\mathcal{A}-\mathcal{B}$ formulation is also presented. A variant of \acrshort{graf}, \acrfull{ugraf} is also presented which can be used as a hashing algorithm. Results on 115 benchmark datasets show that \acrshort{graf} outperforms state-of-the-art bagging and boosting-based algorithms like Random Forest~\cite{rf} and Gradient Boosting~\cite{friedman2001greedy}. The results show that \acrshort{graf} is effective on both binary and multiclass datasets. \acrshort{graf} exhibits both low bias and low variance with the increasing size of the training dataset.   

%% file: ch5applications.tex
\section{Introduction}

In the current day and age, a torrent of data is being generated in every field. Thus to learn from these datasets, it becomes essential to reformulate traditional machine learning algorithms to scale up and be used in distributed data storage settings. The alternative to re-imagining machine learning algorithms is to systematically reduce the data size without losing critical information whenever feasible~\cite{bachem2017practical}. The reduced dataset can then be used to train prediction models for different tasks with similar performance as the whole dataset. One such strategy is to select a vital sample or perform data approximation from the datasets using neighborhood properties or by margin optimization~\cite{nalepa2019selecting,shin2007neighborhood,guo2015fast}.

The analysis and visualization of big datasets are other challenges. Big datasets with high intrinsic dimensionality are harder to visualize using  2D or 3D projections from \acrshort{pca}, \acrshort{tsne}, or other dimensionality reduction algorithms. However, an understanding of a neighborhood can be developed by quantifying the neighborhood of every sample. The quantification measure can identify the amount of overlap in different classes. The samples found in the high overlap areas can then be selected and analyzed in isolation. 

Disjoint regions of space provide us with a comprehensive view of the data distribution without explicitly learning them. This view facilitates multiple types of inferences from the data. For example, regions having samples from multiple classes (impure regions) will have high confusion. When approximating a large dataset, samples from regions with low to no confusion (pure regions) are relatively less important than those from high confusion regions. It is because the high confusion regions are most likely involved in forming the decision boundaries for classification. Thus selecting samples from high confusion regions can approximate the dataset with satisfactory performance. The quantification of confusion in a region can also be utilized to identify similarities or dissimilarities between the samples of multiple classes. 

In this chapter, we present how \acrshort{graf} can be used in scoring samples in a dataset to perform importance sampling. We then utilize unsupervised \acrshort{graf} and \acrshort{combi} to quantify the neighborhood of every sample as a classifiability measure.

\section{GRAF as data approximator}\label{sec:ch5:sensitivity}

We assign a score to every sample to build a data approximator from GRAF. We term this score as the sensitivity or sensitivity score. To assign a score to a sample, we first define the sensitivity of a region. It is defined as the number of weights required to create it. It follows from the idea that regions with higher confusion will require more weights (hyperplanes) to purify them. We define a region with confusion as one in which samples of many different classes reside. We argue that points in these regions are crucial for approximating data, as these points significantly influence defining the decision region.

We define the sensitivity of a point as a function of the number of weights required to put that sample into a pure region. To assign a sensitivity value to every point in a region, we first rank each point arbitrarily and divide the sensitivity associated with a region by the point's rank. Second, we normalize these values class-wise. If the region is big, ranked sensitivity prevents sensitivity scores from being overwhelmed with the points from a single region. On the other hand, class-wise normalization handles an imbalance in the data by assigning higher sensitivities to less populated classes. Formally, we represent the process as follows.

Let us assume that $v:\mathcal{F} \to \mathbb{N}$ maps each region to the number of weights required to pure it. Hence, the importance of each sample $x^{(i)}$ in the region $\Omega_p\in \mathcal{F}$ can be computed as

\begin{align}\label{eq:ch5:rankedSensitivity}
\theta_{p x^{(i)}} = \frac{v(\Omega_p)}{i}\:\:\forall i \in \{1,..,n_p\},\:\:1\leq p\leq P
\end{align}

Equation~\ref{eq:ch5:rankedSensitivity} assigns each sample in dataset an importance value, based on the size of region $\Omega_p$. Assume that the importance of a sample in dataset is given by $\theta_{x^{(i)}}$ $\forall i \in \{1,..,N\}$. Assuming that $X_{j}=\{x^{(k)}:y_{k}=j\:\:\forall k \in \{1,..,N\}\}$ $\forall j \in \{1,..,C\}$ represents a set of samples belonging to a class, the sensitivity of each sample can be computed as

\begin{align}\label{eq:ch5:defsi}
s_{i}=\ln\left(1+\frac{\theta_{x^{(i)}}}{\Theta_{y_{i}}}\right)\:\:\forall i \in \{1,..,N\},\text{ where } \Theta_{j} = \sum_{x^{(k)}\in X_{j}}\theta_{x^{(k)}},\:\:\forall j \in \{1,..,C\}
\end{align}

Assuming that each sample is assigned a sensitivity $s_{i}^{t}$ $\forall t \in \{1,..,T\}\land\forall i \in \{1,..,N\}$, the mean sensitivity of each sample can be defined as

\begin{align}\label{eq:ch5:sensitivityAvg}
\hat{s}_{i} = \frac{1}{T}\sum_{t=1}^{T}s_{i}^{t}
\end{align}

Hence, the probability of each sample can be defined as

\begin{align}\label{eq:ch5:sensitivityProb}
p_{i}=\frac{\hat{s}_{i}}{\sum_{j=1}^{j=N}\hat{s}_{j}},\: \forall i \in \{1,..,N\}
\end{align}

The higher the probability or sensitivity of a sample, the more important it is.

\begin{figure}
	\centering
	\makebox[1 \textwidth][c]{
		\resizebox{1 \linewidth}{!}{
			\includegraphics[width=\linewidth,keepaspectratio]{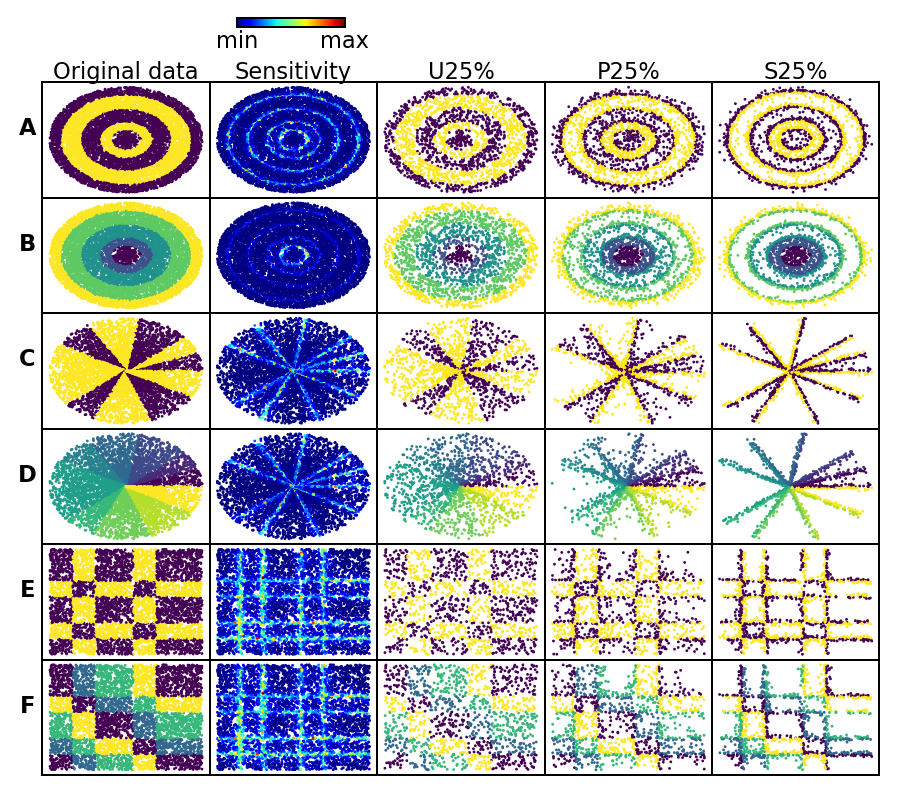}
		}
	}
	\caption{\textbf{Assessment of performance of \acrshort{graf}\'s sensitivity on simulated binary and multi-class datasets.} \textbf{A}, \textbf{B}, and \textbf{E)} represent simulated datasets with binary classes. \textbf{B}, \textbf{D}, and \textbf{F)} represent simulated multi-class datasets. The classes are arranged in different patterns, concentric circles, pie-charts, and XOR representations, in \textbf{A}-\textbf{B)}, \textbf{C}-\textbf{D)}, and \textbf{E}-\textbf{F)}, respectively. For each of these datasets, the distribution of sensitivities computed using \acrshort{graf} has been shown in column \textit{Sensitivity}. A point with higher sensitivity indicates that it is more important for data approximation. The other columns U25\%, P25\%, and S25\%, compare the performances of data approximation using only 25\% of the total samples, sampled using a uniform distribution, distribution defined by \acrshort{graf}\'s sensitivity, and the points with the highest values of sensitivities, respectively. The regions with the most confusion are best approximated using points with the highest sensitivities.   }
	\label{fig:ch5:sensitivity}
\end{figure}

The sensitivities associated with the samples may be used to approximate the complete dataset, for further downstream analyses with high sensitivity points only. A study was designed to assess how well the sensitivity computed using \acrshort{graf} approximates different datasets. To perform this analysis, 6 different datasets were created. Every dataset consists of samples distributed in different patterns (concentric circles, pie charts, and XOR representations). For every pattern, both binary and multi-class versions were generated, as illustrated in Figure~\ref{fig:ch5:sensitivity}. To generate sensitivity scores on each dataset, 200 trees ($L=200$) with complete features space ($M=2$) were generated and sensitivity score ($\hat{s}_{i}$) was computed. The performance of \acrshort{graf}\'s sensitivity has been compared with a uniform distribution for samples. Figure~\ref{fig:ch5:sensitivity} illustrates that when only 25\% of the total points are sampled, samples with the highest sensitivities adequately approximate the regions with the highest confusion.

\begin{figure}
	\centering
	\makebox[1 \textwidth][c]{
		\resizebox{1 \linewidth}{!}{
			\includegraphics[width=\linewidth,keepaspectratio]{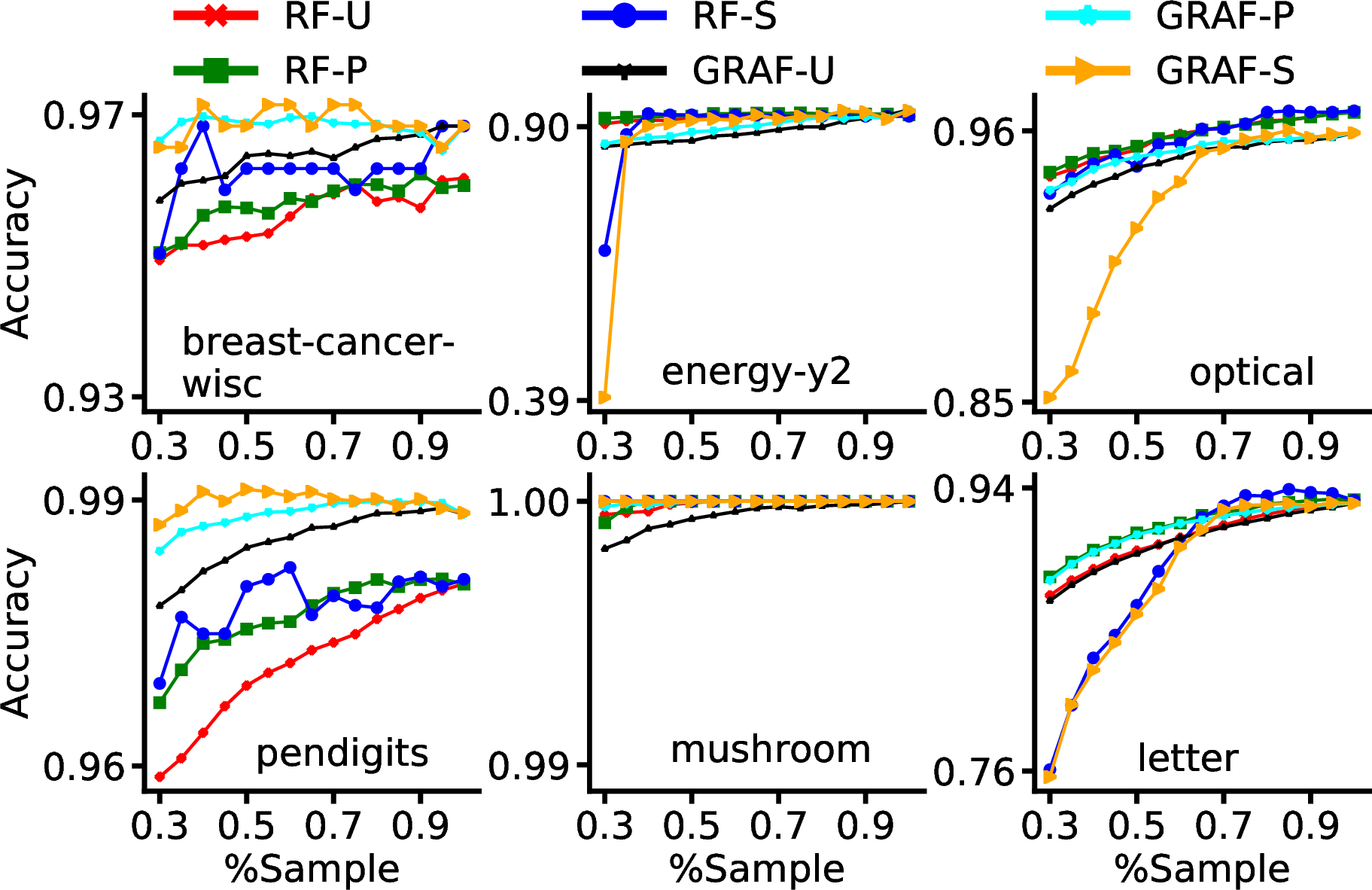}
		}
	}
	\caption{\textbf{Performance evaluation of \acrfull{rf} and \acrshort{graf}}, with increasing fraction of samples used for training, sampled according to uniform distribution (U), their sensitivities (P), and their decreasing order of sensitivities (S). The points sampled using a distribution defined by their sensitivities perform comparable or better when compared with points sampled using a uniform distribution. Also, as points are added in the decreasing order of their sensitivities, the accuracy on the test set converges and reaches its maximum with only a fraction of points with high sensitivities. The trends in results are similar, irrespective of the method used for classification. }
	\label{fig:ch5:combined_acc}
\end{figure}

If points are sampled from two different distributions- 1. uniform, 2. distribution defined by sensitivities associated with points, then the performance of the latter is better than the former (Figure~\ref{fig:ch5:combined_acc}). The performance is measured in terms of the maximal accuracy on a test set that can be achieved by using only a fraction of its samples with the highest sensitivities (Figure~\ref{fig:ch5:combined_acc}). In this aspect, the \textit{breast-cancer-wisc} dataset requires only 50\% of samples to reach within the 1\% of the highest performance of each classifier. Similarly, the \textit{energy-y2} dataset requires only 30\% of samples to reach within the 2\% range of the classifier. Further, optical, pendigits, mushroom, and letter require 70\%, 50\%, 30\%, and 70\% samples to reach within the 1\% of highest classification accuracy, respectively. These trends are observed irrespective of the method \acrfull{rf}~\cite{rf} or \acrshort{graf}. The performance difference between \acrshort{graf} and \acrshort{rf} is due to the data concept complexity, as discussed in Section~\ref{sec:ch4:simulation} and Section~\ref{sec:ch4:perfuci}. GRAF has higher (equatable or lower) performance than other methods on datasets with higher (lower) concept complexity.

This study also enforces the idea that high sensitivity points approximate the decision boundary reasonably well. To perform this experiment, 200 trees ($L=200$) were generated, the number of features ($M$) was chosen as per the tuned model, and sensitivity scores were computed on the resulting trees.

\begin{figure}
	\centering
	\makebox[1 \textwidth][c]{
		\resizebox{1 \linewidth}{!}{
			\includegraphics[width=\linewidth,keepaspectratio]{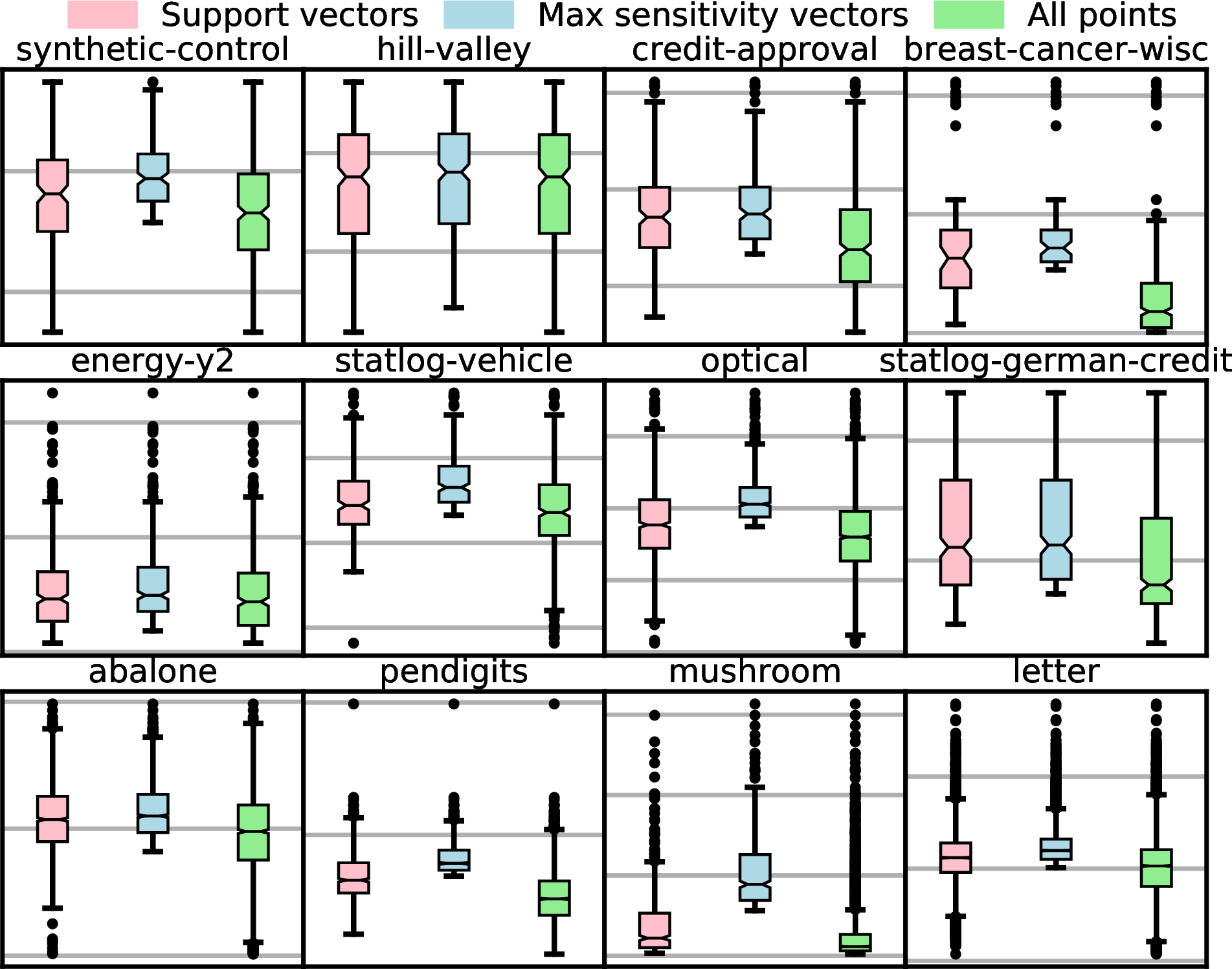}
		}
	}
	\caption{\textbf{An analogy between support vectors and points with high sensitivities.} The distribution of probabilities (\ref{eq:ch5:sensitivityProb}) associated with support vectors has been compared with that of a fraction of points with high sensitivities, and the distribution of probabilities is associated with all points. It can be concluded that points with higher sensitivities coincide with the support vectors with higher values of weights.}
	\label{fig:ch5:combined_box}
\end{figure}

The previous study's extension has shown that high sensitivity points found by \acrshort{graf} are analogous to support vectors. The performance of \acrshort{graf} is compared with two well-known methods used for reducing the samples in the training set for \acrshort{svm}~\cite{nalepa2019selecting}. \acrfull{npps}~\cite{shin2007neighborhood} selects points near the decision boundary by utilizing the property that \textit{"a pattern located near the decision boundary tends to have more heterogeneous neighbors in its class membership"}. A sample has a heterogeneous neighborhood when a few immediate neighbors belong to different classes. The measure for the heterogeneity in the neighborhood of a point is given by (negative) entropy. For points with high heterogeneity (high entropy) in their neighborhood, they are selected from the training set. The performance of \acrshort{npps} algorithm heavily depends on the initial value of the number of clusters $k$. Thus, in the experiments, the value of $k$ was tuned from $2$ to $50$. The reduced set corresponding to that $k$ for which the \acrshort{svm} model had the highest performance on the test data was selected for comparison. The second method for comparison is an ensemble method called \acrfull{svis}~\cite{guo2015fast}. \acrshort{svis} selects points with small values of ensemble margin (\ref{eq:ch4:treeMargin}). A sample with a small margin tends to lie near the decision boundary, and hence, is more informative to build a classifier. In the experiments, an ensemble of 100 decision trees with bagging was created. As suggested by authors~\cite{guo2015fast}, the different bags of datasets were generated by sampling (with replacement) $63.2\%$ of samples from the training set.    

\begin{table*}[!h]
	\makebox[1 \linewidth][c]{
		\resizebox{1.1 \linewidth}{!}{ %
			\begin{tabular}{||lrrr|rr|rr|rrrr||}
				\toprule
				
				&           &       &           & \acrshort{graf}      &               & \acrshort{svis}      &               &       &               &\acrshort{npps}           &               \\
				\cmidrule{5-12}
				&           &       &           &           &\%\acrshort{svm}          &           &\%\acrshort{svm}          &       &               &           &\%\acrshort{svm}          \\
				&           &       &           &           &accuracy on    &           &accuracy on    &       &\%size of      &           &accuracy on    \\
				&\#Train    &       &\%\acrshort{svm}      &\%Overlap  &reduced        &\%Overlap  &reduced        &       &reduced        &\%Overlap  &reduced        \\
				Dataset                 &Samples    &\%\acrshort{sv}s  &Accuracy   &with \acrshort{sv}s   &training set   &with \acrshort{sv}s   &training set   &k      &training set   &with \acrshort{sv}s   &training set   \\
				\midrule
				synthetic-control       &300        &55.67  &99.00      &67.67      &98.00          &61.08      &94.00          &21     &59.00          &70.06      &99.00\\
				
				hill-valley             &303        &95.71  &49.83      &95.52      &50.17          &95.86      &50.83          &49     &59.41          &57.93      &52.48\\
				
				credit-approval         &345        &53.62  &87.54      &74.59      &87.54          &78.92      &87.54          &26     &79.13          &64.86      &88.12\\
				
				breast-cancer-wisc      &350        &17.43  &96.85      &55.74      &96.56          &65.57      &96.28          &30     &20.00          &37.70      &96.56\\
				
				energy-y2               &384        &80.73  &90.89      &84.52      &90.89          &83.87      &90.10          &50     &61.98          &65.81      &71.35\\
				
				statlog-vehicle         &423        &52.72  &79.91      &58.30      &79.91          &79.37      &68.56          &28     &85.34          &74.89      &79.67\\
				
				statlog-german-credit   &500        &60.80  &74.00      &87.83      &75.80          &84.87      &74.80          &7      &51             &53.62      &73.40\\
				
				titanic                 &1101       &43.32  &78.64      &39.83      &78.64          &50.73      &65.09          &46     &24.34          &16.14      &78.64\\
				
				optical                 &1912       &39.33  &98.33      &60.51      &97.75          &66.09      &96.81          &49     &69.61          &90.82      &98.38\\
				
				abalone                 &2089       &68.12  &66.14      &83.91      &64.85          &86.16      &48.75          &7      &59.65          &65.14      &66.04\\
				
				pendigits               &3747       &19.51  &99.52      &51.30      &99.20          &51.85      &95.92          &45     &32.99          &70.59      &97.57\\
				
				mushroom                &4062       &11.18  &100.00     &27.75      &100.00         &21.37      &50.76          &45     &5.15           &15.86      &78.75\\
				
				letter                  &10000      &52.19  &96.53      &67.89      &92.34          &71.60      &92.49          &50     &85.64          &95.65      &96.41\\
				
				\bottomrule
			\end{tabular}
	}}
	\caption{Equivalence between the reduced training set and support vectors. For a given test set, the \acrshort{svm} model is learned using two different sets. First, an \acrshort{svm} model is trained using all the samples in the training set. Its accuracy on the test set is then evaluated (column \textit{\% \acrshort{svm} Accuracy}), and  information about the support vectors is recorded (column \textit{\% \acrshort{sv}s}). Separately, an \acrshort{svm} model is trained using points from the reduced training set (column \textit{\% \acrshort{svm} accuracy on reduced training set}). For \acrshort{graf} and \acrshort{svis}, the size of the reduced training set is the same as that of support vectors. For \acrshort{npps}, the reduced training set consists of samples with high heterogeneity values in their neighborhood (column \textit{\%size of reduced training set}). The size of the neighborhood in \acrshort{npps} is determined by $k$. An analogy between the reduced training set and support vectors is recorded in column \textit{\% Overlap with \acrshort{sv}s} for all three methods. Note that the hyper-parameters for the \acrshort{svm} model in the reduced training set were kept the same as that of the full training set.}
	\label{tab:ch5:supportvector}
\end{table*}

Table~\ref{tab:ch5:supportvector} records the accuracy on a given test set when an \acrshort{svm} model was trained using all the samples in the training set. These results were compared with an \acrshort{svm} model that is trained using only the high sensitivity points of \acrshort{graf}, the points with a low margin in \acrshort{svis}, and the reduced training set of \acrshort{npps}. The size of the reduced training set for \acrshort{graf} and \acrshort{svis} was chosen such that it constituted the same fraction as that of support vectors (\acrshort{sv}s). An analogy between support vectors and the fraction of points with high sensitivity points from \acrshort{graf} has also been illustrated in Figure~\ref{fig:ch5:combined_box}. The \acrshort{svm}'s performance on the reduced training set selected by all three methods is almost similar and is in very close proximity to \acrshort{svm}'s performance when trained on the complete training set.

\subsection{Empirical convergence of sensitivity scores}

The sensitivity score is assigned by generating successive hyperplanes. These planes together represent the distribution of the dataset in space. It indicates that after the generation of sufficient planes, there would be no significant change in the computed score. To prove this claim, an empirical evaluation of the convergence of sensitivity scores was performed.

Let us assume that $T$ hyperplanes were generated to compute the sensitivity score. Following the definition of sensitivity from (\ref{eq:ch5:sensitivityAvg}) and (\ref{eq:ch5:sensitivityProb}), assume that up to $t$ hyperplanes sensitivity is given by

\begin{align}
    \hat{s}_{i}^{t} = \frac{1}{t}\sum_{r=1}^{r=t}s_{i}^{r},
\end{align}

where definition of $s_{i}^{r}$ is given by (\ref{eq:ch5:defsi}).

\begin{align}\label{eq:ch5:intermediatesensitivity}
    p_{i}^{t} = \frac{\hat{s}_{i}^{t}}{\sum_{j=1}^{j=N}\hat{s}_{i}{t}}, \forall i \in \{1,..,N\}
\end{align}

Thus, convergence is computed as

\begin{align}\label{eq:ch5:sensitivityConvergence}
\Delta_{t} = \sqrt{\sum_{i=1}^{i=N}(p_{i}^{t} - p_{i}^{t-1})^{2}}
\end{align}

\begin{figure}[!ht]
	\centering
	\makebox[1 \textwidth][c]{
		\resizebox{1 \linewidth}{!}{
			\includegraphics[width=\linewidth,keepaspectratio]{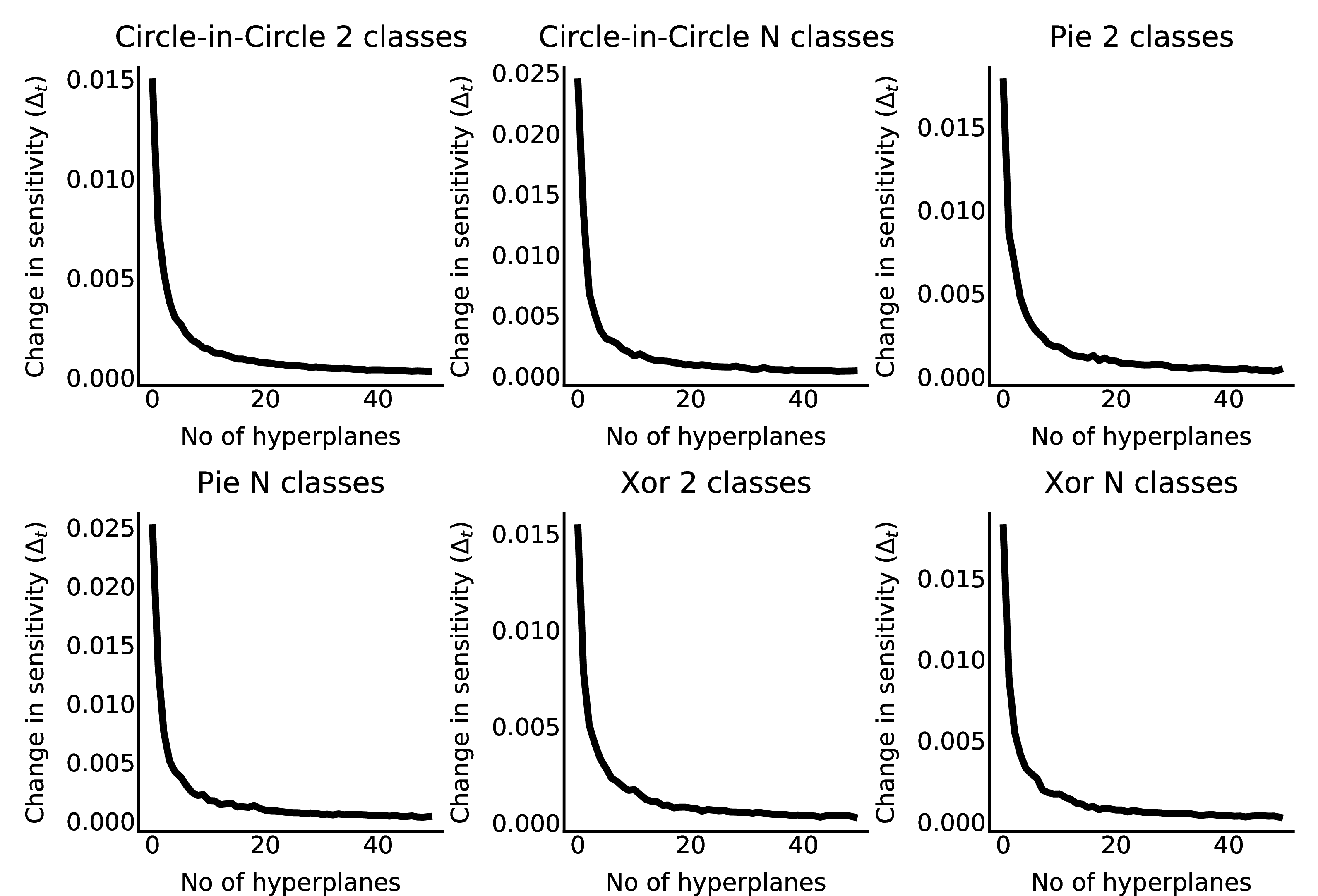}
		}
	}
	\caption{\textbf{Convergence of sensitivity values.} Change in sensitivity score almost reaches $0$ as the number of hyperplanes increases.}
	\label{fig:ch5:sensitivityConvergence}
\end{figure}

Figure~\ref{fig:ch5:sensitivityConvergence} shows the convergence of the sensitivity scores on the six simulated datasets as described in Figure~\ref{fig:ch5:sensitivity}. For all the datasets that change in sensitivity, the score is almost $0$ as the number of hyperplanes goes beyond $40$. 

\section{\acrlong{ugraf} and \acrshort{combi} for classifiability computation}

The classifiability measure estimates the difficulty of classifying a dataset by approximating the confusion in the neighborhood of samples. In general, a classifiability is computed by extracting the $r$-distance neighborhood of all the points in euclidean space. This neighborhood is then used to compute the marginal probabilities of every sample. These marginal probabilities are then used to compute the average joint probabilities for every pair of samples in the neighborhood~\cite{dong2003feature}. However, there are two problems in extracting $r$-distance neighborhood. 1.) For a large dataset, extracting the $r$-neighborhood in euclidean space is time-consuming. 2.) Tuning the value of $r$ for different datasets to extract the minimum required number of neighbors is difficult, but a $k$-nearest neighbor search algorithm can be applied in this context. We utilize \acrshort{ugraf} and \acrshort{combi} together to compute this score. We built a strategy to compute classifiability using an approximate nearest neighbor search algorithm.

\subsection{Per sample classifiability computation}

Let us assume that a dataset $X\in \mathbb{R}^{d}$. Let us assume that the dataset has $N$ samples. The dataset is divided into $C$ classes. Sample labels are represented by $Y_{i} \forall i \in \{1..N\}$. The $n$-neighboorhood of a sample is given by $\zeta_{i} \forall i \in \{1..N\}$. Further, assume that a hash table $H$ is generated using \acrfull{ugraf}. To perform the nearest neighbor search for every sample, we build \acrshort{combi}. 

\begin{algorithm}[!htb]
	\caption{Computation of Per-Sample Classifiability using \acrshort{ugraf} and \acrshort{combi}}
	\label{alg:ch5:persampleclassifiability}
	\textbf{Input}: X: Feature vector of samples, Y: labels of samples, N: Total sample in dataset, C: Total number of classes, n: Number of neareset neighbors, T: Number of ComBI, R: Number of trees in ComBI\\ 
	\textbf{Output}: classifiability: per sample classifiability
	\begin{algorithmic}[1] 
		\STATE Assume a joint probability matrix $[jp]_{(T\times N \times C \times C)}$
		\STATE Assume a classifiability matrix $[clf]_{1\times N}$
		\FOR {i $\in$ 1..T}
            \STATE Hash all the samples using \acrshort{ugraf} (Section~\ref{sec:ch4:ugraf}). Say the hash table is represented by $H_{i}$. 
            \STATE Compute $n$ neatest neighbors of every sample using \acrshort{combi} built on $H_{i}$. Say this list is represented by $NN_{i}$.
            \FOR {j $\in$ 1..N}
                \FOR {c $\in$ 1..C}
                    \STATE $marginal_{ijc} = \frac{\sum_{k=1}^{k=n}\textbf{1}(Y_{NN_{ijk}}==c)}{n}$ 
                \ENDFOR
            \ENDFOR
            
            \FOR {j $\in$ 1..N}
                \STATE $jp_{ij} = \frac{\sum_{k=1}^{k=n}\textbf{transpose}(marginal_{ij}) \times marginal_{i NN_{ijk}}}{n}$
            \ENDFOR
        \ENDFOR
        
        \FOR {i $\in$ 1..N}
            \STATE $avgjp_{i} = \frac{\sum_{k=1}^{k=T}{jp_{ik}}}{T}$
        \ENDFOR
        
        \FOR {i $\in$ 1..N}
            \STATE $clf_{i} = (2 \times \sum_{j=1}^{j=c}avgjp_{ijj}) - \sum_{j=1}^{j=C}\sum_{k=1}^{k=C} avgjp_{ijk} $
        \ENDFOR
        
        \STATE \textbf{return} $[clf]_{1\times N}$
		
	\end{algorithmic}
\end{algorithm}

Let us assume that the marginal probabilities in $n$-neighborhood of every sample is given by $p({c|\zeta_{i}}) \forall c \in \{1..C\}$. Then
\begin{align}
    p({c|\zeta_{i}}) = \frac{\sum_{j=1}^{j=n}\textbf{1}(Y_{\zeta_{ij}} == c)}{n}  
\end{align}

Say the joint probability matrix is given by $[J_{ij}] \forall i,j \in \{1..C\}\times\{1..C\}$ for every sample in neighborhood, then if for a sample $x$ and a sample in its neighborhood $y$, 

\begin{align}
    J_{ij}^{xy} = p(i|\zeta_{x}) p(j|\zeta_{y}) \forall i,j \in \{1..C\}\times\{1..C\}
\end{align}

then, the joint probability matrix is computed as the average of the joint probability of all the smaples in the $n$-neighborhood. 

\begin{align}
    J_{ij}^{x} = \frac{1}{n}\sum_{y=1}^{y=n}J_{ij}^{xy}
\end{align}

Thus, the per-sample classifiability of a sample is computed as the subtraction of the summation of non diagonal element with the summation of diagonal elements. Say for a sample $x$

\begin{align}
    C_{x} = (2 \times \sum_{c=1}^{c=C}J_{cc}^{x}) - \sum_{i=1}^{i=c}\sum_{j=1}^{j=c}J_{ij}^{x}
\end{align}

As we have seen that the precision of nearest neighbors increases as the number of hash tables increases. Thus, the process of classifiability computation is repeated multiple times and the final estimate is computed as the average of all those trials. Assuming that the process is repeated $L$ times and $C_{x}^{l}$ represents the classifiability computed on one trial, then the final estimate of per-sample classifiability is given by

\begin{align}
    C_{x} = \frac{1}{L}\sum_{l=1}^{l=L}C_{x}^{l}
\end{align}

\begin{figure}
	\centering
	\makebox[1 \textwidth][c]{
		\resizebox{0.8 \linewidth}{!}{
			\includegraphics[width=\linewidth,keepaspectratio]{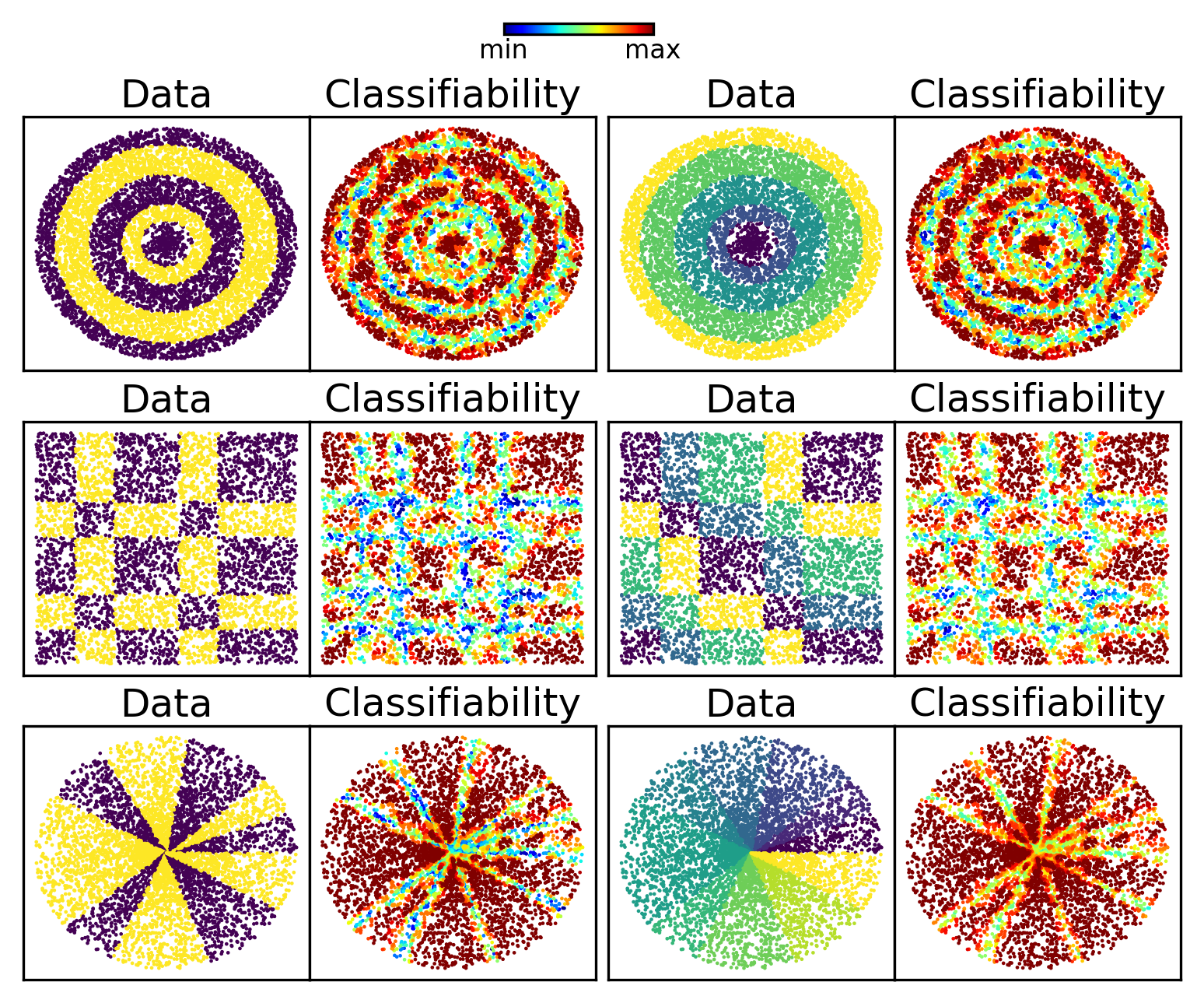}
		}
	}
	\caption{\textbf{Per-sample classifiability on simulated dataset} As expected the samples near the decision boundary have lower classifiability while the inner sample has higher classifiability.}
	\label{fig:ch5:simclassifiability}
\end{figure}

The Algorithm~\ref{alg:ch5:persampleclassifiability} presents the overall algorithm to compute the per-sample classifiability. Figure~\ref{fig:ch5:simclassifiability} presents the per-sample classifiability score on different type of simulated datasets. 

Overall classifiability of the dataset can be computed by averaging the classifiability of every sample.

\subsection{Differentiation between sensitivity and classifiability}

Although in effect, classifiability and sensitivity both can be understood as the mechanism of estimation of the confusion in a neighborhood, both of these estimates serve a different purpose.

\begin{itemize}
    \item Sensitivity scores are designed in such a way that in a neighborhood one sample from a class gets a relatively higher score. Thus from every neighborhood, only one sample becomes the representative of the neighborhood. In classifiability, all the samples get similar classifiability in a neighborhood. 
    \item Class-wise normalization is performed to compute the sensitivity score to make sure the representation of all the classes in the sampled data. However, no such normalization is performed in this formulation.
\end{itemize}

\subsection{Use-case of per-sample classifiability}

Applicability of classifiability is not only limited to estimating classification difficulty. They can be used in feature selection~\cite{dong2003feature}, building decision trees~\cite{li2005classifiability} etc. However, the utility of the confusion region of a sample still remains unexplored. Per sample classifiability combined with spatial locality can be used to identify mislabeled samples or feature errors.  Which can help reduce the data noise. 

Per sample, classifiability can also be used to identify the region of transition in the datasets. This is particularly useful when data collection for multiple classes is performed independently but with the same set of features. Then, all the class data is combined to create a final dataset. Classifiability on the combined dataset then can be used to identify regions where feature values cause labels to change from one class to another class gradually. This idea is explored in detail in Chapter 6 in the cancer patient survivability context.

\section{Conclusion}
In this chapter, \acrshort{graf} is re-envisioned as a Data approximator which assigns a sensitivity score to every sample. To reinforce this, it was shown that the samples with high sensitivity scores can help classifiers reach maximal accuracy with only a fraction of samples. It was also shown that the samples identified by \acrshort{graf} are on par with the support vectors. The empirical convergence of sensitivity scores is also discussed. In the second half of the chapter, an estimate to quantify the confusion surrounding a sample is developed. This estimate is termed classifiability. Then the idea of per-sample classifiability is formalized using \acrshort{ugraf} and \acrshort{combi}. Then differentiation between sensitivity score and classifiability is presented. The chapter concludes with a discussion on the use cases of per-sample classifiability.

We next focus on a very important and interesting application of per-sample classifiability - that to understand the relative carcinogenicity of a cancer mutation. Per-sample classifiability, motivated by the work in this chapter, is a novel thought in this domain. Most biopsy samples are drawn from core tumour regions; but normal tissues from the adjacent non-tumorous regions are rarely extracted. This practice makes it very difficult to correct the tumorous tissue for germline mutations. In the next chapter, We explore a deep learning technique to identify cancerous and non-cancerous mutations in absence of matched normal. Although the pipeline used in the next chapter is somewhat different from the approaches followed thus far in the thesis, the fundamental driving thought behind the next chapter is to explore the notion of per-sample classifiability. This is also relevant in exploring the notion of individual cancer patient survivability.

%% file: ch6CRCS.tex
\section{Introduction}

Cancer is characterized as a pathological condition in which cells divide uncontrollably. There are three main causes of human cancers – i) inherited, which accounts for a relatively small percentage of cancer cases; ii) exposure to environmental mutagens and radiation; iii) random errors caused during \acrshort{dna} replication. According to a recent study, around two-thirds of cancer mutations can be attributed to random errors caused by replication fidelity problems~\cite{tomasetti2017stem}. Cancer-related somatic mutations can be divided into two categories, drivers and passengers, based on how they contribute to cancer development. While driver mutations confer a fitness advantage to cancer cells, passengers, aka. "hitchhikers," don't. Passenger mutations comprise about 97\% of somatic mutations in cancer~\cite{mcfarland2017damaging}. Recent shreds of evidence highlight the indirect and damaging roles of passenger mutations~\cite{mcfarland2017damaging}. While a small number of driver mutations may be frequent and concentrated around driver genes, most cancer-related mutations are indistinguishable from germline variants.

In cancer biology research as well as clinical investigations, \acrfull{wgs} and \acrfull{wes} of \acrshort{dna} have gained widespread acceptance. Over the past few years, the extensive sequencing of cancer genomes resulted in the discovery and cataloging of millions of somatic mutations connected to cancer, which, taken together, allow the identification of cancer-related mutational signatures. The substitution and frameshift mutations with one or two flanking 5' and 3' nitrogenous bases make up most of the mutational signatures. Across different cancer types, these signals are discovered to be differentially enriched. The mutational signatures mainly focus on highly repetitive patterns and not much on the rare mutations which constitute the vast majority~\cite{polak2017mutational,alexandrov2020repertoire}. These signatures are also less generalizable because the subject mutation is located in the middle of the nucleotide string. Although every method currently in use is a significant step toward identifying recurrent patterns that may be tracked across cancer genomes, they are not intended to forecast cancer mutations in contrast to germline or other non-cancer somatic alterations.

Before cell division, accurate replication of \acrshort{dna} is essential to prevent mutagenesis. The fidelity of eukaryotic \acrshort{dna} replication is partially attributable to the recognition and removal of mispaired nucleotides (proofreading) by the exonuclease activity of \acrshort{dna} polymerases \textit{PLOD1} and \textit{POLE}. Church and colleagues identified \textit{POLE} mutations in highly conserved residues, which may significantly impact the proofreading process's disruption. The role of \textit{APOBEC} cytidine deaminases in \textit{APOBEC}-mediated mutagenesis in several cancer types has also been related by numerous research~\cite{roberts2012clustered,roberts2013apobec}. Such rare but frequent observations cumulatively suggest that cancer mutations are exclusive. The investigation of cancer genomes in recent years has mainly concentrated on three directions: i) driver gene identification based on mutational recurrence; ii) evaluation of functional effects of non-synonymous mutations; and iii) identification of mutational signatures. Two main goals of the current study are to demonstrate the usefulness in driver gene identification and patient survival risk assessment. The first goal is to conduct an unbiased investigation of the exclusive nature of cancer mutations compared to germline and non-cancerous alterations. The main objectives of the current study are two – i) an unbiased investigation of the exclusive nature of cancer mutations compared to germline and non-cancerous mutations; ii) demonstrating the applicability in driver gene identification and survival risk stratification in patients.

Unlike gene expressions, which are numeric, variants present the challenge of modeling categorical attributes (four nucleotides) in the context of surrounding nucleotide sequences. Latter is a more complex problem, especially since most cancer mutations are sporadic and observed in a limited number of tumor samples. A limited number of existing deep learning-based approaches enable learning from sequence data. These are used to solve diverse tasks such as unraveling regulatory motifs~\cite{quang2016danq} and prioritizing functional non-coding variants, including \acrlong{eqtl} (\acrshort{eqtl}s) in different pathological conditions~\cite{zhou2015predicting,richter2020genomic}. These approaches are based on the \acrfull{cnn} architecture. We identified two main challenges with the existing \acrshort{cnn}-based approaches: i) It is challenging to capture long-range dependencies by CNN that are typically expected in a \acrshort{dna} sequence; ii) Pooling steps in the CNN abstract the information, making it difficult for the \acrshort{cnn}s to capture subtle differences in the sequences. To this end, we felt the urgent need for a suitable learning framework that fit the requirement of modeling functions and phenotypes associated with coding variants. A significant contribution of our work is to develop a strategy named \acrfull{crcs} for representing coding variants as a finite number of codon switches (total 640 in number). Further, we learned numeric embeddings (vectors of continuous values) for these codon switches, leveraging large volumes of protein-coding genetic variants observed in the population (without any known reference to any disease). Embedding of codon switches unlocks the power of the massive community-scale initiative to process and integrate nearly $\sim$60,000 exome sequencing profiles~\cite{lek2016analysis}.

We constructed a novel deep learning architecture constituting \acrfull{blac} and demonstrated that a significant chunk of cancer mutations are distinguishable from non-cancer mutations~\cite{gupta2020deep,gupta2022new}. We benchmarked \acrshort{blac} to existing deep learning architectures and other generic methods for detecting deleterious mutations and demonstrated its power to score cancer mutations differentially. We validated our findings on independent large-scale mutational data from cancer patients and healthy populations with no reported disorders. Our results highlight the possibility of calling somatic mutation in the absence of matched normal specimens, which has immense clinical value~\cite{sun2018computational}. We identified with \acrshort{blac} a number of putative driver genes on the X chromosome such \textit{DMD}, \textit{RSK4}, \textit{AFF2}, \textit{ODF1} etc. A cumulative score was developed combining mutation level information at the patient level, which showed promise in survival risk stratification in \acrfull{blca}, \acrfull{hcc}, and \acrfull{luad}.

\section{Datasets, Methods, and Experiments}\label{sec:ch6:methods}

\subsection{Description of datasets}

High-quality coding \acrshort{snv}s representing the general population were collected from the \acrfull{exac} browser~\cite{lek2016analysis}(\url{https://console.cloud.google.com/storage/browser/gnomad-public/legacy/exacv1_downloads/release1}). The same data can also be downloaded from the \acrfull{gnomad} website~\cite{gnomad} (\url{https://gnomad.broadinstitute.org/downloads}). An equivalent set of neutral \acrshort{snv}s was downloaded from the \acrfull{dbsnp} after removing genomic alterations that are tagged \textit{pathogenic}~\cite{sherry2001dbsnp}(\url{ftp://ftp.ncbi.nih.gov/snp/latest_release/VCF}). Cancer associated coding variants were downloaded from the \acrfull{cosmic}~\cite{tate2019cosmic}(v89) and \acrfull{cbio}~\cite{cerami2012cbio,gao2013integrative}. 

A list of known driver genes for chromosome X was constructed by combining information from three sources - \acrfull{oncokb}~\cite{oncokb}, \acrfull{intogen}~\cite{intogene}, and \acrfull{cgi}~\cite{cgi}. \acrshort{oncokb} has 44 driver genes, out of which 7 (13) are annotated as oncogenes (tumor suppressors). Among the remaining genes, 23 are not annotated. \textit{MED12} is annotated as both an oncogene and a tumor suppressor. \acrshort{intogen} reports 36 driver genes, out of which \acrshort{oncokb} also reports 25. Amongst these 25 genes, 3 (12) are annotated as oncogenes (tumor suppressors) by \acrshort{oncokb}, while the remaining are not annotated. \acrshort{cgi} reports 7 driver genes, out of which 4 (1) genes are annotated as oncogenes (tumor suppressors). One driver gene is unannotated. \acrshort{oncokb} also reports all genes reported by \acrshort{cgi}. \textit{MED12} is reported by both \acrshort{intogen} and \acrshort{cgi}.

Reference genome (hg19/GRCh37) was downloaded from the \acrshort{ucsc} genome browser~\cite{ucscgenome}. The list of \acrfull{mrna} and their coordinates were obtained from \texttt{kgXref} and the \texttt{knownGene} tables from the \acrshort{ucsc} table browser~\cite{ucsctable}. 

\subsection{Pruning of the coding variants}\label{sec:ch6:coding}

\begin{figure}[!ht]
	
	\makebox[1 \textwidth][c]{
		\resizebox{1 \linewidth}{!}{%
			\includegraphics{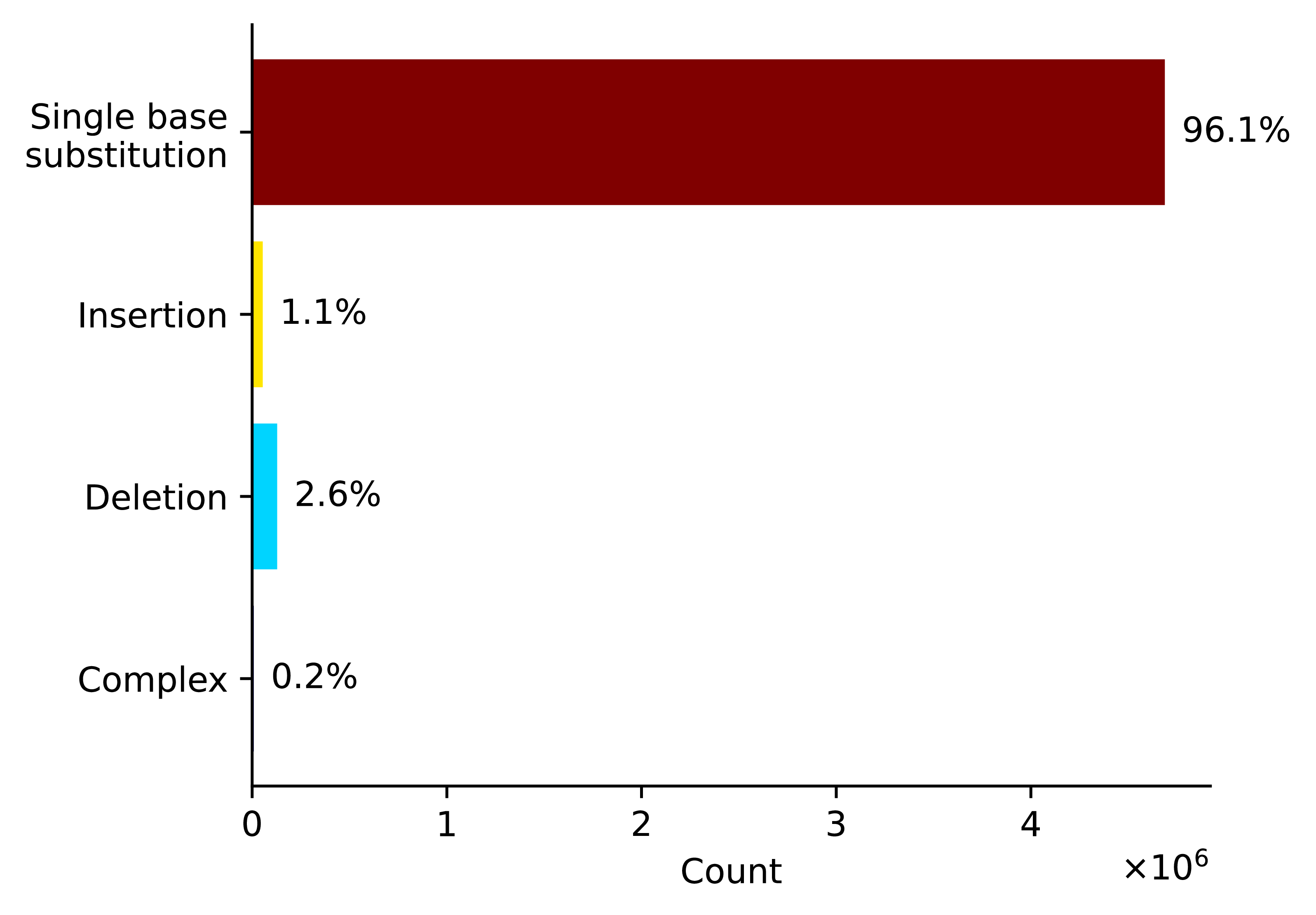}
		}
	}
	\caption{\textbf{Variant distribution in COSMIC data (v89).} Single base substitutions are the most frequent type of mutations in the database. While the complex mutations are the rarer ones.}
	\label{fig:ch6:annotations}
	
\end{figure}

\texttt{knownGene} and \texttt{kgXref} tables were combined, and only protein-coding \acrshort{mrna}s were selected. \acrshort{vcf} files from \acrshort{exac} and \acrshort{cosmic} were scanned, and genomic alterations (\acrshort{indel}s plus \acrshort{snv}s) on the protein-coding region of the genome were analyzed further. Of 4,537,166 (4,664,549) alterations collected from \acrshort{exac} (\acrshort{cosmic}), 107,591 (197,085) alterations were from the X-chromosome. These alterations were mapped to all possible splice variants of the \acrshort{mrna}s, which, for chromosome X, inflated the alteration counts to 289,813 (624,918) from \acrshort{exac} (\acrshort{cosmic}). Since the frequency of insertions, deletions, and complex mutations have a very small contribution to the datasets (Figure~\ref{fig:ch6:annotations}), we restricted the scope of our analysis to \acrshort{snv}s alone. To this end, 285,102 (590,171) \acrshort{snv}s, considering all possible splice variants harboring the \acrshort{snv}s, were retained from \acrshort{exac} (\acrshort{cosmic}). Removal of duplicate \acrshort{snv}s caused a reduction of 21,495/308,198 in these counts corresponding to \acrshort{exac}/\acrshort{cosmic}. After removing duplicates, 40\% variants from \acrshort{exac} were kept aside to learn embeddings. The remaining counts of variants from \acrshort{exac} and \acrshort{cosmic} were 149,566 and 281,973, respectively. For \acrshort{exac}/\acrshort{cosmic}, the count of synonymous, missense, and nonsense variants from chromosome X were 54,541/62,102, 93,647/202,981, and 1,313/16,703, respectively. Since synonymous variants are expected to have minimal effect on cellular fitness\cite{garcia2019somatic}, they were removed from further processing. At this stage, the preprocessed data contained 94,960/219,684 variants from \acrshort{exac}/\acrshort{cosmic} dataset. \acrshort{snv}s specific to \acrshort{dbsnp}~\cite{sherry2001dbsnp}, Met~\cite{priestley2019pan}, and \acrshort{cbio}~\cite{cerami2012cbio,gao2013integrative} were also preprocessed in a similar manner. In \acrshort{dbsnp}, variants marked as pathogenic and likely-pathogenic were removed. After preprocessing, the total X chromosome specific \acrshort{snv}s from \acrshort{dbsnp} were 530,405. Across 287 studies present in \acrshort{cbio}, mutations reported on the X chromosome were filtered. After preprocessing, the mutation count was 374,138 and 2,611 for \acrshort{cbio} and Met data, respectively. 

\subsection{Codon switch sequences}\label{sec:ch6:crcv}

In this article, we present a novel representation for mutations, \textit{viz.}  as codon switches. A codon switch dictionary was created by altering one nucleotide in a codon at a time. This results in 640 codon switches\footnote{Mathematically, total codon switch count is given as $n^c\big({c \choose 1}\times{{n-1}\choose{1}}+{{c}\choose{c}}\big)$ where codon length, $c=3$, and number of nucleotides, $n=4$}. Here we justify the count of the total number of codon switches. A codon is made of 3 nucleotides. If one mutation is introduced to a codon, it can occur at any of these three nucleotides. Every position already contains a specific nucleotide. The change can be made by replacing it with one of the three remaining bases. Therefore, a codon can be transformed into one of the nine possible codons by introducing a single base change. In this way, we obtain $9 \times 64 = 576$ codon switches for all 64 codons. To generalize the applicability of codon switches, we need to also consider unchanged codons as codon switches, where both codons are identical. Therefore, we have $576 + 64 = 640$ codon switches in the dictionary that can seamlessly represent any coding sequence. Each codon switch is assigned a unique numeric code from 0 to 639. To capture mutation identities adequately, we considered the sequence of surrounding codon switches. All protein-coding \acrshort{mrna} sequences (coding regions only) were extracted from the reference genome to construct these codon switch sequences. For each variant, we constructed a codon switch sequence based on the nucleotide triplets as observed in the corresponding reference sequence, except for the single codon switch difference due to the variant itself. For codon switches other than ones harboring variants, we considered identical nucleotide pairs as per the reference sequence.  This is illustrated in Figure~\ref{fig:ch6:panel1}A. For the machine learning task, these codon switch sequences were converted into numeral sequences using their pre-assigned numeric codes. All analyzed variants were processed in this manner for embedding and other machine learning tasks.

\subsection{Continuous embedding of codon switches}\label{sec:ch6:emb}

A \textit{skip-gram}~\cite{skipgram} model with negative sampling was employed to learn continuous representations of codon switches. The \textit{skip-gram} model learns embeddings by training a shallow neural network that attempts to predict a codon switch's context. The word whose context is being learned is referred to as a center codon switch. In general, the context (the surrounding nucleotides) of a codon switch is prohibitively long to predict; thus, we resort to the negative sampling approach. In the said approach, we define a small window (\acrshort{ws}) around it for every codon switch in a sequence, and all codon switches in this region are termed context codon switches or positive samples. Further, some codon switches from outside the windows are randomly selected and termed the negative samples. The rate at which codon switches are sampled is called the negative sampling rate (\acrshort{nsr}).

Theoretically, embeddings are learned by making every codon switch in a sequence a center codon switch. Of note, corner codon switches are also treated as center codon switches, but we look at only one side of the window to get the context. But, practically, in a large dataset with large sequences, the count of center codon switches is extremely high. Thus, making it infeasible to use every instance of a codon switch as a center codon switch. Hence, we performed the subsampling to limit it. Since a codon switch sequence consists of codon switches that do not contain any nucleotide alteration except for one codon switch, the distribution of codon switches is heavily skewed toward the former type of codon switches. Thus, we first systematically squeeze the probability of frequent codon switches and inflate the probabilities of a non-frequent switch. This increases the chances of non-frequent codon switches getting selected as a center codon switches. To systematically adjust the probability of codon switches, we use the following formula:

\begin{align}
\text{probability of selecting a codon switch} = \min\Bigg(1, \Big(1+\sqrt{\frac{f}{\epsilon}}\Big).\frac{\epsilon}{f}\Bigg),
\end{align}
\noindent
where $\epsilon=0.001$ and $f$ are the codon switch frequency in the dataset kept aside for embedding. For each selected center codon switch, $2*ws$ tuples were constructed by pairing it with \acrshort{ws} adjacent codon switches from both sides. Taken together, these tuples constituted the positive category. 
On the other hand, for every center codon switch, negative sampling was performed by pairing the center codon switch with random $\lceil((2*ws + 1)*nsr)\rceil$ or 2 codon switches, uniformly sampled from the codon switch dictionary. We used window size of 3 (\acrshort{ws}) and negative sampling rate (\acrshort{nsr}) of 0.2 for the construction of dataset. For these values, a total of 219,418,024 tuples were generated. Out of this, 182,886,425 were generated as positive samples, and 36,531,599 were generated as negative samples.

In order to learn the 300 sized numeric vectors representing the 640 codon switches, we initialized a $640 \times L$, where $L=300$ sized matrix with random entries. To this end, we also simplify the training procedure of \textit{skip-gram}. We posed the problem of learning codon switch embedding as a classification. To build the dataset, we assigned a class label of 0/1 to all the codon switch pairs in the negative/positive sets (Figure~\ref{fig:ch6:panel1}B). Then a simple neural network was trained to classify between the tuples labeled as 0 or 1 (Figure~\ref{fig:ch6:panel1}C). Network~\ref{alg:ch6:CRCS} shows the network architecture to learn \acrshort{crcs}es. The input to the neural network was the concatenated vectors of length $2L$, corresponding to the pair of codon switches in each tuple. For a pair of embedding, we first compute the dot product of the two and then compute the sigmoid of the resulting value. The binary cross-entropy cost function was optimized on the output of the sigmoid unit. In total, the model has 192,002 trainable parameters. All parameters, except 2, are the learnable parameters from the embedding matrix. The other two parameters are for the last dense layer, where one of the parameters belongs to the neuron's weight and the other one is for the bias of the layer.  The ADAM~\cite{adam} optimizer was used for optimization. The procedure was repeated for $200$ epochs.

\begin{algorithm}
	\floatname{algorithm}{Network}
	\caption{Network architecture to learn the \acrfull{crcs}}
	\label{alg:ch6:CRCS}	
	\begin{algorithmic}[1]
		\STATE index1 $\gets$ embedding array index of first switch in the pair
		\STATE index2 $\gets$ embedding array index of second switch in the pair
		\STATE label $\gets$ label of the switch pair
		\STATE emb1 = EmbeddingMatrix(index1)
		\STATE emb2 = EmbeddingMatrix(index2)
		\STATE dot = \textbf{Dot}([\textit{emb1}, \textit{emb2}])
		\STATE output = \textbf{Dense}(\textit{dot}, neurons=1, activation='sigmoid')
		\STATE cost = \textbf{BinaryCrossEntropy}(\textit{output}, label)
	\end{algorithmic}
\end{algorithm}

\subsection{Cross-chromosome sequence similarity analysis}

To assess the diversity of chromosomes at the amino acid levels, we computed the proportion of unigram, bigram, and trigram of amino acids across all chromosome sequences. Here, a unigram is defined as a single amino acid. There were 21 unique unigrams; among those, 20 were amino acids and one representative unigram corresponding to a stop codon. Similarly, a bigram and trigram are defined as strictly ordered pairs and triplets of amino acids, respectively. In this manner, we obtained 441 unique bigrams and 9261 trigrams (Figure~\ref{fig:ch6:panel2}). These counts also includes pairs and triplets of stop codon. 

\subsection{Variant classification}\label{sec:ch6:classification}

\begin{figure}[!ht]
	
	\makebox[1 \textwidth][c]{
		\resizebox{1 \linewidth}{!}{%
			\includegraphics{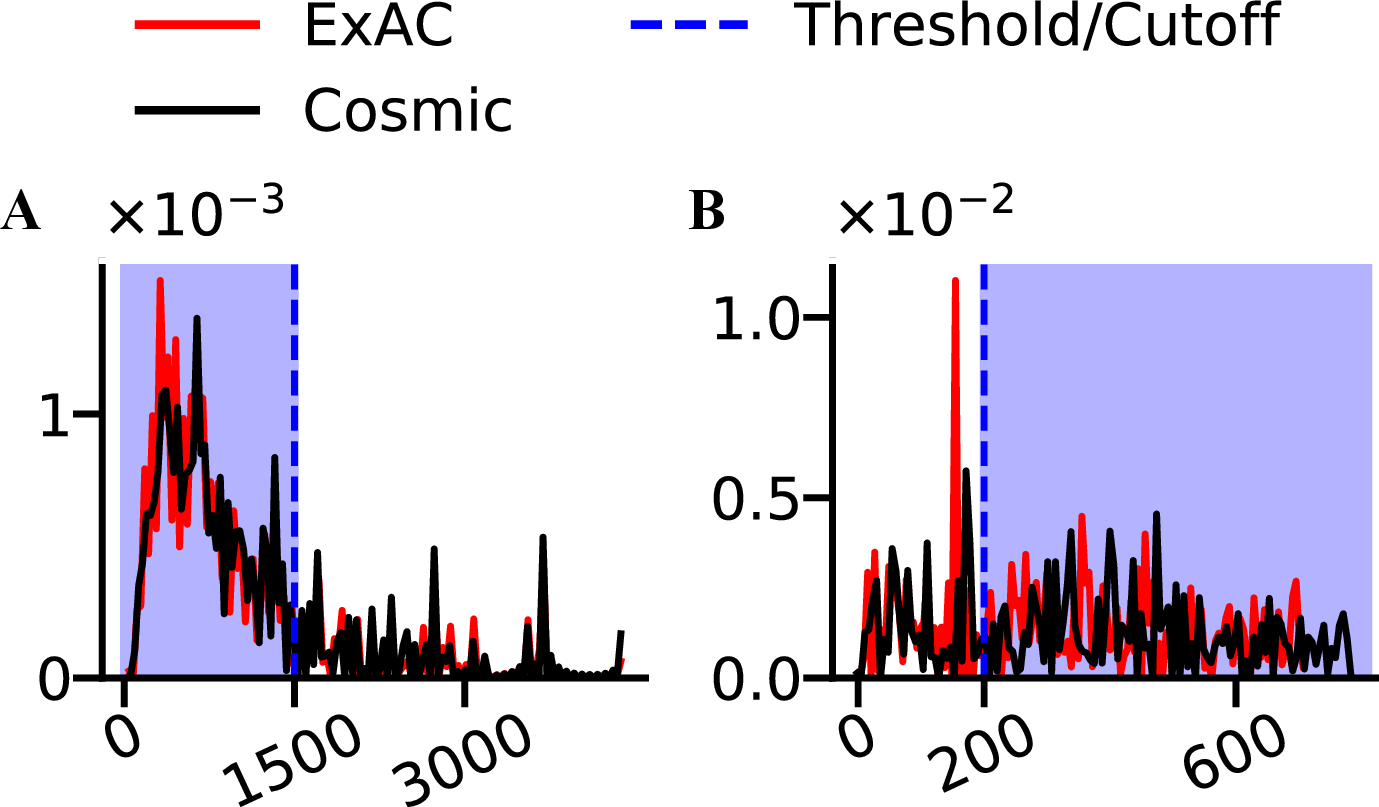}
		}
	}
	\caption{\textbf{Filtering criteria to handle computation overhead.} \textbf{A)} \acrshort{mrna}s whose switch sequences were $\leq 1500$ long were kept for the analysis. \textbf{B)} Genes that have $\geq 200$ mutations were kept for analysis.}
	\label{fig:ch6:filter}
	
\end{figure}

The pre-trained embeddings were used for classifying codon switch sequences. We labeled codon switch sequences stemming from \acrshort{exac} and \acrshort{cosmic}  as 0 and 1, respectively. To reduce the computational overhead, sequences of length less than $1500$ were selected. Only genes with a minimum variant count of $200$ (with alternate splicing) were retained, leading to $332$ genes (Figure~\ref{fig:ch6:filter}). These genes were then randomly split into 4-folds. These folds were created so that there are no common genes in train and validation splits.

A deep neural network was constructed to classify the sequences. The deep neural network consisted of a non-trainable embedding layer, followed by two stacked \acrfull{bilstm} layers~\cite{lstm,birnn}, interleaved with one batch normalization layer~\cite{batchnormalization}. \acrshort{bilstm} layers were followed by another batch normalization layer and a time-distributed dense layer. The time-distributed layer shared weights across all time-states in a sequence. However, the time-states did not communicate with each other. The time-distributed layer was followed by another batch normalization layer and an attention layer~\cite{attention}. The output layer is dense, and its neurons use a sigmoid activation function. In all, the model used 3,877,201 parameters, of which 194,600 were non-trainable or fixed. The embedding layer used pre-trained codon switch embeddings, marked as fixed, whereas other layers were initialized randomly and marked as trainable. The Network was trained by minimizing a binary cross-entropy loss function. The ADAM~\cite{adam} optimizer was used. A schematic of the network architecture is shown in Figure~\ref{fig:ch6:panel3}A and Network~\ref{alg:ch6:ourblac}. The confidence scores for \acrshort{dbsnp} and Met were generated after removing sequences that were part of the training set.

\begin{algorithm}
	\floatname{algorithm}{Network}
	\caption{Customized neural network for sequence classification - \acrfull{blac}}
	\label{alg:ch6:ourblac}	
	\begin{algorithmic}[1]
		\STATE seq $\gets$ Switch sequence
		\STATE label $\gets$ label of the sequence
		\STATE embeddings = \textbf{Embeddings}(seq)
		\STATE bi1 = \textbf{BiLSTM}(\textit{embeddings}, neurons=300, how='cascade')
		\STATE ba1 = \textbf{BatchNorm}(\textit{bi1})
		\STATE bi2 = \textbf{BiLSTM}(\textit{ba1}, neurons=300, how='cascade')
		\STATE ba2 = \textbf{BatchNorm}(\textit{bi2})
		\STATE td = \textbf{TimeDistributedDense}(\textit{ba2}, neurons=100)
		\STATE ba3 = \textbf{BatchNorm}(\textit{td})
		\STATE at = \textbf{Attention}(\textit{ba3})
		\STATE output = \textbf{Dense}(\textit{at}, neurons=1, activation='sigmoid')
		\STATE cost = \textbf{BinaryCrossEntropy}(\textit{output}, label)
	\end{algorithmic}
\end{algorithm}

\subsection{Other methods for mutation annotation}

We compare the performance of our architecture with two other methods, \acrfull{sift}~\cite{vaser2016sift} and \acrfull{pph2}~\cite{adzhubei2010method}, which annotates deleterious mutations. The \acrshort{sift} algorithm's command-line version of the executable (for Linux) was downloaded from \url{https://sift.bii.a-star.edu.sg/sift4g/AnnotateVariants.html}. \acrshort{sift} 4G database of chromosome X was downloaded from \url{https://sift.bii.a-star.edu.sg/sift4g/public/Homo_sapiens/GRCh37.74/}. We combined all 4 test folds into a single dataset to run the predictions. This combined dataset was sorted first on the chromosome, then on position, and then passed it through the executable. The following command was used to annotate the mutation.

\begin{verbatim}
java -jar SIFT4G_Annotator.jar -c -i input_vcf.vcf \
-d sift_db -r output_folder
\end{verbatim}

\textit{SIFT\_SCORE} column from the output file of \acrshort{sift} was used for further analysis. The mutations with a low value of \acrshort{sift} score represent the deleterious mutations. As per recommendation, if the predicted score is below 0.05, the mutation is deleterious. However, we considered these scores as continuous values and performed the analysis. To keep the scores in a similar range as our method, we subtracted \acrshort{sift} scores from 1 before comparing.

We generated predictions from \acrshort{pph2} using the webserver available at \url{http://genetics.bwh.harvard.edu/pph2/}. We used the same dataset used for \acrshort{sift}.

\subsection{Other available embeddings}

To compare the efficacy of \acrshort{crcs} against other embeddings, we downloaded dna2vec~\cite{ng2017dna2vec} embeddings from \url{https://github.com/pnpnpn/dna2vec}. Dna2vec trains the word2vec model on k-mers of the human genome. We extracted 100 length embeddings of every codon (3-mer) of the human genome from the dna2vec model, thus resulting in an embedding matrix of $64 \times 100$. This matrix is fed to the network in the embedding layer of our customized sequence classifier (Section~\ref{sec:ch6:classification}). Codon sequences in place of codon switch sequences were used for training with dna2vec embeddings.

\subsection{Other available architectures}

We also compared the two widely used architectures developed for predicting functional effects of non-coding variants, namely DeepSea~\cite{zhou2015predicting}, DanQ
~\cite{quang2016danq}, and one recently published architecture HeartENN~\cite{richter2020genomic}. DeepSea and HeartENN are pure convolutional neural networks. In contrast, DanQ is a hybrid architecture with convolutional and bidirectional \acrshort{lstm} layers. HeartENN has 90 neurons in the last layer, but we changed it to 919 as in DanQ and DeepSea. Then, to make these architectures suitable for classifying sequences into cancer and non-cancer, we added one more dense layer with a single neuron and sigmoid activation at the end. Networks~\ref{alg:ch6:deepsea},~\ref{alg:ch6:danQ}, and~\ref{alg:ch6:heartenn} give the details about these architectures. Among these models, \texttt{DanQ} has the most parameters (206,177,959), followed by \texttt{DeepSea} (64,921,359). \texttt{HeartENN} has 58,525,559 parameters, out of which 760 are non-trainable. We used binary cross-entropy as the loss function to optimize these networks. We used RMSProp as the optimizer. One-hot encoded protein-coding \acrshort{mrna} sequences are provided as input for training. Since \acrlong{cnn} works with fixed input size, we have padded all the variable length sequences to 4500 (1500 length codon switch sequence) with 0s.

\begin{algorithm}
	\floatname{algorithm}{Network}
	\caption{Modified DeepSea Neural Network}
	\label{alg:ch6:deepsea}	
	\begin{algorithmic}[1]
		\STATE seq $\gets$ One-hot encoded protein coding mRNA Sequence
		\STATE label $\gets$ label of the sequence
		\STATE conv1 = \textbf{Conv1D}(seq, filter=320, kernelsize=8, padding='valid', activation='relu')
		\STATE mp1 = \textbf{MaxPool1D}(\textit{conv1}, poolsize=4, strides=4)
		\STATE d1 = \textbf{Dropout}(\textit{mp1}, 0.2)
		\STATE conv2 = \textbf{Conv1D}(\textit{d1}, filter=480, kernelsize=8, padding='valid', activation='relu')
		\STATE mp2 = \textbf{MaxPool1D}(\textit{conv2}, poolsize=4, strides=4)
		\STATE d2 = \textbf{Dropout}(\textit{mp2}, 0.2)
		\STATE conv3 = \textbf{Conv1D}(\textit{d2}, filter=960, kernelsize=8, padding='valid', activation='relu')
		\STATE mp3 = \textbf{MaxPool1D}(\textit{conv3}, poolsize=4, strides=4)
		\STATE d3 = \textbf{Dropout}(\textit{mp3}, 0.5)
		\STATE f = \textbf{Flatten}(\textit{d3})
		\STATE D1 = \textbf{Dense}(\textit{f}, neuron=919, activation='relu')
		\STATE output = \textbf{Dense}(\textit{D1}, neurons=1, activation='sigmoid')
		\STATE cost = \textbf{BinaryCrossEntropy}(\textit{output}, label)
	\end{algorithmic}
\end{algorithm}

\begin{algorithm}
	\floatname{algorithm}{Network}
	\caption{Modified DanQ Neural Network}
	\label{alg:ch6:danQ}	
	\begin{algorithmic}[1]
		\STATE seq $\gets$ One-hot encoded protein coding mRNA Sequence
		\STATE label $\gets$ label of the sequence
		\STATE conv1 = \textbf{Conv1D}(seq, filter=320, kernelsize=26, padding='valid', activation='relu')
		\STATE mp1 = \textbf{MaxPool1D}(\textit{conv1}, poolsize=12, strides=13)
		\STATE d1 = \textbf{Dropout}(\textit{mp1}, 0.2)
		\STATE b1 = \textbf{BiLSTM}(\textit{d1}, neurons=320)
		\STATE d2 = \textbf{Dropout}(\textit{b1})
		\STATE f = \textbf{Flatten}(\textit{d2})
		\STATE D1 = \textbf{Dense}(\textit{f}, neurons=925, activation='relu')
		\STATE D2 = \textbf{Dense}(\textit{D1}, neurons=919, activation='relu')
		\STATE output = \textbf{Dense}(\textit{D2}, 1, activation='sigmoid')
		\STATE cost = \textbf{BinaryCrossEntropy}(\textit{output}, label)
	\end{algorithmic}
\end{algorithm}

\begin{algorithm}
	\floatname{algorithm}{Network}
	\caption{Modified HeartENN Neural Network}
	\label{alg:ch6:heartenn}	
	\begin{algorithmic}
		\STATE seq $\gets$ One-hot encoded protein coding mRNA Sequence
		\STATE label $\gets$ label of the sequence
		\STATE conv1 = \textbf{Conv1D}(seq, filter=60, kernelsize=8, padding='valid', activation='relu')
		\STATE conv2 = \textbf{Conv1D}(\textit{conv1}, filter=60, kernelsize=8, padding='valid', activation='relu')
		\STATE mp1 = \textbf{MaxPool1D}(\textit{conv2}, poolsize=4, strides=4)
		\STATE b1 = \textbf{BatchNorm}(\textit{mp1})
		\STATE conv3 = \textbf{Conv1d}(\textit{b1}, filter=80, kernelsize=8, padding='valid', activation='relu')
		\STATE conv4 = \textbf{Conv1d}(\textit{conv3}, filter=80, kernelsize=8, padding='valid', activation='relu')
		\STATE mp2 = \textbf{MaxPool1D}(\textit{conv4}, poolsize=4, strides=4)
		\STATE b2 = \textbf{BatchNorm}(\textit{mp2})
		\STATE d1 = \textbf{Dropout}(\textit{b2}, 0.4)
		\STATE conv5 = \textbf{Conv1d}(\textit{d1}, filter=240, kernelsize=8, padding='valid', activation='relu')
		\STATE conv6 = \textbf{Conv1d}(\textit{conv5}, filter=240, kernelsize=8, padding='valid', activation='relu')
		\STATE b3 = \textbf{BatchNorm}(\textit{conv6})
		\STATE d2 = \textbf{Dropout}(\textit{b3}, 0.6)
		\STATE f = \textbf{Flatten}(\textit{d2})
		\STATE D1 = \textbf{Dense}(\textit{f}, neurons=919, activation='relu')
		\STATE output = \textbf{Dense}(\textit{D1}, neurons=1, activation='sigmoid')
		\STATE cost = \textbf{BinaryCrossEntropy}(\textit{output}, label)
	\end{algorithmic}
\end{algorithm}

\subsection{Comparing \acrshort{cbio} predictions with \acrshort{dbsnp} predictions}\label{sec:ch6:cbiodbsnp}

In order to extract the significant genes for different cancer types, we compared the prediction scores generated on the \acrshort{cbio} with the prediction scores generated on the \acrshort{dbsnp}. For the \acrshort{cbio} data, we grouped predictions on cancer type and genes. For the \acrshort{dbsnp} database, the predictions were grouped based on genes alone. The cutoff for the group size was set to 5. We compared the groups obtained using \acrshort{cbio} and \acrshort{dbsnp}.The Mann-Whitney U-test with alternate hypothesis \acrshort{cbio} $>$ \acrshort{dbsnp} was used to determine the statistical significance of genes. For a given cancer type, $P$-values of all genes were collected and corrected using the \textit{holm-sidak} method. The resulting gene sets were used for Gene Ontology analysis~\cite{zhou2019metascape}.

To perform the driver gene analysis using the selected gene sets, we first selected the genes that were present in most cancer types. Genes occurring in $\geq 10$ cancer types were selected for analysis. This resulted in $32$ significant genes. Among the selected cancer types, we removed those cancer types that had $<=5$ genes, resulting in $25$ cancer types.

\subsection{Classifiability for survival analysis}

The \acrshort{cbio}~\cite{cerami2012cbio} and \acrshort{dbsnp}~\cite{sherry2001dbsnp} datasets were used for survival analysis. As discussed in the Section~\ref{sec:ch6:coding} and Section~\ref{res:ch6:network}, the filtering steps applied to select the candidate mutations are i) Synonymous mutations and \acrshort{indel}s were removed. ii) All non-coding mutations were removed. iii) All the mutations that were part of \acrshort{exac} or \acrshort{cosmic} databases were also dropped since these mutations were present in the training data. After these steps, remaining unique mutations collectively spanned across 293 ONCOTREE cancer subtype codes~\cite{kundra2021oncotree} and 14,349 patients from \acrshort{cbio}. Then, these patients were grouped as per their cancer types. Any cancer type having less than 100 patients was also dropped from the analysis. After all the filtering steps, we were left with eight cancer types. We computed the mean classifiability for every patient in these cancer types. For every cancer type, The patients were divided into two groups for every cancer type by thresholding classifiability scores. The optimal threshold for every cancer type was identified using $\chi^2$ statistics~\cite{chang2017determining}. The \textit{survfit} and \textit{surfdiff} functions from survival package in R (v4.1.3) were used to perform the analysis.

\section{Results}\label{sec:ch6:results}

\subsection{Learning numeric representation of mutations}\label{sec:ch6:codingRepresentation}

\begin{figure*}
	\makebox[1\textwidth][c]{
		\resizebox{0.9\linewidth}{!}{
			\includegraphics{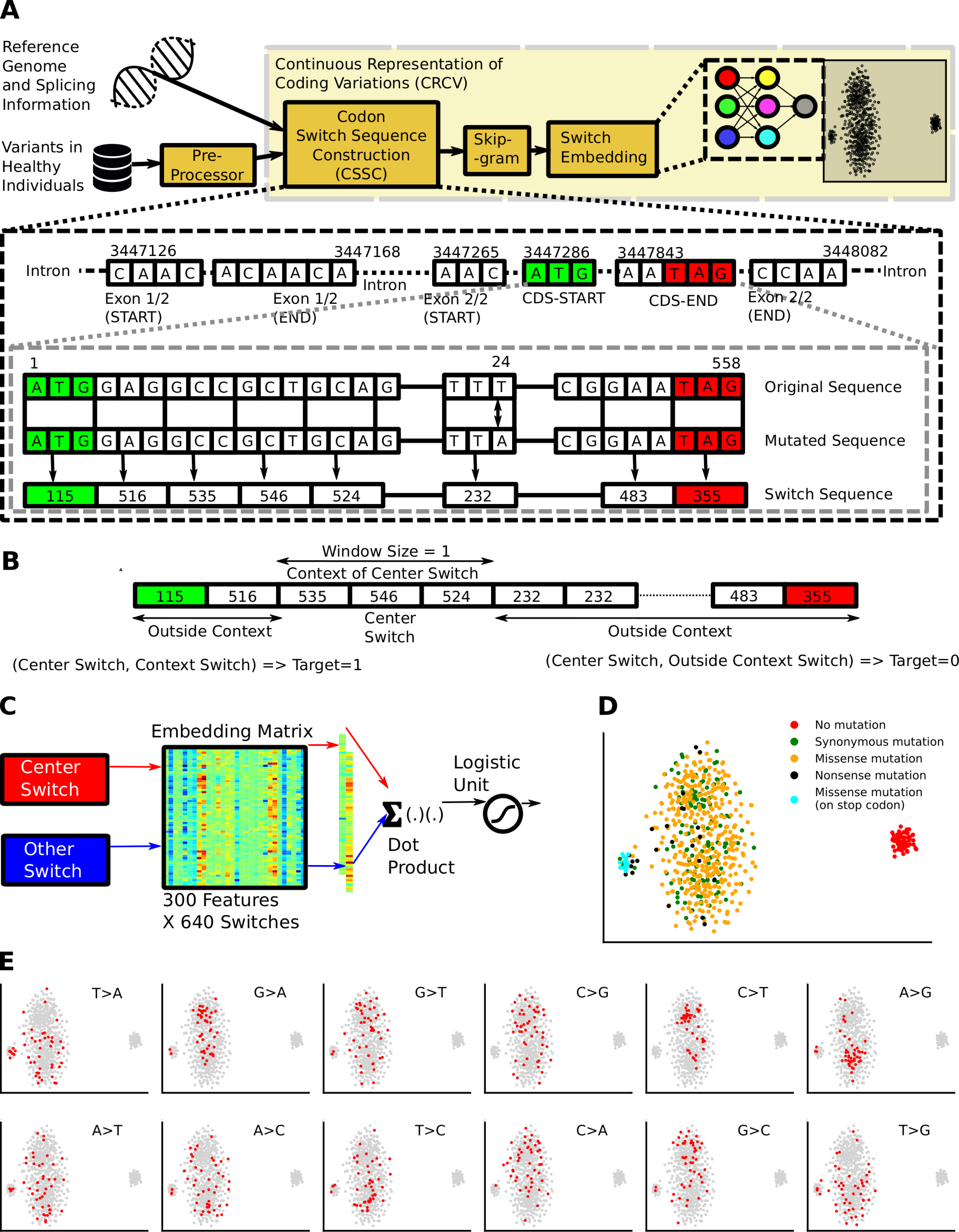}
		}
	}
	\caption{\textbf{An overview of learning \acrfull{crcs}}. \textbf{A)}The procedures include two steps: i) choosing variants that are located in exon regions; and (ii) creating the codon switch sequence. A codon switch is described as a directional pair of codons that includes an alternative codon (a sequence derived from an interest genome) and a reference codon (a sequence derived from the reference genome). Included is a toy example that shows how to build codon switch sequences. A codon switch sequence's index in the codon switch dictionary is indicated by the number next to it. \textbf{B)} A center codon switch is selected probabilistically. Two types of tuples are built for the chosen center codon switch. Tuples belonging to a center codon switch's context window are marked with a 1; a few codon switches from outside the context are also selected; their tuples are marked with a 0; \textbf{C)} A classifier is trained to classify these tuples. Input layer weights of this network behave as codon switch embeddings. \textbf{D)} \acrshort{tsne} plot of learned embeddings. \textbf{E)} Distribution of different codon switches on the \acrshort{tsne} plots. Interestingly, similar codon switches tend to cluster far from opposite codon switches ($G>A$ and $A>G$, $G>T$ and $T>G$, $A>C$ and $C>A$, $C>T$ and $T>C$).}
	\label{fig:ch6:panel1}
\end{figure*}

Long chains of adenine (A), cytosine (C), guanine (G), and thymine (T) bases make up \acrshort{dna} sequences (T). Traditionally, one hot encoding-based presentation of each nucleotide is used in machine learning-based modeling of sequence data. This method involves turning one binary vector's four possible positions to 1 and setting the other three to 0. Although simple to make, such a representation cannot adequately convey the conceptual connection between two nucleotides in a sequence. Recent advancements make it feasible to learn nucleotide embeddings, which are longer and more complex representations of nucleotides~\cite{skipgram}. However, these approaches are not helpful in learning effective embeddings with such a small dictionary (consisting of four nucleotides). Alternatively, one can create embeddings of nucleotide $k$-mers~\cite{ng2017dna2vec,asgari2015continuous}. The dictionary size of $k$-mers representation is $4^k$. Besides having large dictionary sizes, arbitrary $k$-mers do not represent biologically relevant genomic entities.

This work suggests a novel, biologically inspired method for mathematically representing coding variants. Three types of coding mutations/variants exist: synonymous, missense, and nonsense. This categorization is based on the impact that these \acrlong{snv} (\acrshort{snv}s) have on the amount of the amino acids. We factor this by representing coding variants as codon switches. A codon switch is defined as a directional pair of codons, constituting a reference codon (subsequence arising from the reference genome) and an alternative codon (subsequence arising from a genome of interest). The nearby codon switches must be considered because the nucleotides around it influence a variation. Therefore, for modeling, we created codon switch sequences containing a relevant codon switch. Since frameshift and complicated modifications, including double base substitutions, are only sporadic in the data (representing less than 4\% of the total repertoire of cancer-related mutations), we only took the single base-pair substitution into account (Figure~\ref{fig:ch6:annotations}). Figure~\ref{fig:ch6:panel1}A depicts the details of this construction process. Effectively, a codon switch does not necessarily represent an alteration, it may also represent an unaltered amino acid (e.g., ATA$\rightarrow$ATA). A dictionary of codon switches constructed in this way contains a total of 640 codon switches (Section~\ref{sec:ch6:crcv}). Notably, our entire study focuses on coding sequences only.

In this proof-of-concept investigation, we used chromosomal X mutations as the subject. To learn the embeddings, we exclusively used mutations from healthy individuals. Numerical vectors of finite dimension are called embeddings. Codon switch sequences were subjected to \textit{skip-gram} with \textit{negative sampling}~\cite{skipgram} in order to create mutational embeddings. In the field of \acrfull{nlp}, the widely used shallow neural network architecture \textit{skip-gram} is used to represent words numerically while maintaining the semantic similarity of word pairs that occur in the same context across discourses. The \textit{skip-gram} network was changed, and the learning job was presented as a classification task. We created tokens depending on the neighborhood of a codon switch from a codon switch sequence to train the network (Figure~\ref{fig:ch6:panel1}B, C). A total of 68,836 unique coding substitutions from healthy individuals (ExAC~\cite{lek2016analysis}) were used for learning the numeric representation of coding codon switches. Further details on the training of codon switch sequences can be found in Section~\ref{sec:ch6:emb}.

Each of the 640 codon switches in the semantic representation created by training the neural network was represented by a 300-sized numeric vector. Codon switches that share comparable nucleotide contexts organize themselves analogously in the associated vector space because vectors of similar words correlate strongly. Figure~\ref{fig:ch6:panel1}D shows the \acrshort{tsne} projections of the learned embeddings. Interestingly, codon switches without any substitution (identical codons) tend to form a separate cluster. Mutations in the STOP codons form a different cluster of codon switches. Mutations that are both missense and nonsense are grouped into two overlapping clusters. When these two overlapping clusters are examined more closely, it becomes clear that the localization of the codon switches in both clusters is complimentary. This demonstrates how mirrored codon changes have reversible functional effects. Note that codon switches with G $>$ A mutation and A $>$ G mutation are located in different clusters. Codon switches with other complementary mutations also display similar trends (Figure~\ref{fig:ch6:panel1}E).

\subsection{\acrshort{crcs} exposes inherent diversity of chromosomes}

\begin{figure*}
	\makebox[\textwidth][c]{
		\resizebox{1 \linewidth}{!}{
			\includegraphics{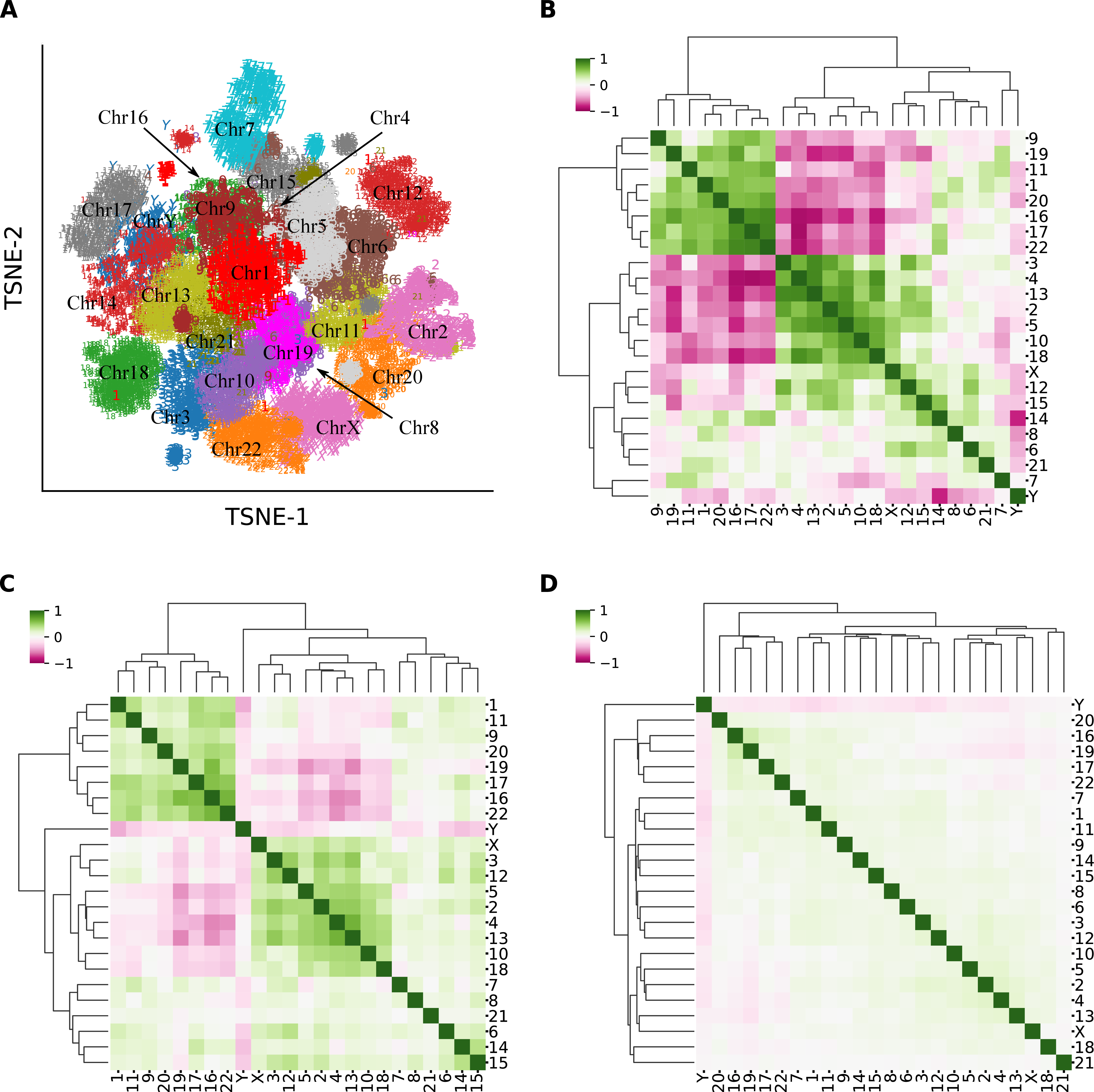}
		}
	}
	\caption{\textbf{\acrshort{crcs} embeddings reveals exclusive nature of chromosomes.} \textbf{A)} \acrshort{tsne} projections of the embeddings learned independently for all the chromosomes. The embeddings are clearly segregated, indicating heterogeneity in nucleotide sequence patterns. \textbf{B)} Spearman correlation of unigram frequencies across chromosomes. Chromosomes are found to give rise to some tight clusters. \textbf{C)} Spearman correlation of bigram frequencies in chromosomes. \textbf{D)} Chromosomes are described as trigrams. Chromosomal similarities fade away with an increase in the sequence length.}
	\label{fig:ch6:panel2}
\end{figure*}

To analyze other chromosomes, we generated \acrshort{crcs}es for all the remaining chromosomes. A \acrshort{tsne} visualization of the chromosome-specific \acrshort{crcs}es highlights heterogeneity manifested by chromosomal nucleotide sequence patterns (Figure~\ref{fig:ch6:panel2}A). To further investigate, we generated the unigrams (individual amino acid), bigrams (strictly ordered consecutive amino acids pairs), and trigrams (strictly ordered consecutive amino acids triplets) of amino acids from the sequences and analyzed chromosomal frequencies. Figure~\ref{fig:ch6:panel2}B shows the similarity of chromosomes in terms of the frequency of individual amino acids. Similar figures are also generated for bigrams and trigrams (Figure~\ref{fig:ch6:panel2}C, D). Clear biases are observed among different chromosomal groups at unigram and bigram levels, suggesting amino acid composition differences. At the trigram level, such chromosomal groups start fading away. This analysis suggests that independent learning of embeddings may be necessary for other chromosomes.

Different chromosomes harbor different sets of genes that are often functionally connected to reduce cell regulatory redundancies. Examples are \textit{HOX} and Odorant Receptor (\textit{OR}) families. \textit{HOX} genes are co-localized in chromosomes in many species, such as Drosophila. In humans, 39 \textit{HOX} genes are present as clusters across four chromosomes~\cite{garcia1994archetypal}. Similarly, a significant fraction of human \textit{OR}s is clustered in Chromosome 11~\cite{malnic2004human}. This could be a strong reason for sequence bias across chromosomes. Chromosomal sequence biases can also be explained by \acrshort{lgd}~\cite{warren2014extensive}Taken together, our analysis unravels inherent differences in nucleotide sequence patterns across human chromosomes, which demands further investigation. It is also apparent that machine learning models should be created in a chromosome-specific manner to enable various genotype-phenotype association studies.

\subsection{Classifying cancerous and non-cancerous mutations}\label{res:ch6:network}

Identifying cancer mutations is vital in various clinical settings, albeit challenging. The most common use case is detecting somatic mutations from tumor specimens in the absence of matched normal tissue~\cite{sun2018computational}. This causes the under-utilization of clinical sequencing outputs. On a separate note, \acrfull{tmb} is estimated by counting cancer-related somatic mutations from cancer specimens. A robust pipeline for cancer mutation detection includes the subtraction of germline variants obtained from matched normal samples. \acrshort{tmb} estimation has been proven to be an efficient way to monitor cancer treatment~\cite{fancello2019tumor}. Due to the challenges involved in obtaining tissue biopsies, it is crucial to assess \acrshort{tmb} using \acrshort{cfdna} from blood, which may include \acrfull{ctdna}. In the absence of matched normal samples, the germline variant databases are used for \textit{in-silico} filtering. These methods are suboptimal and can benefit significantly from the normal-free detection of cancer mutations. We investigated if a classifier can be trained to classify cancerous and non-cancerous mutations.

We used \acrshort{crcs}es to classify codon switch sequences into two categories, namely codon switch sequences harboring cancerous mutations or non-cancerous mutations. Due to the significant computational overhead, we focused on the sex chromosomes for downstream analysis. We note that $\sim$70 protein-coding genes harbored by chromosome Y offer inadequate levels of genetic diversity, thereby trivializing deep learning-based interventions. On the other hand, we obtained about 107,000 high-quality variants across $\sim$800 genes from the ExAC browser for chromosome X. We considered splicing events when populating codon switch sequences for training a custom neural-network architecture for classification. Notably, we generated embeddings for all 640 codon switches independently for each chromosome, and we found substantial heterogeneity, which the chromosomal amino acid composition biases can explain. As such, genome-wide applicability of \acrshort{crcs} warrants independent model building for each specific chromosome (Figure~\ref{fig:ch6:panel2}).

\begin{figure*}
	\makebox[\textwidth][c]{
		\resizebox{1 \linewidth}{!}{
			\includegraphics{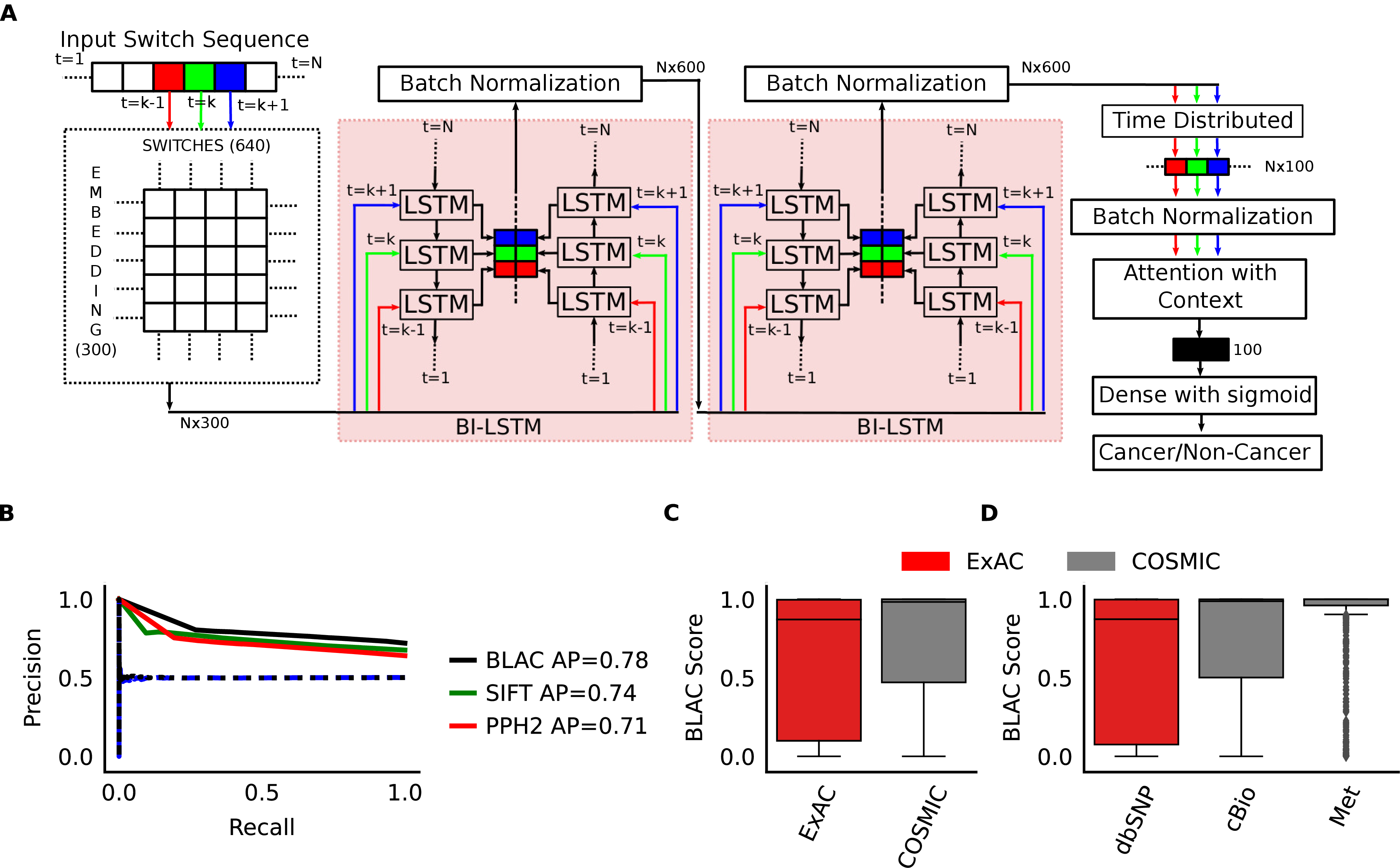}
		}
	}
	\caption{\textbf{Classification of cancerous and non-cancerous variants.} \textbf{A)} Deep learning architecture, used for \acrshort{crcs}-based classification of ExAC/COSMIC variants. \textbf{B)} \acrfull{pr} curve for the \acrshort{blac} after 200 epochs. The red and green curves indicate the performance of SIFT and Polyphen2, respectively. Validation performances were measured on fake alteration classes, constructed by randomly splitting cancer/non-cancer alterations into two equal-size groups. The black dashed line represents the performance of the fake test set created from COSMIC data. Similarly, the blue dashed line is for ExAC data. Both PR curves thus obtained, as expected, collapsed on the 0.5 precision line. \textbf{C)} Boxplots depict the distribution of prediction scores (probability of being a cancer alteration), assigned to the ExAC and COSMIC alterations, in the validation set (across all folds). \textbf{D)} Similar trends are observed for non-pathogenic dbSNP alterations and mutations found in cancer patients from Met and cBioPortal. Scores on these datasets were predicted using the model trained on the full dataset.}
	\label{fig:ch6:panel3}
\end{figure*}

For predicting non-cancerous/cancerous mutations, we trained our custom neural network architecture, \acrfull{blac}, using 34,981/66,165 unique X-chromosome specific substitutions from ExAC/COSMIC databases spanning 332 protein-coding genes (Figure~\ref{fig:ch6:panel3}A). Four-fold cross-validation was used to evaluate the performance of \acrshort{blac}-based detection of cancer mutations. 

We obtained an \acrfull{ap} (i.e., the area under the precision-recall curve) of 0.78 on the four validation sets, indicating predictability of the mutational sub-types (Figure~\ref{fig:ch6:panel3}B). We posed a similar classification problem by randomly splitting the \acrshort{snv} pool obtained from the ExAC browser as a control. As expected, we obtained an \acrshort{ap} of 0.5. With COSMIC alterations, our finding was similar (Figure~\ref{fig:ch6:panel3}B). This strongly supports the conclusion that differential nucleotide contexts surround cancer-related somatic mutations compared to non-cancerous variants. We compared the performance of the trained model with SIFT~\cite{vaser2016sift} and Polyphen2~\cite{adzhubei2010method}. These methods are widely used to predict the deleterious nature of mutations, using sequence homology and amino acids' physical properties. SIFT and Polyphen2 yielded lower values of AP (0.74 and 0.71, respectively), indicating the superiority of our sequence-based approach. Notably, \acrshort{blac} predictions are based on unseen genes (due to our implementation of cross-validation),  whereas SIFT and Polyphen2 use models trained on the entire genome. Under the current experimental setting, SIFT and Polyphen2 enjoy a significant relaxation in terms of the stringency of cross-validation.

Although significantly different, the median distributions of scores on the ExAC and COSMIC datasets (Figure~\ref{fig:ch6:panel3}C) are on the higher end of the spectrum, which has a high false-positive rate at a threshold of 0.5. As a result, there is still a significant area of uncertainty in the probability distribution where it is impossible to distinguish between malignant and non-cancerous mutations. We hypothesized that only some cancer-related mutations occur in an exclusive nucleotide environment since many cancer-related mutations had low probability scores. The mutations with a high likelihood of being malignant are returned when the threshold value is set at 0.9, despite the fact that doing so decreases the model's sensitivity. Table~\ref{tab:ch6:specificity_sensitivity} shows the value of specificity, sensitivity, and F1-score at the threshold of 0.9. It is evident from the table that Polyphen2 has higher specificity while \acrshort{blac} scores have higher sensitivity and F1-score.

\begin{table}[!ht]
	\centering
	\begin{tabular}{llll}
		\toprule
		\textbf{Method} & \textbf{Specificity} & \textbf{Sensitivity} & \textbf{F1-score} \\
		\midrule
		\midrule
		\acrshort{blac} & 0.518 & 0.620 & 0.686\\
        SIFT & 0.556 & 0.561 & 0.633\\
        Polyphen2 & 0.632 & 0.499 & 0.585\\
		\bottomrule
	\end{tabular}
	\caption{Specificity, Sensitivity, and, F1-score values at the threshold of 0.9. This value of threshold was chosen since predictions of all the algorithms are skewed toward high values. These metrics are computed on the predicted scores on mutations reported in ExAC and COSMIC databases.}
	\label{tab:ch6:specificity_sensitivity}
\end{table}

We examined our model trained on ExAX/COSMIC on three separate datasets. After deleting entries marked as pathogenic and likey-pathogenic, we extracted neutral \acrshort{snv}s from the dbSNP database~\cite{sherry2001dbsnp}. We considered somatic \acrshort{snv}s from Met study for a matching cancer alteration pool, a recently published pan-cancer study of solid metastatic tumors~\cite{priestley2019pan}. Met sequenced and analyzed 2,520 Dutch population tumor samples  We adhered to the filtering standards specified in Section~\ref{sec:ch6:coding}. Sequences that overlapped the training set were eliminated, leaving 289,418 non-cancer and 1,151 cancer-related mutations. As a second source of information on cancer mutations, we used cBioPortal for validation. We collected all 287 studies and selected chromosomal X mutations. After applying the aforementioned filtering criteria, the original 246,201 mutation count was reduced to 147,049. As expected, cancer mutations were assigned relatively higher \acrshort{blac} scores (Mann-Whitney U-test $P$-value $<$ $0.01$), thereby underscoring the robustness and cross-demographic reproducibility of our predictions (Figure~\ref{fig:ch6:panel3}D). 

\begin{figure}[!ht]
	
	\makebox[1 \textwidth][c]{
		\resizebox{1 \linewidth}{!}{ %
			\includegraphics{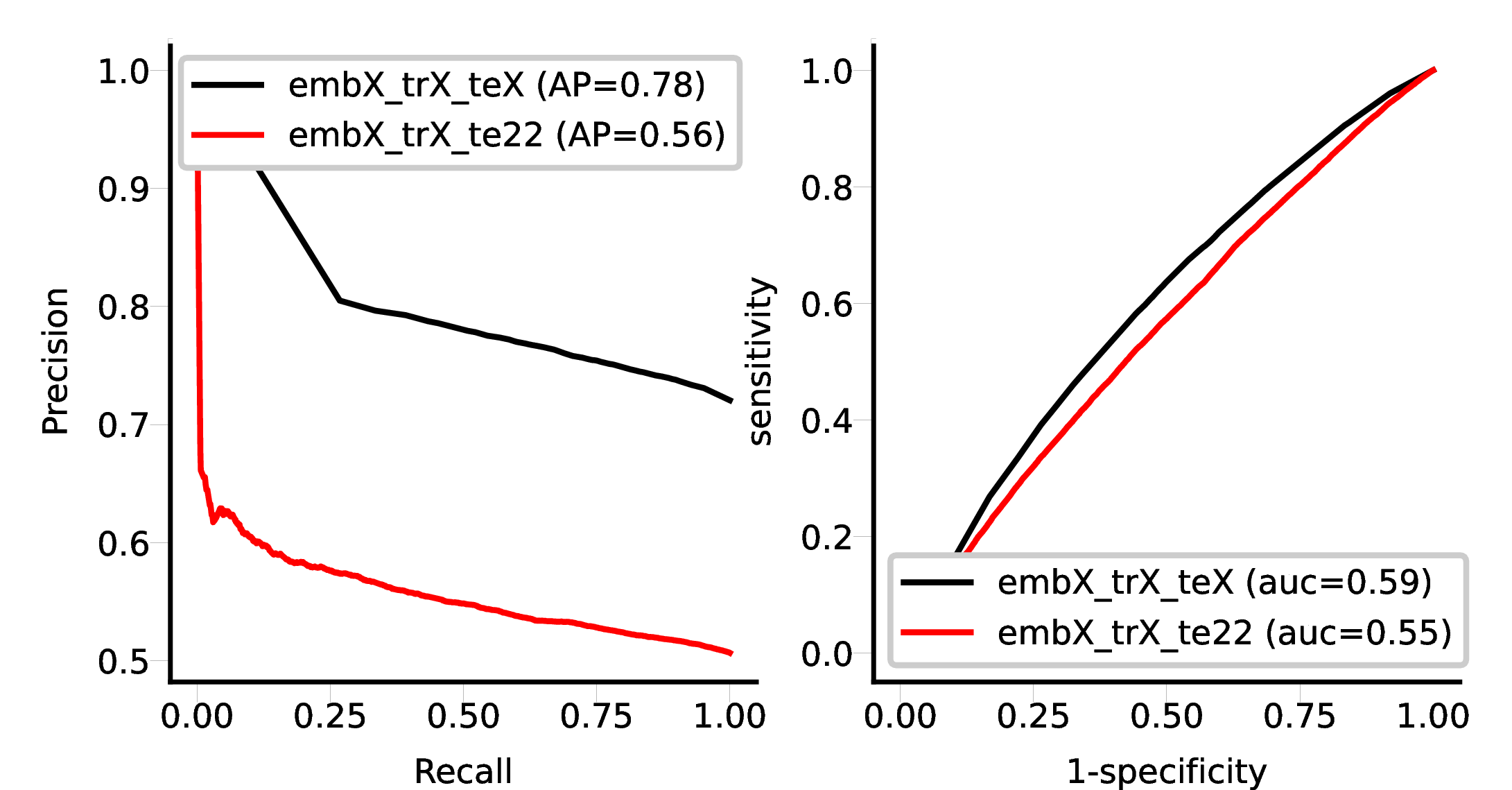}
		}
	}
	\caption{\textbf{Evaluation of model trained on chromosome X against chromosome 22.} As expected, model performance deteriorated. This reduction in performance is due to the fact that the nucleotide distribution in a chromosome is different. Thus a model trained on one chromosome can not be used on the other chromosome without re-training/fine-tuning. Also, the complexity of every chromosome is different, thus same deep learning architecture may not suitable for other chromosomes.}
	\label{fig:ch6:other_combination}
\end{figure}

While interoperability between chromosomes appears intuitive, it might not be optimal for the discussed classification task. As discussed earlier in the section, the embeddings of codon switches are well segregated, indicating apparent heterogeneity (Figure~\ref{fig:ch6:panel2}). To further reinforce this, we predicted the \acrshort{blac} scores on chromosome 22 using the embeddings and classification model trained on chromosome X. As expected, the results on these values were inferior compared to the chromosome-specific model (Figure~\ref{fig:ch6:other_combination}).

\section{Comparison of \acrshort{crcs}-based approach with the existing best practice architectures}

\begin{figure*}
	\makebox[\textwidth][c]{
		\resizebox{1 \linewidth}{!}{
			\includegraphics{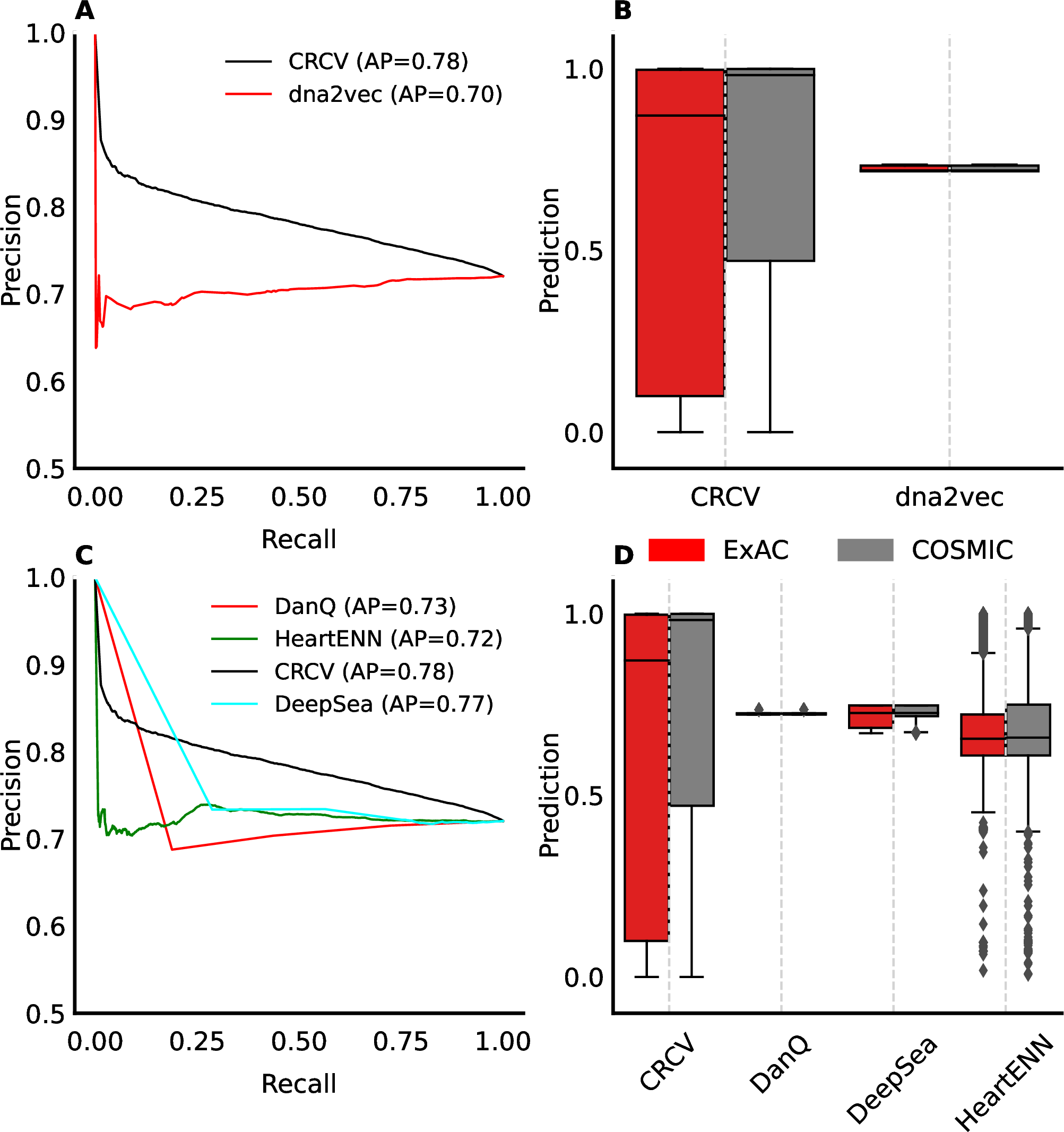}
		}
	}
	\caption{\textbf{Performance comparison of \acrshort{blac} scores with other deep learning architectures.} \textbf{A)} Precision-Recall plot of the predictions obtained from the model trained with \acrshort{crcs} embeddings and dna2vec embeddings. \textbf{B)} Comparison of the distribution of scores obtained from the model. dna2vec does not have any discriminating power (Mann-Whitney U-test $P$-value = $1$). \textbf{C)} Precision-Recall plot of the predictions obtained from other deep learning models, DanQ, DeepSea, and HeartENN. Compared to our proposed model trained with \acrshort{crcs}, other models have inferior performance. \textbf{D)} Comparison of the distribution of scores obtained by models. DanQ does not have any discriminating power (Mann-Whitney U-test $P$-value $= 1$). Other models have a different distribution of scores on ExAC and COSMIC. Mann-Whitney U-test $P$-value for DeepSea and HeartENN is $2.7\times 10^{-20}$ and $9.08\times 10^{-9}$ respectively. Our model with \acrshort{crcs} has the most differentiating power (Mann-Whitney $P$-value is almost near 0).}
	\label{fig:ch6:comparison}
\end{figure*}

As an alternative approach to embedding, we utilized \texttt{dna2vec}~\cite{ng2017dna2vec}, which derives numeric embeddings for variable-length $k$-mers from the reference human genome sequence. Notably, it does not offer a method for learning from \acrshort{snv}s collected by databases like ExAC. Using the \texttt{dna2vec} representation, we encoded the codons from the changed subsequence. Then, we feed these encoded sequences to our novel architecture, \acrshort{blac}. We utilized the same fold as for training with \acrshort{blac} in order to maintain fairness in the comparison. The comparison of the performances of the model trained using these two embeddings is shown in Figure~\ref{fig:ch6:comparison}A, B. It is evident from the figure that dna2vec representation could not discriminate between the cancerous and non-cancerous sequences (Mann-Whitney U-test $P$-value $=$ $1$).

To further validate the performance of our model, three different architectures, namely, \texttt{DanQ}~\cite{quang2016danq}, \texttt{DeepSea}~\cite{zhou2015predicting}, and \texttt{HeartENN}~\cite{richter2020genomic}, were compared. One additional single neuron layer with sigmoid activation (logistic layer) was added at the end to enable these architectures to classify the sequences into cancerous and non-cancerous categories. As prescribed by the authors of these models, one-hot encoded ACTG sequences were provided as input. On the other hand, our model was trained with the \acrshort{crcs} embeddings. Figure~\ref{fig:ch6:comparison}C, D shows the comparison of the performances of these models. Our model performed best (AP=$0.78$) followed by \texttt{DeepSea} (AP=$0.77$). \texttt{DanQ} could not differentiate between cancerous and non-cancerous sequences. (Mann-Whitney U-test $P$-value $=$ $1$). Although \texttt{HeartENN} has differing distribution (Mann-Whitney U-test $P$-value $= 9.08\times 10^{-9}$), it was not able to properly differentiate between the two classes (AP=$0.72$)\footnote{F1-score, specificity, and sensitivity were not reported for these methods since the prediction range for these methods are very small (Figure~\ref{fig:ch6:comparison}B, D). It was not evident which threshold value should be chosen to justify the result. We instead report average precision, which is amortized over different threshold values.}

\subsection{\acrshort{blac} score assists in driver gene exploration}

\begin{figure*}
	\makebox[\textwidth][c]{
		\resizebox{1 \linewidth}{!}{
			\includegraphics{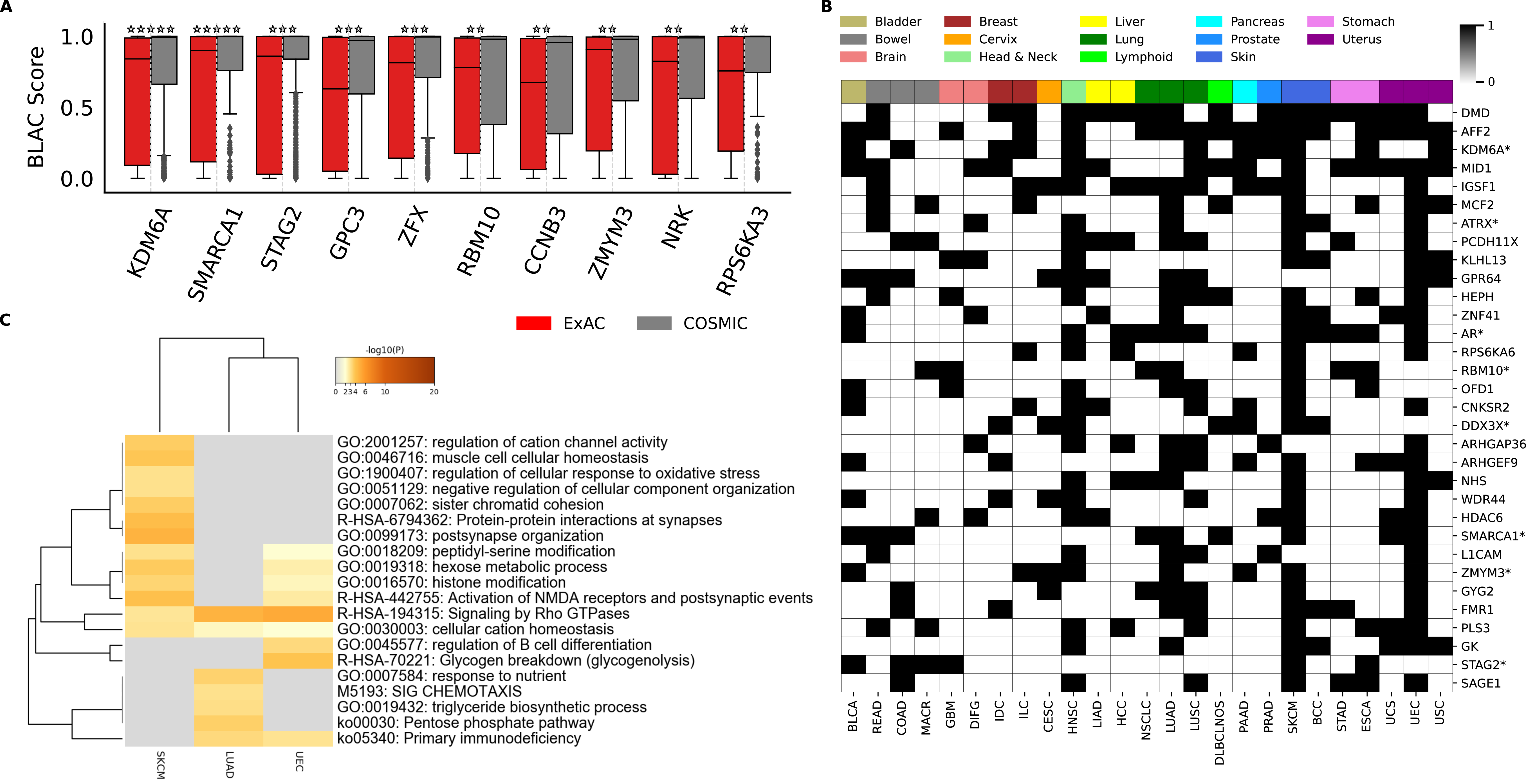}
		}
	}
	\caption{\textbf{Driver gene analysis and exploration.} \textbf{A)} Boxplots show the distribution of prediction scores assigned to ExAC and COSMIC alterations for the known driver genes from the validation set (across all folds). In the figure, 5 stars represent a $P$-value less than $5e^{-15}$. Values in the range $[5e^{-15}, 5e^{-12})$ are represented by 4 stars. Similarly, values in the range of $[5e^{-12}, 5e^{-9})$, $[5e^{-9}, 5e^{-6})$, and $[5e^{-6}, 5e^{-2})$ are represented by 3, 2, and 1 stars, respectively. \textbf{B)} Heatmap shows the genes (in black) that have been marked significant most frequently, across cancer types. For a given cancer type in cBioPortal, a gene was marked significant if the \acrshort{blac} scores of the reported mutations were significantly elevated as compared dbSNP variants. The colors in the top row show the organ of cancer. Gene marked with $*$ are known driver genes. \textbf{C)} Heatmap depicting the cluster-wise enrichment of the prominent biological functions in the indicated cancer types. Of note, the selected cancer types harbored a number of mutational genes identified using the \acrshort{crcs}-based approach. Cancer types that displayed significantly divergent risk groups include \acrfull{skcm}, \acrfull{luad}, and \acrfull{uec}. The scale bar represents the negatively log-transformed (base 10) $P$-values.}
	\label{fig:ch6:panel4}
\end{figure*}

\begin{figure}[!ht]
	
	\makebox[1 \textwidth][c]{
		\resizebox{0.97 \linewidth}{!}{ %
			\includegraphics{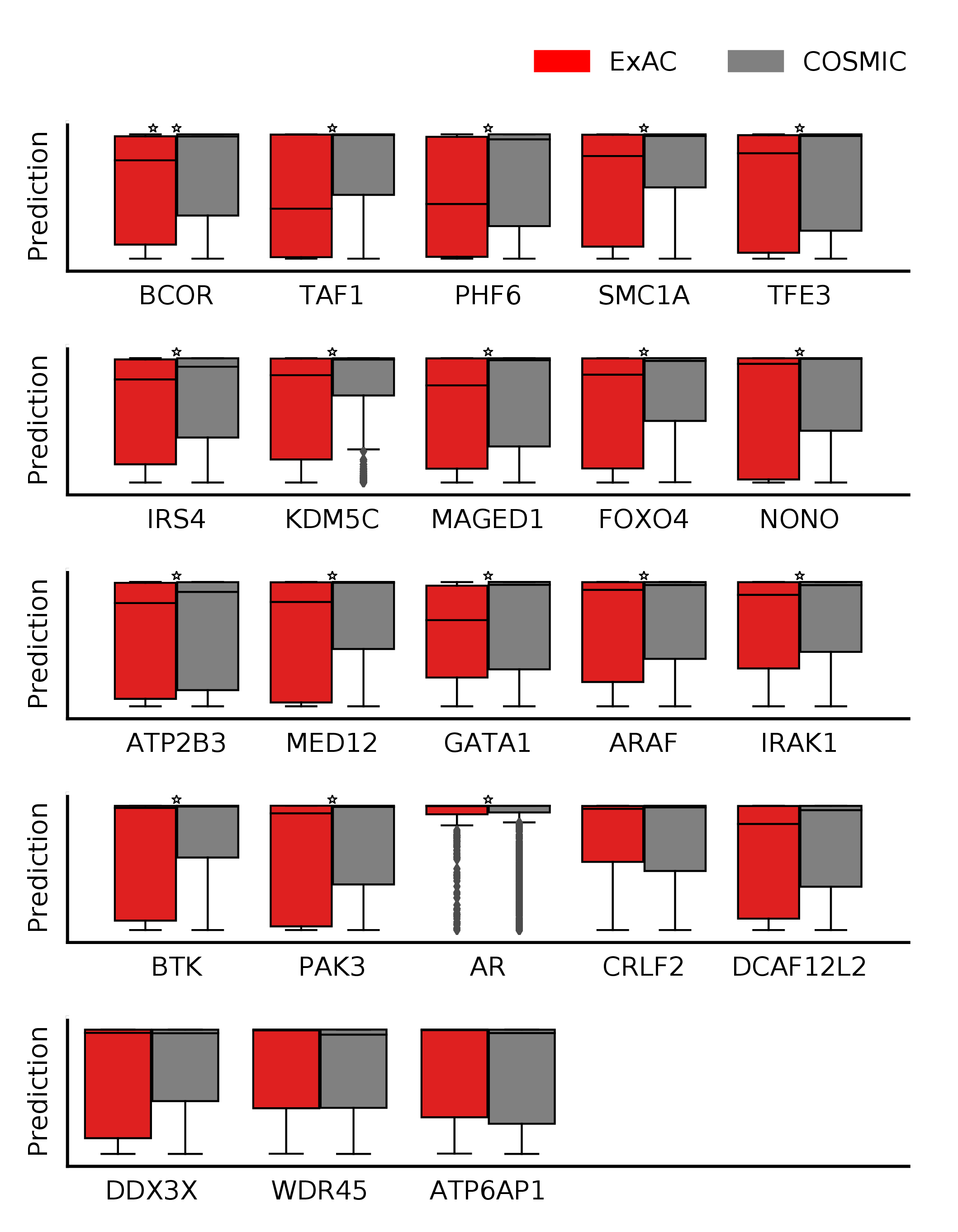}
		}
	}
	\caption{\textbf{\acrshort{blac} score distribution of remaining driver genes.} Boxplots show the distribution of the prediction scores assigned to ExAC and COSMIC alterations for the remaining known driver genes from the validation set (across all folds), except the top 10. Top 10 values are present in Figure~\ref{fig:ch6:panel4}. Stars have the same meaning as in Figure~\ref{fig:ch6:panel4}.}
	\label{fig:ch6:pvalues}
\end{figure}

Driver genes play a pivotal role in the diagnosis and clinical management of cancers. We asked if our model differentiates between driver gene-specific non-cancerous and cancerous mutations. By merging multiple driver gene databases (\acrfull{oncokb}~\cite{oncokb}, \acrfull{intogen}~\cite{intogene}, and \acrfull{cgi}~\cite{cgi}) we obtained 55 potential driver genes on chromosome X, of which 33 were left after filtering. For these 33 driver genes, $\sim$148 and $\sim$680 coding variants were retrieved, on average, from ExAC and COSMIC, respectively. On feeding these variants to our \acrshort{crcs} pipeline, we observed significant differences in the distribution of prediction scores. Figure~\ref{fig:ch6:panel4}A presents the top 10 genes (\textit{KDM6A}, \textit{SMARCA1}, \textit{STAG2}, \textit{GPC3}, \textit{ZFX}, \textit{RBM10}, \textit{CCNB3}, \textit{ZMYM3}, \textit{NRK}, and \textit{RPS6KA3}) based on $P$-values. The distribution of scores for the remaining 23 genes is presented in Figure~\ref{fig:ch6:pvalues}. 

\begin{figure}[!t]
	
	\makebox[1 \textwidth][c]{
		\resizebox{1.1 \linewidth}{!}{ %
			\includegraphics{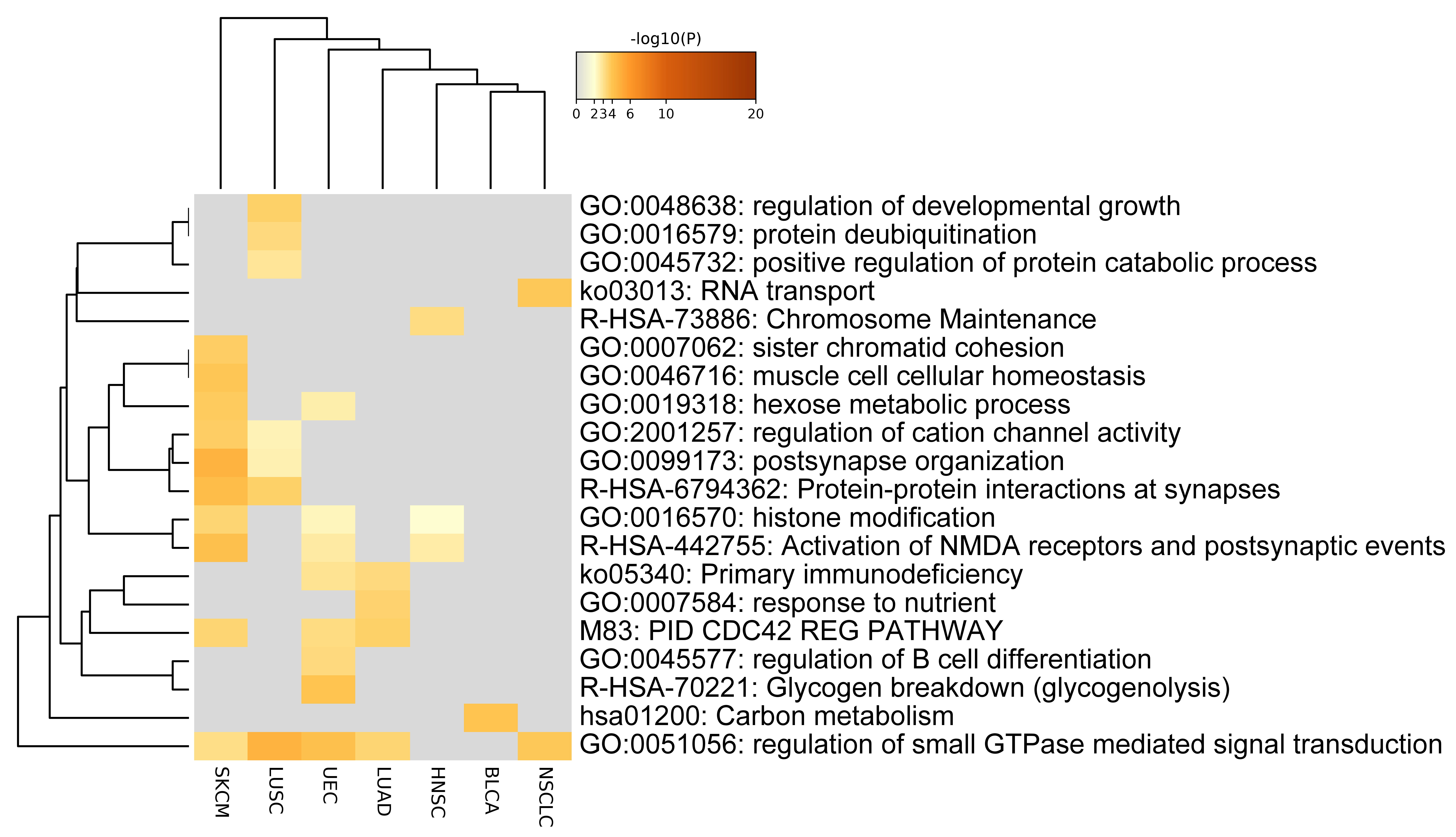}
		}
	}
	\caption{\textbf{Heatmap depicting the cluster-wise enrichment of the prominent biological functions in the indicated cancer types.} Of note, the selected cancer types harbored the number of mutational genes identified using BLAC. Cancer types include Skin Cutaneous Melanoma (SKCM), Lung Adenocarcinoma (LUAD), Undifferentiated Endometrial Carcinoma (UEC), Lung Squamous Cell Carcinoma (LUSC), Head-Neck Squamous Cell Carcinoma (HNSC), Urothelial Bladder Carcinoma (BLCA), and Non-small-cell Lung Carcinoma (NSCLC). The scale bar represents the negatively log-transformed (base 10) $P$-values.}
	\label{fig:ch6:metascape_7}
\end{figure}

We asked if the strength of differential elevation of \acrshort{blac} scores between cancerous and non-cancerous mutations is more pronounced in the case of cancer drivers. For this, we computed the statistical significance of \acrshort{blac} score differences associated with all genes (on Chromosome X) by leveraging cBioPortal (for cancer mutations) and dbSNP (non-cancerous) variant calls (Section~\ref{sec:ch6:cbiodbsnp}). We found adjusted $P$-values (we considered -$\log$10 transformation of the adjusted $P$-values in this case) associated with the known driver genes to be of higher significance than the entire population of X chromosome-specific genes (One-sided Kolmogorov-Smirnov test $P$-value $<$ 0.0493). This indicates that one could use the differential elevation of \acrshort{blac} scores across cancerous and non-cancerous mutations for a given gene as a yardstick for its driver potential. Figure~\ref{fig:ch6:panel4}B reports 32 genes that show significant, cancer-specific \acrshort{blac} score elevation across five or more cancers. Out of 32 genes, 24 were not reported in either of the three databases: \acrshort{oncokb}~\cite{oncokb}, \acrshort{intogen}~\cite{intogene}, and \acrshort{cgi}~\cite{cgi}. Among the genes not cataloged in these three databases, \textit{DMD} is an important candidate. Mutation of the \textit{DMD} gene causes muscular disorders. However, increasing shreds of evidence implicates \textit{DMD} in the development of all major cancer types~\cite{jones2021duchenne}. \textit{RPS6KA6} (aka \textit{RSK4}) has recently been found to play a pivotal role in promoting \acrfull{csc} properties and radioresistance in \acrfull{escc}~\cite{li2020ribosomal}. \acrshort{blac} score-based analyses indicated its potential involvement in pancreas, liver, head-and-neck, and breast cancers. Another intriguing candidate is \textit{OFD1}, a protein involved in ciliogenesis~\cite{tang2013autophagy}. The primary cilium is a thin and long organelle protruding in almost all mammal cell types and is involved in perceiving external stimuli, such as light, odorants, and fluids. The primary cilium also coordinates signaling pathways that convert extracellular cues into cellular responses with the help of receptors and signaling molecules. \textit{OFD1} mutations have been found implicated in Wnt hyper-responsiveness~\cite{fabbri2019primary}. \textit{WDR44}, another enlisted gene, is involved in ciliogenesis~\cite{walia2019akt}. Its role in cancer is still elusive. Genes such as \textit{AFF2}, \textit{MID1}, \textit{PCDH11X}, \textit{MCF2}, \textit{NHS}, and \textit{GYG2} are not reported to have a role in cancer pathogenesis and could be interesting for future validation. Notably, \textit{AFF2} has recently been predicted to have driver roles~\cite{luo2019deepdriver}.

We asked if genes that show differential \acrshort{blac} scores across cancerous and non-cancerous mutations in specific cancer types are functionally interconnected. For this, we used gene ontology analysis by Metascape~\cite{zhou2019metascape}. For the top three cancer types i.e., \acrfull{skcm}, \acrfull{luad}, \acrfull{uec}, harboring the maximum number of genes ($\geq100$) identified by the method discussed in Section~\ref{sec:ch6:cbiodbsnp}. Metascape-based functional enrichment analysis revealed the contribution of identified genes to be largely cancer-specific (Figure~\ref{fig:ch6:panel4}C). For example, the pentose-phosphate pathway~\cite{jin2019crucial,alfarouk2020pentose} and the triglyceride biosynthesis process~\cite{li2020lipid} are highly enriched in \acrshort{luad}. Similarly, the glycogenolysis pathway is enriched in uterine cancer (\acrshort{uec})~\cite{khan2020revisiting}. We relaxed the number of gene cutoff to $\geq40$ and obtained seven cancer types, namely \acrfull{skcm}, \acrfull{lusc}, \acrfull{uec}, \acrfull{luad}, \acrfull{hnsc}, \acrfull{blca}, \acrfull{nsclc}, classified based on the number of genes they possess. Similar to our earlier analysis, we observed cancer-specific pathway enrichments, suggesting functional interconnections between identified genes (Figure~\ref{fig:ch6:metascape_7}). For instance, in the case of \acrshort{blca}, we observed a specific enrichment for the carbon metabolism pathway~\cite{moore2007polymorphisms,massari2016metabolic,newman2017one}. The results suggest that genes that attract more deleterious/driver-like mutations in specific cancers selectively alter different pathways. For example, modifications in the histone pathways are well characterized in multiple cancer types~\cite{kurdistani2007histone,gaspar2018telomere}.

\subsection{\acrshort{blac} enable survival risk stratification in different cancer types}

Characterization of tumor specimens using next-generation sequencing is becoming increasingly common in targeted treatment selection. These processes offer large numbers of alterations per patient. A significant technical difficulty in detecting all somatic mutations from a tissue sample is that it requires the availability of matched normal tissue samples. In practice, paired collection of cancer and normal tissue samples is quite challenging. Even if all somatic mutations are detected, it is hard to conclude unless these are characterized. As such, presently, only a small fraction of these, which are well-characterized, is finally taken into account to devise therapeutic strategies~\cite{sun2018computational}.

\begin{figure*}
	\makebox[\textwidth][c]{
		\resizebox{1 \linewidth}{!}{
			\includegraphics{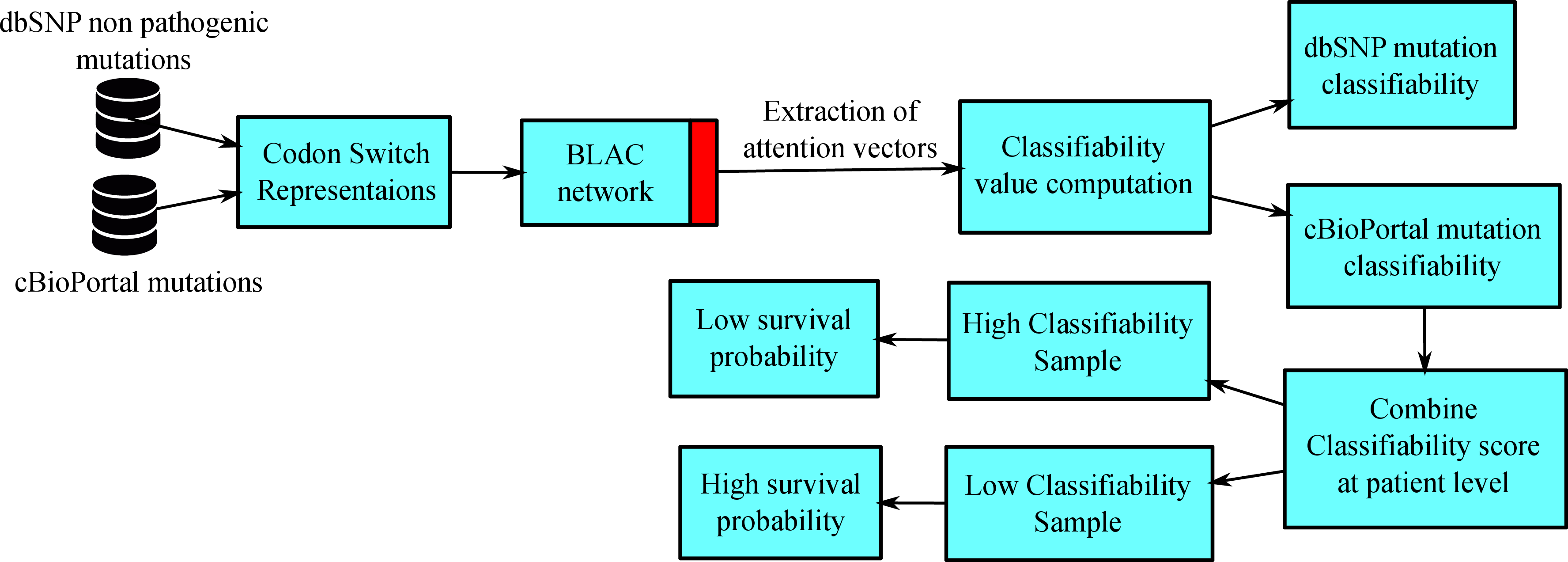}
		}
	}
	\caption{\textbf{Overview of survivability analysis using classifiability}. On the combined dataset of \acrshort{dbsnp} and \acrshort{cosmic}, attention vectors were extracted from the \acrshort{blac} network. These scores are used for classifiability computation. Then groups pertaining to \acrshort{cosmic} were extracted along with their classifiability score and used for survivability analysis.}
	\label{fig:ch6:combinedsurival}
\end{figure*}

Since the advent of massively parallel sequencing platforms, numerous sophisticated methods have been developed for the stratification of patients with differential prognoses. Most of these methods map missense mutations to genes, thereby losing their individualities. For example, Hofree and colleagues, in a seminal paper, mapped somatic mutations to gene networks to cluster tumors (genome sequences) after network smoothing using random walk with restart~\cite{hofree2013network}. Clusters of patients thus obtained indicated significantly differential survival patterns.  Milanese and colleagues leveraged putative functional mutations to predict recurrence in breast cancer. Their approach is also based on mapping mutations to genes~\cite{milanese2021etumormetastasis}. We hypothesized that \acrshort{crcs} could be used for risk stratification using mutation-level information only.

Our construction of the cancer and non-cancer mutation classification problem unavoidably discounts the fact that some cancer-related somatic mutations could indeed be randomly located and hard to differentiate from other non-cancerous somatic mutations. This could be the primary reason for the overlapping \acrshort{blac} scores associated with the two categories. We, therefore, inferred that mutations with extremely high classifiability in \acrshort{cbio} might indicate a higher degree of contribution to the cancer hallmarks.

Figure~\ref{fig:ch6:combinedsurival} presents the details of the survival study. This study was performed on the joint dataset of \acrshort{cbio} and \acrshort{dbsnp}. All kinds of overlap between \acrshort{exac}+\acrshort{cosmic} and these datasets were removed before passing the constructed switch sequences to the pre-trained model on \acrshort{exac}+\acrshort{cosmic} for prediction. The network also outputs the attention vector along with the prediction score. The classifiability measure is computed on these attention vectors, and the classifiability values for the attention vectors corresponding to \acrshort{cbio} were extracted. Since, after removing a few mutations marked as pathogenic and likely-pathogenic, \acrshort{dbsnp} datasets mostly contain mutations found in healthy individuals. Thus, \acrshort{cbio} mutations found in the neighborhood of \acrshort{dbsnp} mutations may have a lower contribution toward cancer hallmarks. Thus, increasing, in turn, patient survivability. However, mutations with higher classifiability are very different compared to \acrshort{dbsnp} mutations. Thus, patients with mutations that are easy to classify will have lower survivability, and patients with mutations that are difficult to classify will have higher survivability.

\begin{figure*}
	\makebox[\textwidth][c]{
		\resizebox{1 \linewidth}{!}{
			\includegraphics{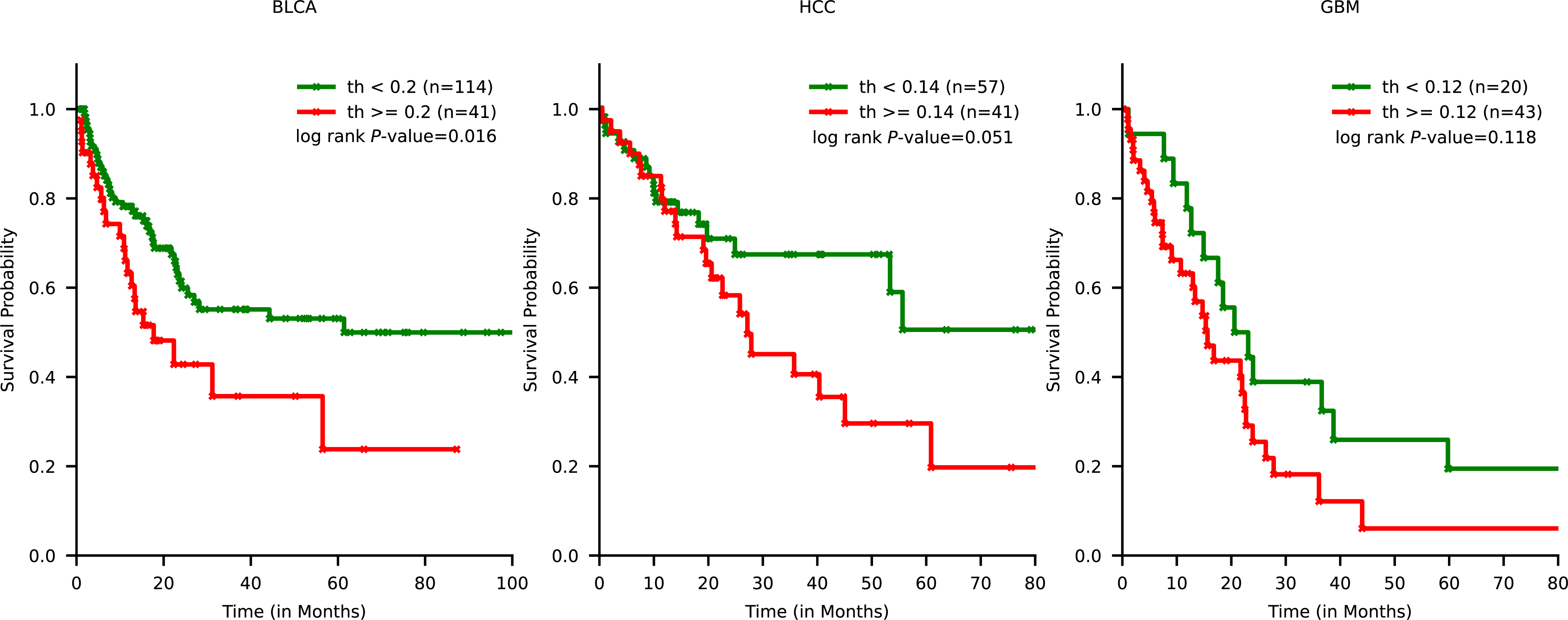}
		}
	}
	\caption{\textbf{Survival risk stratification based on classifiability}. Patients with lower average \acrshort{blac} scores in \acrfull{blca}, a subtype of bladder cancer, has better survival. Similar trends are also visible in \acrfull{hcc}, a subtype of brain cancer, and \acrfull{gbm}, a subtype of lung cancer.}
	\label{fig:ch6:panel5}
\end{figure*}

Out of 8 cancer types that qualified our data filtering criteria, three showed significant (log-rank $P$-value $<$ 0.1, except GBM where it is close to 0.1) survival risk stratification based on this score (Figure~\ref{fig:ch6:panel5}). The three cancer types are – \acrfull{blca}, \acrfull{hcc}, and \acrfull{gbm}. In each case, patients with higher classifiability were mapped to the high-risk group. Significant aberration of the X chromosome has already been reported in \acrshort{hcc}~\cite{liu1993detection,liu2004aberration}. In \acrshort{blca}, similar reports exist highlighting the association of \textit{KDM6A} hypermutation with anti-tumor immune response efficacy~\cite{chen2021significance,sangster2022mutually,kaneko2018x,sidransky1992clonal}. Notably, \acrshort{blac} scores of cancer-associated somatic mutations in \textit{KDM6A} show higher elevation as compared to germline or other non-cancerous somatic mutations (Figure~\ref{fig:ch6:panel4}A).

\section{Discussion}\label{sec:ch6:discussion}

The majority of cancer mutations are thus far understood as hitchhikers. No notable computational work convincingly indicates the contextual difference between cancerous and non-cancerous mutations. This problem has great use in clinics since often matched normal sequences are not available for confident somatic mutation calling. A good example is detecting mutations in \acrlong{cfdna} for diagnostic purposes or measuring the mutational burden post-chemotherapy or immunotherapy~\cite{jiao2020tumor}. Furthermore, current computational approaches for risk stratification of cancer patients under-utilize information captured at the mutation level. These methods typically map missense mutations to genes for further downstream prediction tasks. The current work reports a strategy to address this by learning numeric vector-based representations of mutations (i.e., embeddings) that are more amenable to machine learning tasks.

Cancer mutations have been found to occur at proteins' active areas~\cite{reva2011predicting} and at evolutionarily conserved sites~\cite{walker1999evolutionary}. Additionally, mutagenesis activities like smoking and UV exposure have been linked to specific mutational signatures (i.e., specific changes surrounded by certain surrounding nucleotides)~\cite{poon2014mutation}. However, no holistic approach is visible to discriminate between cancerous and non-cancerous variants. In fact, it is still elusive whether cancer mutations are exclusive in nature. The current work answers this in the affirmative.

There is currently no widely accepted method for embedding individual changes while maintaining their nucleotide sequence context similarity. \texttt{mut2vec}~\cite{kim2018mut2vec}, which gives mutational embeddings at the gene level by utilizing text mining and protein-protein interaction networks are one obvious attempt at this aim. However, No underlying biological process is mimicked by \texttt{mut2vec} strategy. Our method makes it possible to learn semantic representations of mutations from massive amounts of uniformly processed exome sequencing data for the first time.

Our \acrshort{crcs} and \acrshort{bilstm}-based custom deep learning architecture (\acrshort{blac}) can discriminate between cancerous and non-cancerous mutations with a limited number of labeled samples from each category. This is due to the powerful representation learning accrued by the \acrshort{crcs}es by ingesting a large pool of \acrshort{snv}s from tens of thousands of exome sequencing data. The prediction is not black and white as the \acrshort{blac} score spectrum associated with cancerous and non-cancerous mutations overlap substantially. However, we see an apparent elevation in the distribution of \acrshort{crcs} scores for cancer mutations. This provides evidence for the exclusive nature of cancer mutations and the fact that cancer mutations, by and large, differ from germline or other somatic mutations in terms of nucleotide context.

Most of the existing deep learning architectures proposed for solving various tasks using sequence data use \acrfull{cnn} as the building block. There are a few fundamental problems in using a \acrshort{cnn} layer to solve the problem as put forth in the manuscript. i) In the \acrshort{crcs} representation, there is only one codon switch in the sequence containing actual mutational information. Utilizing \acrshort{cnn} followed by a max-pooling on this representation may drop the variability introduced by a single codon switch. Thus, it becomes challenging for the model to differentiate among mutations on an \acrshort{mrna}. ii) A \acrshort{cnn} layer is not able to handle variable-length sequences. The maximum length of the \acrshort{mrna} sequence needs to be fixed at the input, and then other variable-length sequences need to be padded with zeros to be fed into the network. This restriction on the input size limits the generalizability of the model.

\acrshort{lstm}s are intended specifically to capture long-range dependencies, in contrast to \acrshort{cnn}s. A single change in the codon switch sequence can change the state of the \acrshort{lstm} unit, which will ultimately affect the network's prediction. Bi-\acrshort{lstm}s, a variation of \acrshort{lstm}s, examine both ends of the sequence simultaneously, which enhances the network's prediction even more. Additionally, \acrshort{lstm}s are appropriate for variable-length sequences due to their recurrent nature. In addition to these, the numerical representation of the \acrshort{dna} sequence was produced in other proposed networks by concatenating one-hot encoded nucleotides. No semantic information can be captured by such representations. We came to understand that these issues make it impossible to access the information content hidden in variable-length flanking sections surrounding the desired coding variant.

In a nutshell, our findings suggest that cancer-specific \acrshort{snv}s, including passenger mutations, occur with differential nucleotide contexts compared to coding variants observed in healthy populations. One significant advantage of \acrshort{crcs} is that it is not reliant on any clinical or pathological parameters. We predict that the proposed approach could be adopted in attempting a broad range of questions concerning genotype-phenotype interlinking. 

\section{Conclusion}

In this chapter, a new approach to assessing the deleteriousness of cancer mutations is presented. To achieve this, a novel representation of mutations as codon switches is constructed. This mutation representation is then used to construct codon switch sequences. Then a numerical representation of codon switches (or switch embeddings) is constructed using a simplified version of \textit{skip-gram} with negative sampling. These switch embeddings are termed as \acrfull{crcs}. The learned embeddings are then used in the downstream task to annotate every mutation based on its deleteriousness. The developed deep learning model is termed as \acrfull{blac}. The output of \acrshort{blac} is called \acrshort{blac} score and is used to identify driver genes. The deleteriousness annotations for \acrshort{blac} network are benchmarked against multiple state-of-the-art networks and methods. The classifiability measure computed on the attention vectors from \acrshort{blac} network is used to assess the survivability of a cancer patient. 

%% file: conclusion.tex
The thesis presented various approaches to generate inferences from the bins generated by space-partitioning-based hashing functions. There are two major divisions of the thesis, the first division focuses on building a tree-based algorithm to perform faster approximate nearest neighbor in hamming space and uses hashing ideas to build global-partitioning-based hashing classifiers. In the second part, the thesis focuses on a case study of how hashing ideas predict cancer patients' survivability.

Chapter 2 discusses a geometrically motivated novel approach to build an approximate nearest neighbor search algorithm in hamming space \acrshort{combi}. The space-partitioning-based hashing algorithms that assign bit codes to every sample require the search to be performed in hamming space. The bit codes have the inherent property that they can be organized in the form of \acrshort{bsts}. However, a naive arrangement causes many issues from memory and search time points of view. Thus, to mitigate such issues, the \acrshort{combi} draws inspiration from the geometry of the bins in the partitioned space. A \acrshort{combi} is generated from the geometrical view by merging nearby empty bins with the filled bins. The idea was validated with extensive empirical evaluation. A study on the quality of nearest neighbors is also presented. The idea of \acrshort{combi} was then extended to the distributed computing environment. The idea of hashing to understand neighbourhoods motivates the idea of extending it to classification, which is fundamentally connected to labels of known samples in the neighbourhood. These ideas are explored in Chapter 3.

Chapter 3 presents and formalizes the idea of building the hashing classifiers. The chapter starts with a brief introduction to Bayes' classifier. Then a general approach to building a classifier using hashing techniques is discussed. Then examples of three hashing classifiers based on Projection hashing, Sketching, and Binary hash are discussed. A detailed discussion on the pros and cons of these classifiers is presented then a need for a tree arrangement of hashing planes is motivated.

Chapter 4 extends ideas from Chapters 2 and 3. This chapter shows how the tree developed in Chapter 2 can be utilized for classification and how it extends the ideas of hashing classifier by making a Global partitioning-based classifier. The invented algorithm is called \acrfull{graf}. The build classifier results in an extension of the random forest. It also shows how a single tree in \acrshort{graf} can represent boosting algorithms. The developed classifier can also be interpreted as the boosting ensemble, thus bridging the gap between bagging and boosting algorithms. We present extensive empirical evaluation to establish the superiority of methods and identify the type of datasets where the method outperforms. Then an unsupervised version of \acrshort{graf}, \acrshort{ugraf} is developed, which can be used to build guided hashing to capture better neighborhood information.

Chapter 5 discusses some applications of \acrshort{graf}. First, \acrshort{graf} is shown as a data approximator where it assigns a sensitivity score to every sample in the dataset, which can be used to reduce the data size. The higher the sensitivity score, the more important that sample is for data approximation. These sensitivity scores can be used to sample the important samples from the data. Then, an application of the duo of \acrshort{ugraf} and \acrshort{combi} is presented to compute the classifiability of every sample.

Chapter 6 uses the classifiability measure developed in Chapter 5 to estimate the survivability of cancer patients. First, the notion that "there is a difference between the mutational landscape of cancer patients and healthy individuals" is established to facilitate this. A new representation of mutations on a genome is developed to quantify the difference. The numerical descriptors of these representations were then learned using word2vec algorithms. The learned embedding is then used in the downstream tasks to classify codon switch sequences carrying mutations from healthy individuals and cancer patients. The classification model can identify these sequences separately with reasonable confidence. Then the classifiability analysis is performed on the descriptors created by classification models to identify highly potent cancer mutations and correlate them with the patient survivability.

In summary, the thesis attempts to extend the horizon of usability of hashing in the classification and other aspects of machine learning by performing multiple types of interpretation on the bins created during hashing. However, many future directions can stem from the presented work. A brief discussion of some ideas is presented below.

\section{Future work}

\subsection{\acrshort{combi} \& \acrshort{ugraf} for clustering in hamming space}\label{sec:ch7:clustering}

In the previous chapters, \acrshort{combi} has been used as a fast approximate nearest neighbor search algorithm, and another interpretation of \acrshort{combi} resulted in a tree-based classification algorithm \acrshort{graf}. The previous chapter also discussed \acrfull{ugraf} which is closer to \acrshort{combi}. The \acrshort{ugraf} can also be used for clustering in hamming space. The rough sketch of the idea is as follows:

\begin{itemize}
    \item Since there are no empty bins in \acrshort{combi}, thus for any sample, the nearest filled bin will be $1-mutate$ away. This property of \acrshort{combi} can be used to create a $1-mutate$ graph.
    \item The graph construction goes as follows:
    \begin{itemize}
        \item All those samples that are $1-mutate$ away from each other can be put into the adjacent vertices of a graph.
        \item Samples at the $2-mutate$ away can have an edge with one jump and so on.
    \end{itemize}
    \item Then, in these graphs, strongly connected components can be extracted and treated as clusters. However, with only a single tree, many samples may get assigned to the wrong clusters (Section~\ref{sec:ch2:drawback}), since the merging of blank spaces may cause far away samples to share the bit code. Thus, multiple \acrshort{combi} tree needs to be generated, and their $1-mutate$ graph needs to be produced.
    
    \item Then, a consensus graph can be created by adding and normalizing the individual graphs.
    \item From the consensus graph, a binary graph can be constructed by thresholding the edge weights in the consensus graph.
    \item In the resulting binary graph, connected components can be extracted to get clusters.
\end{itemize}

The primary concern here is the consensus algorithm - merging (adding and normalizing) all the graphs generated \acrshort{combi} - does not yield better results. The improper threshold selection leaves the resulting binary graph with very few connected components, thus making clustering difficult. These problems can be solved by developing a better consensus algorithm and thresholding strategy.

\subsection{Pan chromosome \acrshort{blac}}

The work presented in Chapter 6 builds only on chromosome X. This work can be extended to other chromosomes as well and a pan-cancer system can be created using the proposed algorithm in Chapter 6. However, the inherent diversity of chromosomes (Figure~\ref{fig:ch6:panel2} and Figure~\ref{fig:ch6:other_combination}) and the sheer amount of data generated by combining chromosomes into single data may hinder the training process. Thus we may use transfer learning to train on the remaining chromosome and build an independent model for every chromosome. However, this solution also needs more experimentation because \acrshort{blac} architecture for chromosome X may not be complex enough to capture internal variations of chromosomes.

\subsection{Better architectures to handle extreme length variations}

Gene lengths are extremely variable. Their length can vary from a few base pairs to a few thousand base pairs. Learning long-range dependencies on such longer sequences is not a trivial task. Further, as discussed in Section~\ref{sec:ch6:classification}, due to the extreme lengths of genes in chromosomes we only worked with sequences shorter than $1500$ nucleotide (Section~\ref{sec:ch6:classification} and Figure~\ref{fig:ch6:filter}). Even with this artificial limitation, training such networks on all chromosomes requires a tremendous amount of computing power and time. Thus, more research is needed in this direction to be able to learn models with better performance.

\subsection{New horizon to learn better embedding}

word2vec~\cite{skipgram} learns static embeddings. It means that the embeddings do not change with context. For example, if a mutation $A\rightarrow C$ occurs at several locations in a chromosome, its numerical representation will not change once it has been learned. This situation is restrictive because if a mutation has different functionality in different locations in a chromosome, this information would not be present in the embedding.

The human genome has $\sim$3 billion base pairs, and every position is capable of mutations. Also, there are different kinds of mutations, e.g., single base-pair, double base-pair, triple-base pair, insertion, deletion, complex indels, complex indels with substitution, translocation, etc. Thus, it is impossible to learn embedding of a mutation in every position because of almost infinite possibilities. 

In the recent advancements in the \acrfull{nlp} domain, context-aware embedding has been proposed. They give a different numerical representation of a word depending on its surroundings. This strategy can be used to learn positionally or context-aware embedding for mutations. \acrfull{bert}~\cite{devlin2018bert} and its variants are examples of architectures that generate such embedding. But there are a few issues with \acrshort{bert} in the context of genomics:

\begin{itemize}
    \item Although \acrshort{bert} can learn contextual representations, its input size is fixed. \acrshort{bert} has no shared parameters; thus, length extremities and exploding parameter count in \acrshort{bert} input space pose the biggest computation challenge. There has been some interest in creating recurrent transformer architectures~\cite{hutchins2022block} with shared parameters. Of note, BERT is based on transformer architectures~\cite{vaswani2017attention}.

    \item \acrshort{bert} solves two problems to learn embeddings, masked predictions, and next-sequence prediction. The masked-prediction problem of BERT can be understood as the \textit{skip-gram}, but the next-sequence prediction problem will require additional analogy with biology. \acrshort{dna}\acrshort{bert}~\cite{ji2021dnabert} is one attempt to train BERT on the human genome where the authors dropped the next-sequence prediction problem to learn the embedding.
    
\end{itemize}

One possible way to include the next-sequence prediction while training \acrshort{bert} is to assume exon and intron as two sequences and formulate the problem to predict the following intron from a given exon or predict exon from a given intron. However, this idea needs more refinement, and further research is needed to establish the viability of the proposed solution.

\subsection{Extended switch dictionary and their embedding}

In Section~\ref{sec:ch6:crcv} we discussed the steps to construct the codon switch dictionary containing only single point mutation. We obtained 640 unique switches corresponding to single-base substitutions. This number also includes the cases with no mutations. Following similar steps, double and triple base substitutions can be modeled by $1728$ switches each.

\begin{figure}[!ht]
	
	\makebox[1 \textwidth][c]{
		\resizebox{1.1 \linewidth}{!}{%
			\includegraphics{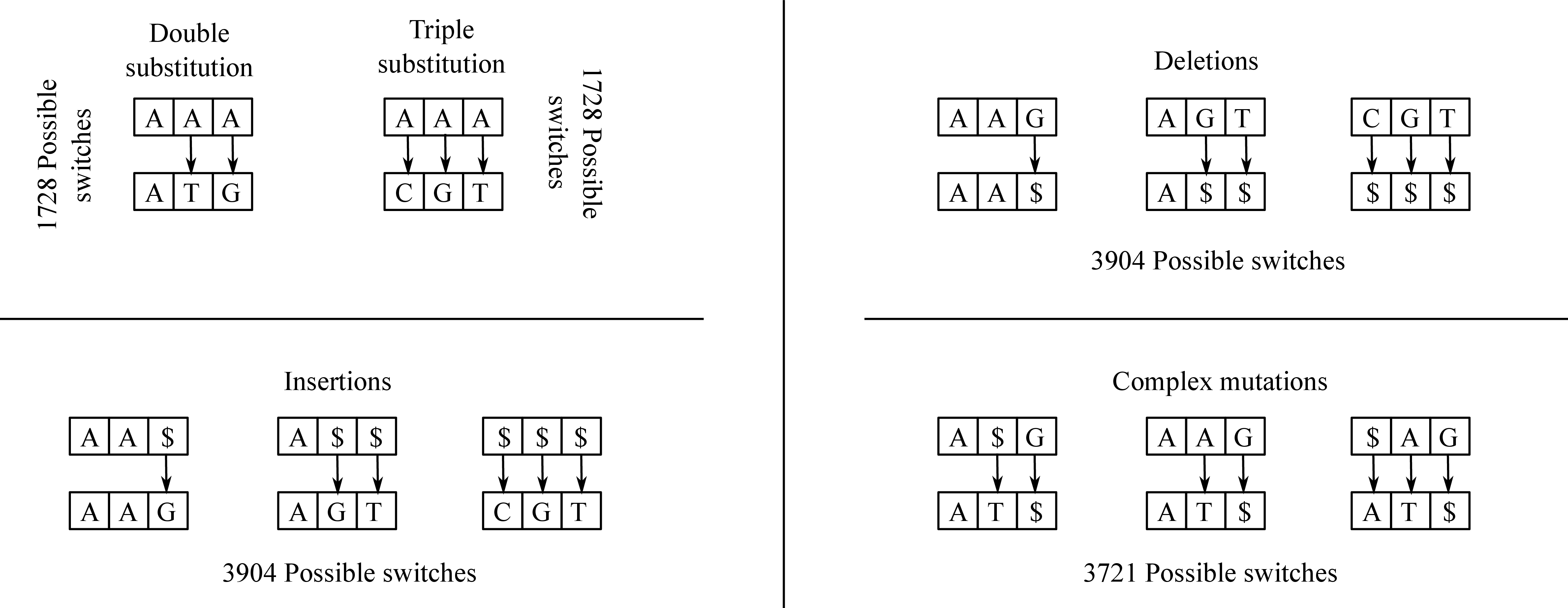}
		}
	}
	\caption{\textbf{Extended codon switch dictionary.} The strategy presented here facilitates the representation of any kind of mutation via these elements. Since switches are constructed on a codon, only three types of substitutions are considered: i) Single base substitution. ii) Double base substitution. iii) Triple base substitution. With the introduction of \$, the representation can handle insertions, deletions, and complex indels.}
	\label{fig:ch7:multimutation}
	
\end{figure}

\begin{table}[!ht]
	\centering
	\begin{tabular}{llll}
		\toprule
		\textbf{Mutation} & \textbf{Total} & \textbf{Available switches} & \textbf{Average frequency}\\
		\textbf{type} & \textbf{switches} & \textbf{in data} & \textbf{of available switches} \\
		\midrule
		\midrule
		Single base substitutions & 640 & 640 & 495\\
		\midrule
		Double base substitutions & 1728 & 66 & 1.5 \\
		(consecutive $\&$ non-consecutive) & & & \\
		\midrule
		Triple base substitutions & 1728 & 53 & 1.7\\
		\midrule
		Insertions & 3904 & 290 & 10\\
		\midrule
		Deletions & 3904 & 357 & 25\\
		\midrule
		Complex Indels & 3721 & 0 & 0\\
		\bottomrule
	\end{tabular}
	\caption{Frequencies of switches in different variant types formed using ExAC dataset.}
	\label{tab:ch7:switchFreq}
\end{table}

To include the insertion and deletions into the dictionary, we need to introduce another character, \$, to represent a space. The reference codon will have \$ at the point of insertions to construct switches that represent insertions. For example, in a nucleotide sequence $AA$, if there is an insertion of $T$ at the beginning, the constructed switch will have the form $\$AA\rightarrow TAA$. Similarly, other switches can be constructed for insertions. Assume that a sequence $TACGTCT$ is inserted between $AA$, then we assume that the reference sequence is $A\$\$\$\$\$\$\$A$ and it will be converted into the codon switch sequence with $A\$\$\rightarrow ATA$, $\$\$\$\rightarrow CGT$, and $\$\$A\rightarrow CTA$ codon switches. All the switches representing deletions can also be constructed by following the same steps, except that the space character is introduced in the altered codon. Switches representing complex insertions and deletions can also be constructed where space character appears in both reference and altered codons. The cumulative number of codon switches rises to $15,625$ after considering complex mutations and indels (Figure~\ref{fig:ch7:multimutation}).

To do a fair analysis, we computed the frequency of different codon switches on the inflated mutation counts from the ExAC DB (Section~\ref{sec:ch6:coding}). We obtained an average of $495$ variants associated with the 640 codon switches representing single base substitution. Notably, all the codon switches with one alteration were present in the datasets. We found at least one variant from the same database for 66 out of the 1728 possible coding switches representing double mutations (consecutive and non-consecutive). The average per switch variants obtained from the database was $\sim$1.5. For the triple substitutions, an average of 1.7 variants were spotted for 53 (out of 1728) switches. Statistics for the remaining variant types can be found in Table~\ref{tab:ch7:switchFreq}.

These double/triple substitutions and insertion and deletion frequencies are not enough to learn the embeddings. Given the sufficient amount of data, the dictionary constructed after all inclusions can be used to learn embeddings. 

\subsection{Application of \acrshort{crcs} in influenza and other diseases}

The study presented the human genome (genotype) in the context of cancer (phenotype). However, the concept generated in this work is generic in nature and can be utilized with any organism to quantify genotype-phenotype association. The possible but not limited use-cases for the \acrshort{crcs} may be:

\begin{itemize}
    \item The representation can be utilized to identify mutations in influenza and other related viruses like SARS-CoV19 which can cause outbreaks.
    \item The representation can be used to measure the drop in the response of vaccines in the presence of certain mutations in the viruses.
    \item This approach can be used to assess the deleteriousness of casual variants associated with neurodegenerative diseases like Alzheimer's and Parkinson’s~\cite{reynaud2010protein}.
    \item The proposed method can also be extended to understand the role of splice site point mutations and their implication in various diseases such as congenital cataracts and Becker muscular dystrophy~\cite{anna2018splicing}.
\end{itemize}

This approach can also be used to train models that can identify previously uncharacterized deleterious mutations that can lead to structural changes to protein and corroborate computational structural biology approaches. I am hopeful that the developed technique may be used to answer many questions related to the genotype-phenotype association and help progress the field of genomics and artificial intelligence.

%% file: publications.tex
\section*{Publications related to thesis chapters}

\begin{enumerate}

\item Gupta P, Jindal A, Ahuja G, Jayadeva, Sengupta D. A new deep learning technique reveals the exclusive functional contributions of individual cancer mutations. Journal of Biological Chemistry. 2022 Jun 24:102177.

\item Gupta P, Jindal A, Jayadeva, Sengupta D. ComBI: Compressed Binary Search Tree for Approximate k-NN Searches in Hamming Space. Big Data Research. 2021 Jul 15;25:100223.
\end{enumerate}

\section*{Co-Authored Publications} 

\begin{enumerate}
\item Jindal A, Gupta P, Jayadeva, Sengupta D. Discovery of rare cells from voluminous single cell expression data. Nature communications. 2018 Nov 9;9(1):1-9.
\item Gupta P, Jindal A, Jayadeva, Sengupta D. Linear time identification of local and global outliers. Neurocomputing. 2021 Mar 14;429:141-50.
\item Jindal A, Gupta P, Sengupta D., Jayadeva Enhash: A Fast Streaming Algorithm for Concept Drift Detection. ESANN 2021 proceedings. Online event, 6-8 October 2021, i6doc.com publ., ISBN 978287587082-7.
\end{enumerate}

\section*{Preprints related to thesis chapters}

\begin{enumerate}
\item Gupta P, Jindal A, Jayadeva, Sengupta D. Guided Random Forest and its application to data approximation. arXiv preprint arXiv:1909.00659. 2019 Sep 2.
\item Gupta, P., Jindal, A. and Sengupta, D. Deep learning discerns cancer mutation exclusivity. bioRxiv:2020.04.09.022731. 2020 Apr 10.
\end{enumerate}

%% file: biodata.tex
\section*{Name: Prashant Gupta}


\section*{Educational Qualifications}

\textbf{Ph.D.} (Completed) \hfill 2023\\
Department of Electrical Engineering\\
Indian Institute of Technology Delhi, Delhi India.\\

\noindent Bachelor of Technology (B.Tech.) \hfill 2013\\
Dept. of Electronics \& Communication Engg. \\
Motilal Nehru National Institute of Technology, Allahabad, Uttar Pradesh, India.

\section*{Areas of Interest}
Machine Learning, Deep Learning, Computational Biology, Bioinformatics

\section*{Industrial Experience}
GenterpretR Inc. \hfill Apr. 2023 - till now\\
Computational Biology \& Machine Learning Advisor.\\

\noindent GenterpretR Inc. \hfill Jan. 2022 - Mar. 2023\\
Chief Technology Officer.\\

\noindent Fintilla Pte. Ltd. \hfill Jan. 2022 - Mar. 2022\\
Deep learning consultant.\\

\noindent NableIT Consultancy Pvt. Ltd. \hfill Jan. 2021 - Jan. 2022\\
Visual Analytics Team Lead.\\

\noindent Amelia, An IPSoft Company \hfill Sept. 2020 - Jan 2021\\
Research \& Development Engineer.\\